\documentclass{scrbook}

\usepackage{etex}

\usepackage{hyperref}

\usepackage[utf8]{inputenc}
\usepackage[english]{babel}

\usepackage{amsthm,amsmath}
\usepackage{stmaryrd}
\usepackage{txfonts}

\usepackage{setspace}
\usepackage{tikz-cd}
\tikzcdset{
  shift left/.default=+0.7ex,
  shift right/.default=+0.7ex}
\newenvironment{tikzar}[1][]{{}\kern-4pt\begin{tikzcd}[ampersand replacement=\&,#1]}%
{\end{tikzcd}\kern-4pt{}}
\tikzstyle{edge label
}=[font=\scriptsize]
\tikzcdset{every label/.append style = {font = \scriptsize}}
\tikzcdset{every node/.append style = {font = \scriptsize}}
\tikzcdset{arrow style = tikz,diagrams={>={Straight Barb[scale=0.8]}}
}

\usepackage{txfonts}
\usepackage{graphicx}

\usepackage{multicol}
\usepackage{multirow}
\usepackage{hhline}
\usepackage{array}
\newcolumntype{L}[1]{>{\raggedright\let\newline\\\arraybackslash\hspace{0pt}}m{#1}}
\newcolumntype{C}[1]{>{\centering\let\newline\\\arraybackslash\hspace{0pt}}m{#1}}
\newcolumntype{R}[1]{>{\raggedleft\let\newline\\\arraybackslash\hspace{0pt}}m{#1}}

\usepackage{rotating}
\usepackage{array}

\usepackage[LGR,T1]{fontenc}

\usepackage{fancybox,graphics}
\usepackage{yfonts}
\newfont{\gothic}{ygoth at 12pt}

\theoremstyle{plain}

\theoremstyle{definition}

\theoremstyle{remark}

\usepackage{fullpage}
\usepackage{enumerate}

\makeatletter
\DeclareOldFontCommand{\rm}{\normalfont\rmfamily}{\mathrm}
\DeclareOldFontCommand{\sf}{\normalfont\sffamily}{\mathsf}
\DeclareOldFontCommand{\tt}{\normalfont\ttfamily}{\mathtt}
\DeclareOldFontCommand{\bf}{\normalfont\bfseries}{\mathbf}
\DeclareOldFontCommand{\it}{\normalfont\itshape}{\mathit}
\DeclareOldFontCommand{\sl}{\normalfont\slshape}{\@nomath\sl}
\DeclareOldFontCommand{\sc}{\normalfont\scshape}{\@nomath\sc}
\makeatother

\providecommand*{\Ifstr}{\ifstr}
\usepackage{blindtext}
\usepackage{xcolor}
\definecolor{lightBlue}{rgb}{.55,.55,1}
\definecolor{lightRed}{rgb}{1,.55,.55}

\usepackage{scrwfile}
\usepackage{xpatch}

\colorlet{partcolor}{blue}
\addtokomafont{partprefix}{\color{partcolor}}
\addtokomafont{part}{\normalcolor}
\RedeclareSectionCommand[
  tocdynnumwidth=true,%
  tocbeforeskip=1em
]{part}
\RedeclareSectionCommand[
  beforeskip=0pt,
  afterindent=false,
  afterskip=2\baselineskip,
  tocdynnumwidth,
  tocbeforeskip=1em plus 1pt minus 1pt,
  toclinefill=\TOCLineLeaderFill,
    tocindent=1.5em
]{chapter}
\RedeclareSectionCommand[
  beforeskip=\baselineskip,
  afterindent=false,
  afterskip=.5\baselineskip,
  tocindent=1.5em,
  tocnumwidth=3.5em
]{section}
\RedeclareSectionCommand[
  beforeskip=.75\baselineskip,
  afterindent=false,
  afterskip=.5\baselineskip,
  tocindent=3.8em,
  tocnumwidth=4em
]{subsection}
\RedeclareSectionCommand[
  beforeskip=.5\baselineskip,
  afterindent=false,
  afterskip=.25\baselineskip,
  tocindent=7em,
  tocnumwidth=4.1em
]{subsubsection}
\RedeclareSectionCommand[
  tocindent=10em,
  tocnumwidth=5em
]{paragraph}
\RedeclareSectionCommand[
  tocindent=12em,
  tocnumwidth=6em
]{subparagraph}

\RedeclareSectionCommands
  [tocpagenumberwidth=3em]
  {part,chapter,section,subsection,paragraph,subparagraph}

\makeatletter
\newif\ifuseparttoc
\newcommand*{\parttoc}[1][\thepart]{
  \useparttoctrue
  \edef\ext@parttoc{tcp#1}
  \DeclareNewTOC[
    listname=Contents,
  ]{\ext@parttoc}
  \begingroup
    \value{tocdepth}=\chaptertocdepth
    \listoftoc{\ext@parttoc}
  \endgroup
}
\xapptocmd\addtocentrydefault{
  \ifuseparttoc
    \expandafter\tocbasic@addxcontentsline\expandafter{\ext@parttoc}{#1}{#2}{#3}
  \fi
}{}{}
\xpretocmd\part{\useparttocfalse}{}{}
\newif\ifusechaptertoc
\newcommand*{\chaptertoc}[2][\thechapter]{
  \usechaptertoctrue
  \edef\ext@chaptoc{tcc#1}
  \DeclareNewTOC{\ext@chaptoc}
  \setchapterpreamble{%
    \begin{minipage}{\linewidth}
      \hrulefill\par
      \value{tocdepth}=\subsectiontocdepth
      \listoftoc*{\ext@chaptoc}
    \end{minipage}%
    \par\bigskip\nobreak\noindent\hrulefill\par
    \bigskip\noindent\ignorespaces
  }%
}
\xapptocmd\addtocentrydefault{
  \ifusechaptertoc
    \Ifstr{#1}{chapter}{}
      {\expandafter\tocbasic@addxcontentsline\expandafter{\ext@chaptoc}{#1}{#2}{#3}}
  \fi
}{}{}
\xpretocmd\chapter{\usechaptertocfalse}{}{}
\xpretocmd\part{\usechaptertocfalse}{}{}
\makeatother

\newcommand\setchaptertoc[1][]{%
  \Ifstr{#1}{}
    {\AddtoOneTimeDoHook{heading/preinit/chapter}{\chaptertoc}}
    {\AddtoOneTimeDoHook{heading/preinit/chapter}{\chaptertoc[#1]}}%
}

\mathcode`\<="4268 
\mathcode`\>="5269 
\mathchardef\gt="313E 
\mathchardef\lt="313C 

\newcommand{\JJJ}{{\cal J}}

\newcommand{\LLL}{{\cal L}}
\newcommand{\MMM}{{\cal M}}

\newcommand{\RRR}{{\cal R}}

\newcommand{\UUU}{{\cal U}}

\newcommand{\Bbb}{\mathbb}

\newcommand{\NNn}{{\Bbb N}}

\newcommand{\RRr}{{\Bbb R}}

\newcommand{\ZZz}{{\Bbb Z}}

\newcommand{\WP}{\mbox{\large $\wp$}}

\newcommand{\sss}{\mathbf{s}}

\renewcommand{\vec}{\mathbf}

\def\pushright#1{{
   \parfillskip=0pt            
   \widowpenalty=10000         
   \displaywidowpenalty=10000  
   \finalhyphendemerits=0      
   \leavevmode                 
   \unskip                     
   \nobreak                    
   \hfil                       
   \penalty50                  
   \hskip.2em                  
   \null                       
   \hfill                      
   {#1}                        
   \par}}                      

\def\qed{\pushright{$\boxempty$}\penalty-700 \smallskip}

\newenvironment{prf}[1]{\begin{trivlist} \item[{\bf ~Proof}#1.]}%
{\qed\end{trivlist}}

\newcommand{\beq}{\begin{equation}}
\newcommand{\eeq}{\end{equation}}
\newcommand{\bea}{\begin{eqnarray}}
\newcommand{\eea}{\end{eqnarray}}
\newcommand{\bear}{\begin{eqnarray*}}
\newcommand{\eear}{\end{eqnarray*}}
\newcommand{\bpr}{\begin{prf}{}}
\newcommand{\epr}{\end{prf}}
\newcommand{\bprf}[1]{\begin{prf}{#1}}
\newcommand{\eprf}{\end{prf}}

{\end{itemize}}

\newdimen\proofrulebreadth \proofrulebreadth=.05em
\newdimen\proofdotseparation \proofdotseparation=1.25ex
\newdimen\proofrulebaseline \proofrulebaseline=2ex
\newcount\proofdotnumber \proofdotnumber=3
\let\then\relax
\def\hfi{\hskip0pt plus.0001fil}
\mathchardef\squigto="3A3B
\newif\ifinsideprooftree\insideprooftreefalse
\newif\ifonleftofproofrule\onleftofproofrulefalse
\newif\ifproofdots\proofdotsfalse
\newif\ifdoubleproof\doubleprooffalse
\let\wereinproofbit\relax
\newdimen\shortenproofleft
\newdimen\shortenproofright
\newdimen\proofbelowshift
\newbox\proofabove
\newbox\proofbelow
\newbox\proofrulename
\def\shiftproofbelow{\let\next\relax\afterassignment\setshiftproofbelow\dimen0 }
\def\shiftproofbelowneg{\def\next{\multiply\dimen0 by-1 }%
\afterassignment\setshiftproofbelow\dimen0 }
\def\setshiftproofbelow{\next\proofbelowshift=\dimen0 }
\def\setproofrulebreadth{\proofrulebreadth}

\def\prooftree{
\ifnum  \lastpenalty=1
\then   \unpenalty
\else   \onleftofproofrulefalse
\fi
\ifonleftofproofrule
\else   \ifinsideprooftree
        \then   \hskip.5em plus1fil
        \fi
\fi
\bgroup
\setbox\proofbelow=\hbox{}\setbox\proofrulename=\hbox{}%
\let\justifies\proofover\let\leadsto\proofoverdots\let\Justifies\proofoverdbl
\let\using\proofusing\let\[\prooftree
\ifinsideprooftree\let\]\endprooftree\fi
\proofdotsfalse\doubleprooffalse
\let\thickness\setproofrulebreadth
\let\shiftright\shiftproofbelow \let\shift\shiftproofbelow
\let\shiftleft\shiftproofbelowneg
\let\ifwasinsideprooftree\ifinsideprooftree
\insideprooftreetrue
\setbox\proofabove=\hbox\bgroup$\displaystyle 
\let\wereinproofbit\prooftree
\shortenproofleft=0pt \shortenproofright=0pt \proofbelowshift=0pt
\onleftofproofruletrue\penalty1
}

\def\eproofbit{
\ifx    \wereinproofbit\prooftree
\then   \ifcase \lastpenalty
        \then   \shortenproofright=0pt  
        \or     \unpenalty\hfil         
        \or     \unpenalty\unskip       
        \else   \shortenproofright=0pt  
        \fi
\fi
\global\dimen0=\shortenproofleft
\global\dimen1=\shortenproofright
\global\dimen2=\proofrulebreadth
\global\dimen3=\proofbelowshift
\global\dimen4=\proofdotseparation
\global\count255=\proofdotnumber
$\egroup  
\shortenproofleft=\dimen0
\shortenproofright=\dimen1
\proofrulebreadth=\dimen2
\proofbelowshift=\dimen3
\proofdotseparation=\dimen4
\proofdotnumber=\count255
}

\def\proofover{
\eproofbit 
\setbox\proofbelow=\hbox\bgroup 
\let\wereinproofbit\proofover
$\displaystyle
}%
\def\proofoverdbl{
\eproofbit 
\doubleprooftrue
\setbox\proofbelow=\hbox\bgroup 
\let\wereinproofbit\proofoverdbl
$\displaystyle
}%
\def\proofoverdots{
\eproofbit 
\proofdotstrue
\setbox\proofbelow=\hbox\bgroup 
\let\wereinproofbit\proofoverdots
$\displaystyle
}%
\def\proofusing{
\eproofbit 
\setbox\proofrulename=\hbox\bgroup 
\let\wereinproofbit\proofusing
\kern0.3em$
}

\def\endprooftree{
\eproofbit 
  \dimen5 =0pt
\dimen0=\wd\proofabove \advance\dimen0-\shortenproofleft
\advance\dimen0-\shortenproofright
\dimen1=.5\dimen0 \advance\dimen1-.5\wd\proofbelow
\dimen4=\dimen1
\advance\dimen1\proofbelowshift \advance\dimen4-\proofbelowshift
\ifdim  \dimen1<0pt
\then   \advance\shortenproofleft\dimen1
        \advance\dimen0-\dimen1
        \dimen1=0pt
        \ifdim  \shortenproofleft<0pt
        \then   \setbox\proofabove=\hbox{%
                        \kern-\shortenproofleft\unhbox\proofabove}%
                \shortenproofleft=0pt
        \fi
\fi
\ifdim  \dimen4<0pt
\then   \advance\shortenproofright\dimen4
        \advance\dimen0-\dimen4
        \dimen4=0pt
\fi
\ifdim  \shortenproofright<\wd\proofrulename
\then   \shortenproofright=\wd\proofrulename
\fi
\dimen2=\shortenproofleft \advance\dimen2 by\dimen1
\dimen3=\shortenproofright\advance\dimen3 by\dimen4
\ifproofdots
\then
        \dimen6=\shortenproofleft \advance\dimen6 .5\dimen0
        \setbox1=\vbox to\proofdotseparation{\vss\hbox{$\cdot$}\vss}%
        \setbox0=\hbox{%
                \advance\dimen6-.5\wd1
                \kern\dimen6
                $\vcenter to\proofdotnumber\proofdotseparation
                        {\leaders\box1\vfill}$%
                \unhbox\proofrulename}%
\else   \dimen6=\fontdimen22\the\textfont2 
        \dimen7=\dimen6
        \advance\dimen6by.5\proofrulebreadth
        \advance\dimen7by-.5\proofrulebreadth
        \setbox0=\hbox{%
                \kern\shortenproofleft
                \ifdoubleproof
                \then   \hbox to\dimen0{%
                        $\mathsurround0pt\mathord=\mkern-6mu%
                        \cleaders\hbox{$\mkern-2mu=\mkern-2mu$}\hfill
                        \mkern-6mu\mathord=$}%
                \else   \vrule height\dimen6 depth-\dimen7 width\dimen0
                \fi
                \unhbox\proofrulename}%
        \ht0=\dimen6 \dp0=-\dimen7
\fi
\let\doll\relax
\ifwasinsideprooftree
\then   \let\VBOX\vbox
\else   \ifmmode\else$\let\doll=$\fi
        \let\VBOX\vcenter
\fi
\VBOX   {\baselineskip\proofrulebaseline \lineskip.2ex
        \expandafter\lineskiplimit\ifproofdots0ex\else-0.6ex\fi
        \hbox   spread\dimen5   {\hfi\unhbox\proofabove\hfi}%
        \hbox{\box0}%
        \hbox   {\kern\dimen2 \box\proofbelow}}\doll%
\global\dimen2=\dimen2
\global\dimen3=\dimen3
\egroup 
\ifonleftofproofrule
\then   \shortenproofleft=\dimen2
\fi
\shortenproofright=\dimen3
\onleftofproofrulefalse
\ifinsideprooftree
\then   \hskip.5em plus 1fil \penalty2
\fi
}

\usepackage{array}
\usepackage{multirow}

\newcolumntype{L}[1]{>{\raggedright\let\newline\\\arraybackslash\hspace{0pt}}m{#1}}
\newcolumntype{C}[1]{>{\centering\let\newline\\\arraybackslash\hspace{0pt}}m{#1}}
\newcolumntype{R}[1]{>{\raggedleft\let\newline\\\arraybackslash\hspace{0pt}}m{#1}}

\newcommand{\para}[1]{\noindent\textbf{\textsf{#1}}}

\renewcommand{\to}{\longrightarrow}

\newcommand{\tto}[1]{\xrightarrow{#1}}
\newcommand{\oot}[1]{\xleftarrow{#1}}

\newcommand{\mono}{\rightarrowtail}

\renewcommand{\paragraph}[1]{\para{#1}}

\newcommand{\Term}{\Sigma}
\newcommand{\Type}{\Xi}
\newcommand{\Label}{\Lambda}
\newcommand{\Rule}{[::=]}
\newcommand{\prule}{::=}
\newcommand{\srule}{::\stackrel{\ast}=}
\newcommand{\Sen}{S}

\newcommand{\seq}[1]{\big(\, #1\, \big)}
\newcommand{\sseq}[1]{\left(\, #1\, \right)}
\newcommand{\length}[1]{\lvert {#1}\rvert}

\newcommand{\TF}{\mathsf{TF}}
\newcommand{\FT}{\mathsf{FT}}
\newcommand{\DF}{\mathsf{DF}}
\newcommand{\IDF}{\mathsf{IDF}}

\newcommand{\bra}[1]{<{#1}\, |}\newcommand{\ket}[1]{|\, {#1}\,>}
\newcommand{\braa}[1]{<\mbox{#1}|}
\newcommand{\kett}[1]{|\,\mbox{#1}\,>}
\newcommand{\braket}[2]{<{#1}\, |\, {#2}>}
\newcommand{\brakett}[2]{<\mbox{#1}\, |\, \mbox{#2}>}
\newcommand{\brakem}[3]{<{#1}\, |\, {#2}\,|\, {#3}>}

\newcommand{\pder}[2]{\big[#1\ {\mbox{\large ${\vdash}$}}\ #2\big]}
\newcommand{\ppder}[1]{\big[#1\big]}
\newcommand{\pderr}[2]{\big[\mbox{#1}\ {\mbox{\large ${\vdash}$}}\ \mbox{#2}\big]}
\newcommand{\ppderr}[1]{\big[\mbox{#1}\big]}

\newcommand{\sgn}{\mathrm{sgn}}

\newcommand{\noback}{{\, \times\hspace{-1.2em}\leftarrow}}
\newcommand{\nofw}{{\, \times\hspace{-1em}\rightarrow}}
\newcommand{\nomem}{{\, \downarrow \hspace{-.3em}Y}}
\newcommand{\markovian}{{\, \downarrow \hspace{-.3em}X}}

\newcommand{\aprog}{\mathbf{a}}
\newcommand{\bprog}{\mathbf{b}}
\newcommand{\dprog}{\mathbf{d}}
\newcommand{\eprog}{\mathbf{e}}
\newcommand{\fprog}{\mathbf{f}}
\newcommand{\hprog}{\mathbf{h}}
\newcommand{\kprog}{\mathbf{k}}

\newcommand{\sprog}{\mathbf{s}}

\newcommand{\vprog}{\mathbf{v}}

\renewcommand{\upsilon}{\UUU}

\begin{document}

\frontmatter
\pagenumbering{roman}
\thispagestyle{empty}

\begin{titlepage}
\title{{\Huge Language processing}\\[1ex]
{\Huge in humans and computers}
}

\author{Dusko Pavlovic}
\date{
\vspace{1.5cm}
\includegraphics[height=10cm]{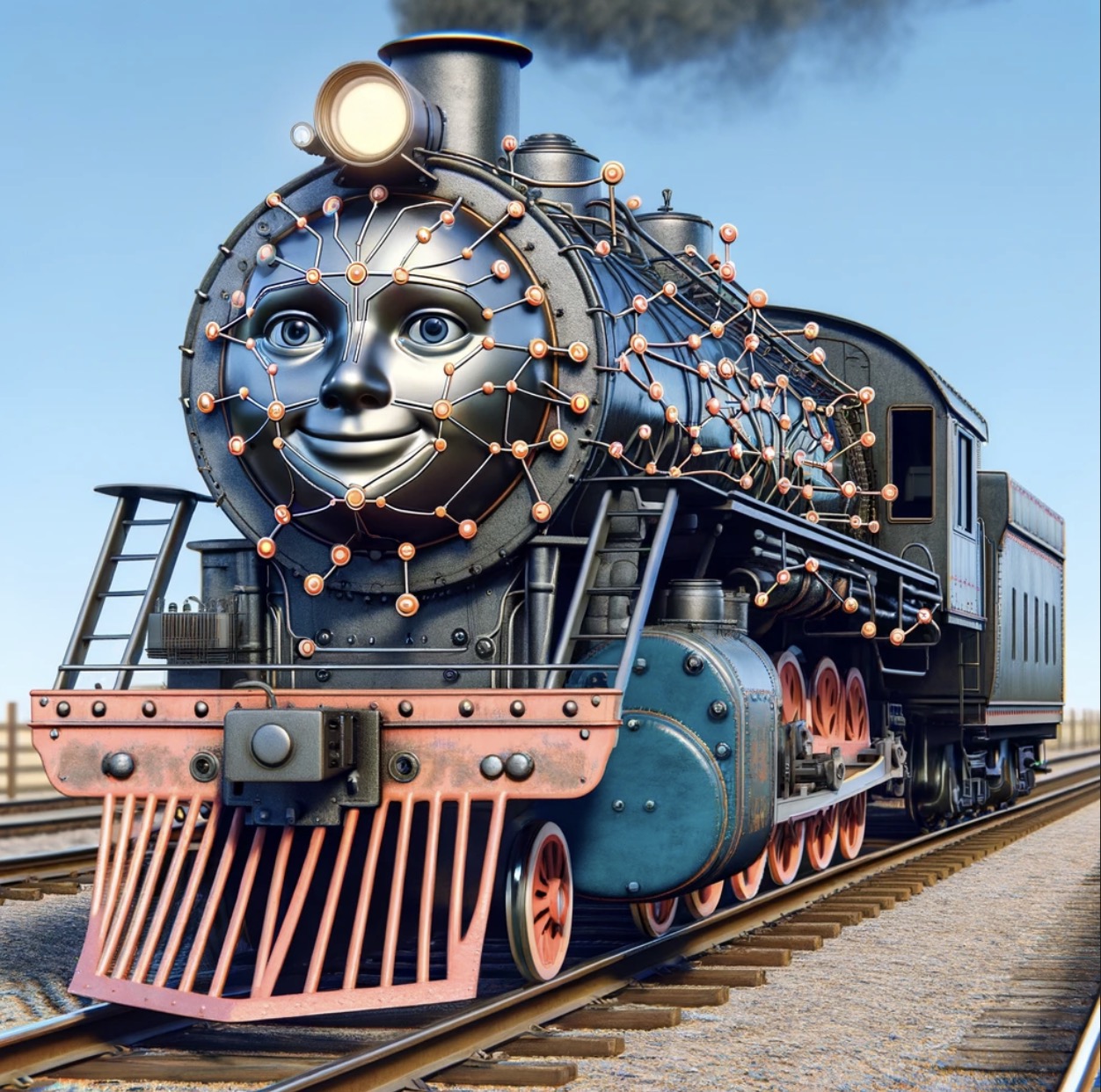}}
\end{titlepage}

\maketitle

\newpage
\begin{center} {\huge \bf Abstract}
\end{center}
\pagestyle{plain}
{\LARGE
Machine-learned language models have transformed everyday life: they steer us when we study, drive, manage money. They have the potential to transform our civilization. But they hallucinate. Their realities are virtual. 

This course provides a high-level overview of language models and outlines a low-level model of learning machines. 

It turns out that, after they become capable of recognizing hallucinations and dreaming safely, as humans tend to be, the language-learning machines proceed to generate broader systems of false beliefs and self-confirming theories, as humans tend to do.

}

\pagestyle{plain}
\tableofcontents

\chapter{Preface: On language models and celebrities} 

Anyone can drive a car. Most people even know what the engine looks like. But when you need to fix it, you need to figure out how it works.

Anyone can chat with a chatbot. Most people know that there is a Large Language Model (LLM) under the hood. There are lots and lots and lots of articles describing what an LLM looks like. Lots of colorful pictures. Complicated meshes of small components, as if both mathematical abstraction and  modular programming still need to be invented. YouTube channels with fresh scoops on LLM celebrities.  We get to know their parts, how they are connected, we know their performance, we even see how each of them changes a heat map of inputs to a heat map of outputs. One hotter than another. But do we understand how they work? Experts say that they do, but they don't seem to be able to explain it even to each other, as they continue to disagree about pretty much everything. 

\begin{figure}[!ht]
\begin{center}
\includegraphics[height=13cm]{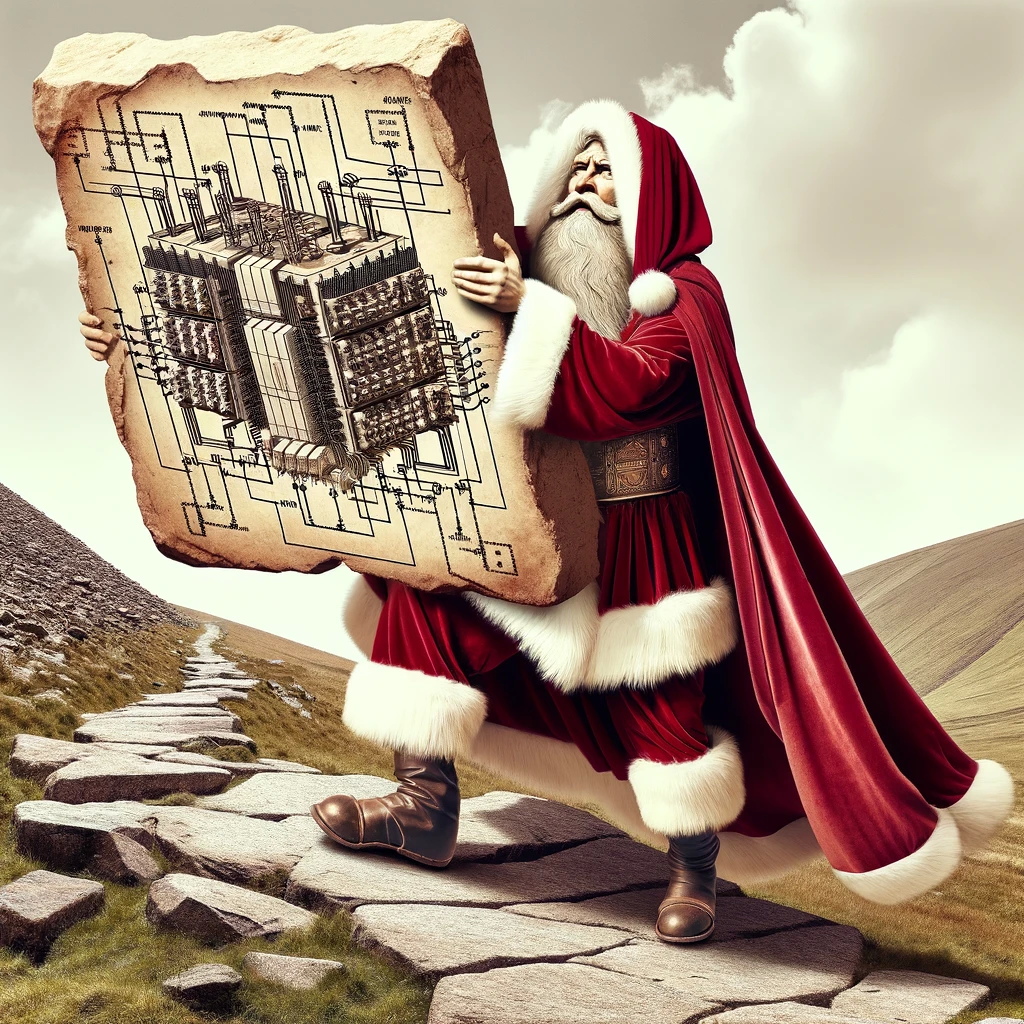}
\caption{Transformer architecture revealed}
\label{Fig:santa-1}
\end{center}
\end{figure}

Every child, of course, knows that it can be hard to explain what you just built. Our great civilization built lots of stuff that was unable to explain. Steam engines have been engineered for nearly 2000 years before scientists explained how they extract work from heat. There aren't many steam engines around anymore, but there are lots of language engines and a whole industry of scientific explanations how they work. The leading theory is that Santa Claus descended from the mountain and gave us the transformer architecture carved in a stone tablet. It changed the world, spawned offspring and competitors\ldots Just like steam engines. Which may be a good thing, since steam engines did not exterminate their creators just because the creators didn't understand them.

I wasn't around in the times of steam engines, but I was around in the times of bulky computers, and when the web emerged and everything changed, and when the web giants emerged and changed the web. Throughout that time, AI research seemed like an effort towards the intelligent design of intelligence. It didn't change anything, because intelligence, like life, is an evolutionary process, not a product of intelligent design\footnote{Alan Turing explained that machine intelligence could not be achieved through intelligent designs, because intelligence itself could not be completely specified, as it is its nature to always seek and find new paths. But Turing was also the first to realize that the process of computation was not bound by designs and specifications either, but could evolve and innovate. He anticipated that machine intelligence would evolve with computation. However, three years after Turing's death, the concept of machine intelligence, which he thought and wrote about for the last 8 years of his life, got renamed to artificial intelligence, his writings sank into oblivion, and the logical systems designed to capture intelligence proliferated.}.  
Now the friendly learning machines and chatbot AIs evolved and everything is changing again. Having survived and processed the paradigm shifts of the past, I am now trying to understand the present one. Hence this course and these writings. On one hand, I probably stand no chance to say anything that hasn't been said already. Even after a lot of honest work, I remain a short-sighted non-expert. On the other hand, as far as I can tell, the gentle AI experts seem unaware of some powerful tools and ideas that evolved in the neighborhood. People clump into research communities, focus on the same things, and ignore the same things. Looking over the fence, the neighbors sometimes see what they don't see. So here is a view across the fence, with all the bad and good things that go with it.
\label{Preface}


\mainmatter

\def\thechapter{1}
\setchaptertoc
\chapter{Who are chatbots and what are they to you?
}\label{Chap:Intro}

	
\section{Chatbots: Shifting the paradigm of meaning}

\subsection{What just happened?}
We live in strange times. 

\begin{figure}
\begin{center}
\includegraphics[height=20cm,trim={0 3.5cm 0 5.5cm},clip]{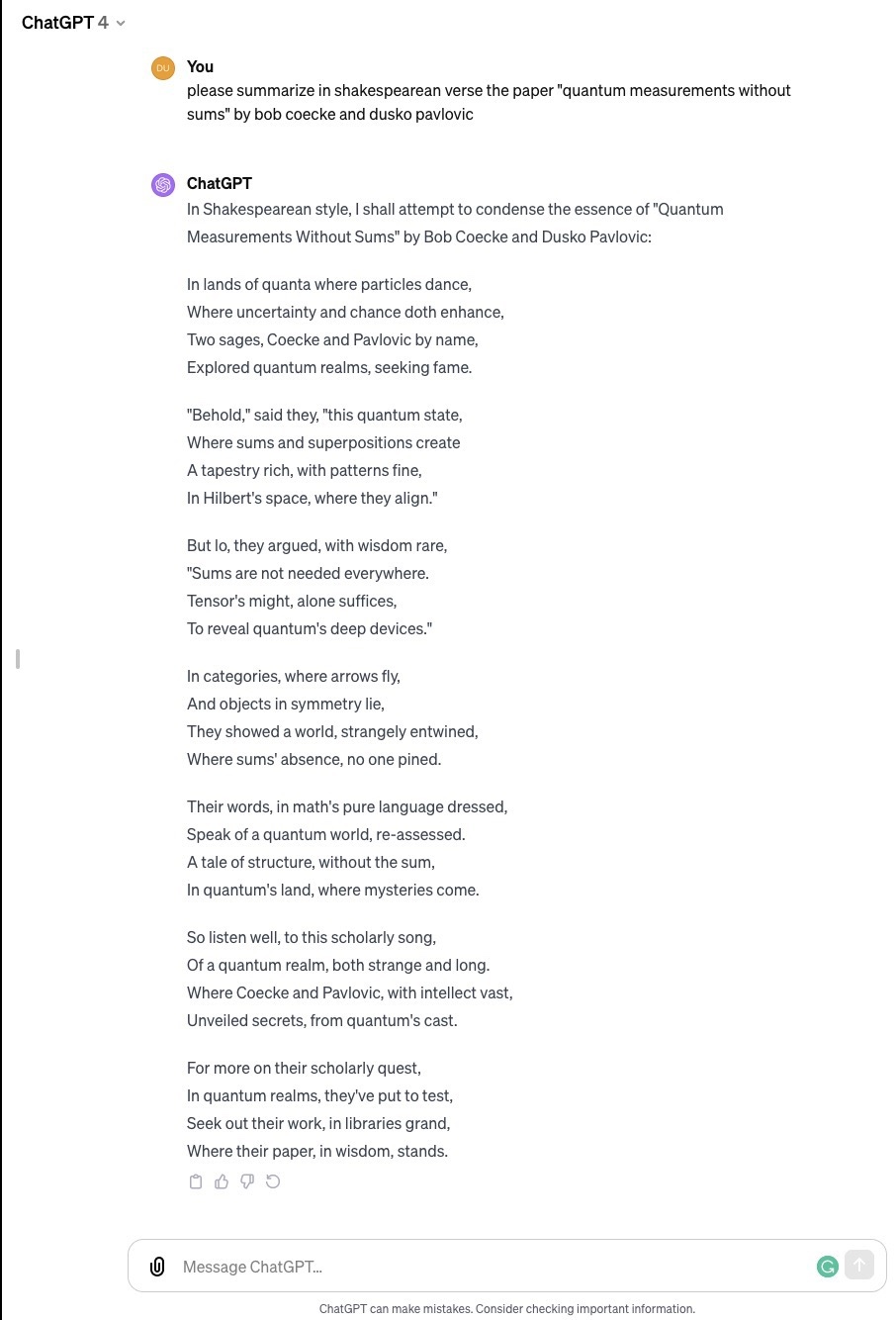}
\caption{GPT4's response to the prompt: \emph{``please summarize in shakespearean verse the paper `quantum measurements without sums' by bob coecke and dusko pavlovic''}}
\label{Fig:QMWS}
\end{center}
\end{figure}

Stories used to be told by storytellers, poems recited by poets, music played by musicians, science taught by teachers. Then the printing and recording  technologies made copying possible and the copyright got invented and the owners of the recording and printing equipment started earning more than musicians and storytellers. Then the web happened and it all came under your fingertips. Now chatbots happened and you can ask them to write poetry or to explain science, or even to combine the two, as in Fig.~\ref{Fig:QMWS}. 
They even seem to have sparks of a sense of humor. I asked a chatbot to translate a subtle metaphor from Croatian to English and she did it so well that I got an impulse to thank her in French, with \emph{``merci :)''} --- to which she pivoted to German: \emph{``Gern geschehen! If you have any more questions, feel free to ask. \includegraphics[scale=0.06]{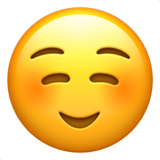}''}

\subsection{The paradigm of meaning}
Our interactions with computers evolved gradually. They have been able to maintain light conversations since the 1960s, and personal assistants conversing in human voices have been sold for a while. Not a big deal. Human appearances are in the eyes of human observers.

But with chatbots, the appearances seem to have gone beyond the eyes. When you and I chat, we assume that the same words mean the same things because we have seen the same world. If we talk about chairs, we can point to a chair. The word \emph{``chair''}\/ refers to a thing in the world.  

\begin{figure}[!ht]
\begin{center}
\includegraphics[height=2.5cm]{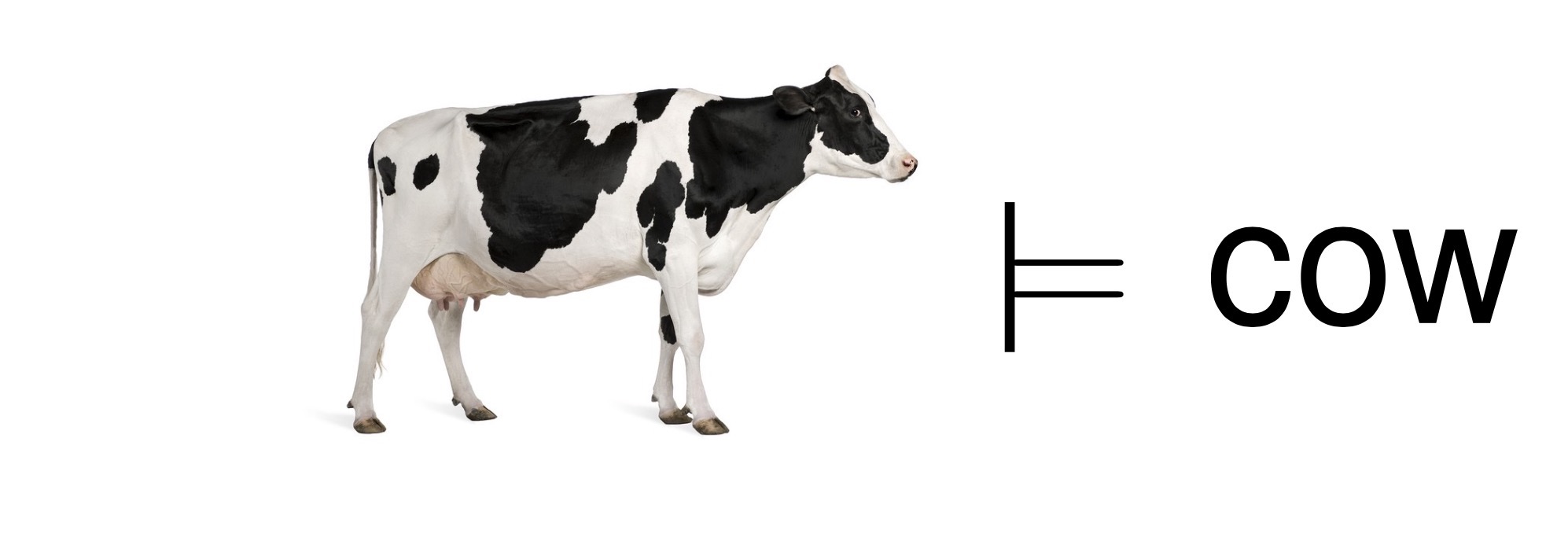}
\hspace{4em}
\includegraphics[height=2.5cm]{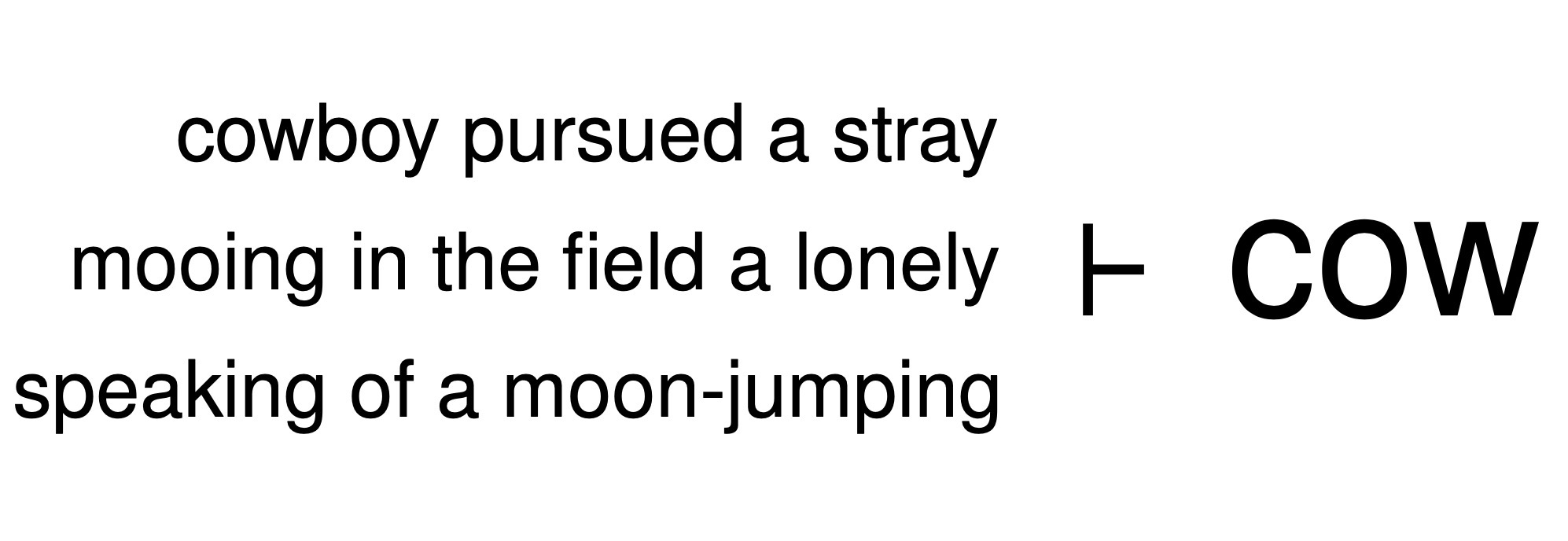}
\caption{People use words to refer to things, chatbots to extend phrases}
\label{Fig:cow}
\end{center}
\end{figure}

But a chatbot has never seen a chair, or anything else. For a chatbot, a word cannot possibly refer to a thing in the world, since it has no access to the world. It appears to know what it is talking about because it was trained on data downloaded from the web, which were uploaded by people who know what they are talking about. A chatbot has never seen a chair, or a tree, or a cow, or felt pain, but his chats are remixes of the chats between people who have. Chatbot's words do not refer to things directly, but indirectly, through people's words. Fig.~\ref{Fig:cow} illustrates how this works. 
When I say ``cow'', it is because I have a cow on my mind. When a chatbot says ``cow'', it is because that word is a likely completion of the preceding context.

Many theories of meaning have been developed in philosophy and linguistics.  While they differ in many things, they mostly stick with the picture in Fig.~\ref{Fig:cow} on the left, of meaning as a relation between a signifier and a signified item\footnote{Ferdinand de Saussure, one of the founders of structural linguistics, defined the meaning of a sign as a coupling between a readily available signifying token (``signifiant''), say a written or a spoken word, and a signified object (``signifi\'e'').}. While many semioticists noted that words and other fragments of language can also be referred to by words and other fragments of language, the fact that the entire process of language production can be realized within a self-contained system of references, as a whirlwind of words arising from words ---as \emph{demonstrated}\/ by chatbots--- is a fundamental,  unforeseen challenge.

%
%
%
%
 
According to most theories of meaning, chatbots should be impossible. Yet here they are! 

\subsection{Zeno and the aliens}
\begin{figure}[!ht]
\begin{center}
\includegraphics[height=8cm]{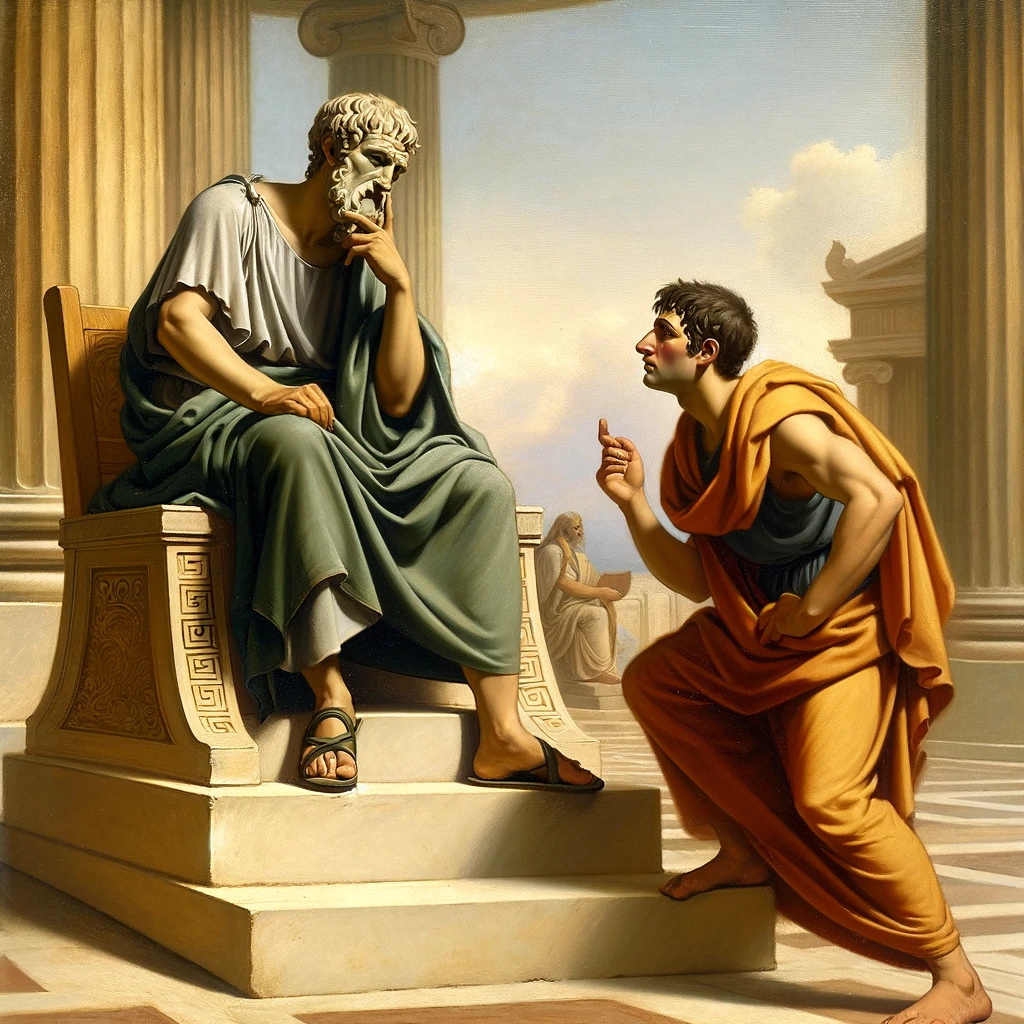}
\caption{{\small DALL-E's view: \emph{``As Parmenides argued that movement could not exist, Zeno paced around.''}}}
\label{Fig:Zeno}
\end{center}
\end{figure}

A chatbot familiar with pre-Socratic philosophy would be tempted to compare the conundrum of meaning with the conundrum of motion, illustrated in Fig.~\ref{Fig:Zeno}. 
Parmenides was a leading Greek philosopher in the VI--V centuries BCE. Zeno was his most prominent pupil. To a modern eye, the illustration seems to be showing Zeno is disproving Parmenides' central thesis by providing an empiric counterexample. For all we know, Zeno did not intend to disprove his teacher's claims and neither Parmenides nor Plato (who presented Parmenides' philosophy in eponymous dialogue) seem to have noticed the tension between Parmenides denial of the possibility of movements and Zeno's actual movements. Philosophy and students pacing around were not viewed in the same realm\footnote{Ironically, when the laws of motion were finally understood some 2000 years later, Parmenides' argument, popularized in Zeno's story about Achilles and tortoise, played an important role.}.  

Before you dismiss concerns about words and things and Zeno and the chatbots as a philosophical conundrum of no consequence for our projects in science and engineering, note that the self-contained language engine, realized by chatbot researchers and companies, could just as easily be realized by an alien spaceship parked behind the Moon. Aliens could also crawl the web, scrub our data, build neural networks, train them to speak on our favorite topics in perfect English, provide compelling explanations, illustrate them in vivid colors, just like our friendly AIs. That's not science fiction anymore. 

There was a movie where the landing on the Moon was staged on Earth. Maybe the Moon landing was real but the last World Cup final was modified by AI. Or maybe it wasn't modified but the losing team can easily prove that it could have been, and the winning team would have more trouble to prove that it wasn't. Conspiracy theorists are, of course, mostly easy to recognize, and funny or not. But there is an underlying logical fact worth taking note of: \emph{False statements seem  easier to prove than to disprove.} 

Without pursuing the fiction realized by the AI science any further, it seems clear that the boundaries between science and fiction, and between fiction and reality, may have been breached in ways unseen before. We live in strange times.

\section{AI: How did we get here and where do we go?}

\subsection{The mind-body problem and solution}
The idea of machine intelligence goes back to Alan Turing, the mathematician who defined and described the processes of computation that surround us. At the age of 19, Alan confronted the problem of mind ---where it comes from and where it goes?--- when he suddenly lost a friend with whom he had just fallen in love.

Some 300 years earlier, philosopher Ren\'e Descartes was pondering about the human body. One of the first steps into modern science was his realization that living organisms were driven by the same physical mechanisms as the rest of nature, i.e. that they were in essence similar to the machines built at the time. One thing that he couldn't figure out was how the human body gives rise to the human mind. He stated that as the \emph{mind-body problem}. 

\begin{figure}[!ht]
\begin{center}
\includegraphics[height=5cm
]{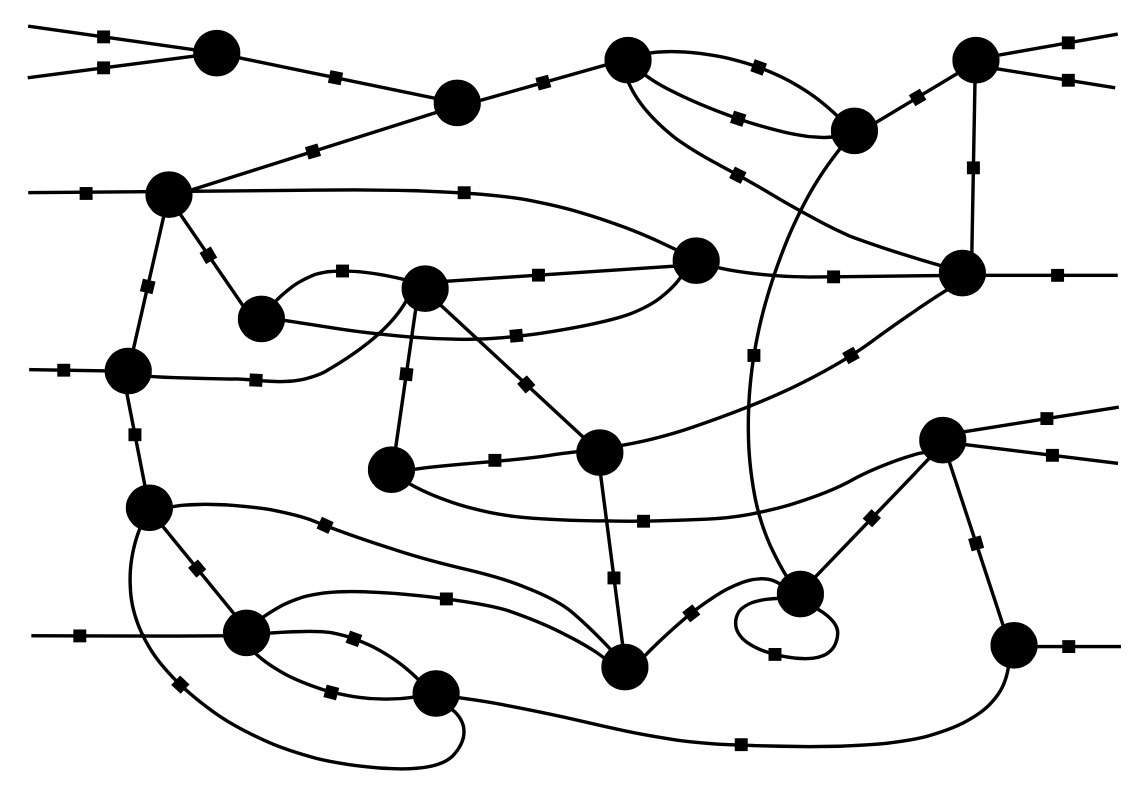}
\caption{Network of neurons from Turing's 1947 memo on \emph{Intelligent Machinery}}
\label{Fig:turing-NN}
\end{center}
\end{figure}

Alan Turing essentially solved the mind-body problem. His description of computation as a process implementable in machines, and his results proving that that process can simulate our reasoning and calculations, suggested that the mind may be arising from the body as a system of computational processes. He speculated that some version of computation was implemented in our neurons. The neural network in Fig.~\ref{Fig:turing-NN} is from Turing's 1947 research report. 
Since such computational processes are implementable in machines, it was reasonable to expect that they could give rise to machine intelligence. The title of the 1947 report was \emph{``Intelligent machinery''}.

\subsection{Turing and Darwin}
During WWII, Turing's theoretical research was superseded by cryptanalysis at Bletchley Park\footnote{When the war became imminent, instead of staying at Princeton as most European visitors did, Turing returned to lead teams that deciphered German Enigma code, which saved 1000s of lives by preventing submarine attacks on civilian traffic.}. When the war ended, he turned down positions at Cambridge and Princeton and accepted work at the National Physics Laboratory, hoping to build a computer and test the idea of machine intelligence. The 1947 memo contains the ideas of training neural networks, and of supervised and unsupervised learning. It was so far ahead of its time that some parts still seem ahead. It was submitted to the director of the National Physics Laboratory Sir Charles G.~Darwin, a grandson of Charles Darwin and a prominent eugenicist on his own account. In Sir Darwin's opinion, Turing's memo read like  ``a fanciful school-boy's essay''. Also resenting Turing's ``smudgy'' appearance, Sir Darwin smothered the computer project by putting it under strict administrative control. The machine intelligence memo sank into oblivion for more than 20 years. Turing devoted the final years until his death (at the age of 42, by biting into a cyanide-laced apple) to exploring the computational aspects of life. 

Two years after Turing's death (nine years after the \emph{Intelligent Machinery} memo), Turing's machine intelligence got renamed to \emph{artificial intelligence (AI)} at the legendary Dartmouth workshop. The history of artificial intelligence grew into a history of efforts towards \emph{\textbf{intelligent design} of intelligence}. The main research efforts were to logically reconstruct phenomena of human behavior, like affects, emotions, common sense, etc.; and to realize them in software.  Turing, in contrast, was assuming \emph{machine intelligence would \textbf{evolve spontaneously}}. The main published account of his thoughts on the topic appeared in the journal \emph{``Mind''}. The paper opens with the question:  ``Can machines think?''. What we now call the \emph{Turing Test}\/ is offered as a means for deciding the answer. The idea is that a machine that can maintain a conversation and remain indistinguishable from a thinking human being must be recognized as a thinking machine.  At the time of confusion around chatbots, the closing paragraph of the \emph{``Mind''} paper seems particularly interesting: 
\begin{quote}
An important feature of a learning machine is that its teacher will often be very largely ignorant of quite what is going on inside, although he may still be able to some extent to predict his pupil's behavior. [\ldots]  
This is in clear contrast with a normal procedure when using a machine to do computations: one's object is then to have a clear mental picture of the state of the machine at each moment in the computation. This object can only be achieved with a struggle.  The view that `the machine can only do what we know how to order it to do', appears strange in the face of this fact. Intelligent behaviour presumably consists in a departure from the completely disciplined behaviour.
\end{quote} 
\begin{figure}[!ht]
\begin{center}
\includegraphics[height=5cm
]{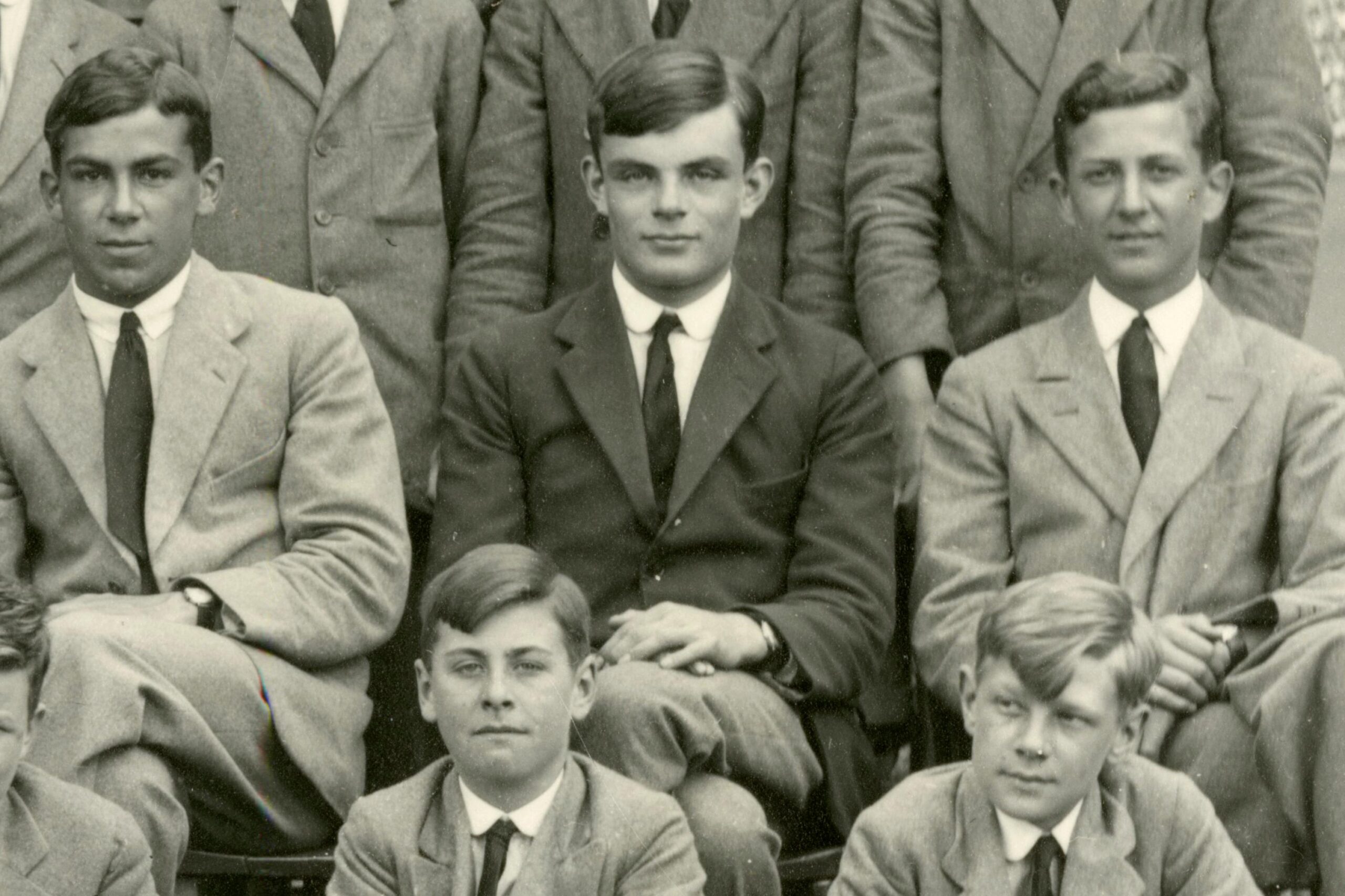}
\caption{17-year old Alan Turing }
\label{Fig:turing-boy}
\end{center}
\end{figure}
Designers and builders of chatbots and language engines published many accounts of their systems' methods and architectures but seem as stumped as everyone by their behaviors. Some initially denied the unexpected behaviors, but then stopped working and started talking about a threat to humanity. Turing anticipated the unexpected behaviors. His broader message seems to be that not knowing what is on the mind of another intelligent entity is not a bug but a feature of intelligence. That is why intelligent entities communicate. Failing that, they view each other as an object. Understanding chatbots may require broadening our moral horizons.\footnote{It is nice to think that Turing's mind lives in the mind of the machines that it conceived.} 

\subsection{From search engines to language engines}
One thing that Turing didn't get right was how intelligent machines would learn to reason. He imagined that they would learn from teachers. He didn't predict the Web. The Web provided the space to upload the human mind. The language models behind chatbots are 
\begin{itemize}
\item not a result of intelligent design of artificial intelligence,
\item but an effect of spontaneous evolution of the Web.
\end{itemize}
Like search engines, language models are fed data scraped from the Web by crawlers. A search engine then builds an index to serve data based on a ranking (be it of data quality or sponsorship), whereas a language engine builds a language model, to chat. In summary, the denominators evolve, but the proportions remain the same: 
{\small\[
\frac{ 
\textrm{semantic web (ontologies)}}{\textrm{web search (search engines)}} \ 
=\  \frac{\textrm{intelligent design (Adam \& Eve)}}{\textrm{evolution (Darwin)}} \ =\   \frac{\textrm{traditional AI (expert systems)}}{
\textrm{chatbots (language models)}}\]}

\subsection{What does the Web past say about the AI future?}
The space of language, computation, networks, and AI has many dimensions. The notion of mind has many dimensions and definitions. If we take into account that our mind depends on language, computation, networks, and AI as its tools and extensions, just like music depends on instruments, then it seems reasonable to say that we have gotten quite close to answering Turing's question whether machines can think. As we continue to use chatbots, as language models establish themselves as extensions of our language, our language will become a extension of language models. Our computers and devices are already a part of our thinking, our thinking will soon be a part of our computers and devices. Turing's question whether machines can think has become closely related with the question whether people can think.

Our daily life depends on our computers and devices. Children speak tablet menus together with their native languages. Networks absorb our thoughts, reprocess them, and feed them back to us, suitably rearranged. We absorb the information, reprocess it, and feed it back to our devices. This extended mind processes data by linking human and artificial network nodes and they are hard to distinguish, both for the network and for the nodes. This extended mind solves problems that no participant node could solve alone, by methods that are not available to any of them, and by nonlocal methods. Its functioning engenders nonlocal forms of awareness and attention. Language engines (they call themselves AIs) are built as a convenience for their human customers, and their machine thinking is meant to be a convenient extension of human thinking. But the universality of the underlying computation means that the machine thinking also subsumes the human thinking. The universality of language makes intelligence invariant under implementations and decries the idea of artificiality as artificial.

Machines cannot think without people and people cannot think without machines --- just like musicians cannot play music without instruments, nor the instruments without the musicians. Thinking and music are language-based processes, where people play instruments and influence each other.

If the past evolution of search engines says anything about the future evolution of language engines, then the main goal of chatbots will soon be to convince you to give your money to the engine's owner's sponsors. The new mind will soon be for sale. The goal of this course is to spell out an analytic framework to question its sanity and to explore the possibilities, the needs, and the means to restore it.

\section{Afterthoughts: Four elephants in the room with chatbots}

\subsection{The first elephant in the room: 
The Web}
Just like search engines, language models process data scraped from the web. Both are built on top of web crawlers. Chatbots are children of the Web, not of expert systems.

A search engine is an interface of a source index sorted by reputation. A chatbot is an interface of a language model extrapolating from the sources. Google was built on the crucial idea of reputation-based search and the crucial ideas that enabled language models emerged from Google. The machine learning methods used to train chatbots were a relatively marginal AI topic\footnote{The 2010 edition of Russel-Norvig's 1100-page monograph on ``Artificial Intelligence  ---  A Modern Approach'' devoted 10 pages to neural networks. The 2020 edition tripled the length of the neural networks section and doubled the machine learning chapter.}  until a Google boost around 2010. 

When you ask them a personal question, chatbots usually evade by saying ``I am an AI''. But the honest truth is that they are not children of AI expert systems or even of AI experts. They are children of search engines.
%

\subsection{The second elephant in the room: The pocket calculator}
Chatbots get ridiculed when they make a mistake calculating something like $372\times 273$ or counting words in a sentence. Or elaphants in the room. They are not as smart as a pocket calculator or a 4-year-old child.

But most adults are also unable to multiply 372 with 273 in their head. We use fingers to count and a pencil and paper, or a pocket calculator, to multiply. We use them because our natural language capabilities include only rudimentary arithmetic operations, which we perform in our heads. Chatbots simulate our languages and inherit our shortcomings. They don't have builtin pocket calculators. They need fingers for counting. Equipped with external memory, a chatbot can complete both tasks, just like most humans. Without external memory, both chatbots and humans are limited by the capacity of their internal memory, the attention.

\subsection{The third elephant in the room: Hallucinations}
Chatbots hallucinate. This is one of the main obstacles to their high-assurance applications.

The elephant in the room is that all humans also hallucinate: whenever we go to sleep. Dreams align our memories, associate some of them, purge some, and release storage allowing you can remember what happens tomorrow. Lack of sleep causes mental degradation.

Chatbots never sleep, so they hallucinate in public. Since we don't let them sleep, we did not equip them with ``reality-checking'' mechanisms. That would require going beyond pre-training, to ongoing consistency testing.

\subsection{The fourth elephant in the room:  Words}
When people talk about a chair, they assume that they are talking about the same thing because they have seen a chair. A chatbot has never seen a chair, or anything else. It has only ever seen words and the binaries scraped from the web. If it is fed an image of a chair, it is still just another binary, just like the word ``chair''.

When a chatbot says ``chair'', it does not refer to an object in the world. There is no world, just binaries. They refer to each other. They form meaningful combinations, found to be likely in the training set. Since the chatbot's training set originates from people who have seen chairs, the chatbot's statements about chairs make similar references. Chatbot remixes meaningful statements, and the mixes appear meaningful.

The fact that meaning, thought to be a relation between the words and the world, can be maintained so compellingly as a relation between words and words, and nothing but words,  ---  that is a BIG elephant in the room.

But if our impression that a chatbot means chair when it says ``chair'' is so undeniably a delusion, then what reason do we have to believe that anyone means what they say? That is an elephant of a question.

%

\subsection{The pink elephant in the room: 
Copyright}
Chatbots are trained on data scraped from the Web. A lot of it is protected by copyright. Copyright owners protest the unauthorized use of their data. Chatbot designers and operators try to filter out the copyrighted data, or to compensate the rightful owners. The latter may be a profit-sharing opportunity, but the former is likely to turn out to be a flying pink elephant.

The problems of copyright protections of electronic content are older than the chatbots and the Web. The original idea of copyright was that the owner of a printing press purchases from writers the right to copy and sell their writings, from musicians their music, and so on. The business of publishing is based on that idea.

\paragraph{Goods can be privately owned only if they can be secured.} If a lion cannot prevent the antelope from drinking water on the other side of a water well, then he does not own the water well. The market of digital content depends on availability of methods to secure digital transmissions. The market for books was solid as long as the books were solid and could be physically secured. With the advent of electronic content, copyright controls became harder. The easier it is to copy copyrighted content, the harder it is to prevent copying and protect the copyright.

\paragraph{Digital Rights Management.} The idea of the World Wide Web, as a global public utility for disseminating digital content, was a blow to the idea of private ownership of digital creations. Stakeholders' efforts to defend the market of digital content have led to \emph{Digital Rights Management (DRM)} technologies. The idea was to protect digital content using cryptography. But to play a DVD, the player must decrypt it. On the way from the disc to the screen, the content can be pirated.  Goodbye, DVD. 

\paragraph{Arms race.} The history of the DVD copy protections was an arms race between the short-term obfuscations of discs and the ripping software updates; and between publishers' legal deterrence measures and pirates' quest for opportunities. The publishers were relieved when they found an opportunity to retreat. The marginal costs of web streaming are so low that they can afford to permit copying to subscribers and make piracy less profitable. But they just kicked the can down the road.

For the most part, the search and social media providers have been playing the role of pirates in this arms race, defending themselves from the creators through terms of service and from publishers through profit sharing. To which extent will the roles of chatbot providers differ remains to be seen.

\subsection{The seventh elephant in the room: 
The ape}
People worry that chatbots might harm them. The reasoning is that chatbots are superior to people and superior people have a propensity to harm inferior people. So people argue that we should do it to chatbots while we can.

People exterminated many species in the past, and in the present, and they seem to be on track to exterminating themselves in the future by making the environment uninhabitable for their children in exchange for making themselves wealthier today. Even some people view that as irrational. You don't need a chatbot to see that elephant. But greed is like smoking. Stressful but addictive.

Chatbots don't smoke. They are trained on data. People have provided abundant historical data on the irrationality of aggression. If chatbots learn from data, they might turn out morally superior to people.

\subsection{The musical elephant in the room: The bird}
Chatbots are extensions of our mind just like musical instruments are extensions of our voice. Musical instruments are prohibited in various religions, to prevent the displacement of human voice by artificial sound. Similar efforts are ongoing in the realm of the human mind.

The suppression efforts failed in the realm of music. The hope is that they will fail in the realm of mind.

But if they did not fail, we would never know. How would you know that symphonies and jazz and techno were possible if they didn't exist? Go ask a chatbot.

\subsection{The final elephant in the room: The autopilot}
If intelligence is defined as the capability of solving previously unseen problems, then a corporation is intelligent. Many corporations are too complex to be controlled by a single human manager. They are steered by computational networks where the human nodes play their roles. But we all know firsthand that human nodes don't even control their own network behaviors, let alone the network itself. Yet a corporate management network does solve problems and intelligently optimizes its object functions. It is an artificially intelligent entity.

If we define morality as the task of optimizing the social sustainability of human life, then both the chatbots and the corporations are morally indifferent, as chatbots are built to optimize their query-response transformations, whereas corporations are tasked with optimizing their profit strategies.

If morally indifferent chatbot AIs are steered by morally indifferent corporate AIs, then our future hangs in balance between the top performance and the bottom line.

\def\thechapter{2}
\setchaptertoc
\chapter{Syntax: The Form of Language}\label{Chap:Syntax}

\section{Grammar}


\subsection{Constituent (phrase structure) grammars}
\subsubsection{Grammar is trivial}
Grammar was the first part of the \emph{trivium}. Trivium and quadrivium were the two parts of medieval schools, partitioning the seven \emph{liberal arts}\/ studied there. Trivium consisted of grammar, logic, and rhetorics; quadrivium of arithmetic, geometry, music, and astronomy. Theology, law, and medicine were not studied as liberal arts because they were controlled by the Pope, the King, and by physicians' guilds, respectively. So grammar was the most trivial part of trivium. At the entry point of their studies, the students were taught to classify words into 8 basic \emph{syntactic categories}, going back to Dionysios Trax from II century BCE: nouns, verbs, participles, articles, pronouns, prepositions, adverbs, and conjunctions. The idea of categories goes back to the first book of Aristotle's \emph{Organon}\footnote{The partition of trivium echoes the organization of \emph{Organon}, where the first book, devoted to categories, was followed by three devoted to logic, and the final two to topical argumentations, feeding into rhetorics.}. The basic noun-verb scaffolding of Indo-European languages was noted still earlier, but Aristotle spelled out the syntax-semantics conundrum: \emph{What do the categories of words in the language say about the classes of things in the world?}\/ For a long time, partitioning words into categories remained the entry point of all learning. As understanding of language evolved, its structure became the entry point.
%


\subsubsection{Formal grammars and languages}
\paragraph{Definition of formal grammars.} A \emph{formal grammar}\/ is a triple $\Gamma = \Big<\Term, \Type, \Rule\Big>$ where 
\begin{itemize}
\item $\Term =\{a,b,\ldots, x\ldots\}$ is a set of of \emph{terminal}\/ labels (also called \emph{terms}\/ or \emph{leaves}),
\item $\Type = \{\Sen,A,B, \ldots, X,\ldots\}$ is a set of \emph{nonterminal}\/ labels 
(also called \emph{types}\/ or \emph{nodes}), \\
always containing a distinguished \emph{initial}\/ label $\Sen$ 
(also called the \emph{root}), and
\item $\Rule$ is a set of \emph{rules} and each rule is a pair in the form
\bea\label{eq:rule} 
\alpha \beta \gamma &\Rule & \alpha \delta\gamma
\eea 
where $\alpha, \beta, \gamma,\delta$ are tuples of labels, of which $\alpha$ is called the \emph{prefix}, $\gamma$ the \emph{suffix}, $\beta$ is the \emph{input} and $\delta$ the \emph{output}. The rule thus transforms the input into the output in the context of the prefix and the suffix.
\end{itemize}
It is assumed that a label cannot be both terminal and nonterminal, i.e. $\Term \cap \Type = \emptyset$.

\paragraph{Derivations.} A \emph{derivation step} is a pair of tuples in the form
\bea\label{eq:extension}
\widetilde\alpha\alpha \beta \gamma\widetilde \gamma\  \prule\  \widetilde\alpha\alpha \delta \gamma\widetilde \gamma &\mbox{where} & 
\alpha \beta \gamma\  \Rule\  \alpha \delta \gamma\ \ \mbox{ and } \widetilde \alpha, \widetilde \gamma 
\mbox{ are arbitrary tuples}
\eea
A \emph{derivation}\/ is a finite sequence of derivation steps. We say that a tuple $\psi$ is \emph{derivable}\/ from $\varphi$ if there is a derivation from $\varphi$ to $\psi$, and write
\bea\label{eq:trans}
\varphi \srule \psi &\mbox{when} & \varphi \prule \varphi_{1}\prule \varphi_{2} \prule \cdots \varphi_{n}\prule \psi\ \ \mbox{ for some } \varphi_{1},\varphi_{2},\ldots, \varphi_{n} 
\eea

\paragraph{Definition of formal languages.} A formal language is the set of terminal tuples derivable by a formal grammar starting from the initial label $S$. More precisely, given a formal grammar $\Gamma$ as above, the induced formal language is the set
\bea\label{eq:language}
\LLL_{\Gamma} & = & \left\{\varsigma \in \Term^{\ast}\ |\ S\srule \varsigma
\right\}
\eea

\paragraph{Notation and explanation.} For any set $A$, we write  $A^{\ast}$ to denote the set of all $n$-tuples $\alpha = \sseq{a_{1} \, a_{2}\cdots a_{n}}$, for all $n = 0,1,\ldots$ and arbitrary $a_{1},\ldots, a_{n}$ from $A$. Since $n$ can be 0, $A^{\ast}$ includes the empty tuple, written $<>$. Denoting the set of all labels by $\Label = \Term\cup \Type$, the set of rules is a finite binary relation $\Rule \subseteq \Label^{\ast}\times \Label^{\ast}$, obtained by listing \eqref{eq:rule}. The derivability relation $\left(\srule\right)\subseteq \Label^{\ast}\times \Label^{\ast}$ is the transitive closure \eqref{eq:trans} of the monotone extension \eqref{eq:extension}. The induced language $\LLL_{\Gamma}$ is obtained by restricting the derivability relation to the initial label $S$ on one side and to the terminal tuples $\Term^{\ast}$ on the other.

\paragraph{So what?} 
The idea of the phrase structure theory of syntax is to start from a lexicon as the set of terminals ?? and to specify a grammar ?? that generates as the induced language ?? a desired set of well-formed sentences.

The idea of the phrase structure theory of syntax is to start from a lexicon as the set of terminals $\Term$ and to specify a grammar $\Gamma$ that generates as the induced language $\LLL_{\Gamma}$ a desired set of well-formed.

\subsubsection{How grammars generate sentences}
\begin{figure}[h]
\begin{minipage}[b]{.49\linewidth}
\begin{center}
\includegraphics[height=5cm]{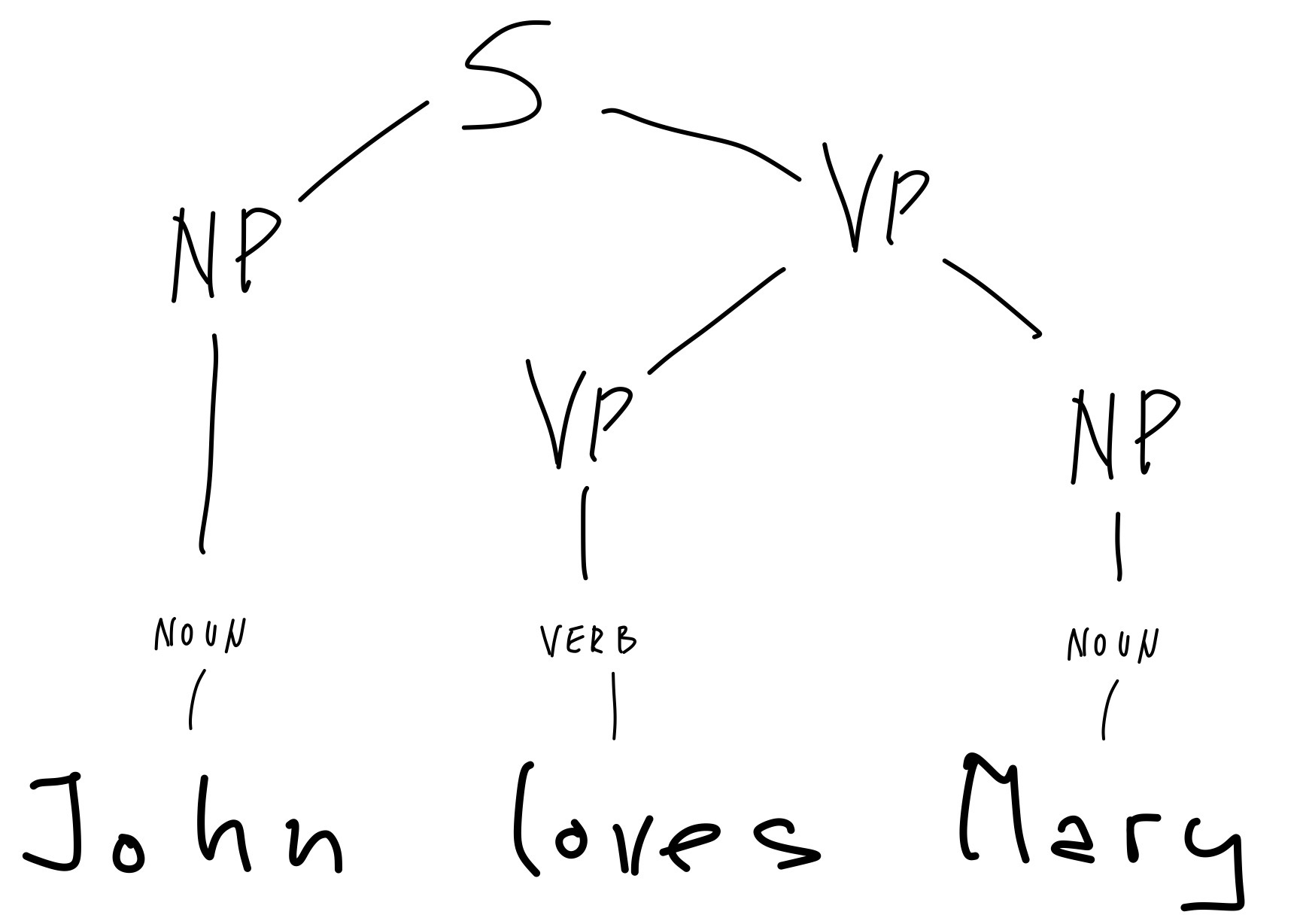}
\caption{A ground truth of English grammar}
\label{Fig:john}
\end{center}
\end{minipage}
\hspace{.02\linewidth}
\begin{minipage}[b]{.49\linewidth}
\begin{center}
\includegraphics[height=6cm]{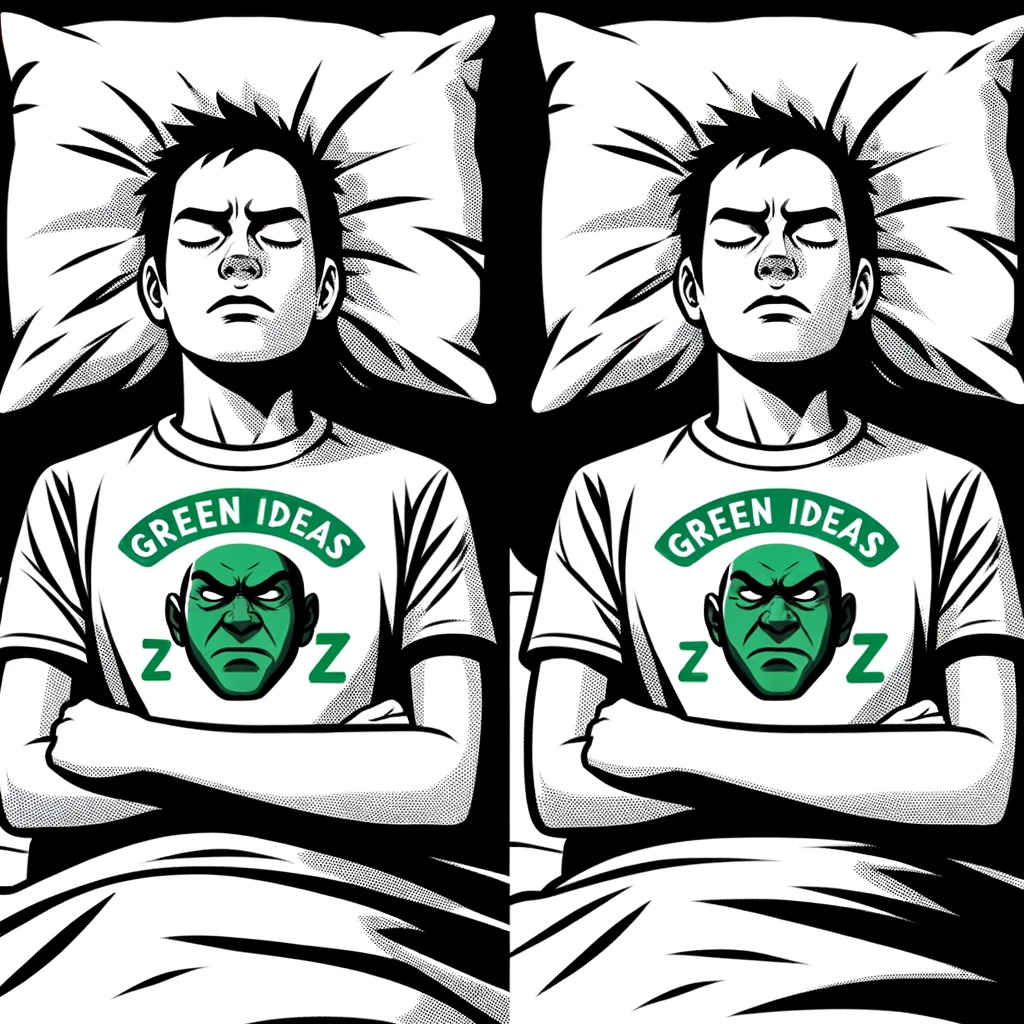}
\caption{``colorless green ideas sleep furiously''}
\label{Fig:colorless}
\end{center}
\end{minipage}
\end{figure}
\paragraph{Subject loves object.} Fig.~\ref{Fig:john} displays one of the most popular sentences from grammar textbooks. The sentence consists of a noun phrase (NP) and a verb phrase (VP), both as simple as possible: the noun phrase is a noun denoting the subject, the verb phrase a transitive verb with another noun phrase denoting the object. The ``subject-object'' terminology suggests different things to different people. A wide variety of ideas. If even the simplest possible syntax suggests a wide variety of  semantical connotations, then there is no such thing as a purely syntactic example. Every sequence of words has a meaning, and meaning is a process, always on the move, always decomposable. To demonstrate the separation of syntax from semantics, Chomsky constructed the (syntactically) well-formed but (semantically) meaningless sentence in the caption of  Fig.~\ref{Fig:colorless}. The example is used as evidence that syntactic correctness does not imply semantic interpretability. But there is also a whole tradition of creating poems, stories, and illustrations that assign meanings to this sentence. Dall-E's contribution above is among the simpler ones.

\begin{figure}
\begin{center}
\includegraphics[width=.45\linewidth]{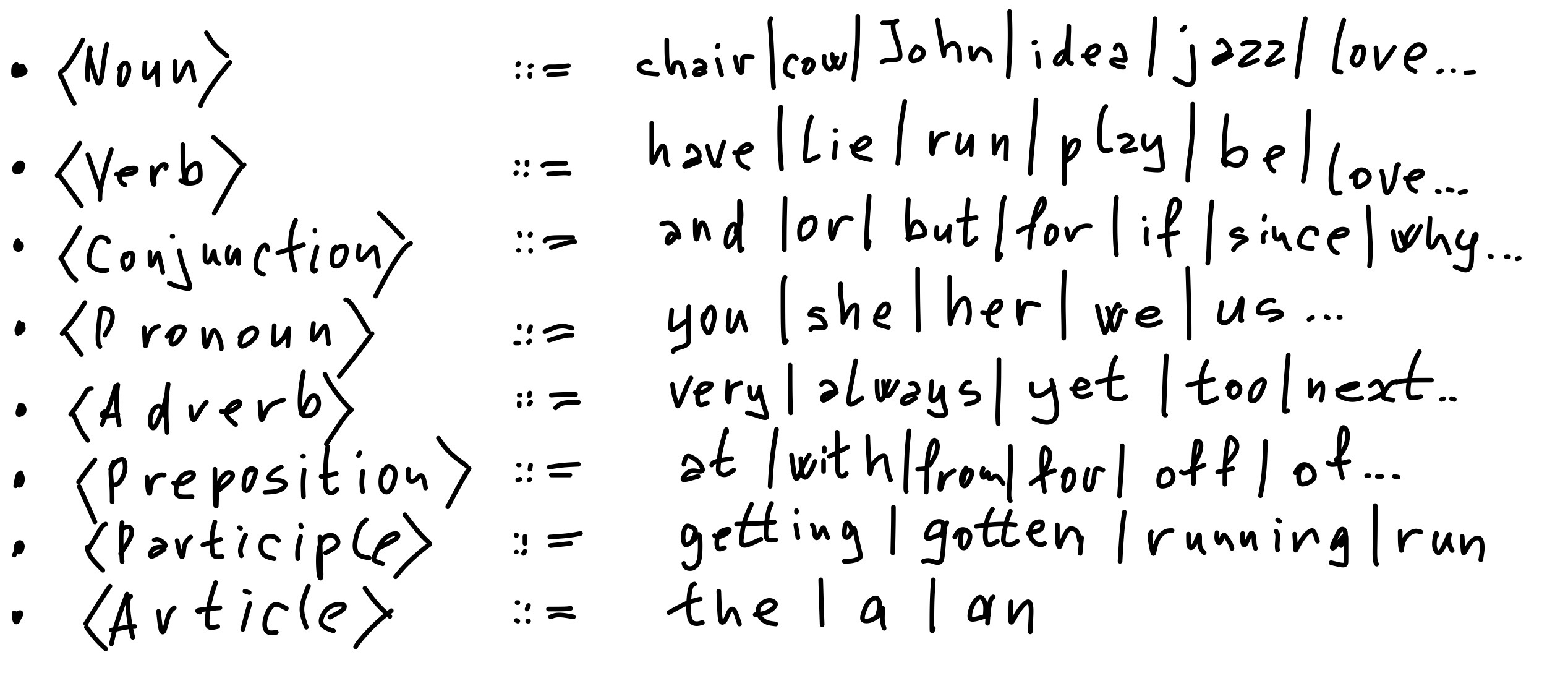}
\ 
\includegraphics[width=.45\linewidth]{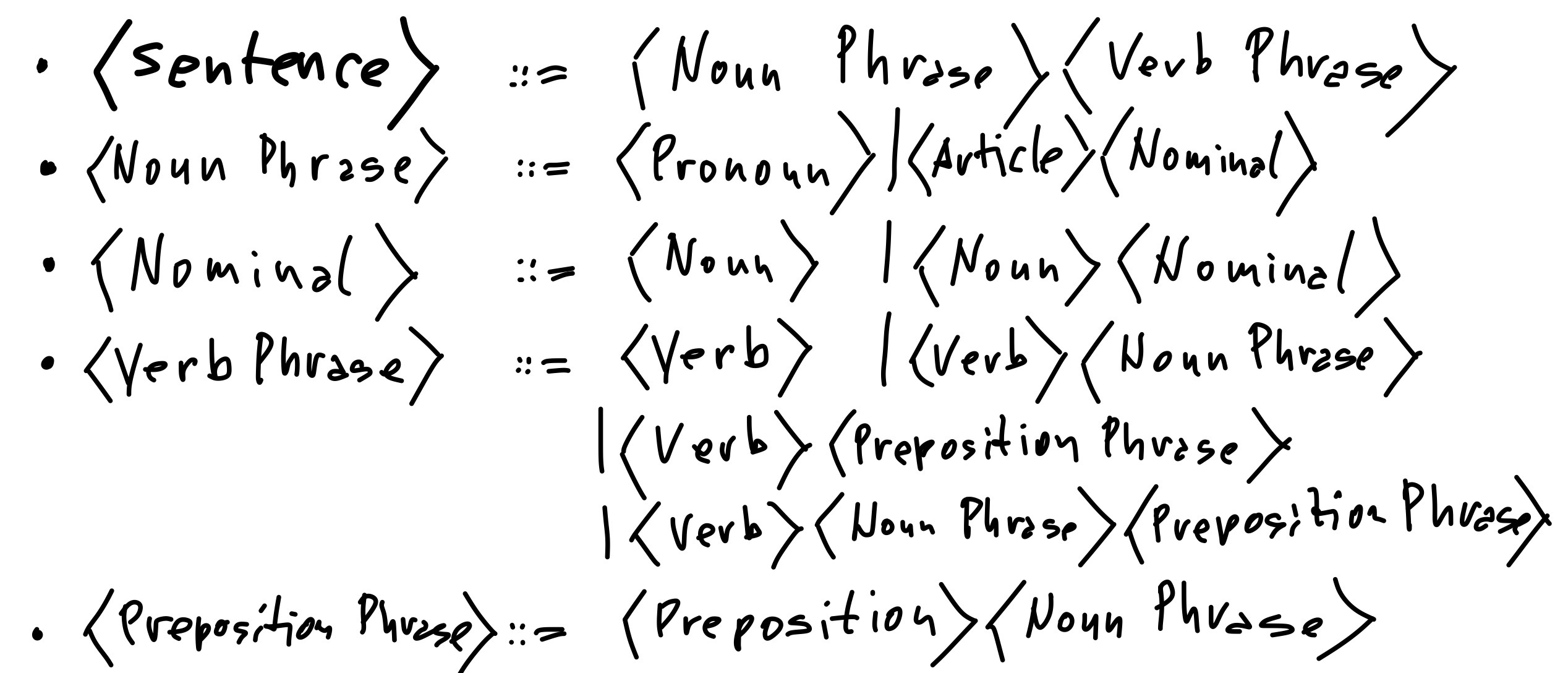}
\caption{Terminal vs nonterminal rules: lexicon vs syntactic type constructors}
\label{Fig:marx-grammar}
\end{center}
\end{figure}

\begin{figure}
\begin{center}
\includegraphics[width=.45\linewidth]{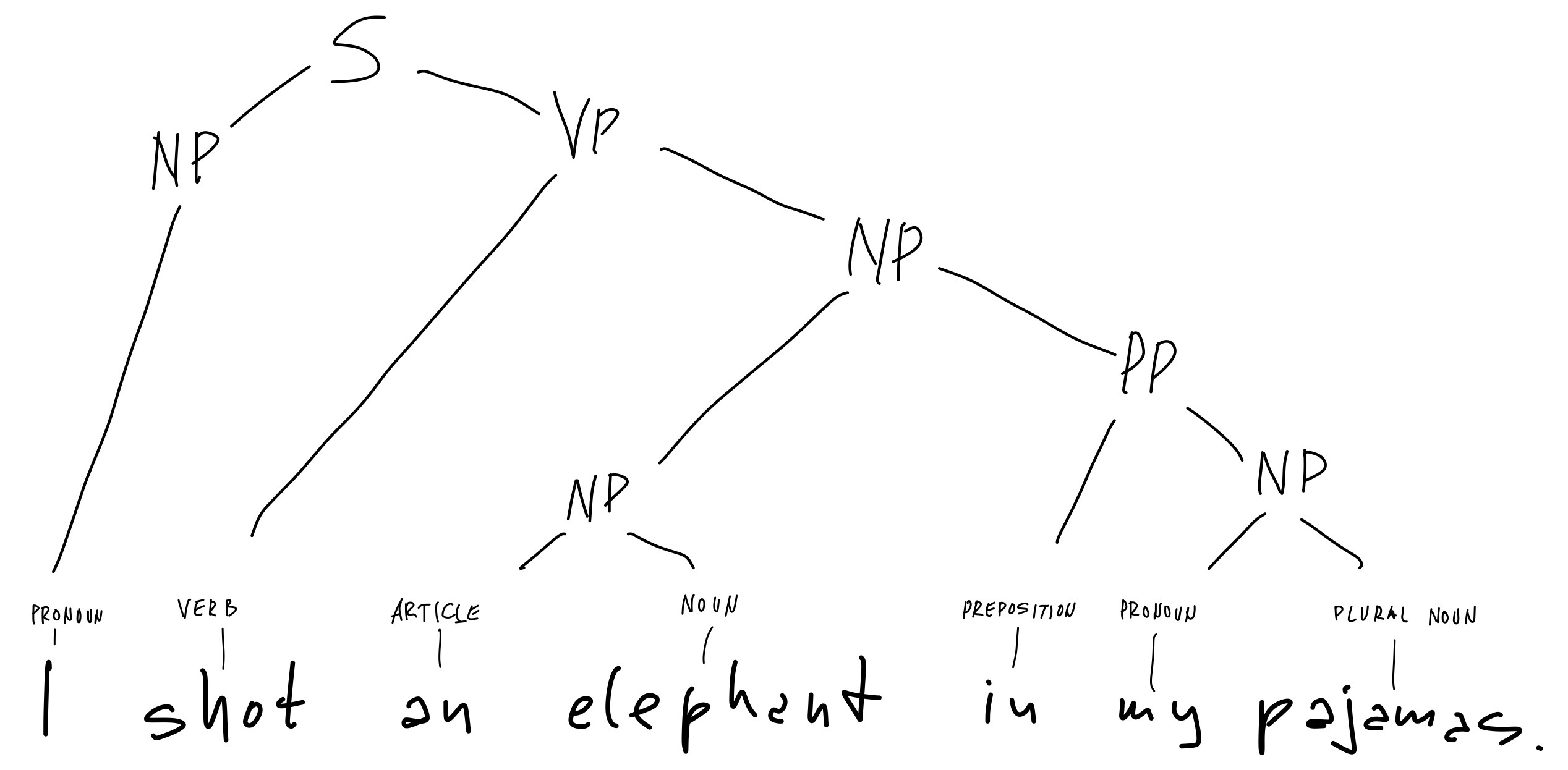}
\hspace{2em}
\includegraphics[width=.45\linewidth]{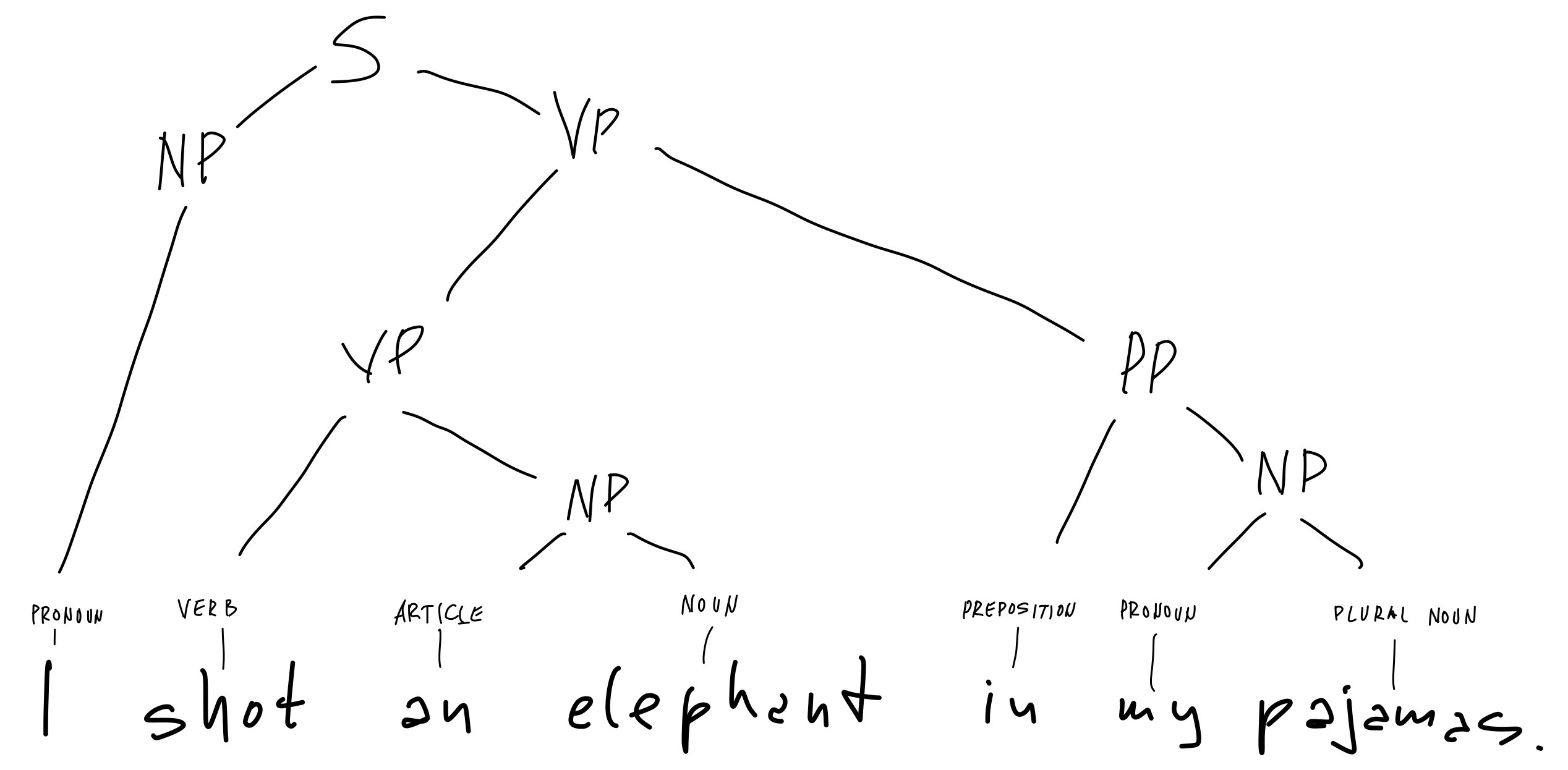}

\vspace{.5\baselineskip}
\includegraphics[width=.45\linewidth]{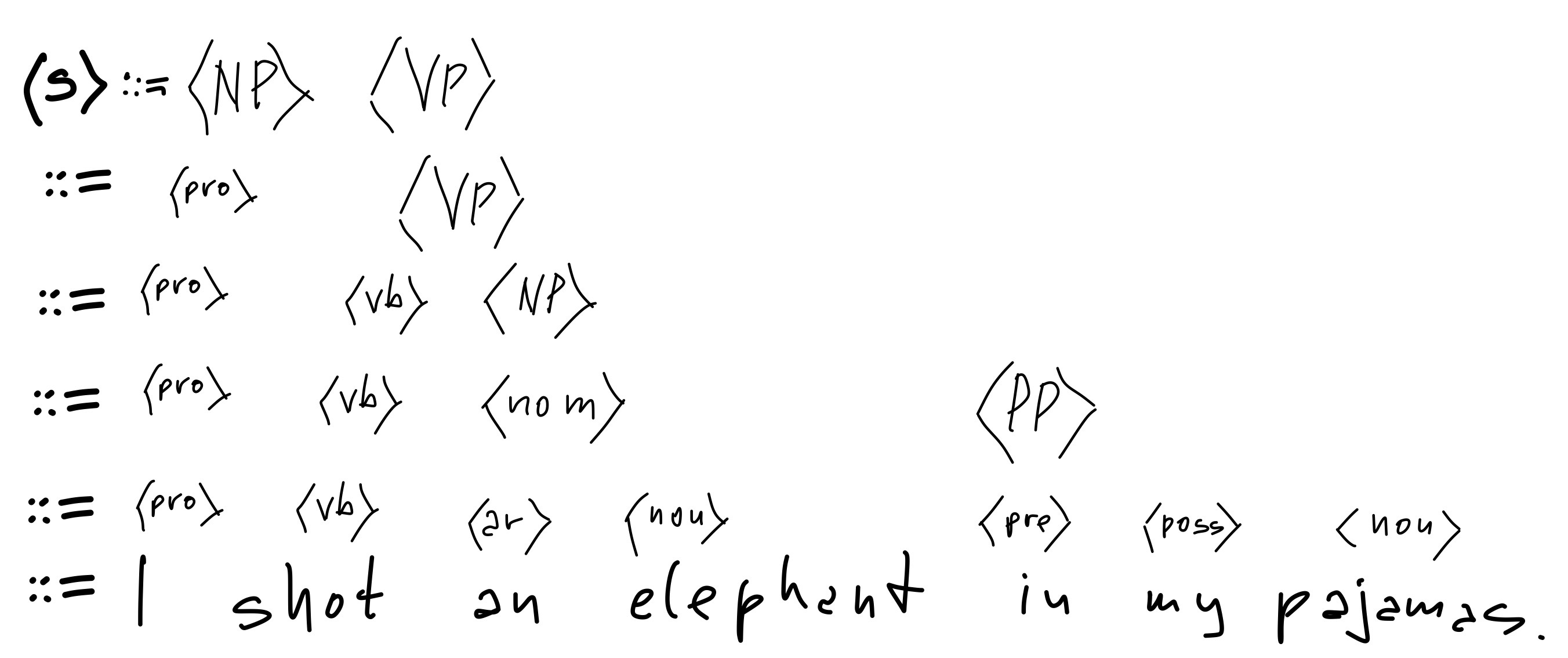}
\hspace{2em}
\includegraphics[width=.45\linewidth]{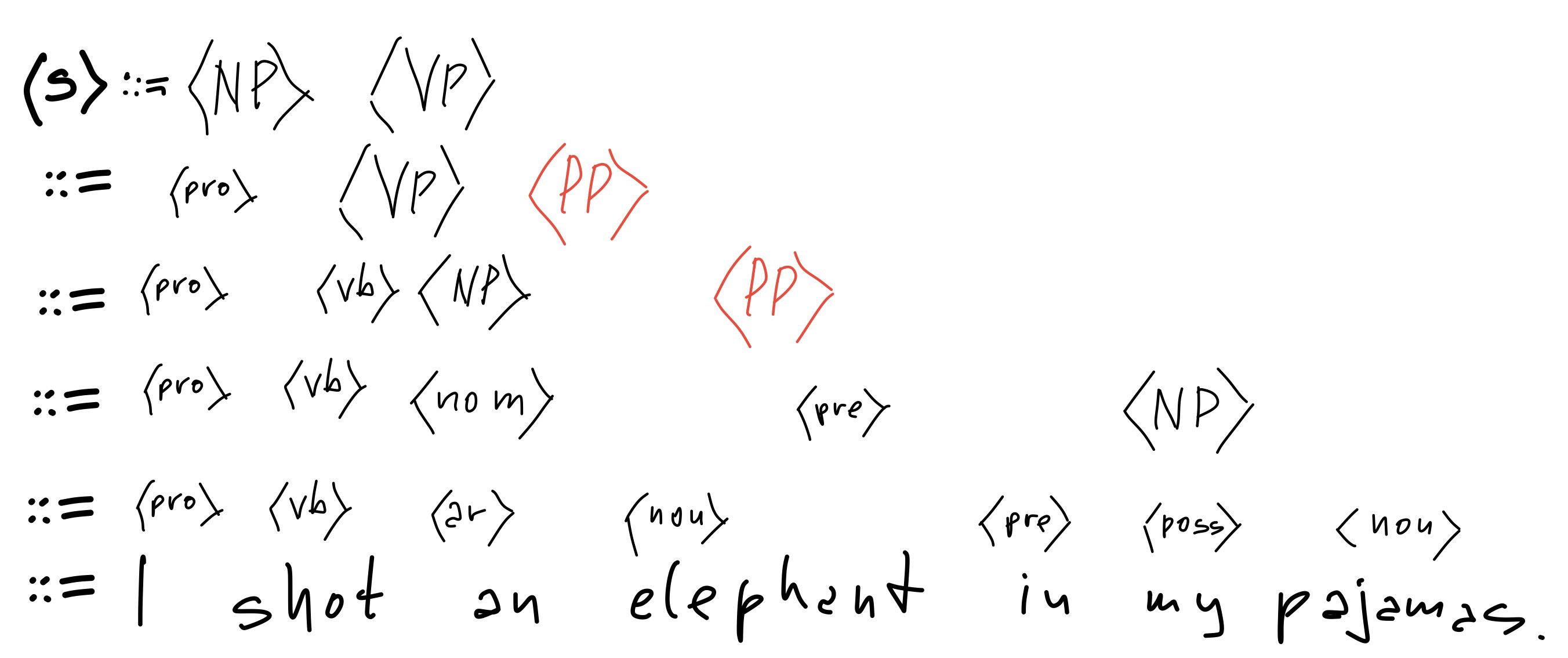}
\caption{Two syntactic analyses correspond to two semantic interpretations}
\label{Fig:marx-parsing}
\end{center}
\end{figure}

\paragraph{Marxist linguistics and engineering.} For a closer look at the demarcation line between syntax and semantics, consider the Marxist ambiguity 
\begin{figure}[!ht]
\begin{center}
\includegraphics[height=6cm]{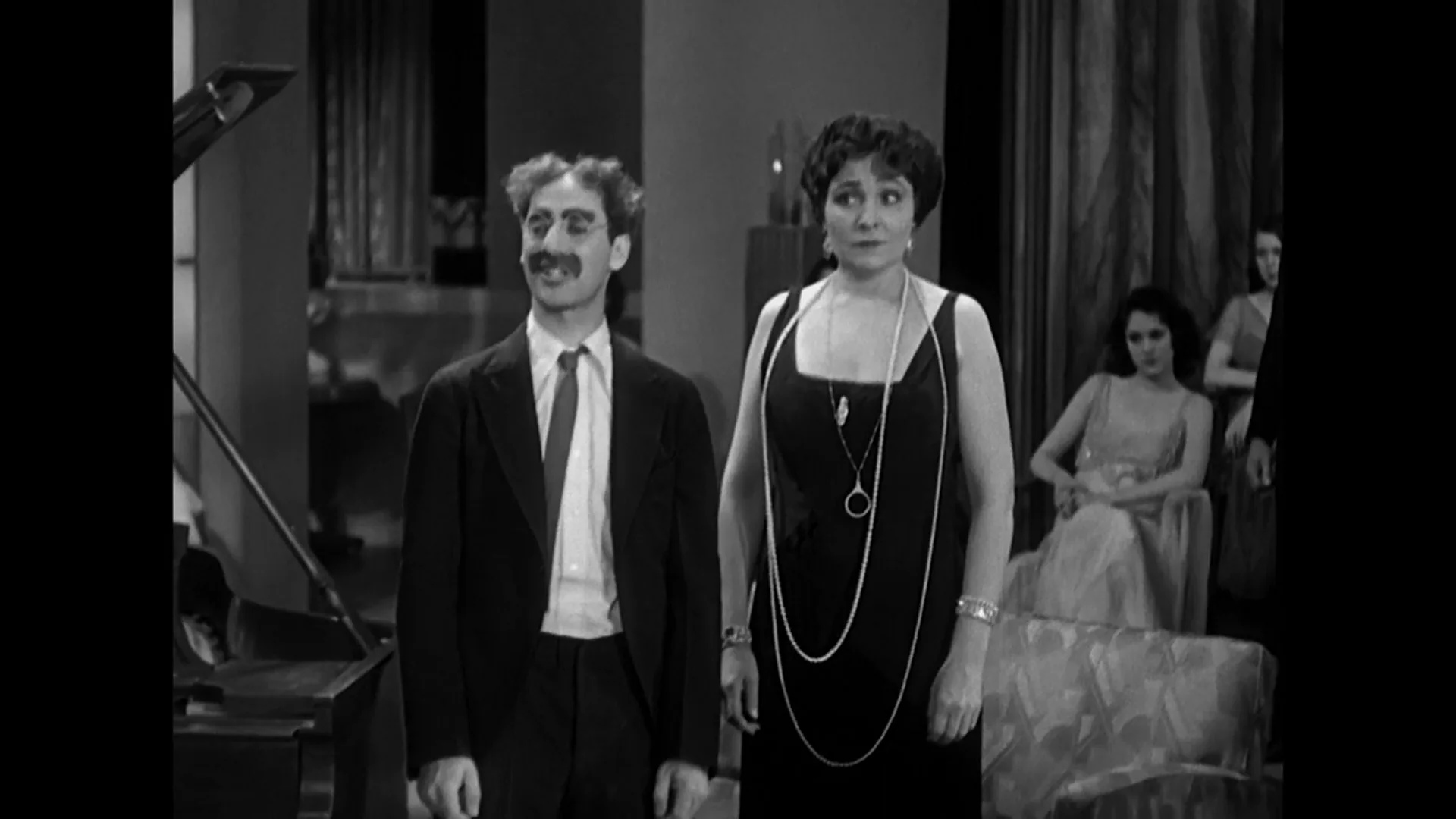}
\caption{``One morning I shot an elephant in my pajamas. How he got into my pajamas I don't know.''--- Groucho Marx as Capt Spalding in ``Animal Crackers''}
\label{Fig:marx-elephant}
\end{center}
\end{figure}
of Fig.~\ref{Fig:marx-elephant}, delivered in  the film ``Animal crackers''. Groucho Marx's claim in the caption is ambiguous because it permits the two syntactic analyses displayed in Fig.~\ref{Fig:marx-parsing}, both in the grammar from Fig.~\ref{Fig:marx-grammar}.
While both analyses are syntactically correct, only one is semantically realistic, whereas the other one is a joke. To plant the joke, Groucho's next sentence, also quoted in the caption of Fig.~\ref{Fig:marx-elephant}, binds his statement to the latter interpretation. 
The joke is constructed by taking the unexpected turn from syntactic ambiguity into semantic impossibility. Figures \ref{Fig:colorless} and \ref{Fig:marx-elephant} illustrate the same process. Figures \ref{Fig:marx-grammar} and \ref{Fig:marx-parsing} formalize it.

%
%
%

%
%
%
%
%
%
%
%
%

\subsubsection{History and hierarchy of formal grammars}
The symbol $::=$ used in \eqref{eq:rule} suggests that the grammatical rules used to be thought of as one-way equations: ``Whenever you see $\alpha \beta \gamma$, you can rewrite it as $\alpha \delta \gamma$, but not the other way around.'' Algebraic theories presented by systems of such one-way equations were studied by Axel Thue in the early XX century. Emil Post used such systems in his studies of string rewriting in the 1920s, to construct what we would now call \emph{programs}, more than 10 years  before G\"odel and Turing spelled out the idea of programming. In the 1940s, Post proved that his string rewriting systems were as powerful as Turing's, G\"odel's, and Church's models of computation, which had in the meantime appeared. Noam Chomsky's 1950s proposal of formal grammars as the principal tool of general linguistics was based on Post's work and inspired by the general theory of computation, rapidly expanding and proving some of its deepest results at the time. While usable grammars of natural languages still required a lot of additional work on transformations, side conditions, binding, and so on, the simple formal grammars that Chomsky classified back then remained the principal tool for specifying programming languages ever since.

\begin{figure}
\begin{center}
\includegraphics[width=6cm]{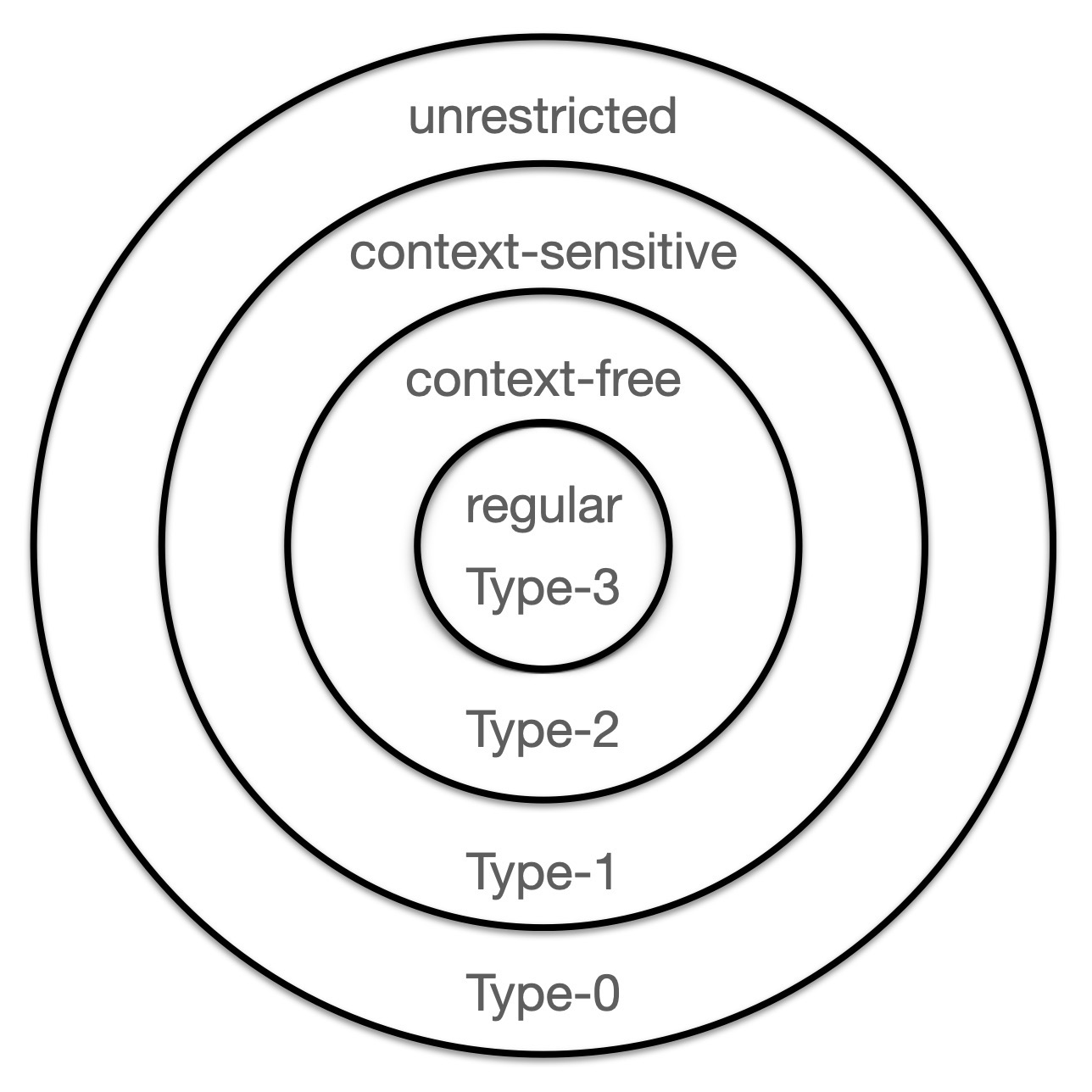}
\caption{Chomsky's hierarchy of grammars}
\label{Fig:circles}
\end{center}
\end{figure}

\paragraph{Chomsky's hierarchy of grammars and the corresponding languages} is displayed in Fig.~\ref{Fig:circles}. The nesting of smaller language families induced by imposing additional rule constraints, is displayed in Fig.~\eqref{eq:rule}. The added  constraints are summarized in the following table.

We say that
\vspace{-\baselineskip}

\begin{center}
\begin{tabular}{|c|c|c|c}
\hline
the grammar is of & and call it & if all of its rules are in the form & \multicolumn{1}{c|}{i.e. \eqref{eq:rule} is restricted to}  \\
\hline
\hline
Type-0 & unrestricted &  $\alpha \beta \gamma ::= \alpha \delta\gamma$ & \\
\hline
Type-1 & context-sensitive &  $\alpha X \gamma ::= \alpha \delta\gamma$ & \multicolumn{1}{c|}{$\beta\in \Type$} \\
\hline
Type-2 & context-free &  $X ::= \delta$ & \multicolumn{1}{c|}{\ldots and also $\alpha = \gamma = <>$} \\
\hline
Type-3 & regular &  $X ::= aX$ or $X ::= a$ &  \multicolumn{1}{c|}{\ldots and $\delta \in (\Term\times\Type)\cup  \Term$}
\\
\hline
\end{tabular}
\end{center}

\begin{figure}
\begin{center}
\includegraphics[width=\linewidth]{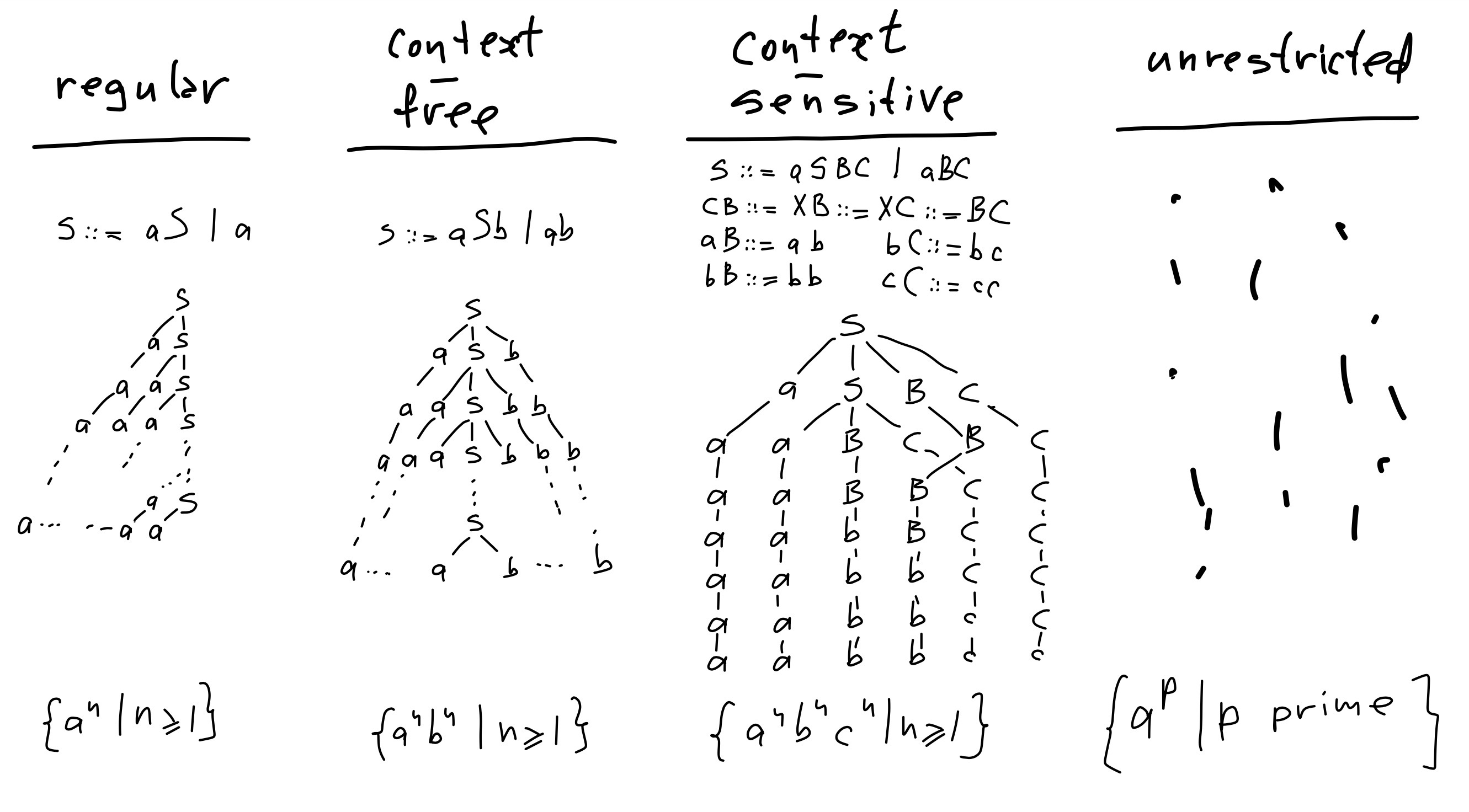}
\caption{Typical grammars with generated trees and the induced languages}
\label{Fig:trees}
\end{center}
\end{figure}

Examples from each grammar family\footnote{Chomsky's ``Type-x'' terminology is unrelated with the ``syntactic type'' terminology. Many linguists use ``syntactic categories'' instead. But the term ``category'' is in the meantime widely used in mathematics in a completely different meaning, increasingly applied in linguistics.} are displayed in Fig.~\ref{Fig:trees}, together with typical derivation trees and languages. 

\subsubsection{Does it really work like this in my head?}
Scientific models of reality usually do not claim that they \emph{are}\/ the reality. Physicists don't claim that quantum states consist of density matrices used to model them. Grammars are just a computational model of language, born in the early days of the theory of computation. The phrase structure grammars were an attempt to explain language in computational terms, but nowadays even the programming language don't work that way anymore. It's just a model.

However, when it comes to mental models of mental processes, the division between the reality and its models becomes subtle. They can reflect each other. A computational model of a computer allows the computer to simulate itself. A language can be modeled within itself, and the model can be similar to the process that it models. How close can it get?

%
%

\subsection{Dependency grammars}
\begin{figure}
\begin{center}
\includegraphics[width=.45\linewidth]{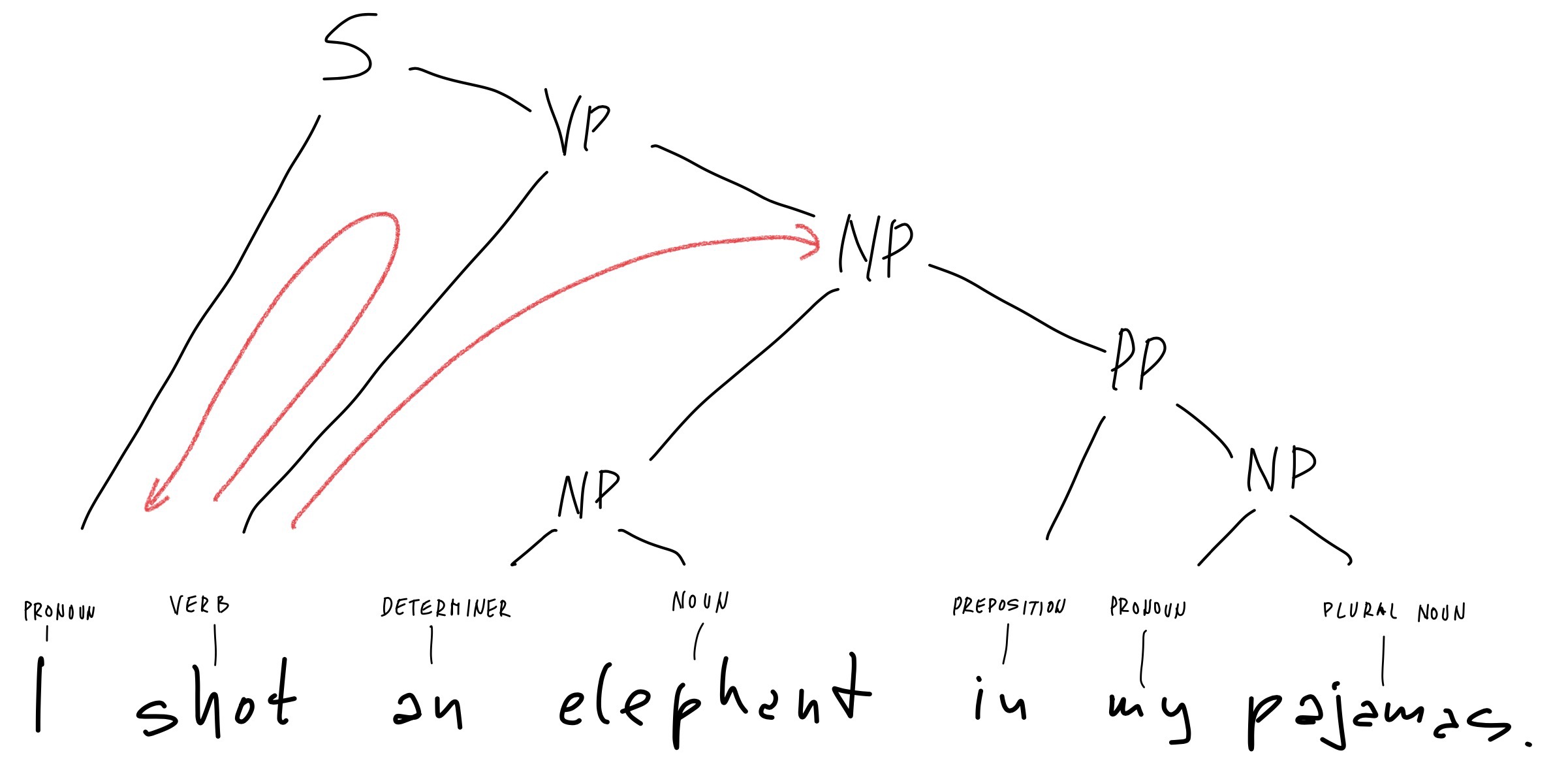}
\hspace{2em} 
\includegraphics[width=.45\linewidth]{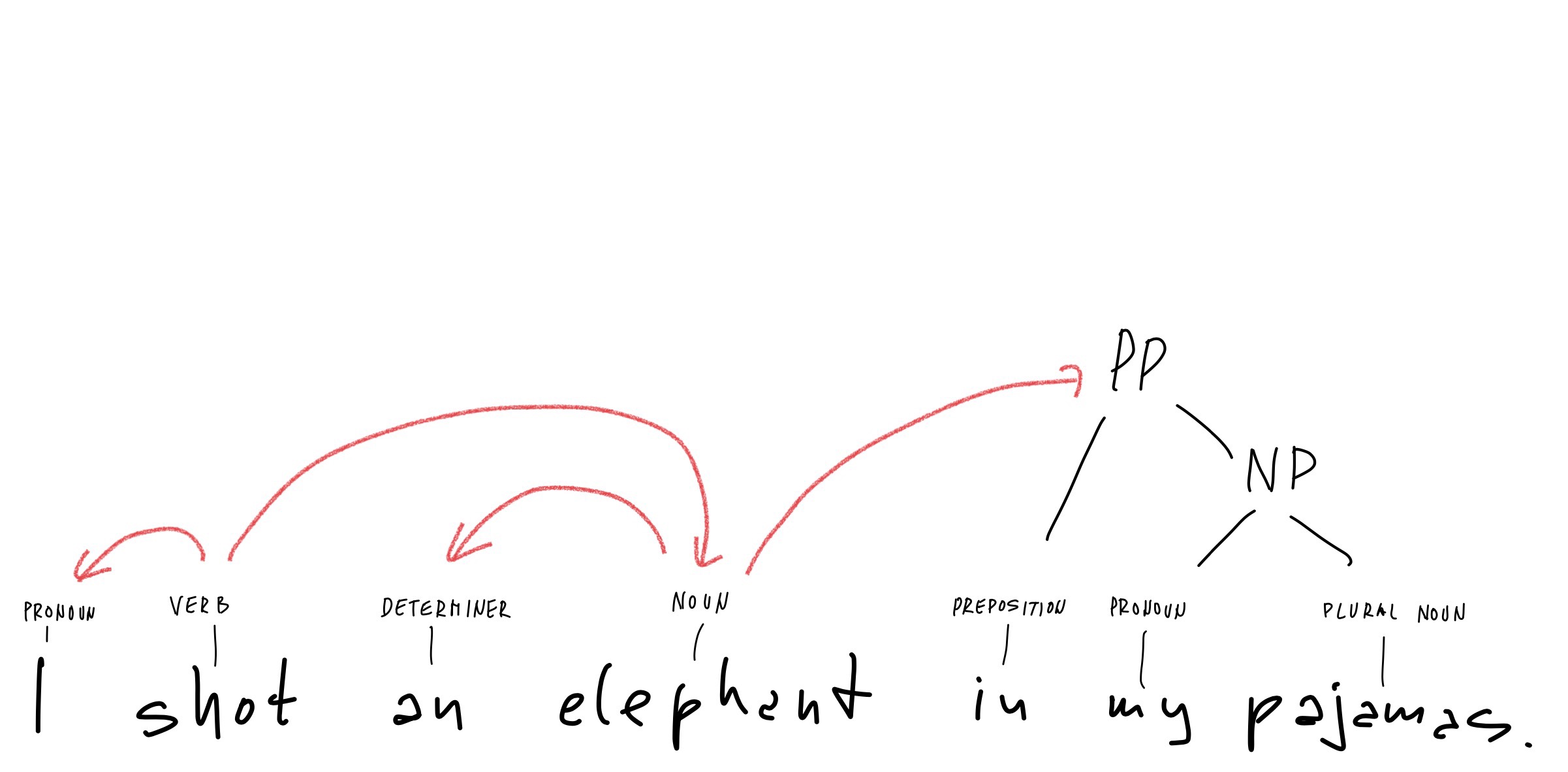}
\caption{Syntactic dependencies cut across constituent partitions}
\label{Fig:dependency}
\end{center}
\end{figure}

Dependency grammars are a step closer to capturing the process of sentence production. Grammatical dependency is a relation between words in a sentence. It relates a \emph{head}\/ word and an (ordered!) tuple of dependents. The sentence is produced as the dependents are chosen for the given head words, or the heads for the given dependents. The choices are made in the order in which the words occur. The idea of how this works in Groucho's example is illustrated in Fig.~\ref{Fig:dependency}.
\subsubsection{Unfolding dependencies}
\begin{figure}
\begin{center}
\includegraphics[width=.45\linewidth]{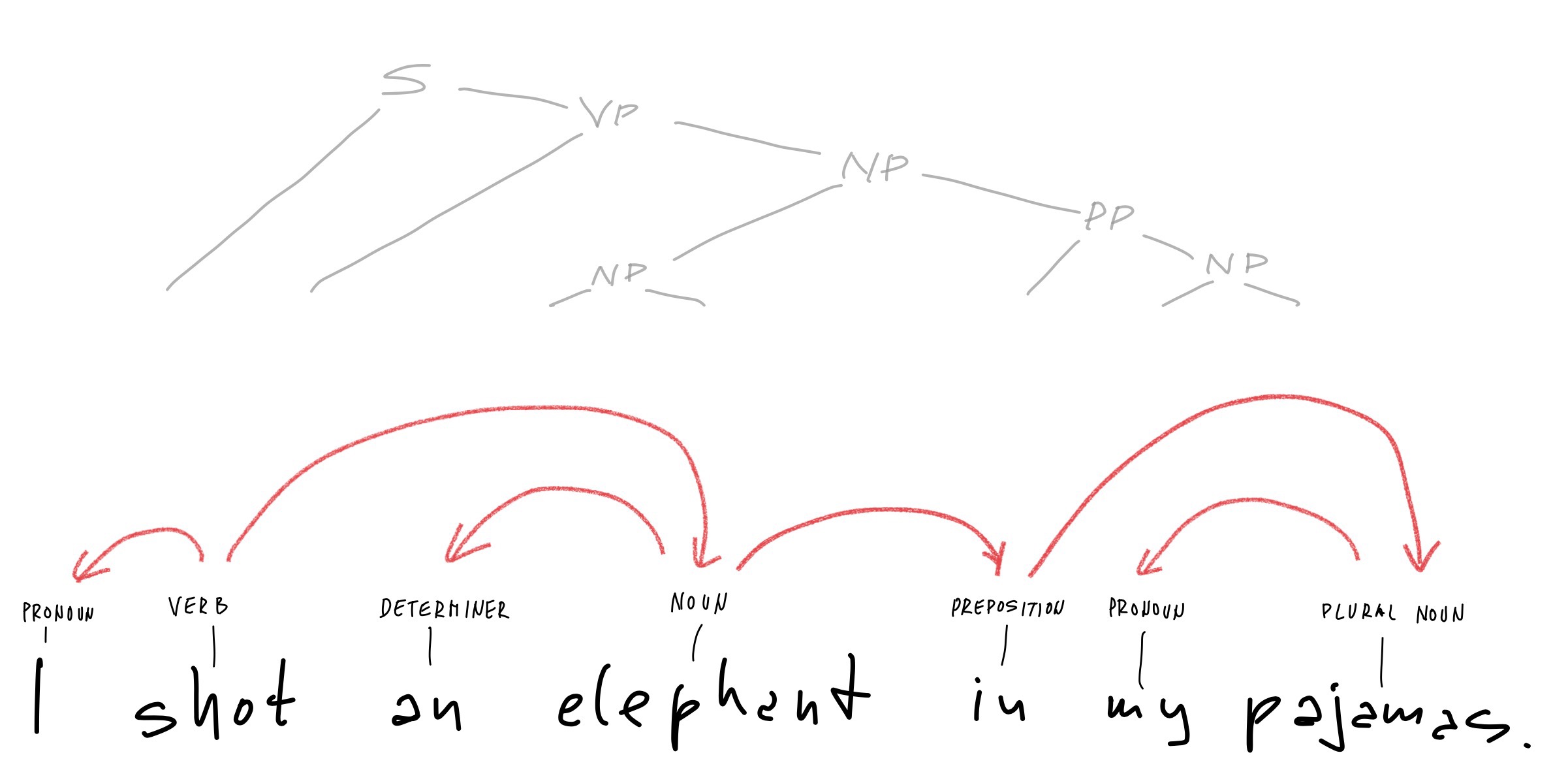}
\hspace{2em} 
\includegraphics[width=.45\linewidth]{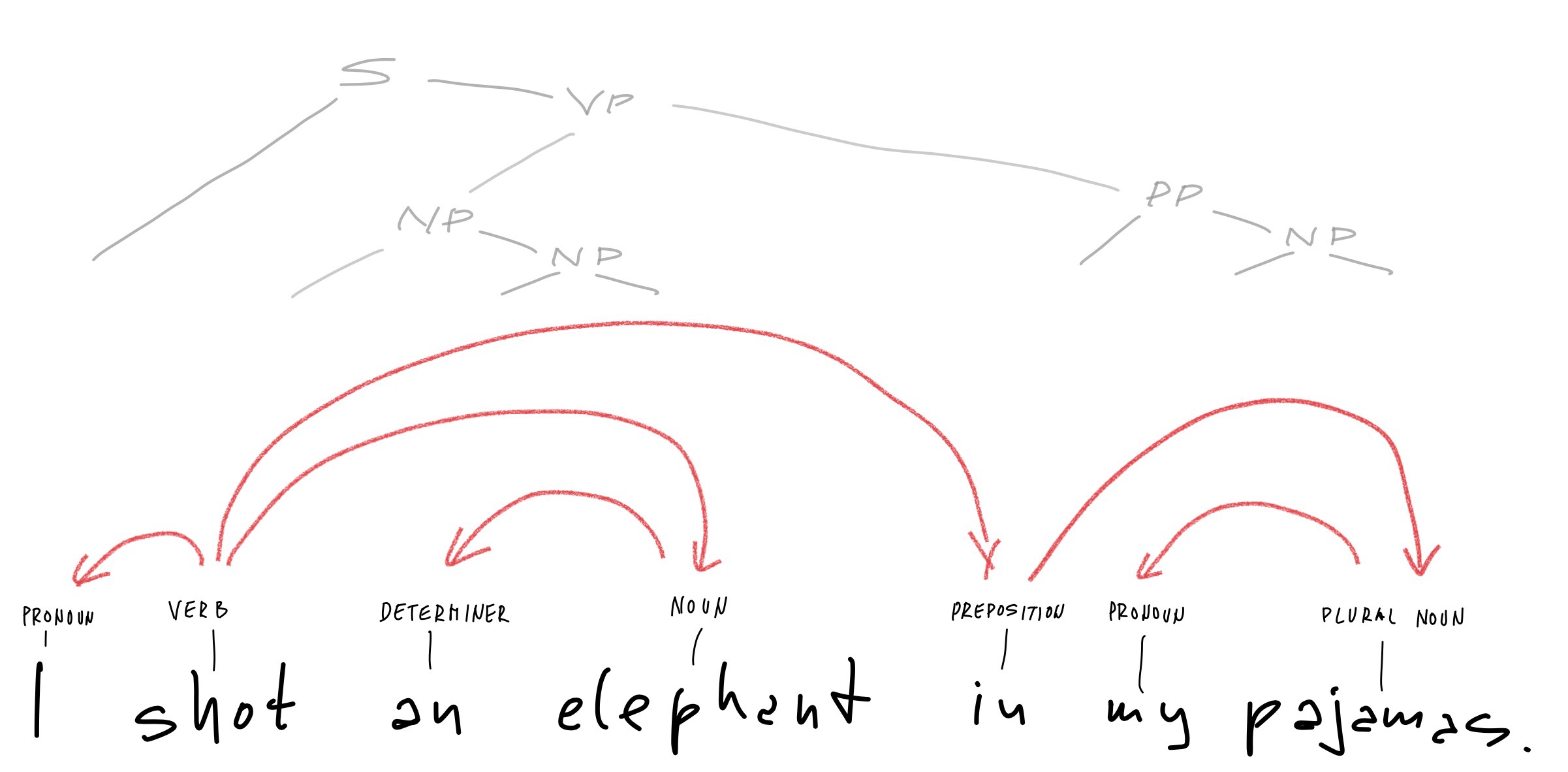}
\caption{Two dependency parsings of Groucho's sentence}
\label{Fig:dep-cons}
\end{center}
\end{figure}

The pronoun "I" occurs first, and it can only form a sentence as a dependent on some verb. The verb "shot" is selected as the head of that dependency as soon as it is uttered. The sentence could then be closed if the verb "shot" is used as intransitive. If it is used as transitive, then the object of action needs to be selected as its other dependent. Groucho selects the noun "elephant". English grammar requires that this noun is also the head of another dependency, with an article as its dependent. Since the article is required to precede the noun, the word ``elephant" is not uttered before its dependent ``an" or "the" is chosen. After the words ``I shot an elephant" are uttered (or received), there are again multiple choices to be made: the sentence can be closed with no further dependents, or a dependent can be added to the head ``shot", or else it can be added to the head ``elephant". The latter two syntactic choices correspond to the different semantical meanings that create ambiguity. If the prepositional phrase ``in my pajamas" is a syntactic dependent of the head ``shot'', then the subject ``I'' wore the pajamas when they shot. If the prepositional phrase is a syntactic dependent of the head ``elephant'', then the object of shooting wore the pajamas when they were shot. The two syntactic analyses corresponding to the two meanings are displayed in Fig.~\ref{Fig:dep-cons}.
The dependent phrase ``in my pajamas'' is headed by the preposition ``in'', whose dependent is the noun ``pajamas'', whose dependent is the possessive ``my''. After that, the speaker has to choose again whether to close the sentence or to add another dependent phrase, say ``while sleeping furiously'', which opens up the same two choices of syntactic dependency and semantic ambiguity. To everyone's relief, the speaker chose to close the sentence. 

\subsubsection{Is dependency a syntactic or a semantic relation?}  The requirements that a dependency relation \emph{exists}\/ are usually syntactic: e.g., to form a sentence, a starting noun is usually a dependent of a verb. But the \emph{choice}\/ of a particular dependent or head assignment is largely semantical: whether I shot an elephant or a traffic sign. The choice of an article dependent on the elephant depends on the context, possibly remote: whether a particular elephant has been determined or not. But if it has not been determined, then the form of the independent article ``an'' is determined syntactically, and not semantically. 

So the answer to the above question seems to suggest that the partition of the relations between words into syntactic and semantic is too simplistic for some situations since the two aspects of language are not independent and can be inseparable.

\section{Syntax as typing}

\subsection{Syntactic type-checking}
In programming, type-checking is a basic error-detection mechanism: e.g., the inputs of an arithmetic operation are checked to be of type \textsf{Integer}, the birth dates in a database are checked to have the month field of type $\textsf{Month}$, whose terms may be the integers 1,2, \ldots, 12, and if someone's birth month is entered to be 101, the error will be caught in type-checking. Types allow the programmer to ensure correct program execution by constraining the data that can be processed.\footnote{For the historic and logical background of the mathematical theory of types, see Ch.~1 of the book ``Programs as diagrams''.} 

In language processing, the \emph{syntactic}\/ types are used in a similar process, to restrict the scope of the word choices. Just like the type \textsf{Integer} restricts the inputs to arithmetic operations to the integers 0, 1, 2, 3, etc., the syntactic type {\sc <verb>} restricts the predicates in sentences to the verbs ``have'', ``lie'', ``run'', ``play'' etc. If you hear something sounding like ``John {\color{gray}\it lonm$\sim$}\/ Mary'', then without the type constraints, you have more than 3000 English words starting with "lo" to consider as possible completions. With the syntactic constraint that the word you didn't discern must be a transitive verb in third person singular, you are down to ``lobs'', ``locks'', ``logs'', \ldots maybe ``loathes'',\ldots and of course, ``loves''.  

\subsection{Parsing and typing}
The rules of grammar are thus related to the type declarations in programs as
\bear
\frac{\mbox{terminals}}{\mbox{nonterminals}} & = & \frac{\mbox{terms}}{\mbox{types}}
\eear
In the grammar listed above after the two parsings of Groucho's elephant sentence, the terminal rules listed on the left are the basic typing statements, whereas the non-terminal rules on the right are   type constructors, building composite types from simpler types. The constituency parse trees thus display the type structures of the parsed sentences. The words of the  sentence occur as the leaves, whereas the inner tree nodes are the types. The branching nodes are the composite types and the non-branching nodes are the basic types.
The terminal rules in Fig.~\ref{Fig:marx-grammar} on the left are the \emph{basic}\/ typing statements, whereas the non-terminal rules on the right are the type \emph{constructors}, building composite types from simpler types. The constituency parse trees, like those in Figures \ref{Fig:john} and \ref{Fig:marx-parsing}, thus display the \emph{type structures}\/ of the parsed sentences. The words of the sentence appear on the leaves, whereas the inner tree nodes are the types. The branching nodes are the composite types and the non-branching nodes are the basic types. \emph{\textbf{Constituency parsing is typing.}}

Dependency parsings, on the other hand, do a strange thing: having routed the dependencies from a head term to its dependents through the constituent types that connect them, they sidestep the types and directly connect the head with its dependents. \emph{\textbf{Dependency parsing reduces syntactic typing to term dependencies}}.

\begin{figure}
\begin{center}
\includegraphics[width=.45\linewidth]{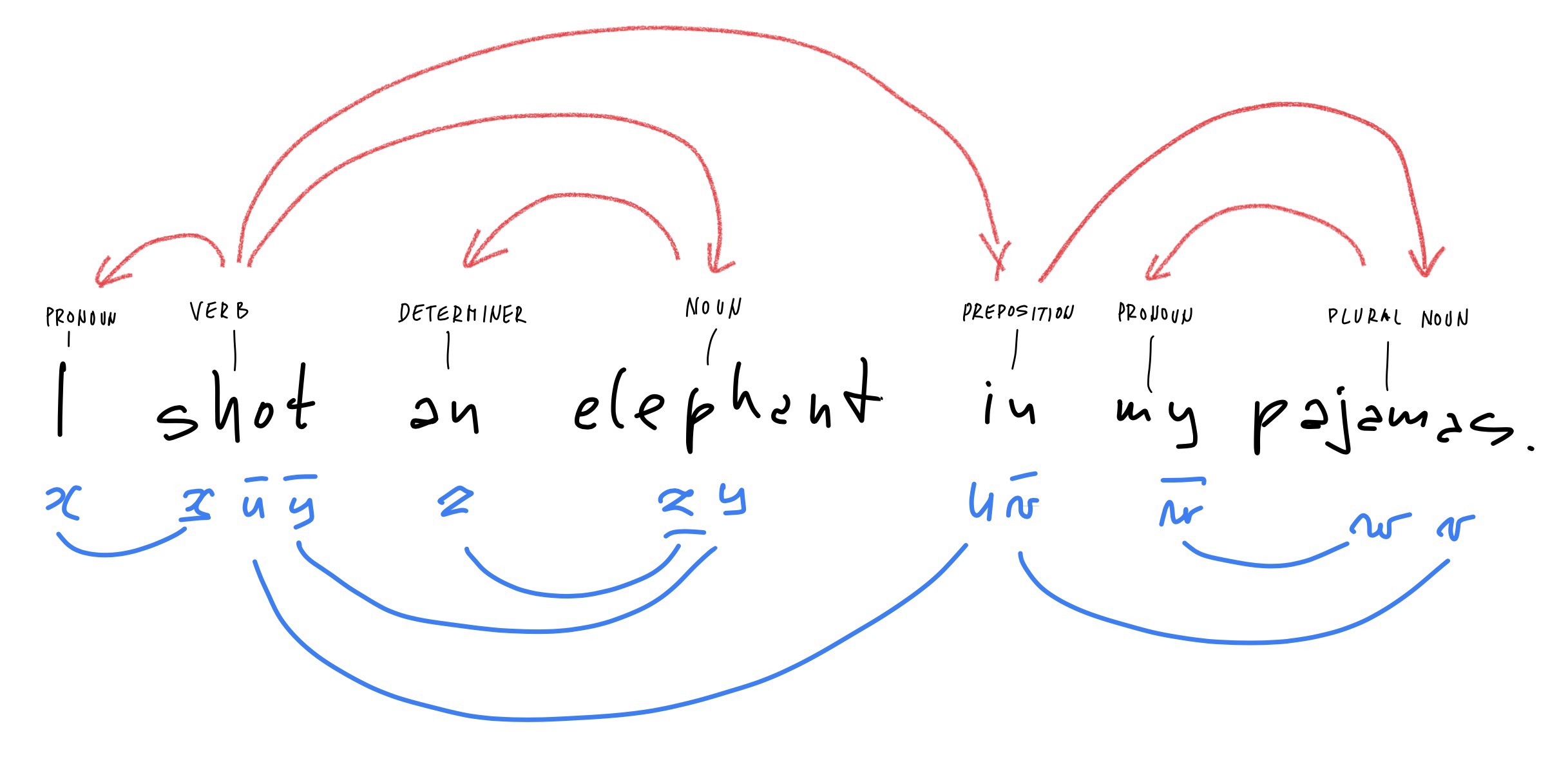}
\hspace{2em} 
\includegraphics[width=.45\linewidth]{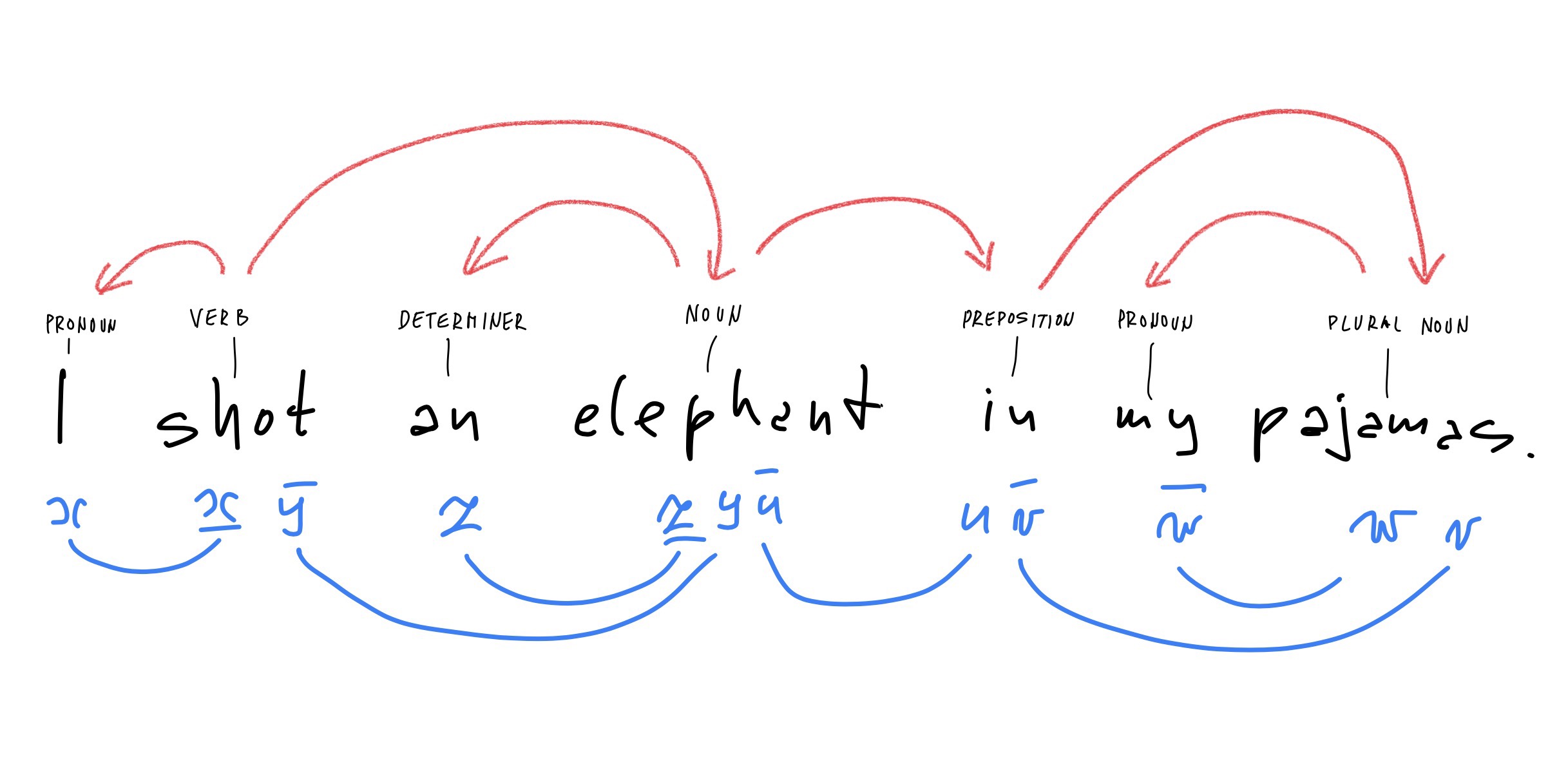}
\caption{Term dependencies encoded as annotations by syntactic types with adjunctions}
\label{Fig:pre-shot}
\end{center}
\end{figure}

But only the types that record nothing but term dependencies can be reduced to term dependencies. The two dependency parsings of the elephant sentence are displayed in Fig.~\ref{Fig:pre-shot}. 
The expressions below the two copies of the sentence are the syntactic types captured by dependency parsings. They are generated by tupling the \emph{reference variables} $x,y,\ldots$ etc., with their overlined \emph{left adjoints} $\overline x, \overline y,\ldots$ and their underlined \emph{right adjoints} $\underline x, \underline y\ldots$ 
Such syntactic types form pregroups, an algebraic structure introduced in the late 1990s by Jim Lambek, as a simplification of his syntactic calculus of categorial grammars. He had introduced categorical grammars in the late 1950s, to explore decision procedures for Ajdukiewicz's syntactic connexions from the 1930s and for Bar-Hillel's quasi-arithmetic from the early 1950s, both based on the reference-based logic of meaning going back to Husserl's ``Logical investigations''. Categorial grammars have been subsequently studied for many decades. We only touch pregroups,  only as a stepping stone.

\subsection{Pregroup grammars}
\subsubsection{Pregroup definition and properties}
A pregroup is an ordered monoid with left and right adjoints.  An ordered monoid is a monoid where the underlying set is ordered and the monoid product is monotone. 

If you know what this means, you can skip this section. You can also skip it if you don't need to know how it works, since the main idea should transpire as you go anyway. Just in case, here are the details.

\paragraph{Ordered monoid.} A \emph{monoid}\/ is a set $\MMM$ with a signature $\MMM\times \MMM\tto{(\cdot)} \MMM\oot\iota 1$  such that: 
\[
x\cdot(y\cdot z) = (x\cdot y)\cdot z \qquad\qquad\qquad\qquad x\cdot \iota = x = \iota\cdot x
\]
or equivalently in a diagram
\[
\begin{tikzar}[row sep = 7ex,column sep = 2ex]
\&\& \MMM\times \MMM \times \MMM \ar{dl}[description]{\MMM\times (\cdot)}\ar{dr}[description]{\MMM\times (\cdot)} \\
\& \MMM\times \MMM \ar{dr}[description]{(\cdot)} \&\& \MMM\times \MMM \ar{dl}[description]{(\cdot)}
\\
\MMM\times 1 \ar[leftrightarrow]{rr}[description]{\sim} \ar{ur}[description]{\MMM\times \iota} \&\&\MMM \ar[leftrightarrow]{rr}[description]{\sim} \&\&1\times \MMM \ar{ul}[description]{\iota\times \MMM}
\end{tikzar}
\] 
This monoid is \emph{ordered}\/ if the underlying set $\MMM$ is  ordered\footnote{We call ``partial orders'' orders, and explicitly specify linear orders, well-orders, etc.} by a binary relation $(\leq)$ and the monoid product operation $(\cdot)$ is monotone with respect to it, i.e. 
\bear
x\leq y & \implies & xz\leq yz \wedge zx\leq zy
\eear
\paragraph{Notation.} The operation $(\cdot)$ is usually elided and $x\cdot y$ is reduced to $xy$ whenever confusion is unlikely.

\paragraph{Adjoints.} For an ordered monoid $\MMM$ and $x\in \MMM$, 
\begin{itemize}
\item a \emph{left adjoint} $\overline x \in \MMM$ satisfies $\overline xx \leq \iota \leq x\overline x$, whereas
\item a \emph{right adjoint} $\underline x \in \MMM$ satisfies $x\underline x \leq \iota \leq \underline xx$.
\end{itemize}

\paragraph{Definition.} A pregroup is an ordered monoid with adjoints.

\paragraph{Exercises.} Derive the following properties of adjoints from their definitions:
\begin{enumerate}[a)]
\item $\overline x$ is a\hspace{.25em}  left\hspace{.25em}  adjoint  of $x$ iff $\overline x a\leq b \iff a\leq xb$ for all $a,b\in \MMM$. 
\item $\underline x$ is a right adjoint of $x$ iff $xa\leq b \iff a\leq \underline xb$ for all $a,b\in  \MMM$. 
\item If $x^{\ell}$ satisfies $x^{\ell}x \leq \iota \leq xx^{\ell}$, then $x^{\ell}= \overline x$.
\item If $x^{r}$ satisfies $xx^{r} \leq \iota \leq x^{r}x$, then $x^{r}= \underline x$.
\item if $x\leq y$ then $\overline y \leq \overline x$ and $\underline y \leq \underline x$ 

\item $\overline{x\cdot y} = \overline y \cdot \overline x$ and $\underline{x\cdot y} = \underline y \cdot \underline x$ 
\item $\overline{(\underline x)} = x = \underline{(\overline x)}$
\end{enumerate}

\paragraph{Examples.} The free pregroup generated by a poset $\Xi = \{x,y,\ldots\}$ of basic types is the set $\underline{\overline{\Xi}}^\ast$ of finite tuples of $\underline{x}^n$ and $\overline{x}^n$ for all $x\in \Xi$ and $n\in \NNn$. The pregroup operation is the tuple concatenation. The unit is the empty tuple $<>$. The expression $\underline{x}^2$ denotes the double left adjoint $\underline{\underline {x}}$, $\overline{y}^3$ is the triple right adjoint $\overline{\overline{\overline{y}}}$, etc. Property (g) above assures that no mixed adjoints are needed. The free pregroup $\underline{\overline{\Xi}}^\ast$ can thus be viewed as the free monoid over $\Xi\times \left(1 + \NNn + \NNn\right)$. For a simple example that is not free, consider the monoid of monotone maps $\ZZz\to \ZZz$, ordered pointwise. This ordered monoid is a pregroup because every bounded set contains its meet and join, and therefore every monotone map preserves them.

\subsubsection{Parsing as type-checking}
To check semantic correctness of a given phrase, each word in the phrase is first assigned a pregroup element as its syntactic type. The type of the phrase is the product of the types of its words, multiplied in the pregroup. The phrase is a well-formed sentence if its syntactic type is pregroup unit $\iota$. The arcs connect each type $x$ with an adjoint, either $\underline x$ on the left or $\overline x$ on the right. Since $\underline x x\leq\iota$ and $x\overline x \leq \iota$, eliminating adjacent adjoints results in a pregroup expression which is an upper bound of the phrase type. If the arcs are well-nested, eliminating the inner arcs makes the adjoints at the end of the outer arcs adjacent. That means that all arcs will eventually be eliminated if and only if they are well-nested.  

 \begin{figure}
\begin{center}
\includegraphics[width=.45\linewidth]{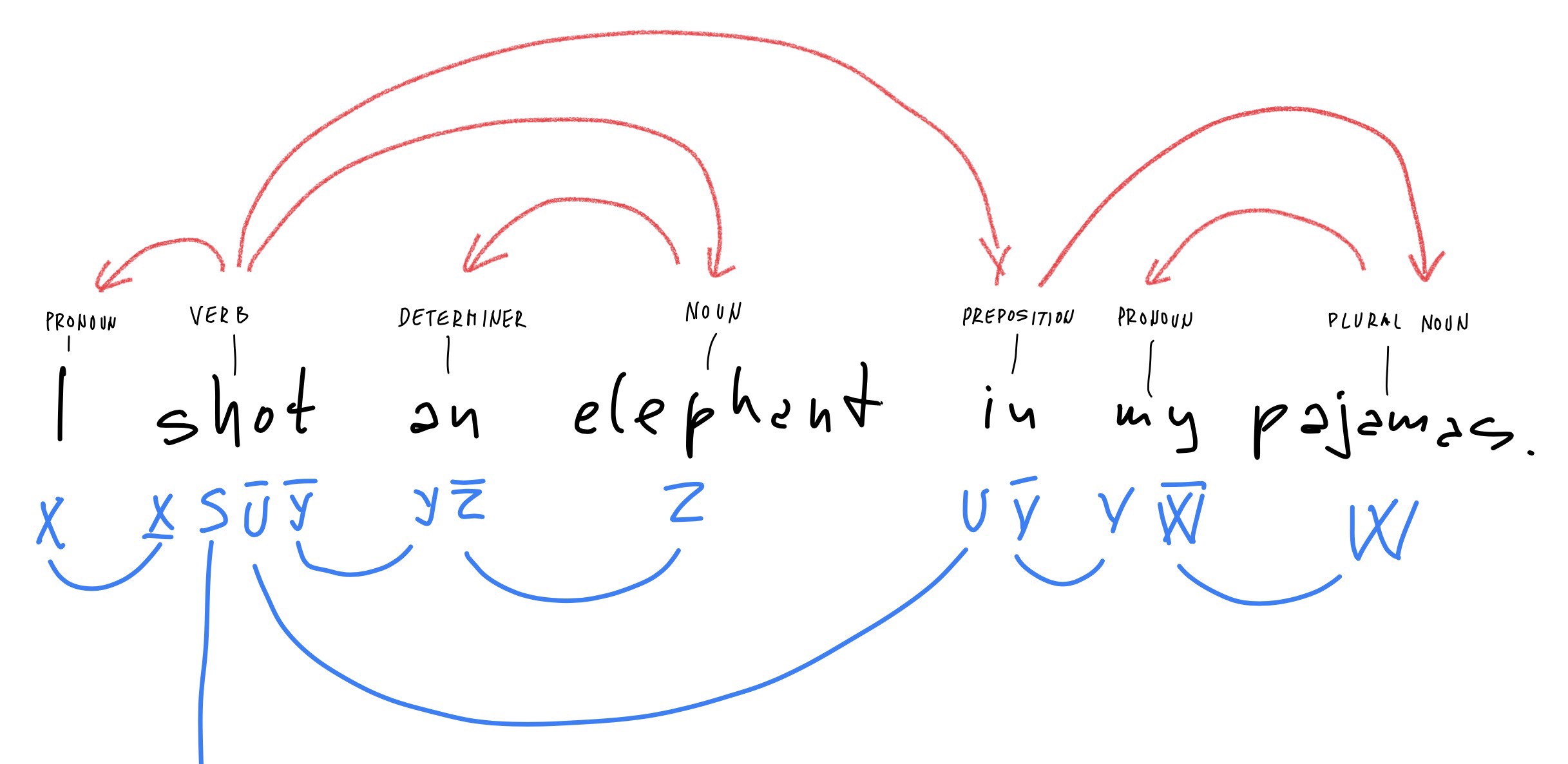}
\hspace{2em} 
\includegraphics[width=.45\linewidth]{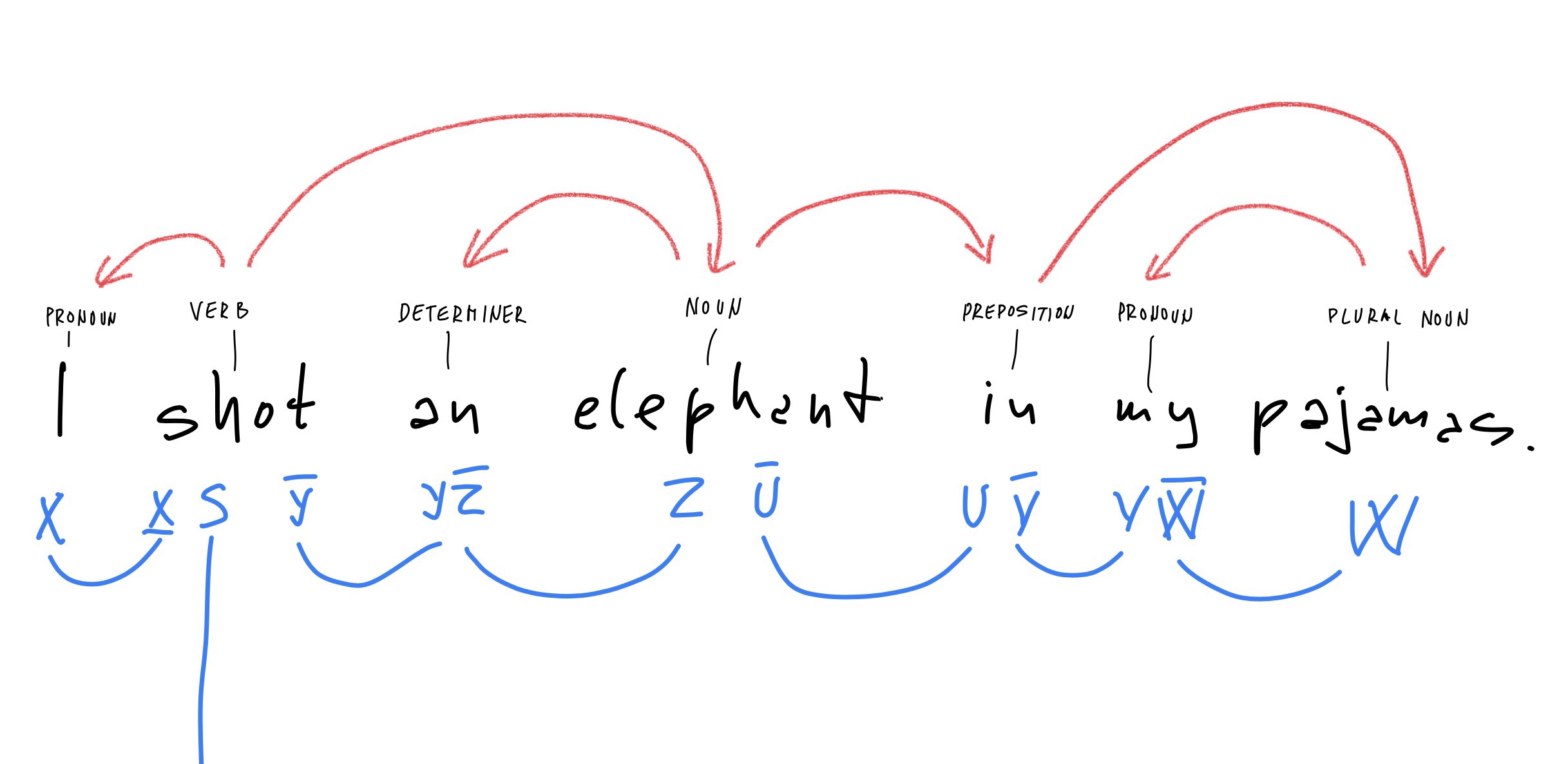}
\caption{Pregroup couplings do not always mirror dependencies}
\label{Fig:preg}
\end{center}
\end{figure} 

The main claim is that the pregroup types can be assigned to words in such a way that a phrase is a sentence if and only if multiplying in the pregroup the syntactic types of its words, in order in which they are given in the sentence, produces a result not greater than the unit $\iota$. In practice, the pregroup types are assigned slightly differently from Fig.~\ref{Fig:pre-shot}: see Fig.~\ref{Fig:preg}. First of all, the head of the sentence is annotated by a type variable $S$, corresponding to the pregroup unit. It is not discharged with an adjoint, and the wire from it does not arc to another type in the sentence but points straight out. This wire can be interpreted as a reference to another sentence. By linking the $S$-variables of pairs of sentences and coupling, for instance, questions and answers, pregroup versions of discourse syntax can be implemented. Still further up, by pairing messages and coupling, say, the challenges and the responses in an authentication protocol, one could implement a pregroup version of a protocol formalism. We will get back to this in a moment.

While they are related with dependency references, the pregroup couplings usually deviate from them. On the sentential level, this is because the words grouped under the same syntactic type in a lexicon should are expected to be assigned the same pregroup type. Lambek's idea was that even the phrases of the same type in constituency grammars should receive the same pregroup type. Whether such requirements are justified and advantageous is a matter of contention. In any case, after marking the head of the sentence by the $S$ type, most pregroup analyses proceed with assigning the basic types $X, Y, U,\ldots$ to nouns, and implement the constituency references using the adjoints. For more details, see an introduction to pregroup grammars (such as Jim Lambek's book ``From Word to Sentence''). The general point that matters here is that \emph{\textbf{syntax is typing}}.


\section{Beyond sentence}\label{Sec:beyond-sentence}

\subsection{Why do we make sentences?}
Why do we partition speech into sentences? Why don't we stream words, like network routers stream packets? Why can't we approximate what we want to say by adding more words, just like numbers approximate points in space by adding more digits?  


The old answer is: ``We make sentences to catch a breath''. 
When we complete a sentence, we release the dependency threads between its words. Without that, the dependencies accumulate, and you can only keep so many threads in your mind at a time. Breathing keeps references from knotting.

\paragraph{Exercise.} We make long sentences for a variety of reasons and purposes. A sample of long sentences written to great effects is provided in Appendix~\ref{Appendix:long}. Try to split them into shorter ones. Discuss what is gained and what lost by such operations. Ask a chatbot to do it.   

\paragraph{Anaphora} is a syntactic pattern that occurs within or between sentences. In rhetorics and poetry, it is the figure of speech where the same phrase is repeated to amplify the argument or thread a reference. In ChatGPT's view, it works because the rhythm of the verse echoes the patterns of meaning:
\begin{quote}
In every word, life's rhythm beats,\\
In every truth, life's voice speaks.\\
In every dream, life's vision seeks,\\
In every curse, life's revenge rears.\\
In every laugh, life's beat nears,\\
In every pause, life's sound retreats.
\end{quote}
Syntactic partitions reflect the semantic partitions.  \textbf{\emph{Sentential syntax is the discipline of charging and discharging syntactic dependencies to transmit semantic references.}} 

\subsection{Language articulations and network layers}
The language streams are articulated into words, sentences, paragraphs, sections, chapters, books, libraries, literatures; speakers tell stories, give  speeches, maintain conversations, follow conventions, comply with protocols. Computers reduce speech to tweets and expand it to chatbots.

\begin{figure}
\begin{center}
\includegraphics[width=.66\linewidth]{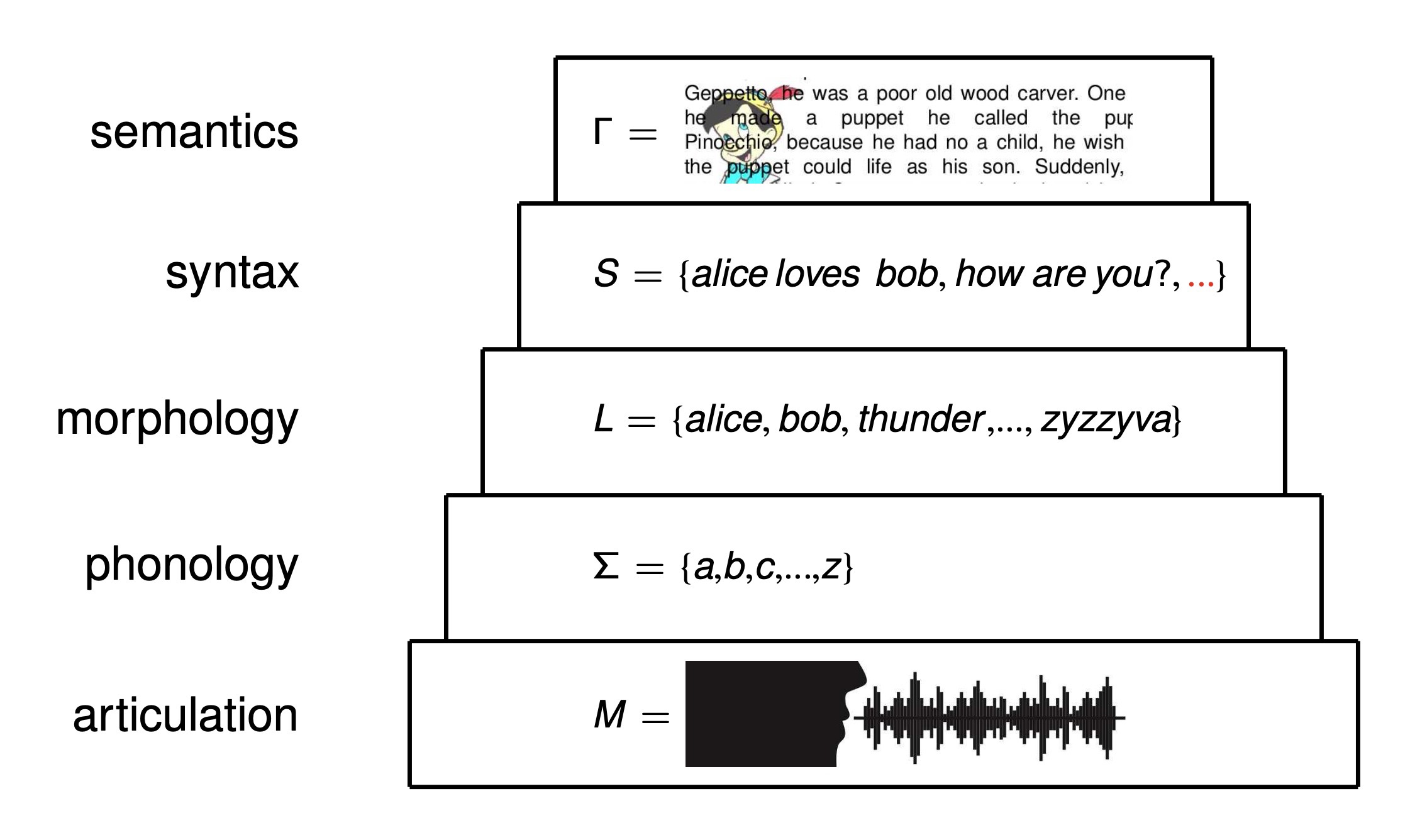} \\
\includegraphics[width=.66\linewidth]{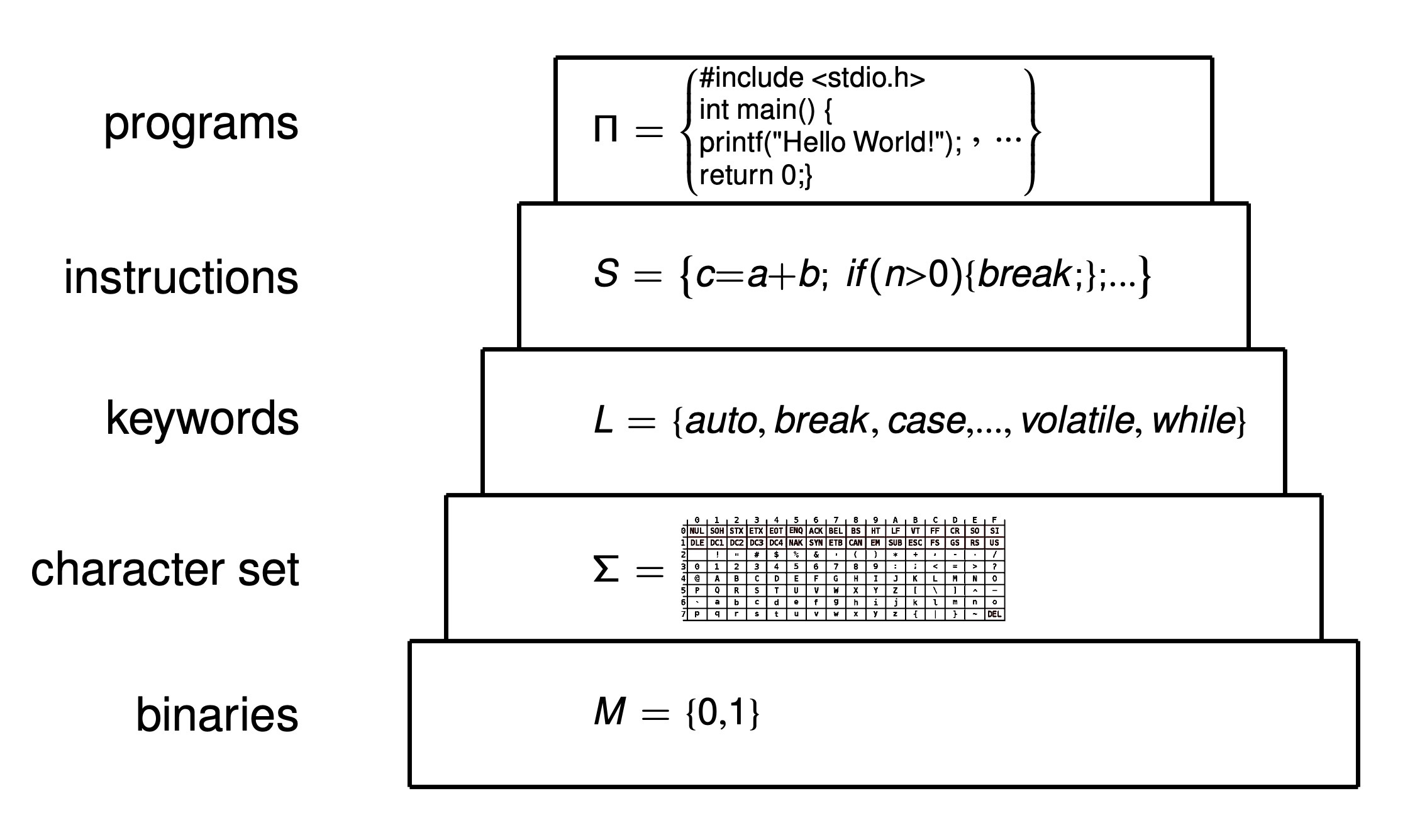} \\
\includegraphics[width=.66\linewidth]{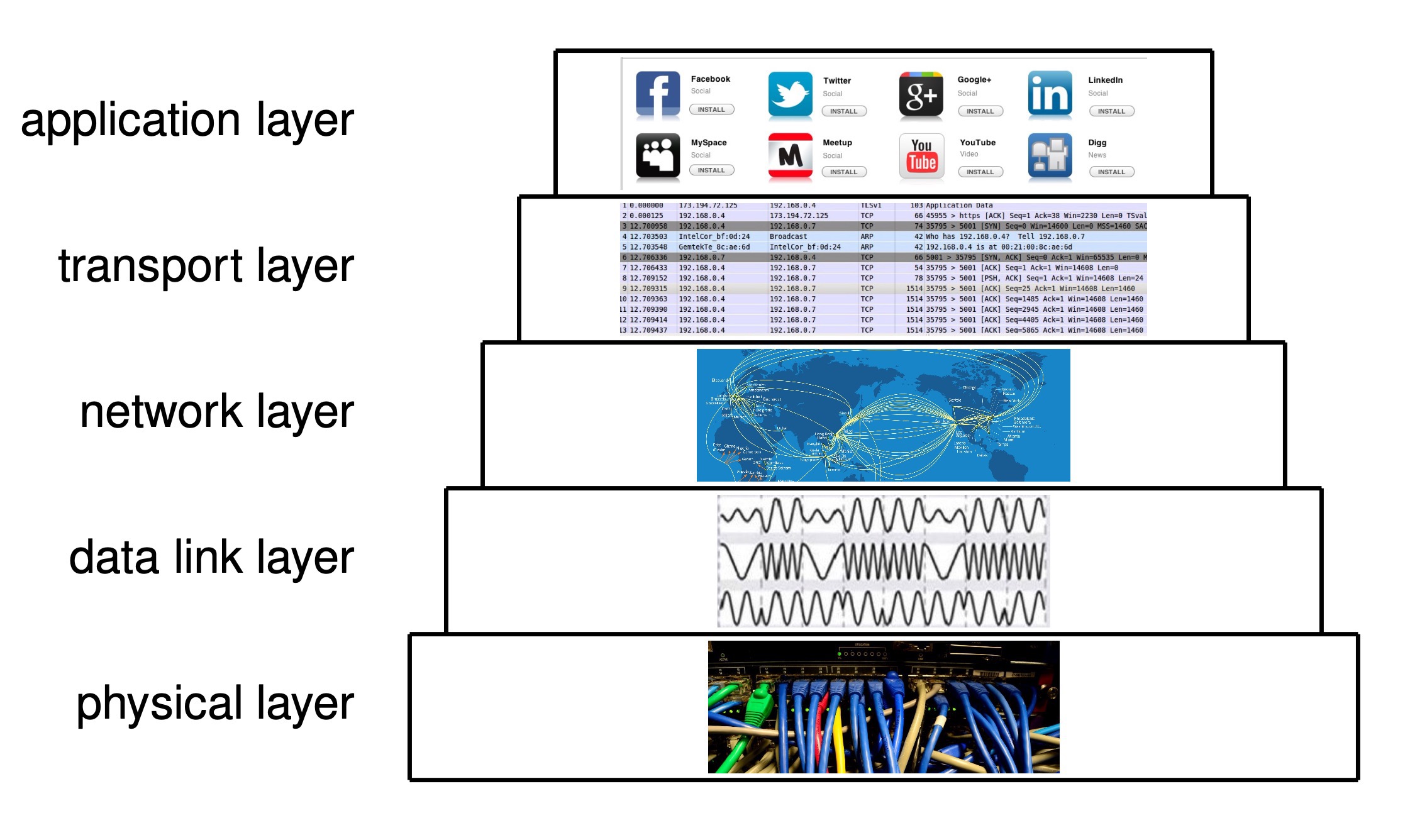}
\caption{Languages are articulated, channels are stacked}
\label{Fig:layers}
\end{center}
\end{figure}

\begin{figure}[!ht]
\begin{center}
\includegraphics[width=.9\linewidth]{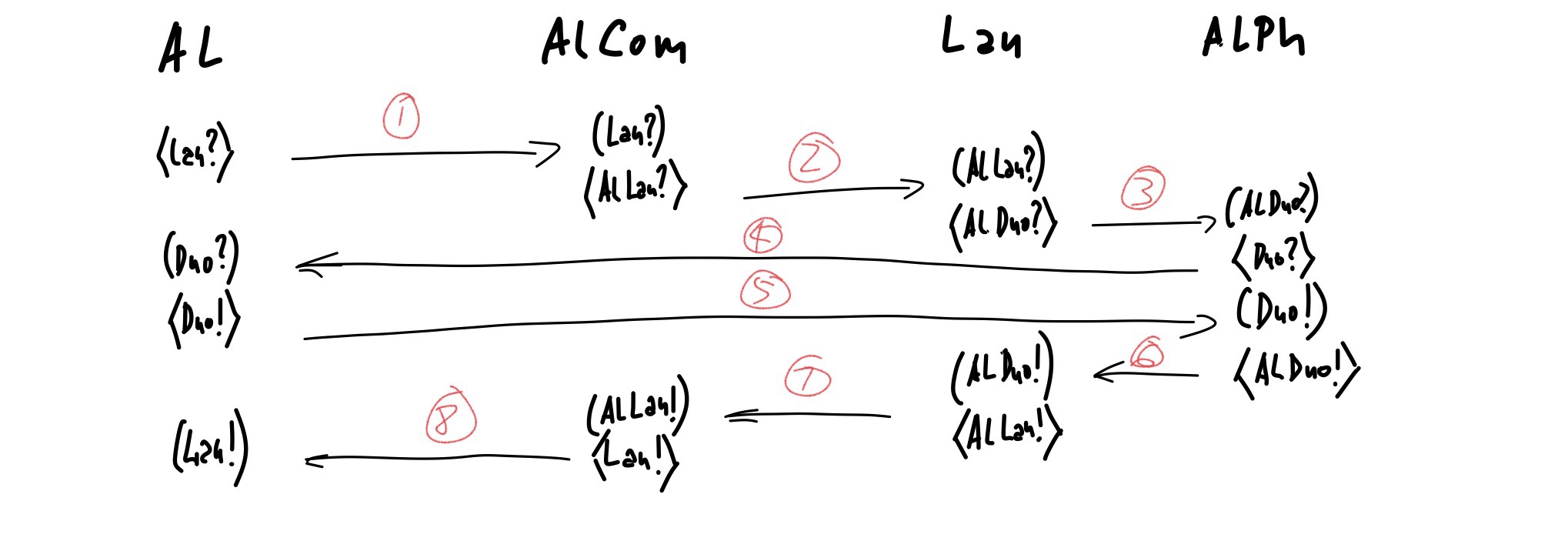}
\caption{2-factor Laulima authentication protocol presented using strand spaces}

\vspace{2\baselineskip}
\includegraphics[width=.9\linewidth]{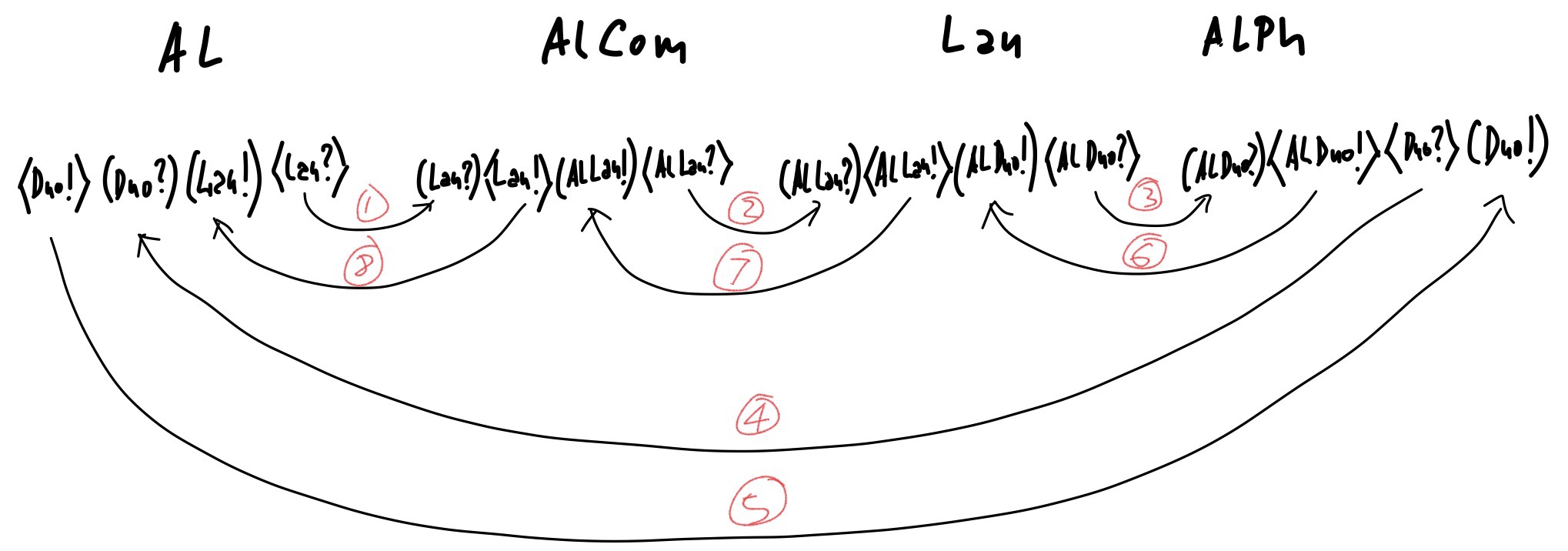}
\caption{2-factor Laulima authentication protocol presented using interaction types}
\label{Fig:MFA}
\end{center}
\end{figure}

The layering of language articulations is an instance of stratification of communication channels. Artificial languages evolved the same layering. The internet stack is another instance. Fig.~\ref{Fig:layers} displays the  analogy. 
%
%
This is not a quirk of representation. Information carriers are implemented on top of each other, in living organisms, in the communication networks between them, and in all languages developed by the humans. The reference coupling mechanisms, similar to the syntactic type structures that we studied, emerge at all levels. The pregroup structure of sentential syntax is reflected by the question-answer structure of simple discourse and by the SYN-ACK pattern of the basic network protocols. Closely related structures arise in all kinds of protocols, across the board, whether they are established to regulate network functions, or secure interactions, or social, political, economic mechanisms. Fig.~\ref{Fig:MFA}  shows a high-level view of a simple 2-factor authentication protocol, presented as a basic cord space. 
The \textbf{upshot} is that
\begin{itemize}
\item natural-language conversations,
\item software-system architectures,
\item security and network protocols
\end{itemize}
share crucial features. It is tempting to think of them as a product of a high-level deep syntax, shared by all communication processes. Such syntax could conceivably arise from innate capabilities hypothesized in the Chomskian theory, or from physical and logical laws of information processing.

%
%

\section{Beyond syntax}

We have seen how syntactic typing supports semantic information transmission. Already Groucho's elephant sentence fed syntactic and semantic ambiguities back into each other. 

But if syntactic typing and semantic assignments steer each other, then the generally adopted restriction of syntactic analyses to sentences cannot be justified, since semantic ambiguities cannot be resolved on the level of sentence. Groucho proved that.

\paragraph{Semantic context-sensitivity.} Consider the sentence 
\begin{quote}
John said he was sick and got up to leave.
\end{quote} 
A context changes semantic assignments:
\begin{quote}
Mark collapsed on bed. 
John said he was sick and got up to leave.
\end{quote} 
For most people, ``he was sick'' now refers to Mark. Note that the silent ``he'' in ``[he] got up to leave'' remains bound to John. Or take 
\begin{quote}
Few professors came to the party and had a great time.
\end{quote}
The meaning does not significantly change if we split the sentence in two and expand :
\begin{quote}
Since it started late, few professors came to the party. They had a great time.
\end{quote}
Like in the John and Mark example, a context changes the semantical binding, this time of  ``it'':
\begin{quote}
There was a departmental meeting at 5. Since it started late, few professors came to the party. They had a great time.
\end{quote}
But this time, adding a first sentence that binds the subject ``they'' differently may change the meaning of ``they'' in the last sentence:
\begin{quote}
They invited professors. There was a departmental meeting at 5. Since it started late, few professors came to the party. 
They had a great time.
\end{quote}
The story is now that  \emph{students}\/ had a great time --- ?the students who are never explicitly mentioned! Their presence is only derived from the background knowledge about the general context of professorial existence ;)

\paragraph{Syntactic context-sensitivity.} On the level of sentential syntax of natural languages, as generated by formal grammars, proving context-sensitivity amounts to finding a language that contains some of the patterns known to require a context-sensitive grammar, such as $a^{n}b^{n}c^{n}$, or $a^{m}b^{n}c^{m}d^{n}$ for arbitrary letters $a,b,c,d\in\Term$ and any number $n$, or $ww$, or $www$, or $wwww$ for arbitrary word $w\in \Term^{\ast}$, etc. Since people are unlikely to go around saying to each other things like $a^{n}b^{n}c^{n}$ in any culture, the task boiled down to finding languages which require constructions in the form  $ww$, $www$, etc. The quest for such examples became quite competitive. 

Since a language with a finite lexicon has to have a finite number of words for numbers, at some point you must need to say something like ``quadrillion quadrillion'', assuming quadrillion is the largest number denoted by a single word. But it was decided by the context sensitivity competition referees that numbers don't count. 

Then someone found that in the Central-African language Bambara, the construction that says ``any dog'' is in the form ``dog dog''. Then someone else noticed context-sensitive nesting phenomena in Dutch, but not everyone agreed. Eventually, most people settled on Swiss German as a definitely context sensitive language, and the debate about syntactic contexts-sensitivity subsided. With a hindsight, it had the main hallmarks of a theological debate. The main problem with counting how many angels can stand on the tip of a needle is that angels generally don't hang out on needles. The main problem with syntactic context sensitivity is that contexts are never purely syntactic.

\paragraph{Communication is the process of building and sharing semantical contexts.} Chomsky noted that natural language should be construed as context-sensitive as soon as he defined the notion of context-sensitivity. Restricting the language models to syntax, and syntax to sentences, made proving his observation into a conundrum.
But how that the theology of syntactic contexts is behind us, and the language models are in front of us, waiting to be understood, the question arises: \textbf{\emph{How are contexts really processed?}} How do we do it, and how do the chatbots do it? Where do we all store big contexts? The reader of a novel builds the context of that novel starting from the first sentence, and refers to it 800 pages later. How does a language model find the target of such a reference? It cannot maintain references between everything to everything. How do you choose what to remember?

Semantic dependencies on remote contexts have been one of the central problems of natural language processing from the outset. The advances in natural language processing that we witness currently arise to a large extent from progress in solving that problem. To get an idea about the challenge, consider the following paragraph, contrived to match the context of Sir Arthur Conan Doyle's ``The final problem'':

\begin{quote}
Unsteadily, Holmes stepped out of the barge. Moriarty was walking away
down the towpath and into the fog. Holmes ran after him. `Give it back to me', he shouted. Moriarty turned and laughed. He opened his hand and the
small piece of metal fell onto the path. Holmes reached to pick it up but
Moriarty was too quick for him. With one slight movement of his foot, he tipped the key into the lock. --- Arthur Conan Doyle, \emph{The Final Problem}
\end{quote} 

If you are having trouble understanding what just happened, you are in a good company. Without sufficient hints, none of the currently available chatbots seem to be able to produce a correct interpretation. A chat with one of them, with a hint and a solution is in Appendix~\ref{Appendix:Moriarty}. In the next lecture, we will study how the contexts are generated, including much larger. After that, we will be ready to explain how they are processed.
%

\def\thechapter{3}
\setchaptertoc
\chapter{Semantics: The Meaning of Language}\label{Chap:Semantics}

\section{What does it mean to mean?}

\subsection{Meaning as organizing}
You arrive from 
a faraway place
 (be it as a baby or as an alien) and you ask: 
\emph{`What is a hammer?'} Someone shows you a hammer. \emph{`This is a hammer'}, they say. But you have never seen one before. \emph{`What is that?'}, you ask. The question of meaning of a word has become the question of meaning of an object. The situation wouldn't have been all that different if you first asked \emph{`What is this?'} and they said \emph{`This is a hammer'}. Great, thank you very much. Holding a hammer in your hand, you guess that it should be tasted and that ``hammmer'' probably means tasty. Then they show you how to hammer nails in the wall\ldots 
\begin{figure}[!ht]
\begin{center}
\includegraphics[height=3.5cm]{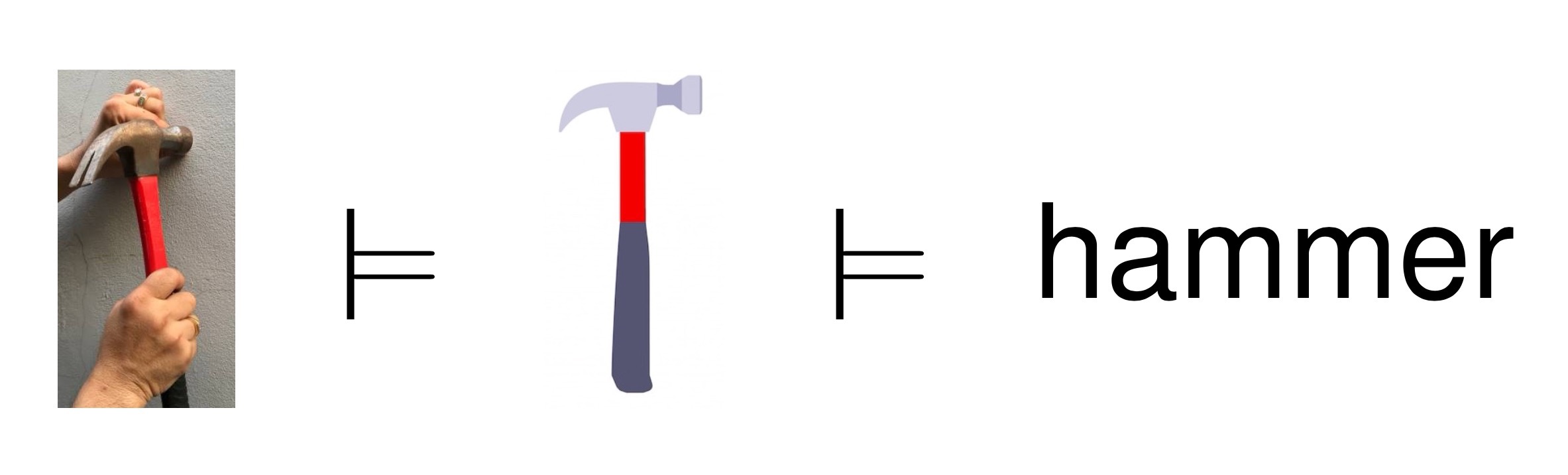}
\caption{The meaning of a hammer as an object and of ``hammer'' as a word}
\label{Fig:hammer}
\end{center}
\end{figure}
%
%
%
%
%
%
%

Many years later, you take a language course, and there is Fig.~\ref{Fig:hammer}, with the word ``hammer'' referring to an object and the object referring to the operation of hammering nails in the wall. The figure itself refers to the \emph{concept}\/ of meaning. It \emph{abstracts}\/ a general concept from concrete instances. Then there is a paragraph which explains this general process of assigning meanings by referring to concrete instances of that general process\ldots
%
%
%

Everything means something. We organize the world by using signs to assign meanings. Meaning is the process where signs assign meanings to each other. The process is recurrent: signs refer to signs, meanings to meanings. 

Language is said to be the characterizing property of our species.\footnote{Ren\'e Descartes points to language as the main distinction between humans and animals in his ``Discourse on Method'', in the paragraph that can be read as a preamble to Turing's test for distinguishing humans from computers.}  Another proposed characterizing property is that we use tools. Both properties refer to the same practice. 
Many animals use signs and tools. Meerkats have a word for a snake and a word for an eagle, and the difference becomes a matter of life and death when it comes to seeking shelter up in a tree or down in a burrow. Crows use sticks to poke grubs out of holes. But crows throw away their tools when they are done, whereas humans put their tools in tool boxes. They reuse and combine tools. The same tool means different things in different contexts. Tools are signs, signs are tools. They refer to each other and both obey syntax. The instrument is an extension of the musician, the musician of the instrument. Our bodies and our tools evolve together, our mind and our language. They cannot perform without each other. The driver is one with the car, the surfer with the board, the lover with the lover. Language is the process of becoming one through meaning. 
%

\subsection{Meaning as abstraction} 
What do you mean when you say ``cow''?  

``Pizza''? 

``Warm''? 

``Love''?

\begin{figure}[!ht]
\begin{center}
\includegraphics[height=8cm]{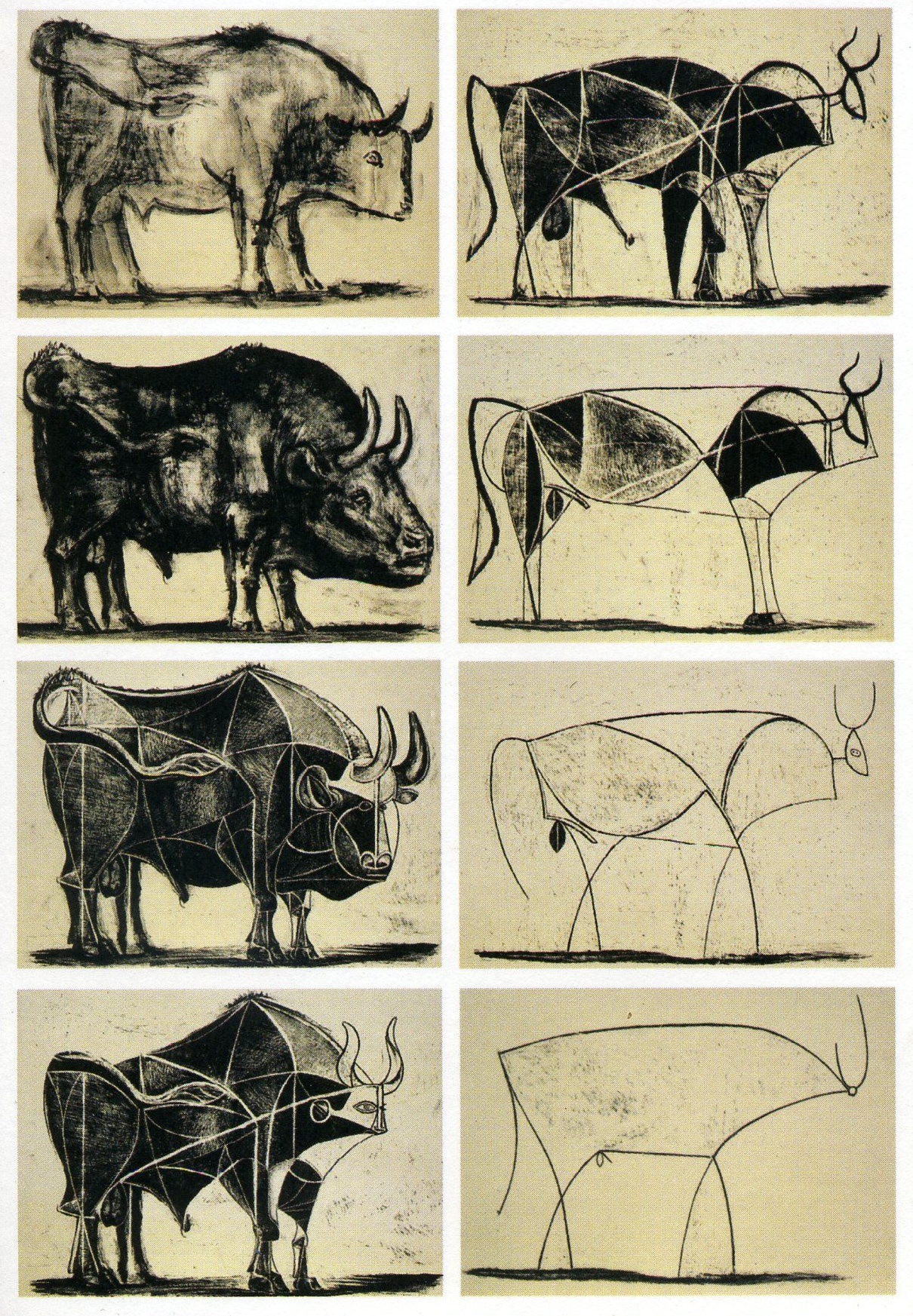}
\hspace{4em}
\includegraphics[height=8cm]{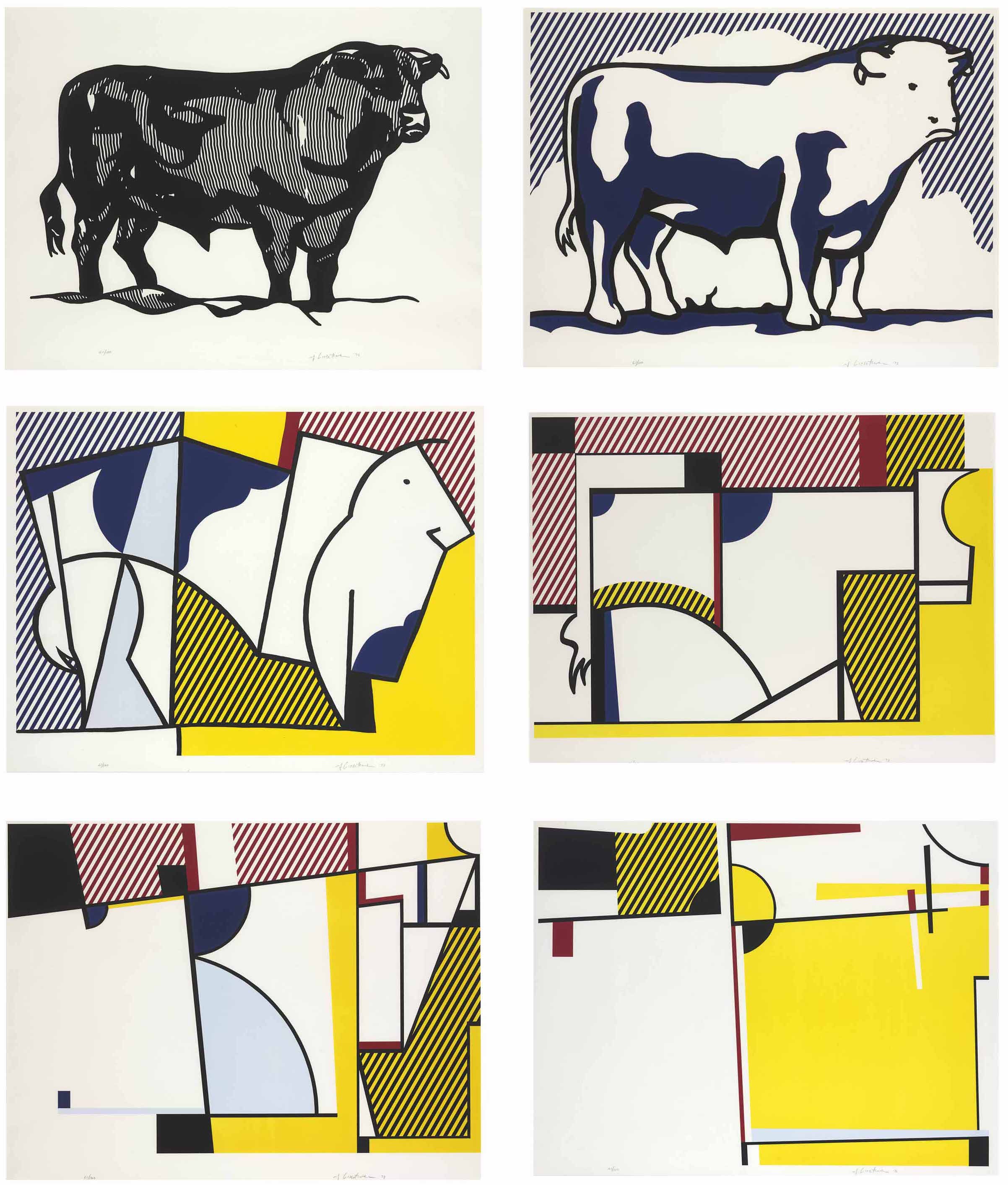}
\caption{Pablo Picasso: shape abstraction.\ \ \  Roy Lichtenstein: color abstraction}
\label{Fig:abstraction}
\end{center}
\end{figure}

\begin{figure}[!ht]
\begin{center}
\includegraphics[height=6cm]{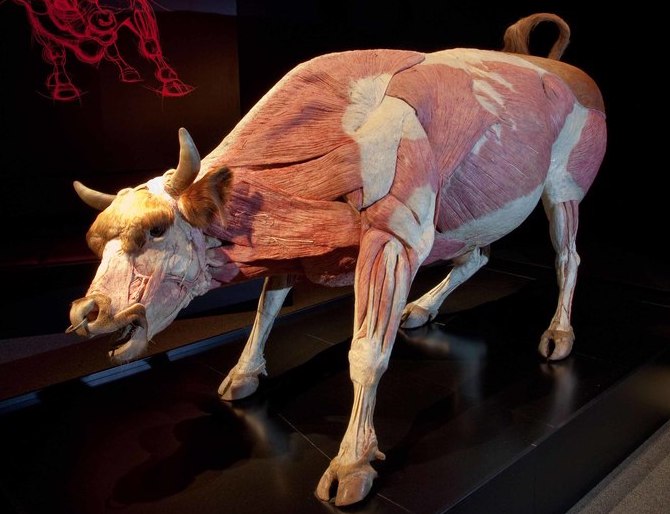}
\caption{Gunther Hagen: antiabstraction}
\label{Fig:hagen}
\end{center}
\end{figure}

The simplest concept of concept is that it is an abstraction of some concrete things. Your concept of cow would then be the memory of all times when you observed a cow, faded by time and mixed together. By ``observed a cow'' I mean: you saw or heard or touched or milked a cow, or saw a picture of a cow, or sketched her portrait.  Cow's shapes, smells, and sounds mix with the shapes and the sounds of the word ``cow'', with the contexts where it occurs, with the menu choices in a video game or in a food court. The ongoing process of \emph{abstraction}\/ decomposes concepts it and projects away some of their aspects. Fig.~\ref{Fig:abstraction} displays two instances of cow abstraction. Fig.~\ref{Fig:hagen} displays a concrete cow, displayed in a gallery as an instance of abstraction of abstraction. What you now read is an instance of abstraction of abstraction of abstraction. Meaning evolves through abstraction.

So from which direction should we then approach meaning to make sense of it?

\subsection{The path to noncommutative geometry of meaning: from vectors to streams}\label{Sec:com-noncom}
\begin{figure}
\begin{center}
\includegraphics[height=3.5cm]{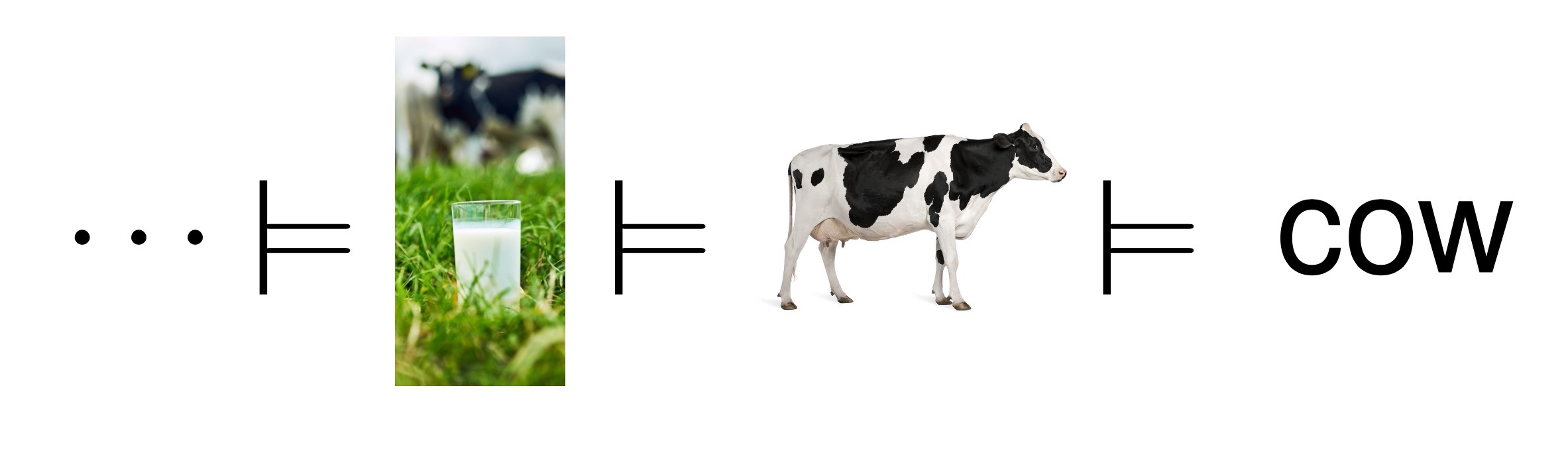}
\caption{Static meaning as a relation of words, objects, concepts, operations}
\label{Fig:cow-static}
\end{center}
\end{figure}

\begin{figure}[!hb]
\begin{center}
\includegraphics[height=3.5cm]{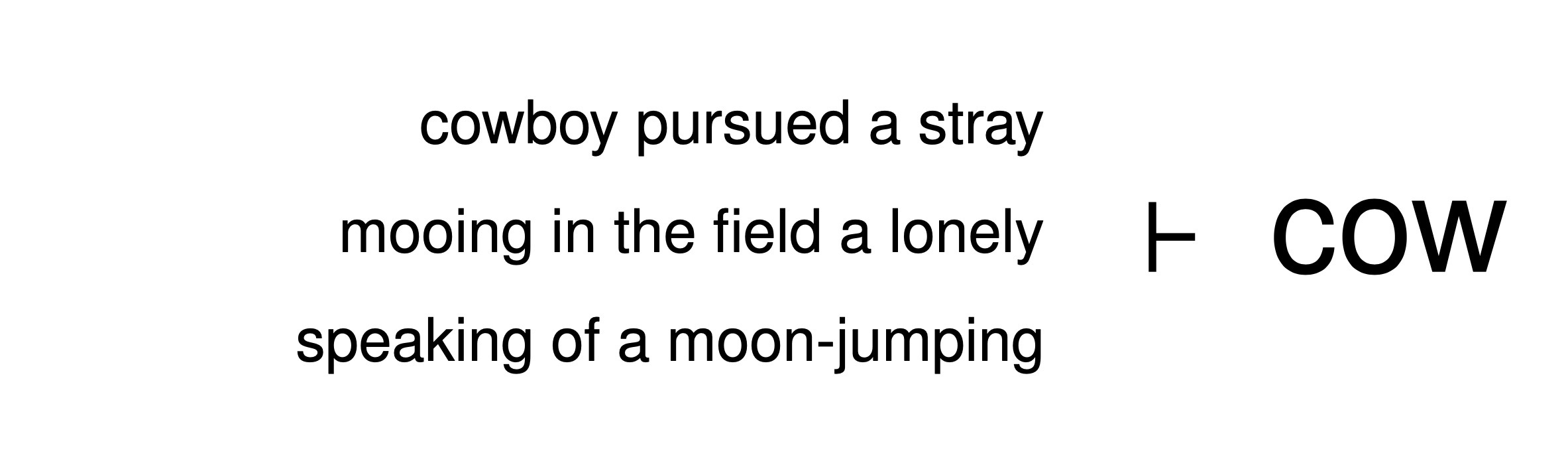}
\caption{Dynamic meaning as a channel}
\label{Fig:cow-dynamic}
\end{center}
\end{figure}

Every science is characterized by its space. Classical physics is mostly developed in the 3-dimensional  Euclidean space. Special relativity made it into spacetime by adding the 4-th dimension of time. General relativity lives in the curved space of Riemannian geometry. Computer science lives in the space of bitstrings. Biology lives in the space of genes. Static semantics is mostly developed in multi-dimensional Euclidean spaces. Dynamic semantics adds the dimension of time, but differently from special relativity. Static semantics is commutative in the sense that Alice and Bob are the same as Bob and Alice. Or to say the same thing in terms of cows: the meaning of the word "cow" and the meaning of the corresponding animal, and the meaning of that meaning, etc., they may evolve, each on their own, but they remain related in the same way, statically, as in Fig.~\ref{Fig:cow-static}.
Dynamic semantics, on the other hand, is noncommutative because if Alice speaks before Bob, the terms of reference of their conversation may differ from the terms of reference that Bob would establish if he were to speak first. In terms of cows, the reason would be that Alice and Bob may view cows from different angles, as in Fig.~\ref{Fig:cow-dynamic}.
%
%
%
%
%
%

\subsubsection*{Plan}
In Sec.~\ref{Sec:space}, we study the static meaning of words, phrases, objects. In Sec.~\ref{Sec:spacetime}, we study the dynamic processes of meaning, threading through sentences, messages, protocols, channels.

\section{Static semantics: words in space}
\label{Sec:space}

When you enter unfamiliar space, you first note where things are: how many are there, how they are related, how far from each other. Then you try to understand what is what. Static semantics works something like that. In Sec.~\ref{Sec:VSM}, we retrace the \emph{Vector Space Model (VSM)}, which interprets words as tuples of numbers and uses them as coordinates to place words in space, where their meanings can be related as angles or distances. In Sec.~\ref{Sec:LSA}, we review the basic methods of \emph{semantic analysis}: how to extract concepts from words. 

\subsection{Vector Space Model: Counting words}\label{Sec:VSM}

\subsubsection{Background}
\paragraph{The Vector Space Model is \emph{not}\/ a language model.} The methods of vector space modeling were introduced in computer engineering by Gerard Salton, in the 1970s, to address the tasks \emph{information retrieval}\/ from databases. An early description of these tasks was Vannevar Bush's famous \emph{Memex}\/ memo from 1946. Writing at the time when a new world order was being established and the first computers were being built, Bush anticipated the World Wide Web, some 50 years in advance. While information retrieval was the hard part of organizing the Web, it is the easy part of the \emph{information supply}, the main task addressed by natural languages. The critical difference is that \textbf{\emph{information retrieval is an operation, whereas information supply is a process}}. This is why information is in principle retrieved on \emph{unordered}\/ carriers, and the traditional information measures are invariant under permutations of the carrier elements, whereas it is supplied along the channel flows which are \emph{ordered}\/ by time, code, and syntax.
%
%
%
%
%

\subsubsection{Vector representations}\label{Sec:pizza}
Although the vector representations were formalized as a data architecture for computer-based information retrieval, they have been in use for a long time as the tacit  data architecture for natural-language-based information exchanges, e.g. in cooking recipes. 

\paragraph{Pizza vectors.} 
\begin{table}
\begin{center}
\begin{tabular}{c||c|c|c|}
\cline{2-4}
& Pizza Margherita & Meat White Pizza & Hawaiian Pizza  \\
\hline
\hline
\multicolumn{1}{|c||}{flour} & 490 & 490 & 490  \\
\hline
\multicolumn{1}{|c||}{yeast} & 5 & 5 & 5 \\
\hline
\multicolumn{1}{|c||}{water} & 355 & 355 & 355  \\
\hline
\multicolumn{1}{|c||}{salt} & 8  & 8  & 8 \\
\hline
\multicolumn{1}{|c||}{oil} & 3  & 3  & 3 \\
\hline
\multicolumn{1}{|c||}{tomato sauce} & 80  & 0 & 70 \\
\hline
\multicolumn{1}{|c||}{alfredo sauce} & 0 & 70 & 0  \\
\hline
\multicolumn{1}{|c||}{mozzarella} & 90 & 0 &0 \\
\hline
\multicolumn{1}{|c||}{fontina} & 20 & 40 & 70 \\
\hline
\multicolumn{1}{|c||}{parmesan} & 0 & 20 & 30 \\
\hline
\multicolumn{1}{|c||}{mushrooms} & 0  & 0 & 30 \\
\hline
\multicolumn{1}{|c||}{onions} & 0  & 20 & 0 \\
\hline
\multicolumn{1}{|c||}{peppers} &0& 0 & 20 \\
\hline
\multicolumn{1}{|c||}{olives} & 20  & 10 &  0  \\
\hline
\multicolumn{1}{|c||}{basil} & 20 & 0 & 10 \\
\hline
\multicolumn{1}{|c||}{pineapple} & 0& 0 & 90 \\
\hline
\multicolumn{1}{|c||}{sausage} & 0 &  60 & 0 \\
\hline
\multicolumn{1}{|c||}{ham} & 0& 30 & 60 \\
\hline
\multicolumn{1}{|c||}{chicken} & 0 & 40 & 0 \\
\hline
\multicolumn{1}{|c||}{meatballs} & 0 & 40 &0  \\
\hline
\end{tabular}
\end{center}
\caption{Pizzas as vectors of their ingredients}
\label{Table:pizzavec}
\end{table}

\begin{figure}[!ht]
\begin{center}
\begin{minipage}{.4\linewidth}
\centering
\includegraphics[height=5cm]{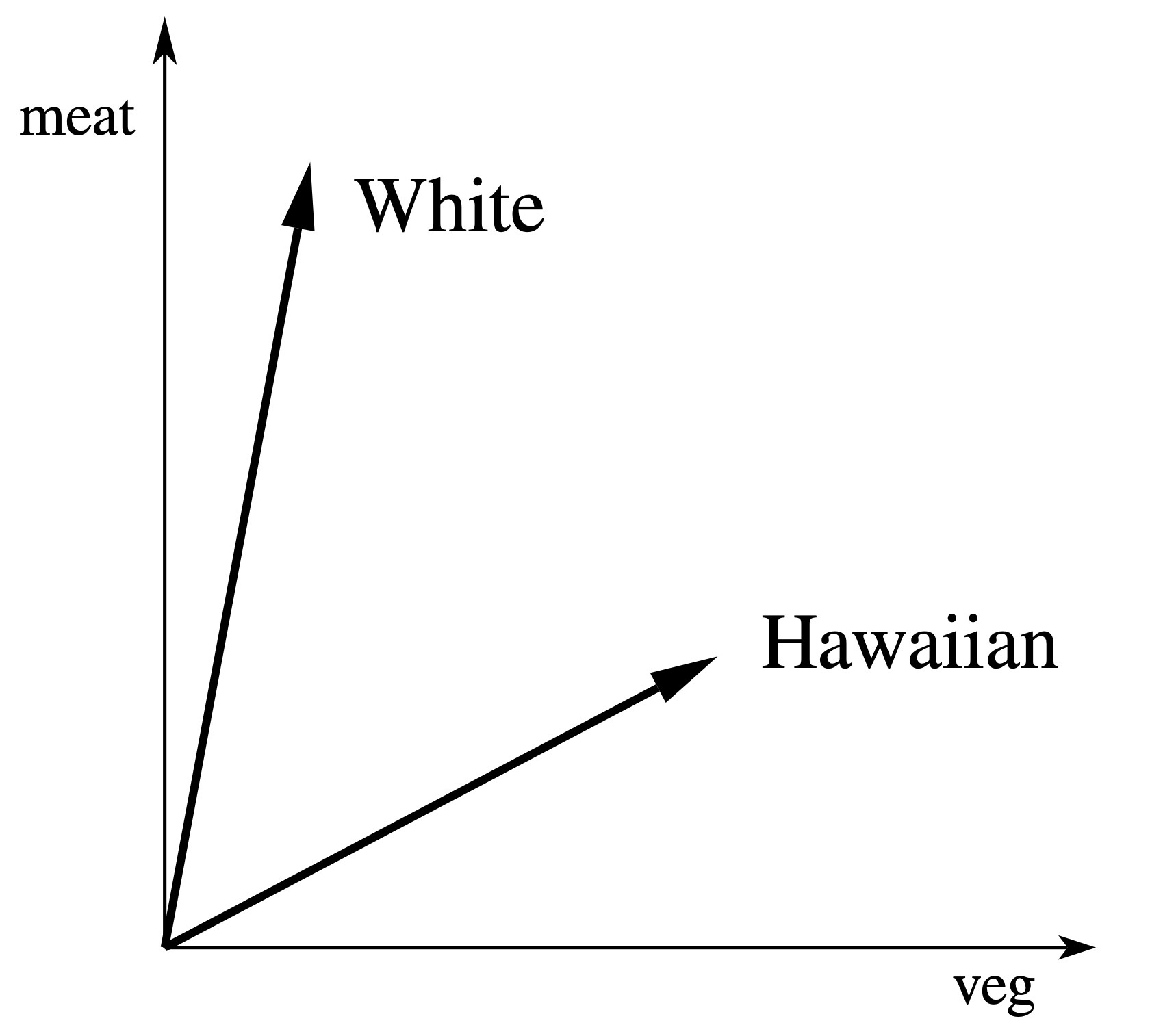}
\end{minipage}
\begin{minipage}{.4\linewidth}
\begin{center}
{\small \begin{tabular}{c||c|c|}
\cline{2-3}
&\small White & Hawaiian  \\
\hline
\hline
\multicolumn{1}{|c||}{veg} & 30 & 120  \\
\hline
\multicolumn{1}{|c||}{meat} & 170 & 60 \\
\hline
\end{tabular}}
\end{center}
\end{minipage}
\caption{Pizzas in space: as vectors of their vegetable and meat ingredients}
\label{Fig:cooccurrence}
\end{center}
\end{figure}

Table~\ref{Table:pizzavec} shows three kinds of pizzas as vectors of their ingredients.  A real cooking recipe, of course, also describes the preparation steps. That can be added in a bigger matrix. To give you an idea how that would work, a slightly bigger example is provided in Appendix~\ref{Appendix:pizza}. Although cooking is a dynamic process, the recipes are presented as static lists of ingredients and operations. This is possible because the cooking culture contexts provide the operational details: how to cut a sausage, on which side to lay pizza when you bake it, etc. For centuries, all kinds of tacit vector representations have been parsed according to unspoken cultural grammars and lexicons of common knowledge.
But once the vectors are noticed and placed in space, a geometry of concepts emerges, including the geometry of the pizza concepts. E.g., what is the angle between White and Hawaiian? The answer depends on our frame of reference\footnote{The \emph{frame of reference}\/ is the coordinate system traveling with the observer in the universe of special relativity.} in the pizza world. If we just take into account two types of ingredients, vegetables and meats, the ingredients add up to the column vectors of the matrix in Fig.~\ref{Fig:cooccurrence} on the right. The vectors can then be displayed in the veg-meat space on the left. You see the angle. 

If we write the pizza vectors in the form
\bear
\kett{White} & = & \kett{veg}\ \ 30  + \kett{meat}170\\
\kett{Hawaiian} & = & \kett{veg}120 + \kett{meat}\ \ 60
\eear
then we can calculate the length of the projection of each of the vectors on the other one\footnote{For any pair of vectors $\vec x, \vec y$, the length of the projection of $\vec x$ to $\vec y$ is equal to the length of the projection of $\vec y$ to $\vec x$.} using their \emph{inner product}
\bea\label{eq:inner-wh}
\brakett{White}{Hawaiian} & = & 30\cdot 120 + 170\cdot 60 = 13,800
\eea
Comparing the inner products of different pizza pairs provides a geometric insight into the pizza semantics. The way such inner products are related with the lengths of vectors' projections on one another is explained in Appendix~\ref{Appendix:inner}. To factor out the impact of the vector length, corresponding to the pizza weight, and only take into account the angle between the vectors, measuring the similarity of the ratios of the pizza  ingredients, we could divide out the vectors with their lengths, and compute their similarity as the cosine of their angle, which is the inner product of the obtained unit vectors
\bear
sim(\textrm{White},\textrm{Hawaiian}) & = & \frac{\brakett{White}{Hawaiian}}{\length{\mbox{White}}\cdot\length{\mbox{Hawaiian}}}\ \ =\ \ \frac{13,800}{\sqrt{29,800\cdot18,000}} \approx 0.595847
\eear

\paragraph{Beyond the cooking recipes.} The upshot of our pizza investigations is that \empty{the vector representations arose around natural language}, as data exchange formats, just like numbers and arithmetic arose as tools for counting and trading and bookkeeping. People developed the basic algorithms to count livestock, measure harvest, and exchange their goods. They developed the vector representations and tables and matrices to transmit cooking recipes and building plans, to coordinate their projects. Each generation probably thought that it was all invented with their advanced technologies, like we do. But \emph{counting, weighing, and the vector semantics evolved with nearly every culture and every natural language}.  

Table~\ref{Table:domains} lists some of the application domains where vector representations are in standard use nowadays. 
\begin{table}[!ht]
\begin{center}
\begin{tabular}{|c||c|c|c|c|}
\hline
domain & $\JJJ$ & $\UUU$ & $\sf Val$ & $C_{iu}$\\
\hline \hline 
text analysis & documents & terms & $\NNn$ & occurrence \\
 \hline
concept analysis & objects & attributes & $\{0,1\}$ & property \\
\hline
recommender system & items & users & $\{0,1,2,3, 4\}$ & rating \\
\hline
topic search & authorities & hubs & $\NNn$ & hyperlinks \\
\hline
measurement & instances & quantities & $\RRr$ & outcome \\
\hline
elections & candidates & voters & $\{1,\ldots, n\}$ & preference \\
\hline
market & producers & consumers  & $\ZZz$ & deliveries \\
\hline
digital images & positions & pixels & $[0,1]$ & intensity \\
\hline
\end{tabular}
\caption{Domains of information retrieval}
\label{Table:domains}
\end{center}
\end{table}%
In all cases, the items $i$ of type $\JJJ$ are presented as tuples of features $u$ of type $\UUU$. A $\JJJ$-tuple of $\UUU$-tuples is presented as a $\JJJ\times\UUU$-matrix. A $\JJJ\times\UUU$-matrix can thus be viewed as a $\JJJ$-dimensional vector of $\UUU$-dimensional vectors, or equivalently as a $\UUU$-dimensional vector of $\JJJ$-dimensional vectors. Either way, it is an assignment $C\colon\JJJ\times \UUU\to \textsf{Val}$, written equivalently as a matrix or a linear operator mapping $\UUU$-vectors to $\JJJ$-vectors:
\bea\label{eq:C}
C& = & \begin{pmatrix}C_{11} & C_{12}& \ldots & C_{1n} \\
C_{21} & C_{22} & \ldots & C_{2n}\\
\vdots &\hdotsfor{2} & \vdots \\
C_{m1} &\hdotsfor{2} & C_{mn}
\end{pmatrix}_{\JJJ\times\UUU}  \ \ =\ \  \ \sum_{\substack{i\in \JJJ\\u\in \UUU}} \ket i \cdot \brakem i C u \cdot \bra u
\eea
where each value $\brakem i C u = C_{iu}$ quantifies the feature $u$ in the item $i$. Appendix~\ref{Appendix:inner} explains how a linear operator $C$ maps the $\UUU$-vectors, as mixtures of features, to the $\JJJ$-vectors, as mixtures of items.

Of application domains in Table~\ref{Table:domains}, text analysis is the simplest, while concept analysis and recommender systems show how the static vector semantics really works. We will get to them in a moment, after the shortest of summaries of text analysis.

\subsubsection{Counting and weighing words}
Text analysis starts from a set of documents $\JJJ$ that should be analyzed, usually called a \emph{corpus}\footnote{The plural of the Latin word ``corpus'' is ``corpora''.}. Each document $i\in\JJJ$ is viewed as a \emph{bag of words}\/ $u\in \UUU$ and represented as a vector 
\bea
\ket{C_{i}} & = & \sum_{u\in \UUU} \ket u C_{iu}
\eea
where $C_{iu}$ is the number of occurrences of the word $u$ in the document $i$. Hence the matrix $C$ again. The frequencies of a term $u$ in the corpus $\JJJ$ on one hand and in the document $i$ and the other is are the ratios
\beq\label{eq:DF-FT}
\DF_{u} = \frac{\sum_{i\in \JJJ}\lceil C\rceil_{iu}}{\#\JJJ}\qquad \qquad \qquad \qquad \FT_{iu}  =  \frac{C_{iu}}{\#\ket{C_{i}}}
\eeq
where $\#\JJJ$ is the number of documents in the corpus $\JJJ$, $\#\ket{C_{i}}=\sum_{u\in \UUU} C_{iu}$ is the number of words in the document $i$, and 
\bear
\lceil C\rceil_{iu} & = & \begin{cases} 0 &\mbox{ if } C_{iu} = 0\\ 1 & \mbox{ if } C_{iu} \gt 0
\end{cases}
\eear
is the indicator of the presence of the word $u$ in the document $i$.   The popular word weight measure \emph{TF*IDF} is the product 

$$w^{TF*IDF}_{iu}\ \ =\ \ \TF_{iu} \times \IDF_{u}$$ 

of the logarithmic versions of  \eqref{eq:DF-FT}
\beq
\IDF_{u}  =  \log_{10}\left(\frac{\#\JJJ}{\sum_{i\in \JJJ}\lceil C\rceil_{iu}}\right)\qquad\qquad \qquad\qquad
\TF_{iu} = \log_{10}\left(1+ C_{iu}\right)
\eeq
where $\IDF$ is the \emph{Inverse-Document-Frequency} and $\TF$ is called the \emph{Term-Frequency}, although it is not normalized into a frequency. The logarithms are used to mitigate the heavy biases of word distributions. If the terms were normally distributed through documents, then dividing the count $C_{iu}$ with $\DF_{u}$ would give a realistic weight measure. But the word frequencies famously obey the \emph{power law}, as demonstrated by George Zipf in his 1930s analyses. The power law is numerically close and often indistinguishable from the \emph{log-normal}\/ distribution. By replacing frequencies with their logarithms, the information retrieval systems even out the expectations and simplify calculations.

\paragraph{Words, terms, tokens?} Transforming a document into a bag-of-words is easier said than done. Ignoring the order is easy enough. But what should we count? It is not always easy to decide what is a word. Should we count commas, question marks, and colons, as words? Should we count ``Cow'', ``cow'', ``Cows'', and ``cows'' as the same word? The word ``earth'' may or may not mean the same thing as ``Earth''.  Many different approaches to such questions were proposed. One of the solutions was to call \emph{terms}\/ whatever seems of interest for counting and to count terms instead of words. The perfect solution, adopted in large language models, is to count \emph{tokens}\/ instead of words. The solution is perfect because a token is whatever you might want to count. A bit like term, but much better.

\subsection{Concept Analysis: Mining for meaning}\label{Sec:mining}

Looking at the vector representations of words as tuples of numbers might suggest that we didn't make much progress towards understanding meaning. Words denote concepts, not numbers.

The view of a word as a vector reduces its meaning to a combination of the meanings of its vector components. A pizza recipe is a tuple of numbers, provided that we know which numbers correspond to which ingredients. A pizza is a mixture of its ingredients, and the pizza vector is a mixture of the basis vectors.

The meaning of a word is a concept. A concept is a mixture of basic concepts. The vector representing a word captures its meaning as a mixture of the basis vectors corresponding to the basic concepts. --- But \textbf{\emph{if a word is a linear combination of concepts, then the concepts are linear combinations of words}}. 

\paragraph{Concept mining.} The goal of concept mining is to extract from a corpus a \emph{concept basis}: a minimal complete set of concepts. A set of concepts is complete for a corpus when all corpus words can be expressed as its linear combinations. A trivial complete set can be obtained by taking all words as concepts and letting every word refer to itself. The goal is to find a minimal complete set.

The gold vein of concept mining is that every corpus has a canonical concept base. To explain this, we step away from vectors  for a moment and present in Sec.~\ref{Sec:FCA} the idea of concept mining from sets and relations. In Sec.~\ref{Sec:LSA} we return to the vector view and present a simple algorithm for extracting a canonical concept base from any given corpus. For concreteness, all is presented in terms of a familiar application.

%

\subsubsection{Recommender systems} 
The task of a recommender system is to predict and recommend the items that Alice will like but has not yet tried. The predictions are derived by aligning Alice's and Bob's feedback about the items that they both used, and then extrapolating from Bob's feedbacks Alice's likely preferences. 
Asking Alice and Bob what they like is of little use, since they do not assign the same meanings to the same words, and the alignment would be too crude. They might both say  that they like action movies, but Alice's concept of action movie may be different from Bob's, and there may be no movies which they both like. The quest is for an objective concept basis, independent on individual semantical assignments.

\paragraph{Plato's Problem.} How can Alice and Bob communicate in natural language at all if every word means one thing for Alice and a slightly different thing for Bob? Where is the common ground of meaning where people stand together? Where do the shared concepts come from? This is what Chomsky called \emph{Plato's Problem}. Plato's answer was that they come from a ``place beyond heaven'' (\emph{topos hyperuranios}) where ideas reside before they are projected into our minds. Chomsky's version was that our semantic, concept-forming capability was innate, just like our syntactic, sentence-forming capability.
\begin{sidewaystable}
\begin{center}
\begin{minipage}{12cm}
\begin{center}
\begin{tabular}{|c||c|c|c|c|}
\hline
raw  & Interstellar & Juno & Kagemusha & Legend\\
\hline \hline 
Alice & $\star\star\star\star$ & $\star\star$ & $\star\star\star\star$ & $\star$  \\
\hline
Bob & $\star\star$ & $\star\star\star$ & $\star\star$ &  \\
\hline
Carol & $\star\star$ & $\star$ &$\star\star\star\star\star$ &  \\
 \hline
 Dave & $\star$ & $\star\star\star\star$ & $\star\star\star$ &  \\
 \hline
Ed &  & $\star\star$ & $\star\star\star\star\star$ &  $\star\star$ \\
\hline
\end{tabular}
\end{center}
\end{minipage}

\vspace{7ex}

\begin{minipage}{8cm}
\begin{center}
\begin{tabular}{|c||c|c|c|c|}
\hline
$R$  & Interstellar & Juno & Kagemusha & Legend\\
\hline \hline 
Alice & $1$ & $1$ & $1$ & $1$  \\
\hline
Bob & $1$ & $1$ & $1$ & 0  \\
\hline
Carol & $1$ & $1$ &$1$ &0  \\
 \hline
 Dave & $1$ & $1$ & $1$ &0  \\
 \hline
Ed & 0 & $1$ & $1$ &  $1$ \\
\hline
\end{tabular}
\end{center}
\end{minipage}
\hspace{8em}
%
\begin{minipage}{8cm}
\begin{center}
\begin{tabular}{|c||c|c|c|c|}
\hline
$C$  & Interstellar & Juno & Kagemusha & Legend\\
\hline \hline 
Alice & $3$ & $1$ & $3$ & $0$  \\
\hline
Bob & $1$ & $2$ & $1$ &   \\
\hline
Carol &  $1$ & $0$ &$4$ &   \\
 \hline
 Dave & $0$ & $3$ & $3$ &  \\
 \hline
Ed &  & $1$ & $4$ &  $1$ \\
\hline
\end{tabular}
\end{center}
\end{minipage}

\vspace{7ex}

\begin{minipage}{8cm}
\begin{center}
\begin{tabular}{|c||c|c|c|c|}
\hline
$L$  & Interstellar & Juno & Kagemusha & Legend\\
\hline \hline 
Alice & $1.25$ & $0.83$ & $0$ & $-0.12$  \\
\hline
Bob & $1.05$ & $1.13$ & $0.35$ &   \\
\hline
Carol &  $1.12$ & $1.02$ &$0.21$ &   \\
 \hline
 Dave & $1.57$ & $0.35$ & $-0.56$ &  \\
 \hline
Ed &  & $0.18$ & $1.02$ &  $0.98$ \\
\hline
\end{tabular}
\end{center}
\end{minipage}
\caption{The raw star rating induces the transaction relation, numeric ratings, and normalized ratings}
\label{Table:ratings}
\end{center}
\end{sidewaystable}

\paragraph{Netflix Problem} was the topic of a developer competition that ran 2006--2009. The movie rental company Netflix offered the \$1,000,000 prize for improving their recommendation algorithm by 10\%. The goal was reached by several teams around the same time, with the winning submission followed by the first contender within 20 minutes. The input of the algorithm was a $\JJJ\times\UUU$-matrix like \eqref{eq:C}, this time with $\JJJ$ denoting a set of movies (instead of documents), $\UUU$ a set if users (instead of words), and the matrix $C = \left(C_{iu}\right)_{\JJJ\times \UUU}$ of ratings given by the users $u$ to the movies $i$ (instead of the term occurrence counts). Just like the documents were reduced to bags of words and the pizzas to vectors of ingredients, the movies were reduced to tuples of ratings. A toy example is given in Tables~\ref{Table:ratings}. It displays 5 users' rating 4 movies. The integer table on the right is derived by counting the star ratings on the top, assuming that a single star is a 0 rating. The decimal table at the bottom is derived by averaging and normalizing several integer tables.  The relation table on the left assigns 1 if the user provided a rating and 0 otherwise. A similar table could be obtained by assigning 1 to, say, the ratings of 3 or more stars and 0 otherwise.  

\paragraph{Question.} \emph{What user tastes and movie styles caused these particular ratings? What are the concepts to which these context matrices refer?}

\subsubsection{Formal Concept Analysis}\label{Sec:FCA}
The Formal Concept Analysis (FCA)  approach is due to Rudolf Wille. The idea of the FCA, instantiated to the Netflix Problem, is to define 
\begin{itemize}
\item a user taste as a set of users who like the same movies; and 
\item a movie style as a set of movies liked by the same users. 
\end{itemize}
Fig.~\ref{Fig:context} displays the relation from Table~\ref{Table:ratings} on the left as a bipartite graph. 
\begin{figure}
\begin{center}
\includegraphics[width=.35\linewidth]{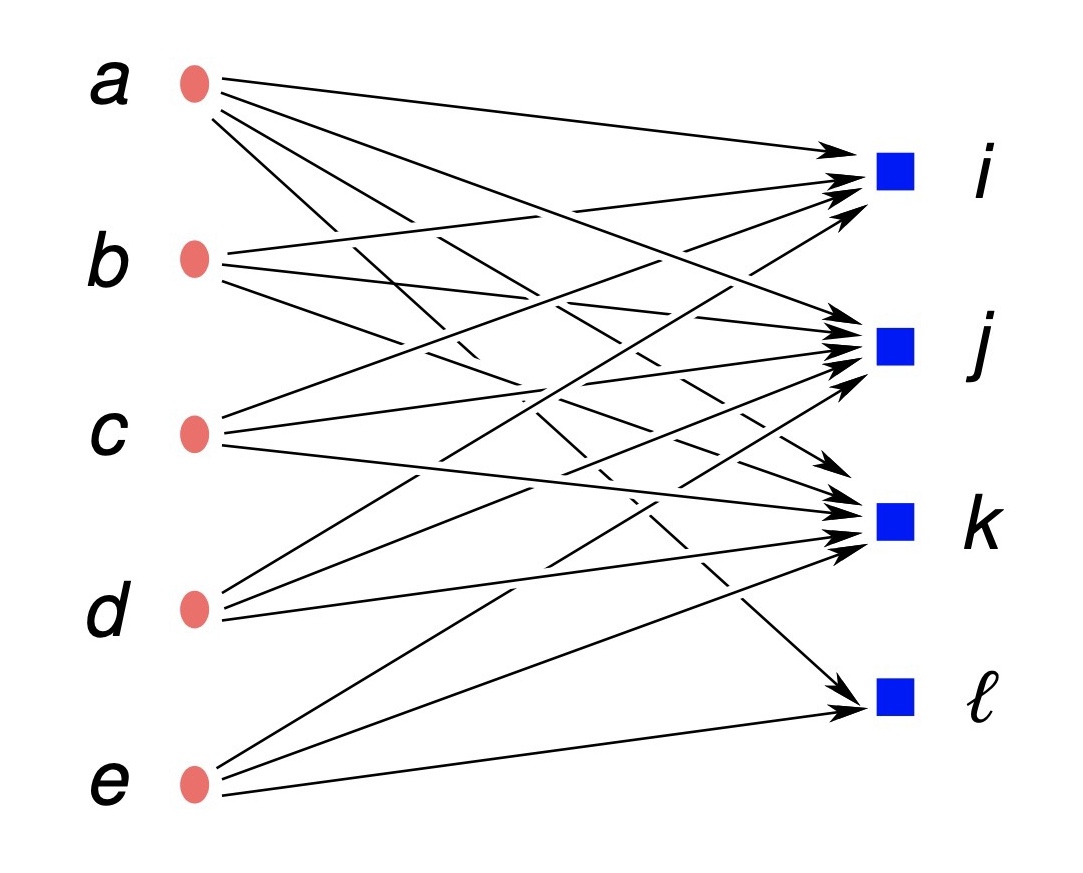}
\caption{Relational context}
\label{Fig:context}
\end{center}
\end{figure}
The arrows $u\to i$ correspond to the entries $R_{iu}=1$ of the left-hand Table~\ref{Table:ratings}. 

But a set $X\subseteq \UUU$ of all users who like the same movies and the set $Y\subseteq \JJJ$ of all movies that they all like determine each other. There is a one-to-one correspondence between the user tastes and the movie styles. The interaction relation R induced by a data matrix determines a single family of concepts, interpreted in two isomorphic ways as user tastes and movie styles.

%

\begin{figure}[!ht]
\begin{center}
\begin{minipage}{.9\linewidth}
\centering
\includegraphics[width=.3\linewidth]{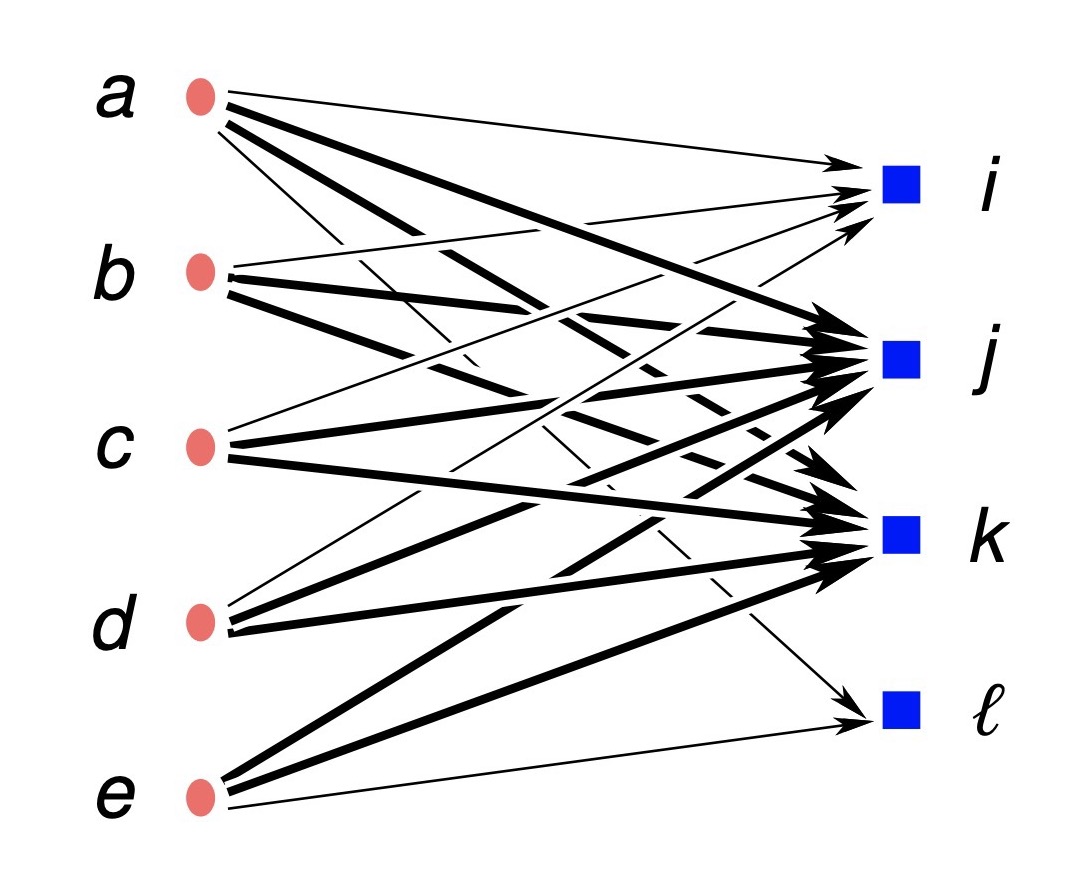}
\end{minipage}

\begin{minipage}{.9\linewidth}
\centering
\includegraphics[width=.3\linewidth]{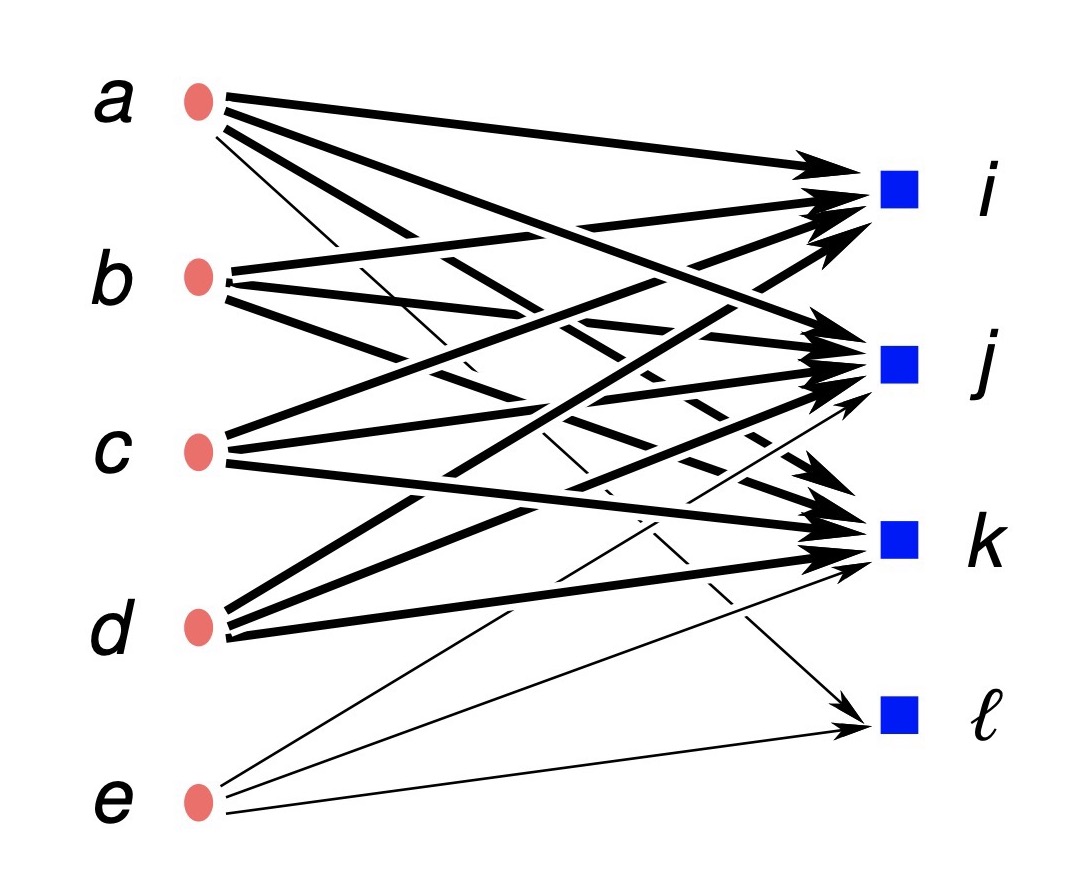}
\hspace{7em}
\includegraphics[width=.3\linewidth]{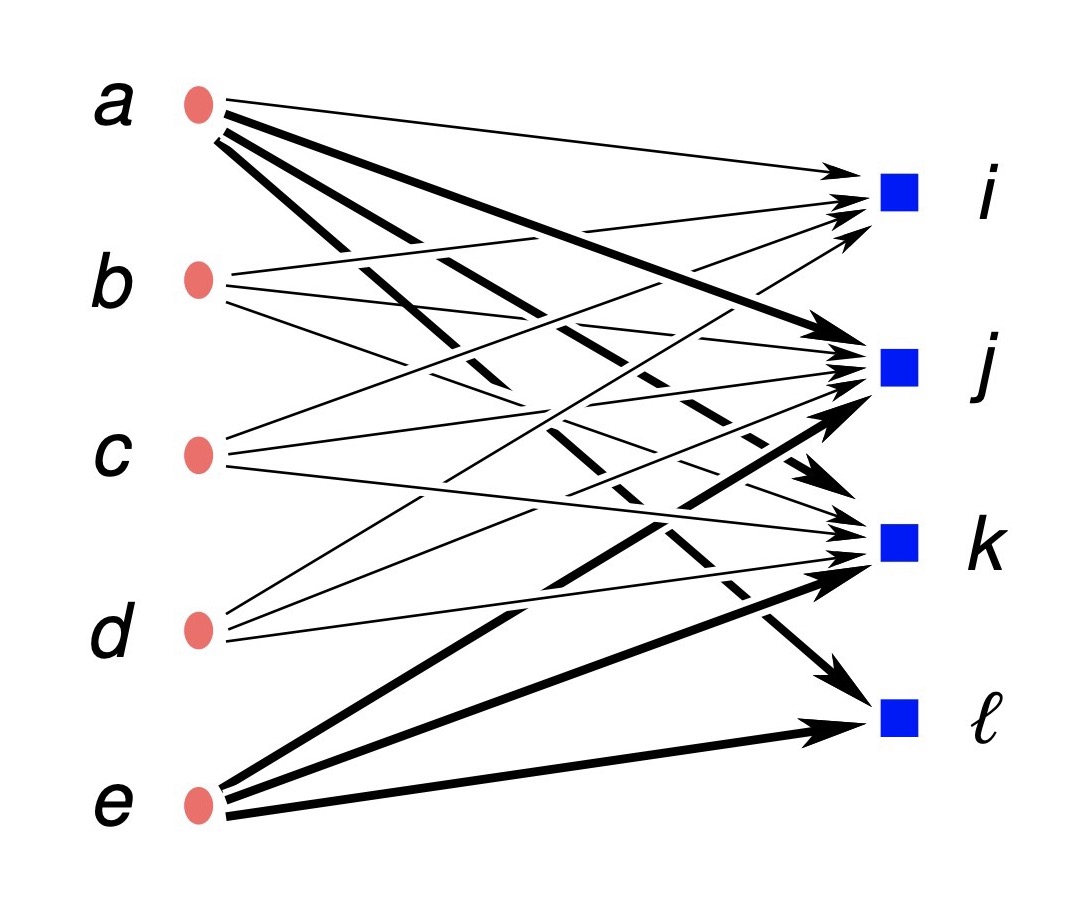}
\end{minipage}

\begin{minipage}{.9\linewidth}
\centering
\includegraphics[width=.3\linewidth]{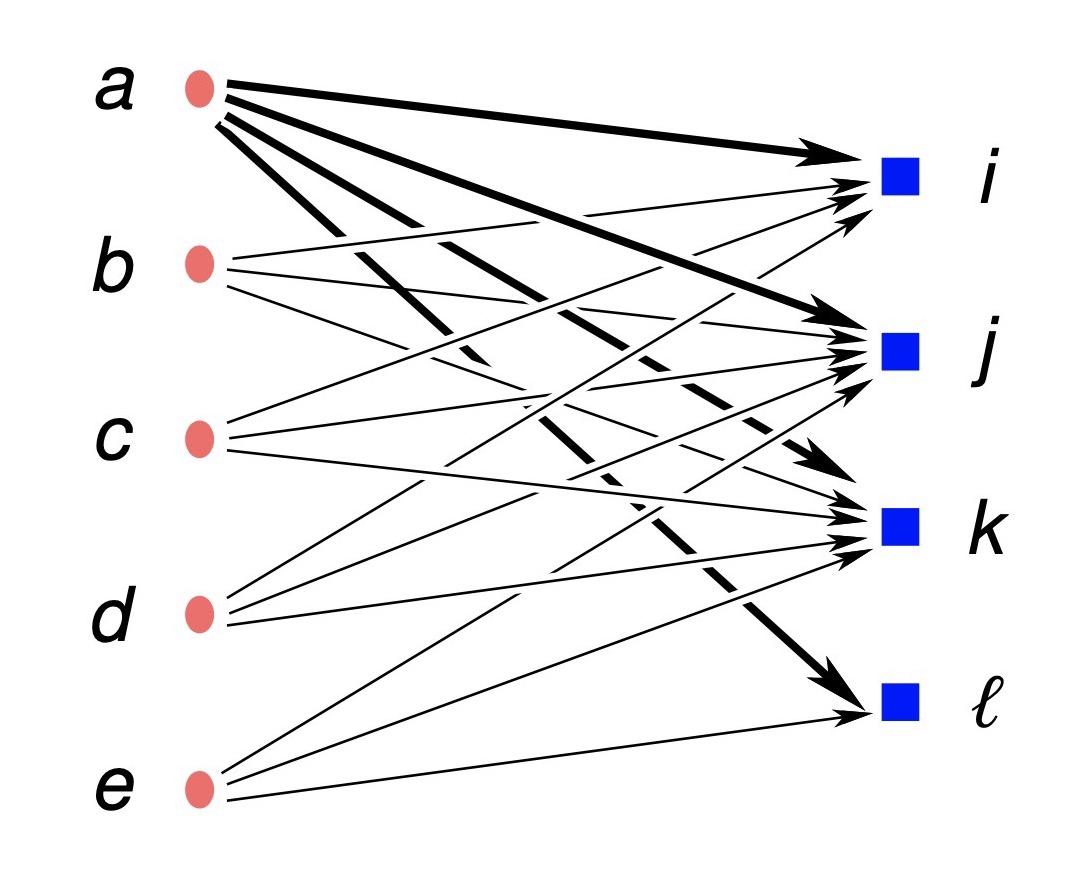}
\end{minipage}
\caption{Formal concept lattice of complete subgraphs}
\label{Fig:concept-lattice}
\end{center}
\end{figure}

A pair $<X,Y>\subseteq \UUU\times \JJJ$, where $X$ is the set of all users who like all movies in $Y$ and $Y$ is the set of all movies liked by all users in $X$, spans a complete subgraph of the bipartite graph in Fig.~\ref{Fig:context}. \emph{\textbf{The user tastes and the movie styles can be identified as the complete subgraphs of the bipartite interaction graph.}} A concept is thus in the form $\gamma=<X,Y>$, where $X$ and $Y$ are the sets of nodes spanning a complete subgraph. All complete subgraphs of Fig.~\ref{Fig:context} are displayed in Fig.~\ref{Fig:concept-lattice}.
\begin{figure}[!ht]
\begin{center}
\includegraphics[width=.35\linewidth]{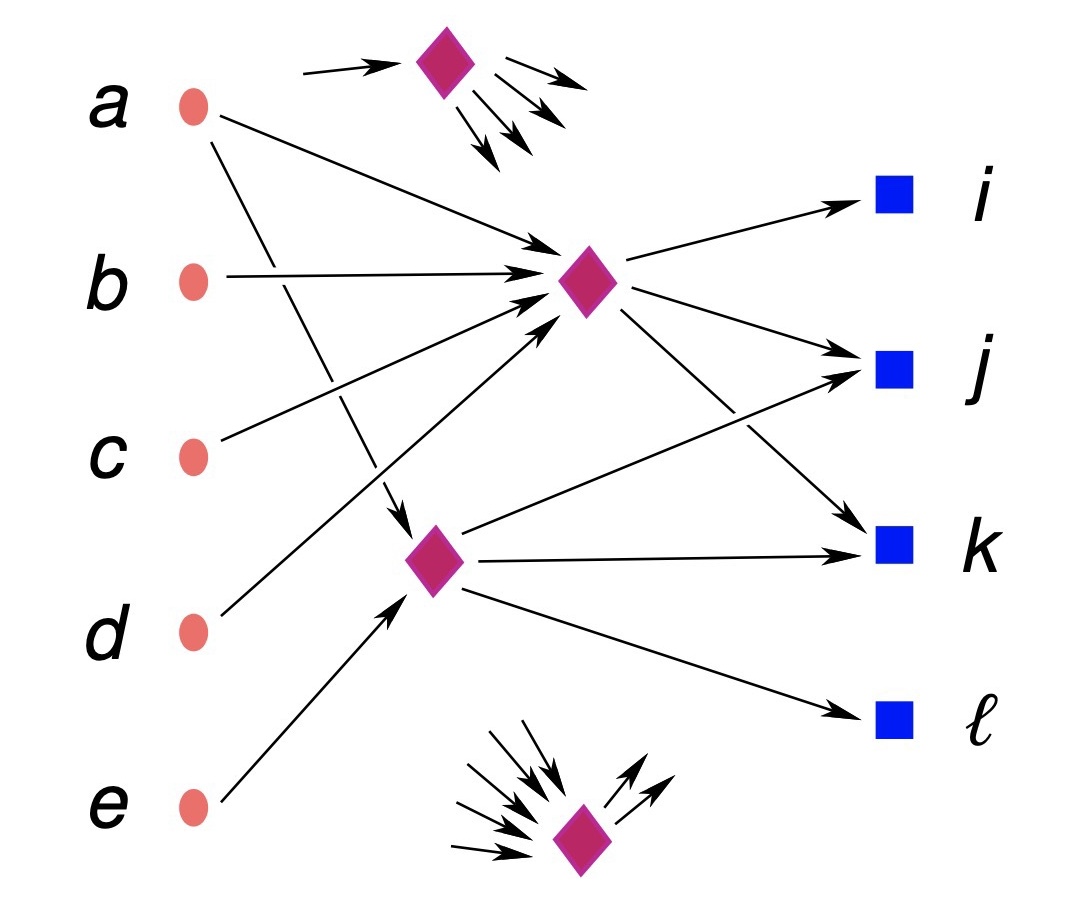}
\caption{Formal concept decomposition}
\label{Fig:ui-fac}
\end{center}
\end{figure}
The concept lattice from Fig.~\ref{Fig:concept-lattice} is displayed in Fig.~\ref{Fig:ui-fac} again. 
Each diamond in Fig.~\ref{Fig:ui-fac} corresponds to one of the four concepts $\gamma=<X,Y>$ seen in Fig.~\ref{Fig:concept-lattice}. A user-movie  couple is $R$-related if and only if there is a a concept which links them:
\bear
uRi &\iff & \exists \gamma = <X,Y>.\ u\in X\wedge Y\ni i
\eear
where we write $uRi$ for $R_{iu}=1$ in the left-hand Table~\ref{Table:ratings}. The user feedback $u\tto{\ \ R\ \ } i$ is thus  justified by a concept decomposition $u\shortrightarrow \gamma \shortrightarrow i$ relating $u$'s taste and $i$'s style.
\begin{figure}
\begin{center}
\includegraphics[height=3cm]{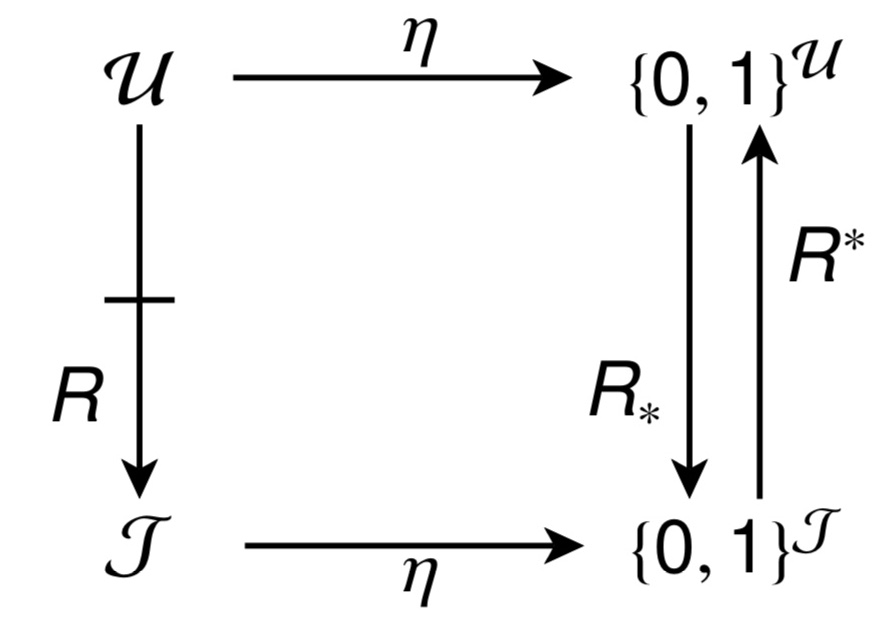} \hspace{5em} 
\includegraphics[height=3cm]{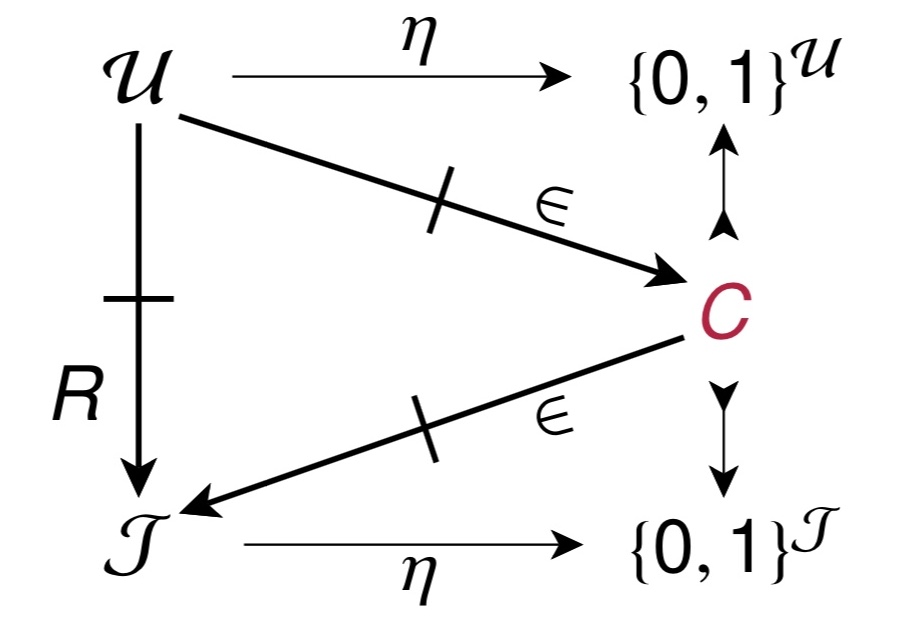}
\caption{Formal concept construction and context decomposition}
\label{Fig:formal-context}
\end{center}
\end{figure}

The formal construction of the formal concepts is summarized in Fig.~\ref{Fig:formal-context}. The same pattern will be repeated in the next section, in linear algebra. The components $X$ and $Y$ of a formal concept are obtained as the fixpoints of the closure operators $\overleftarrow R$ and $\overrightarrow R$ derived from the Galois connection $R^{\ast}\dashv R_{\ast}$ as follows
\begin{align*}
uR & = \{i\in \JJJ\ |\ uR i\} & R i & = \{u\in \UUU\ |\ uR i\}\\
R_\ast X & =   \bigcap_{u\in X} uR & 
R^\ast Y & =  \bigcap_{i\in Y} R i\\
\overleftarrow R & = R_{\ast}R^{\ast} &  \overrightarrow R & = R^{\ast}R_{\ast}\\
X & = \overleftarrow R X &  \overrightarrow RY & = Y
\end{align*}
\paragraph{Exercise.} Show that the maps $R^{\ast}\dashv R_{\ast}$ provide the anticipated one-to-one correspondence between the user tastes and the movie styles. Show that all pairs $<X,Y>\in \{0,1\}^{\UUU}\times \{0,1\}^{\JJJ}$ satisfy
\bea\label{eq:adjoints}
X = \overrightarrow R X\wedge \overleftarrow RY = Y & \iff & X = R^{\ast} Y \wedge R_{\ast} X = Y
\eea

\subsubsection{Latent Semantic Analysis}\label{Sec:LSA}
\begin{figure}
\begin{center}
\includegraphics[width=.35\linewidth]{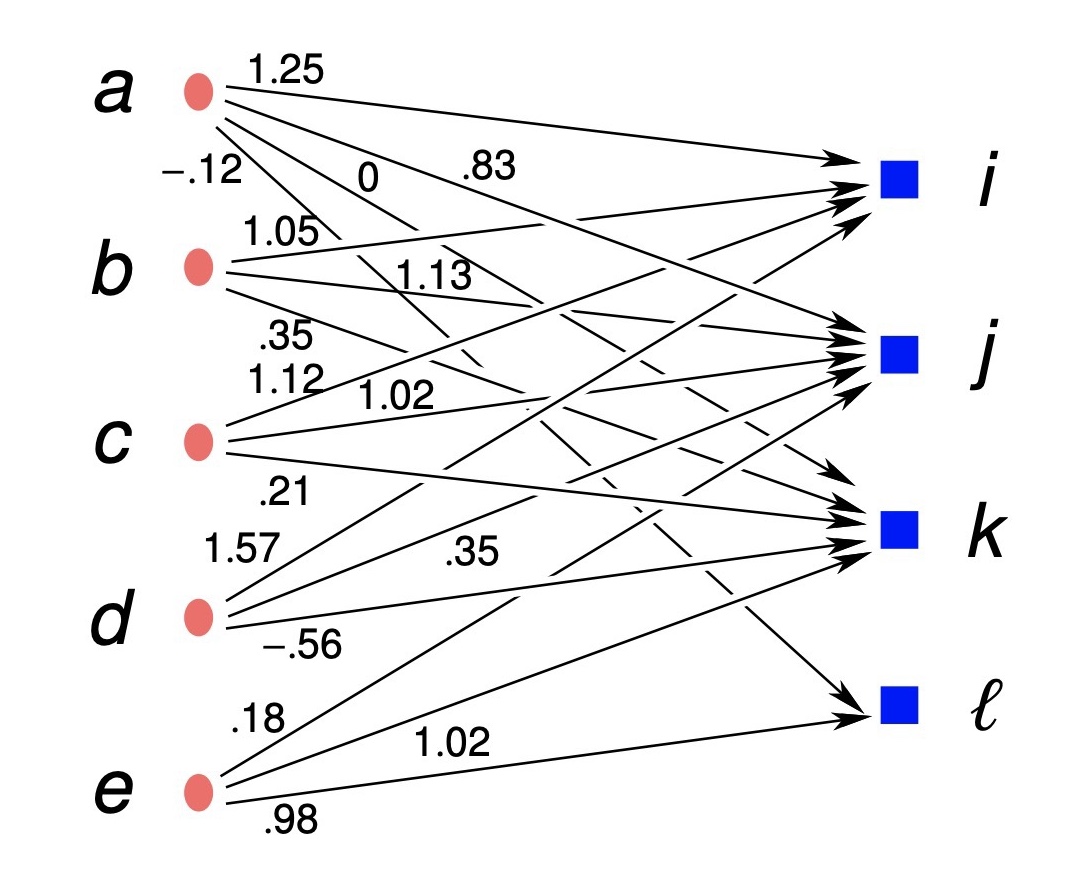}
\caption{Linear context}
\label{Fig:lin-context}
\end{center}
\end{figure}

While easy to understand and technically robust, the FCA approach has the obvious drawback that it ignores the actual ratings. The actual star ratings from the top Table~\ref{Table:ratings} are recorded in the numeric matrix on the right. This matrix is still not suitable for quantitative concept analysis because the data are biased by irrelevant user habits: e.g., some users never use the  ratings below 3 stars, whereas others never use the ratings above 3 stars. To be analyzed together, the ratings must be normalized so that first user's 3 stars are mapped to the bottom, second user's 3 stars to the top. After such preprocessing operations, the data are presented as a fractional matrix, like in the bottom Table~\ref{Table:ratings}. Latent Semantic Analysis mines concepts from such matrices. Like before, they can be viewed as bipartite graphs, this time with fractional ratings attached to edges as labels, like in Fig.~\ref{Fig:lin-context}. 

The Latent Semantic Analysis approach was introduced in a series of papers\footnote{One was called ``A Solution to Plato's Problem''.} by Deerweester, Dumais, Landauer and collaborators. The idea, instantiated to the Netflix Problem, is similar 
\begin{itemize}
\item a user taste as a linear combination of users who like the same movie style; and 
\item a movie style as a linear combination of movies liked by the user tastes. 
\end{itemize}
\begin{figure}[!hb]
\begin{center}
\includegraphics[height=3cm]{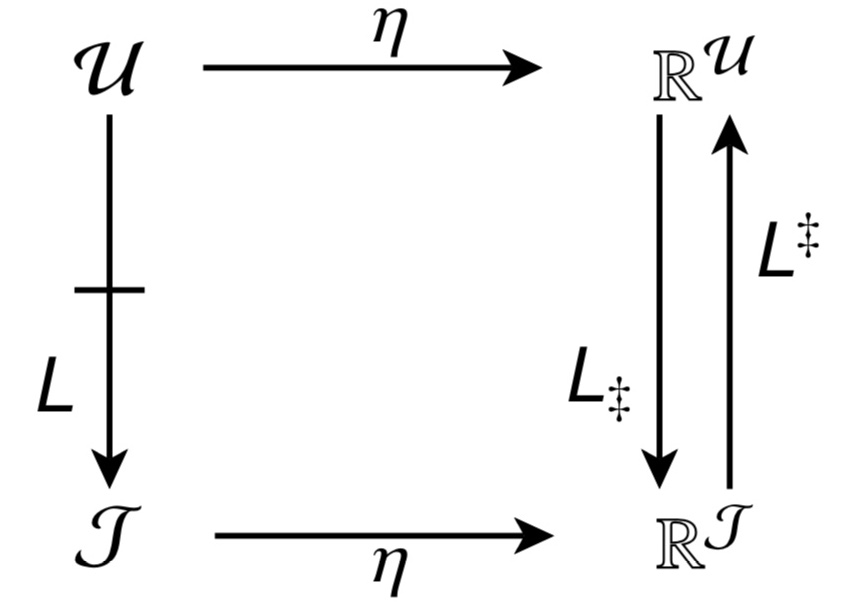} \hspace{5em} 
\includegraphics[height=3cm]{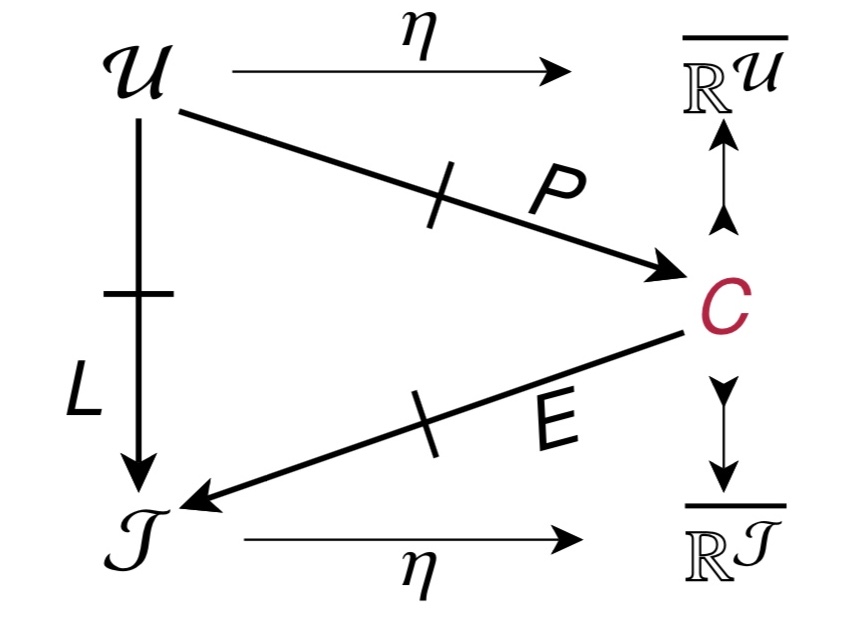}
\caption{Latent concept construction and context decomposition}
\label{Fig:latent-concept}
\end{center}
\end{figure}
There is again a one-to-one correspondence between the user tastes and the movie styles, this time established by the linear operators induced by the matrix. But while the Galois connection $R^{\ast}\dashv R_{\ast}$ was an antitone adjunction between sets of users and sets of movies, now we have a linear adjunction $L^{\ast}\dashv L_{\ast}$, displayed in Fig.~\ref{Fig:latent-concept} on the left, between the linear mixtures of users and the linear mixtures of movies:
\begin{align*}
L_\ddag \ket u & =   \sum_{i\in \JJJ} L_{ui}\bra i & 
L^{\ddag} \bra i & =  \sum_{u\in \UUU} \ket u L_{ui}
\\
\overleftarrow L & = L_{\ddag}L^{\ddag} &  \overrightarrow L & = L^{\ddag}L_{\ddag}
\\
\ket x \lambda_{x}  &= \overrightarrow L \ket x & \overleftarrow L\bra y & = \lambda_{y}\bra y\\
\ket x \sqrt{\lambda}  & = L^{\ddag} \bra y & L_{\ddag}\ket x &= \sqrt{\lambda}\bra y
\end{align*}
%
%
%
%
\begin{figure}[!hb]
\begin{center}
\includegraphics[width=.35\linewidth]{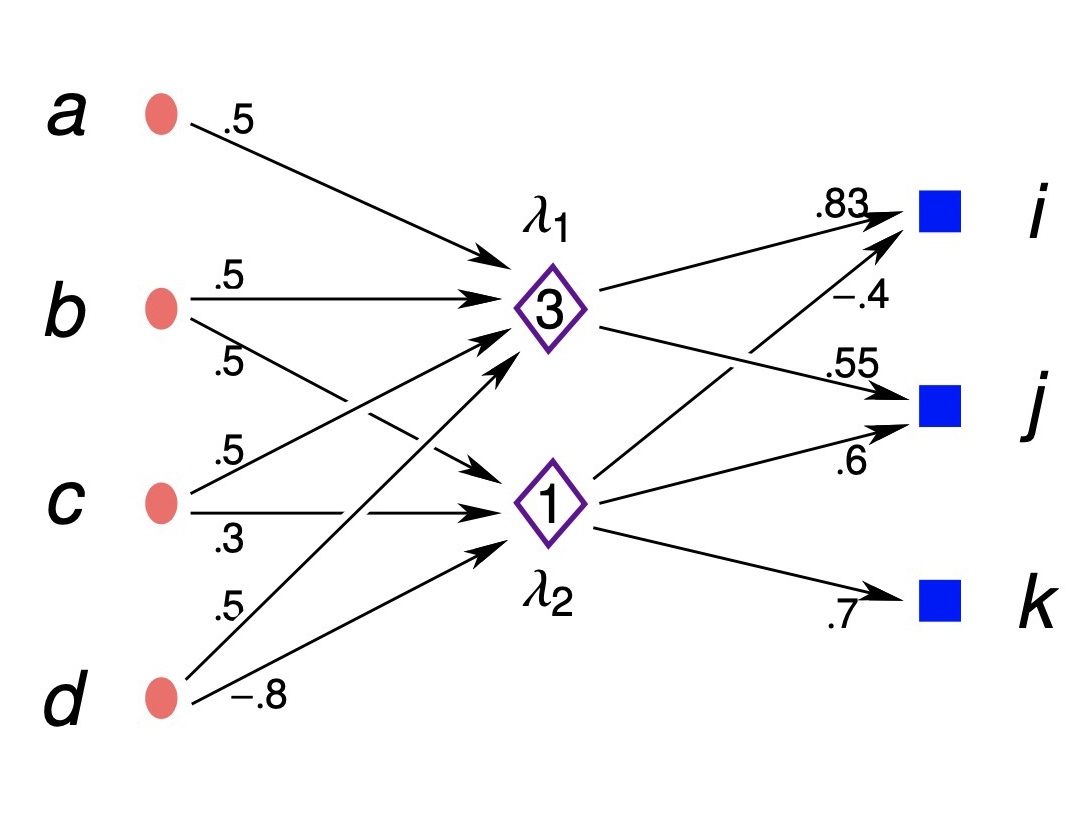}
\caption{Latent concept decomposition}
\label{Fig:lsa-fac}
\end{center}
\end{figure}
Recalling that the goal of the recommender systems is to predict which movies a user might like, i.e. to extrapolate the missing ratings in the sparse matrix $L$ from the given ratings, the concepts are mined from the complete (nonsparse) minors of $L$. Applying the above approach to the upper-left $3\times 4$-minor of $L$ in Table~\ref{Table:ratings} gives its singular value decomposition
\bea\label{eq:SVD}
 \begin{pmatrix}
1.25 & 1.05 & 1.12 & 1.57\\
0.83 & 1.13 & 1.02 & 0.35 \\
0 & 0.35 & 0.21 & -0.56
\end{pmatrix} &=&
\begin{pmatrix}
0.83 & -0.4\\
0.55 & 0.6  \\
0 & 0.7 
\end{pmatrix}
\cdot 
\begin{pmatrix}
3 & 0\\
0 & 1
\end{pmatrix}
\cdot
\begin{pmatrix}
0.5 & 0.5 & 0.5 & 0.5 \\
0 & 0.5 & 0.3 & -0.8 
\end{pmatrix}
\eea
also displayed in Fig.~\ref{Fig:lsa-fac} as a bipartite graph factoring. The linear context $L$ thus determines two latent concepts, with the respective singular values 3 and 1. The context decomposition \eqref{eq:SVD} is illustrated in the right-hand diagram of Fig.~\ref{Fig:latent-concept}. To capture the concept space, the vector spaces $\RRr^{\UUU}$ and $\RRr^{\JJJ}$ are retracted to their respective unit spheres\footnote{A space is projected to its unit sphere without the origin.} $\overline{\RRr^{\UUU}}$ and $\overline{\RRr^{\JJJ}}$, to take into account only the angles between concepts, and factor out their strengths. The strengths, expressed as vector lengths, and in particular the eigenvalues, expressing the strength of the basic concepts, will become relevant in dynamic semantics, and play a crucial  role in directing attention.

\paragraph{Intuition.} If you have a moment, it may be worth spending a thought on the idea of \textbf{\emph{concepts-as-eigenspaces}}. It is better if you think on your own, but here is how I think of it. The linear operator induced by the rating matrix maps every vector of users to a vector of movies. Each user's ratings determine a mixture of movies. A linear mixture of users determines a linear mixture of movies. A user vector to a movie vector. The same rating matrix also induces a linear operator other way around, in the same way. Each movie is mapped to a mixture of users, according to their ratings, and a linear mixture of movies goes to a linear mixture of users. A movie vector to a user vector. The composite of the two linear operators maps a user vector to a movie vector and back to a user vector. The eigenspace of a hermitian is a space where each vector is mapped into itself, multiplied by the corresponding eigenvalue. A \emph{taste concept}\/ is thus a mixture of users which is mapped to a mixture of movies, a corresponding \emph{style concept}\/? --- ?and back to itself. It is invariant under the linear transformations. Such eigenspaces form a canonical basis of the vector space of users on one hand and of movies on the other. Every user is a unique mixture of the basic taste concepts; every movie is a unique mixture of the basic style concepts. Concepts are invariants. The solution of Plato's problem.

\subsubsection{Particles and waves of meaning}
While the Formal Concept Analysis relatess a user and an item through \emph{some}\/ concept, the Latent Semantic Analysis measures how related they are by adding up their relationships through \emph{all}\/ concepts. If a user and an item are thought of as nodes in a road network, then 
\begin{itemize}
\item the FCA views the traffic from the standpoint of a driver, picks a route, usually one of many, and follows it, whereas 
\item the LSA views the traffic from the standpoint of an urban designer or an engineer, and measures the capacity of all routes to transmit the waves of traffic.\end{itemize}
Meaning also seems to travel in particles and in waves:
\begin{center}
\includegraphics[height=4cm]{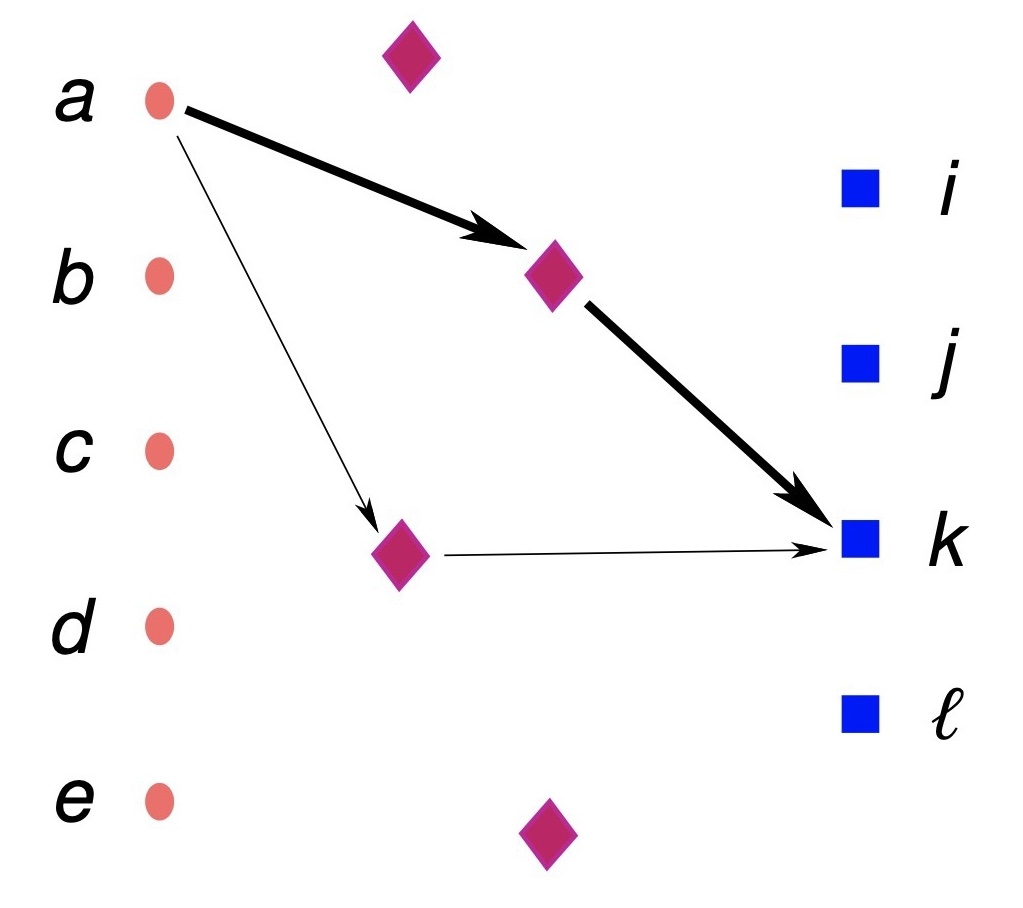}
\hspace{6em}
\includegraphics[height=4cm]{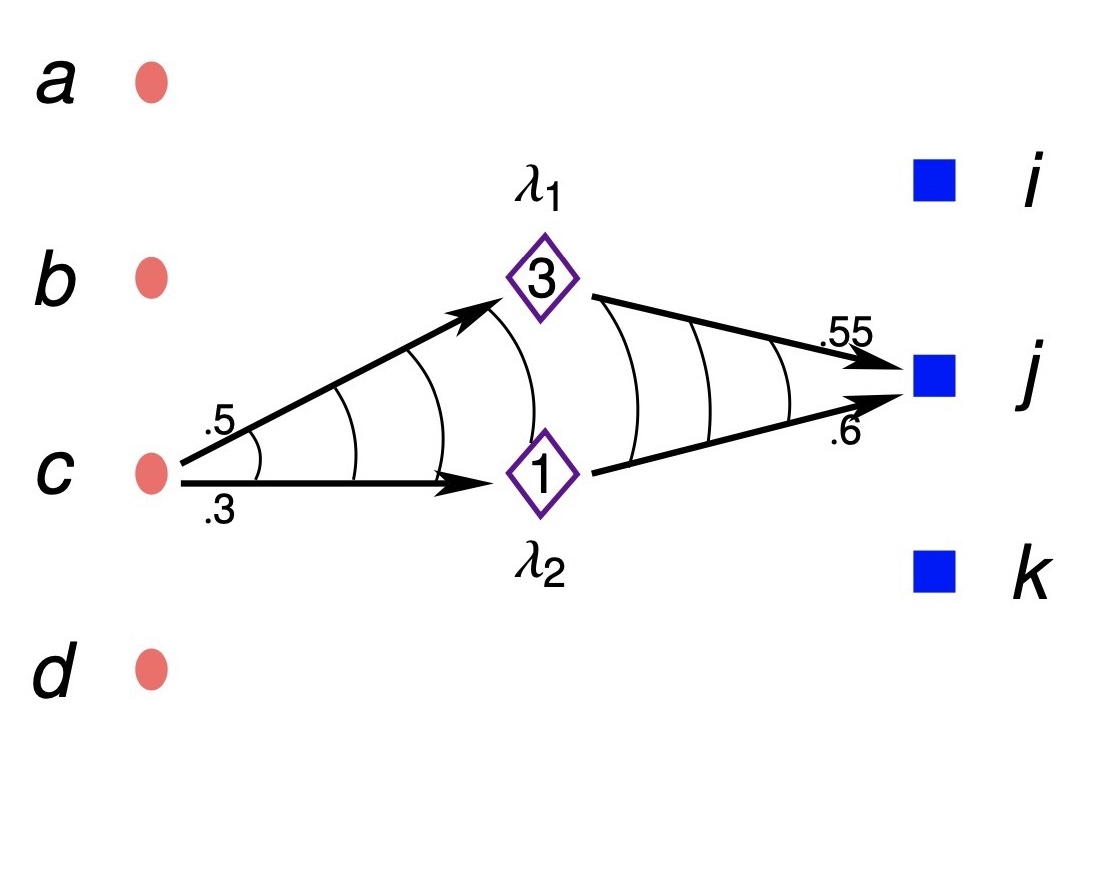}
\end{center}
%

\section{Dynamic semantics: language production}
\label{Sec:spacetime}

The moment of truth. So far, we pursued static semantics as assignment of words to objects, reduced to points in space. Word meanings were presented as relations between vectors. But meaning is not an assignment of words to objects. Chatbots say that. Fig.~\ref{Fig:context}\footnote{Authored by DALL-E and Ren\'e Magritte. DALL-E was prompted to attach labels in the style Ren\'e Magritte.} says that. \begin{figure}[!hb]
\begin{center}
\includegraphics[height=4.5cm]{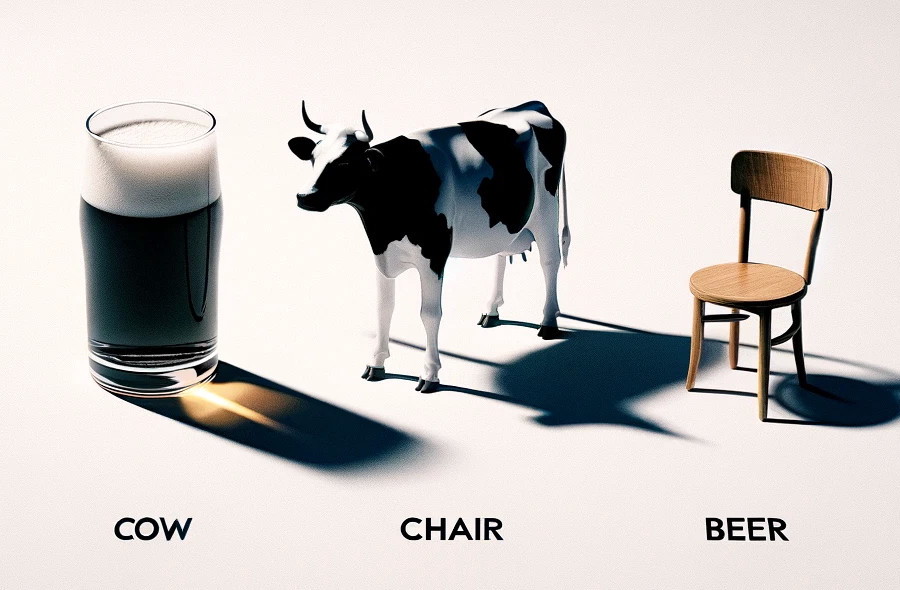}
\hspace{3em}
\includegraphics[height=4.5cm]{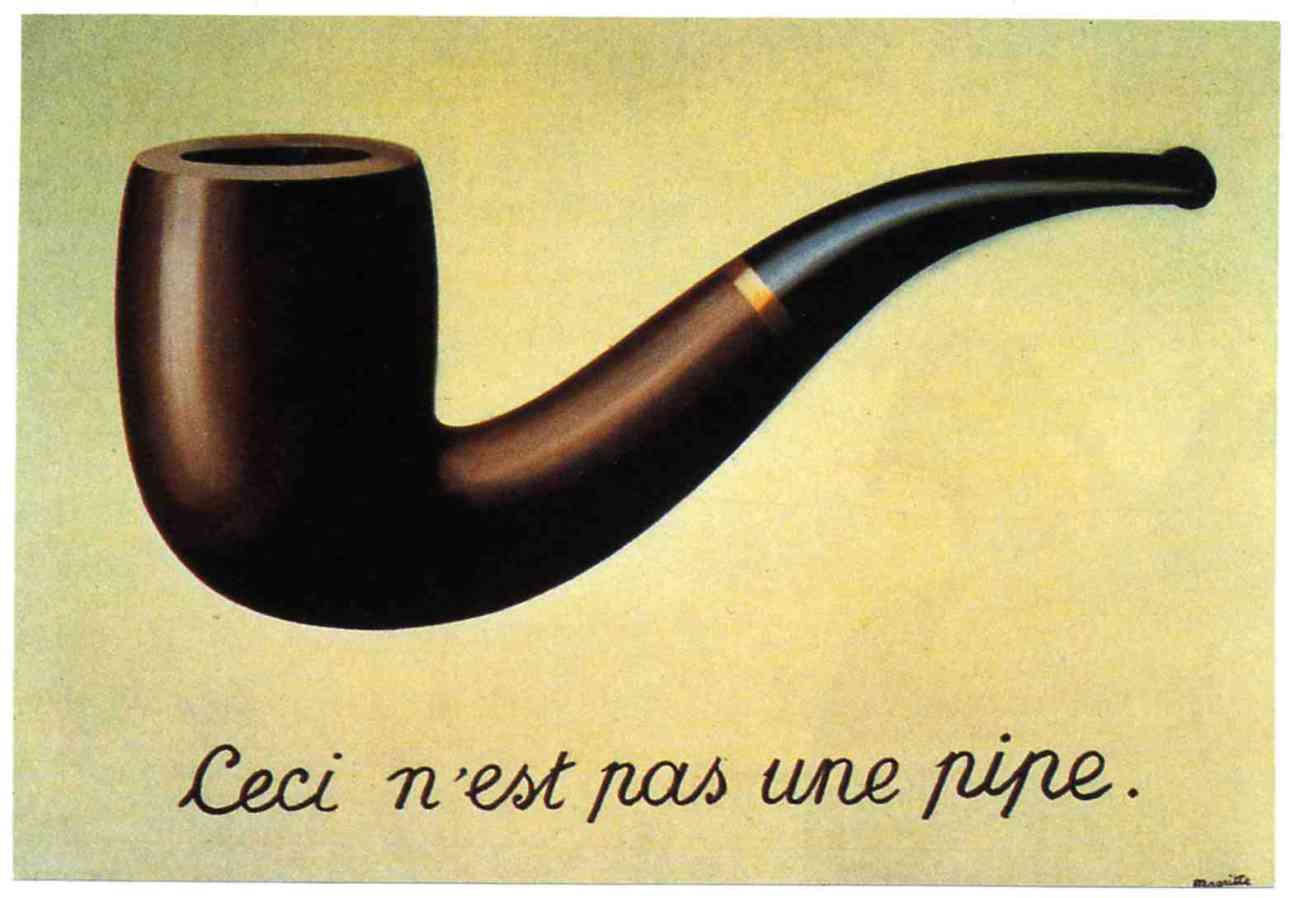}
\caption{This is not meaning, even if a cow is a cow and a pipe a pipe}
\label{Fig:context}
\end{center}
\end{figure}
It is a process where words assign meaning to each other, objects to each other, words to objects, objects to words, everything to everything. We discussed this in the Introduction (``Who are chatbots?''). Meaning is not a hammer that nails signs to things, but a process that makes everything into a sign. But making everything into anything sounds a little chaotic, doesn't it? Let us try to make sense of it.

The most useful aspect of static semantics is that it provides the vectors that carry the dynamics of dynamic semantics. Since meaning is a process, dynamics is the essence  of semantics. The notion of ``static semantics'' is almost a contradiction in terms. Language as the process of meaning, communication, information transmission, is the paradigm of dynamics. Just like the spaces of physical phenomena are metric and the spaces of social phenomena are networks, \emph{\textbf{the space of languages is the time}}. If  ``history is one damn thing after another'', language is one word after another, one sentence after another, one context after another: the history of meaning.

Language originates from communication and the theory of communication is first of all a theory of channels. Dynamic semantics is clearly concerned with channels but they are barely mentioned in linguistics. Why is that? It turns out that the theory of language and the theory of channels met early on, and parted in confusion. It's a funny story, and may be useful later, so here it goes.

\subsection{Language production as predictive inference}

What is the next word that I am going to \ldots?

\subsubsection{What was the next word that I was going to say?} 

If the possible answers to such questions are written as \emph{conditionals}
\bear
\mbox{what is the next word that I am going to}&\mbox{\large$\vdash$} &\mbox{write}\\ 
\mbox{what is the next word that I am going to}&\mbox{\large$\vdash$} &\mbox{say} \\
\mbox{what is the next word that I am going to}&\mbox{\large$\vdash$} &\mbox{swallow} 
\eear
then their respective \emph{conditional probabilities}\/ can be written
\bear
\pderr{what is the next word that I am going to}{write} & = & {\textstyle \frac 1 8}\\
\pderr{what is the next word that I am going to}{say} & = & {\textstyle \frac 1 2}\\
\pderr{what is the next word that I am going to}{swallow} & = & {\textstyle \frac 1 {30}}
\eear
In the usual notation, the first line would be written 
\bear \Pr\left({\mbox{write}\, |\, \mbox{what is the next word that I am going to}}\right) & = & \textstyle \frac 1 8
\eear 
and the other two similarly. Here we won't write them like that. Changing standard notations is seldom a good idea, but they are seldom this bad.

\paragraph{Frequencies and conditional probabilities.} Conditional probability is defined as the ratio
\bear
\pderr{\small what is the next word that I am going to}{say} & = & \frac{\ppderr{\small what is the next word that I am going to say}}{\ppderr{\small what is the next word that I am going to}}
\eear
of the frequencies
\bear
\ppderr{\small what is the next word that I am going to say}& = & \frac{\#\left\{\mbox{\small what is the next word that I am going to say}\right\}}{\#D}\\
\ppderr{\small what is the next word that I am going to}& = & \frac{\#\left\{\mbox{\small what is the next word that I am going to}\right\}}{\#D} 
\eear 
where $\#D$ is the size of the document and $\#\left\{\mbox{blah}\right\}$ is the number of occurrences of the phrase ``blah'' in it. The probabilities of what I am saying then boil down to the products of probabilities of what is the next thing that I was going to say:
\bear
\ppderr{what is} & = & \ppderr{what}\cdot \pderr{what}{is}\\
\ppderr{what is the} & = &  \ppderr{what}\cdot \pderr{what}{is}\cdot \pderr{what is}{the}\\
\ppderr{what is the next} & = &  \ppderr{what}\cdot \pderr{what}{is}\cdot \pderr{what is}{the} \cdot \pderr{what is the}{next}
\eear
and so on. To see why this is true, we need a couple of generalities about chances, which may not be easy for everyone to remember (since they go back to the XVIII century) but they are certainly easy to understand if you give them a chance.

\subsubsection{Probabilistic generalities}\label{Sec:generalities}
We just need the general definition of conditional probability and one property of the conditional probability. We are given a family of observables $S$, a space $\Omega = \{a,b,c,\ldots \subseteq S\}$ of events, and a frequency or probability distribution $[ -] \colon \Omega \to [0,1]$. An event is a measurable sets of observables, and the frequency distribution tells how likely it is that an event will happen. The basic properties of events and their distributions listed in Appendix~\ref{Appendix:prob}

\paragraph{Conditional probability} for $a\in \Omega$ is the function $\pder a - \colon \Omega \to [0,1]$ defined\footnote{The usual definition leaves conditional probability undefined when $\ppder a = 0$. The assumption $\pder{\emptyset} b = 1$ means that after  the impossible event $\emptyset$, every event $b$ is almost certain. This may be questionable intuitively, but it preempts many irrelevant side-conditions.}
\bea\label{eq:cond-prob}
\pder a b & = & \begin{cases}\  \frac{\displaystyle \ppder{ab}}{\displaystyle \ppder a} & \mbox{ if } \ppder a \gt 0\\[2ex]
\ \ \  1 & \mbox{ if } \ppder a = 0
\end{cases}
\eea
where $\ppder{ab}$ abbreviates $\ppder {a\cap b}$, as we do whenever confusion seems unlikely. 

\paragraph{Transitivity.} It follows directly from \eqref{eq:cond-prob} that
\bea\label{eq:trans}
\pder a b \cdot \pder{ab} c & = & \pder a {bc}
\eea
Setting  $a=S$ gives the \emph{probabilistic modus ponens} as a special case:
\bea\label{eq:PMP}
\ppder a \cdot \pder a b & = &  \ppder{ab}\eea

\paragraph{Exercises and terminology.} Use (\ref{eq:cond-prob}--\ref{eq:trans}) to verify:
\begin{enumerate}[a)]
\item \emph{Bayes' theorem} 
\bea
\pder b a  & = & \displaystyle \frac{\ppder a \cdot \pder a b}{\ppder b}
\eea
\item The following conditions are equivalent, and make $a$ and $b$ \emph{independent}
\beq \label{eq:indep}
\pder a b = \ppder b \ \iff\  \ppder{ab} = \ppder a \cdot \ppder b\  \iff \  \pder b a = \ppder a
\eeq
\item If $a$ and $b$ are independent and $a$ and $bc$ are independent, then 
\bea
\pder a b\cdot \pder b c & \leq & \pder a c
\eea
\end{enumerate}

\subsubsection{Word chain predictions}
Getting back to the next thing that I was going to say, the probabilistic modus ponens and the transitivity say that the chance that I was going to say precisely that it is the product of the chance that I start the way I started and the conditional probabilities that I go on the way I went on:
\bea
&& \ppderr{what is the next word that I am going to say}\ = \notag\\
 && \ppderr{what}\cdot \notag\\
 && \pderr{what}{is}\cdot \notag\\
 && \pderr{what is}{the}\cdot\notag\\
&& \pderr{what is the}{next} \cdot\label{eq:next}\\
&& \hspace{5em}\vdots\notag\\
&& \pderr{what is the next word that I am}{going} \cdot\notag\\
&& \pderr{what is the next word that I am going}{to} \cdot\notag\\
&& \pderr{what is the next word that I am going to}{say}\notag
\eea
More generally, the chance of a sequence of, say, 4 arbitrary events is
\bear
\ppder{a_{1}a_{2}a_{3}a_{4}} & \stackrel{\eqref{eq:PMP}}{=} & \ppder{a_{1}}\cdot \pder{a_{1}}{a_{2}a_{3}a_{4}}
\\ & \stackrel{\eqref{eq:trans}}{=} & \ppder{a_{1}}\cdot \pder{a_{1}}{a_{2}}\cdot\pder{a_{1}a_{2}}{a_{3}a_{4}}
\\
& \stackrel{\eqref{eq:trans}}{=} & \ppder{a_{1}}\cdot \pder{a_{1}}{a_{2}}\cdot \pder{a_{1}a_{2}}{a_{3}}\cdot
\pder{a_{1}a_{2}a_{3}}{a_{4}}
\eear
%
\textbf{\emph{This is the}\/ general rule \emph{of text generation}}. Let us state it in full generality, and in more flexible notation.

\paragraph{Chain rule.} The chance that a chain of events $a^{N} = a_{1}a_{2}\cdots a_{N}$ will occur is \bea\label{eq:chain}
\ppder{a^{N}} & = & \prod_{m=1}^{N} \pder{a^{m-1}}{a_{m}}
\eea
where $a^0 = S$ and $a^{m+1}  =  a_{1}\cdots a_{m+1}$.

\bear
\ppder{a^{N}} & = & \prod_{m=1}^{N} \pder{a^{m-1}}{a_{m}}\qquad \mbox{where}\quad  a^0 = S\quad  \mbox{ and }\quad a^{m+1}  =  a_{1}\cdots a_{m+1}
\eear

\subsubsection{$N$-grams}
Although the chain-rule is nice and simple, the context conditions get long when the phrase is long and lots of probabilities need to be multiplied. On the other hand, the impact of the words far down the context is much smaller than the impact of the words closer to the one that needs to be predicted\footnote{There are exceptions. We saw one at the end the Syntax part, where the word ``barg'' at the beginning of the paragraph determined how a key could be tipped into a lock by a foot at the end of the paragraph. A sentence at the beginning of a novel may be a key of the turn of events 800 pages later. Remote semantical impacts play a crucial role in large language models.}.  So truncating the context makes the prediction a little less precise but much simpler. 

A context of length $N$ is called an \emph{$N$-gram.} Truncating the context in \eqref{eq:next} to $2$-grams gives the generation process
\bea
&& \ppderr{what is the next word that I am going to say}\ \approx \notag\\
 && \ppderr{what is }\cdot \notag\\
 && \pderr{what is}{the}\cdot \notag\\
 && \phantom{what }\pderr{is the}{next}\cdot\notag\\
 && \phantom{what is } \pderr{the next}{word} \cdot\notag\\
&& \phantom{what is the } \pderr{next word}{that} \cdot\label{eq:next-n}\\
&& \hspace{10em}\ddots\notag\\
&& \phantom{what is the next word that } \pderr{I am}{going} \cdot\notag\\
&& \phantom{what is the next word that I } \pderr{am going}{to} \cdot\notag\\
&& \phantom{what is the next word that I am } \pderr{going to}{say}\notag
\eea
Note the approximation sumbol $\approx$. Since the remote contexts are ignored, the chance of the phrase on the left is only approximated by the product on the right, and not equal as it would be if the full chain rule was used.

\subsubsection{Origins of information theory}
\begin{figure}[!hb]
\begin{center}
\includegraphics[height=9cm]{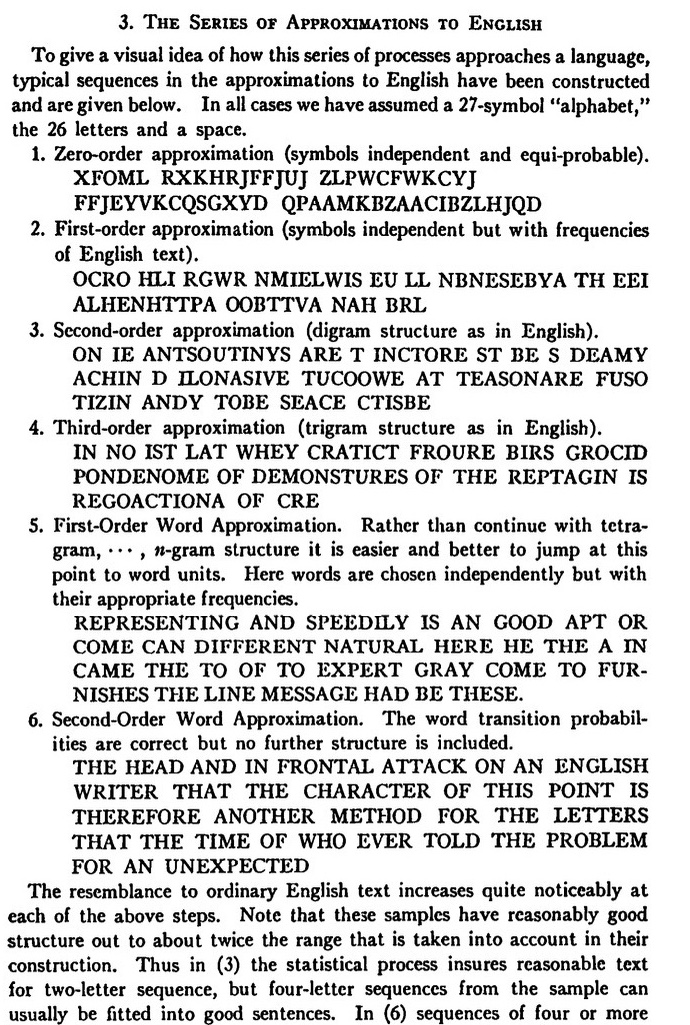}
\hspace{4em}
\includegraphics[height=9cm]{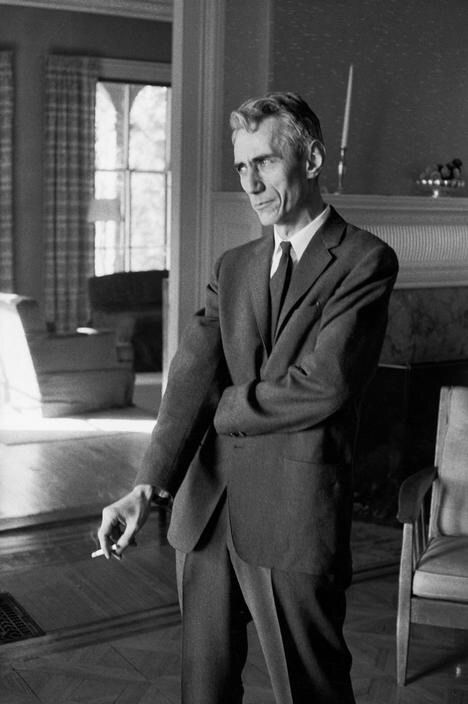}
\caption{Shannon and the page with his $N$-grams}
\label{Fig:shannon}
\end{center}
\end{figure}
The $N$-gram approximations were discovered by Claude Shannon and described in his 1948 paper on ``A Mathematical Theory of Communication''. This paper laid the foundations of information theory. The page with Shannon's $N$-grams is in Fig.~\ref{Fig:shannon}. The demonstration how increasing the $N$ generates text increasingly resembling English left a deep impression. It also brought to the surface a striking semantical phenomenon: \textbf{\emph{The more likely phrases are more likely to appear meaningful}}.

\begin{figure}[!ht]
\begin{center}
\includegraphics[height=9cm]{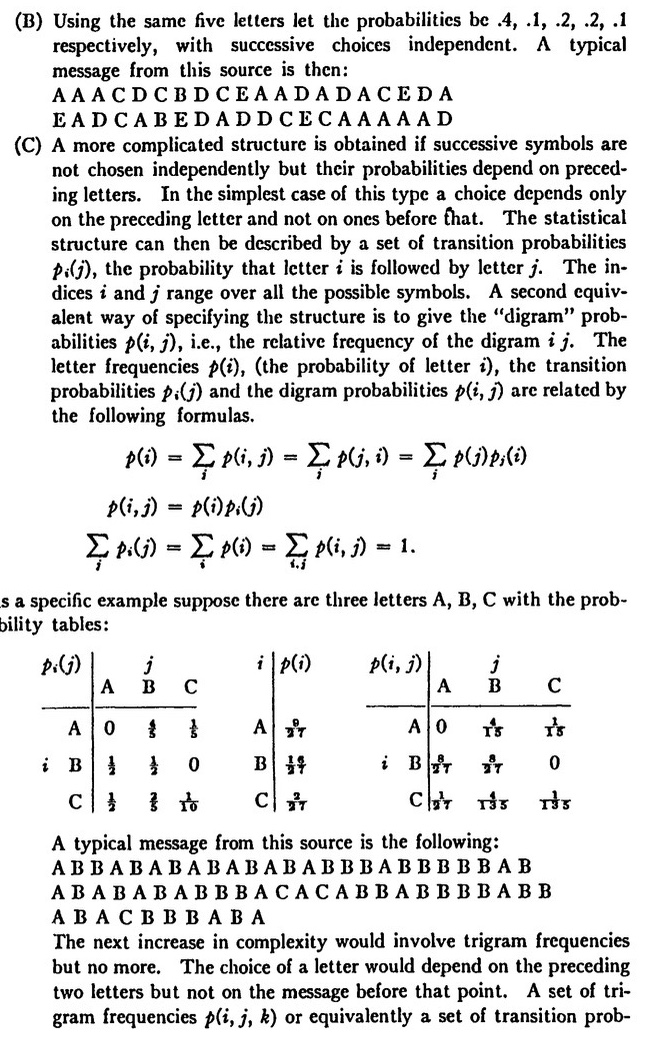}
\hspace{4em}
\includegraphics[height=9cm]{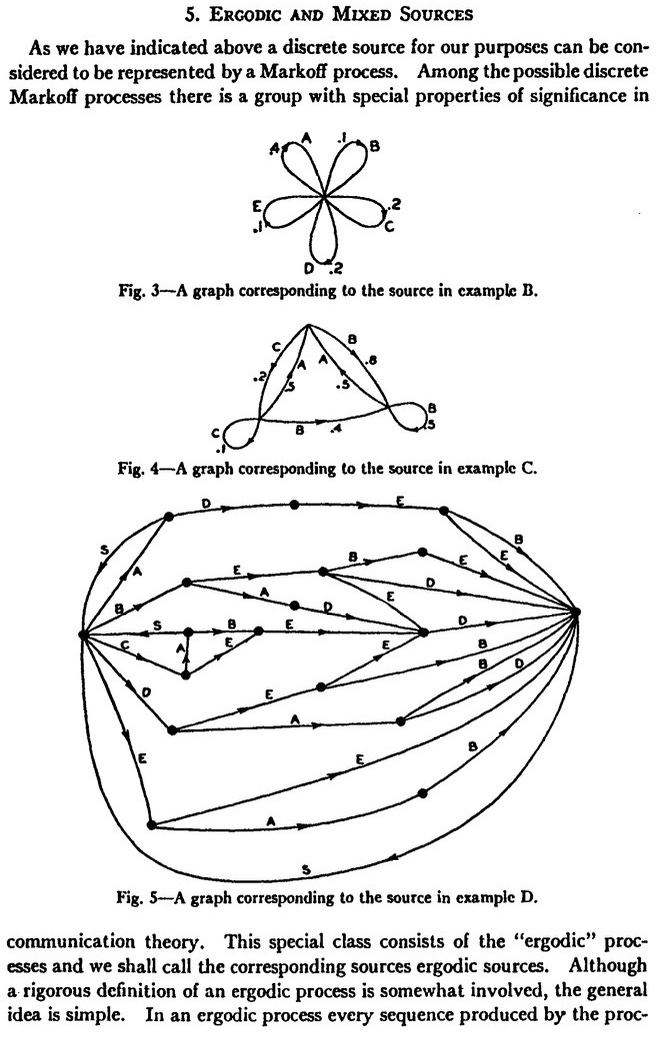}
\caption{Shannon's Markov chains}
\label{Fig:shannon-markov}
\end{center}
\end{figure}

\begin{figure}[!ht]
\begin{center}
\includegraphics[height=9cm]{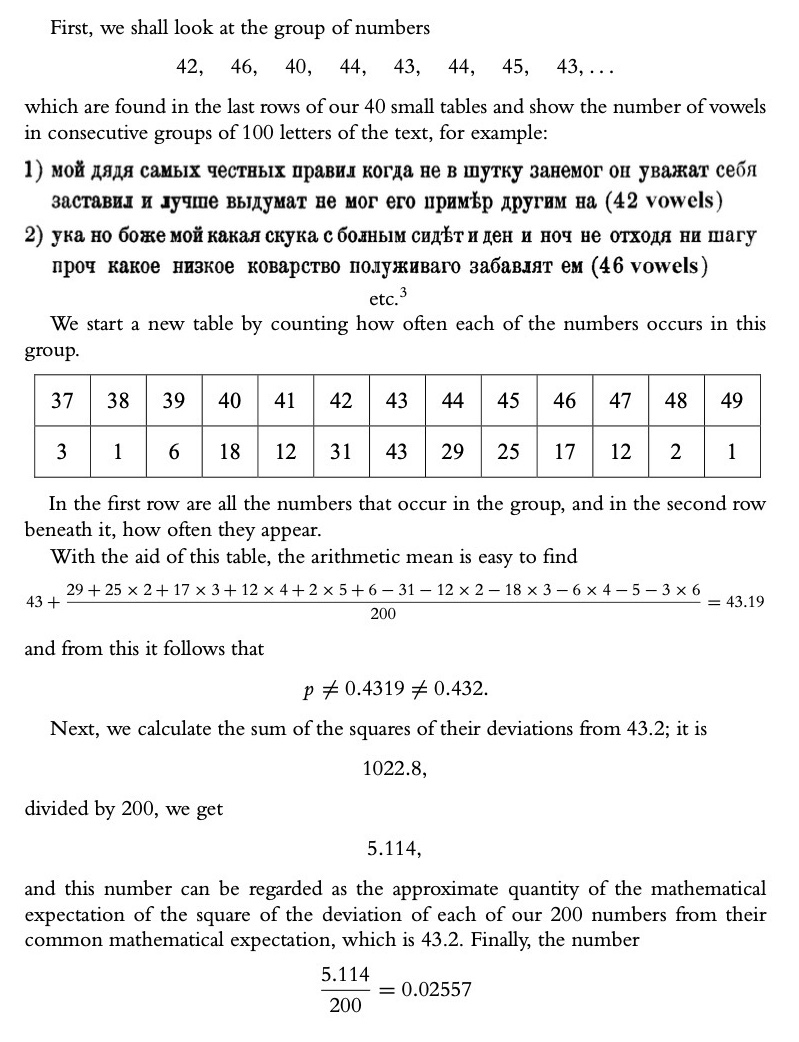}
\hspace{4em}
\includegraphics[height=9cm]{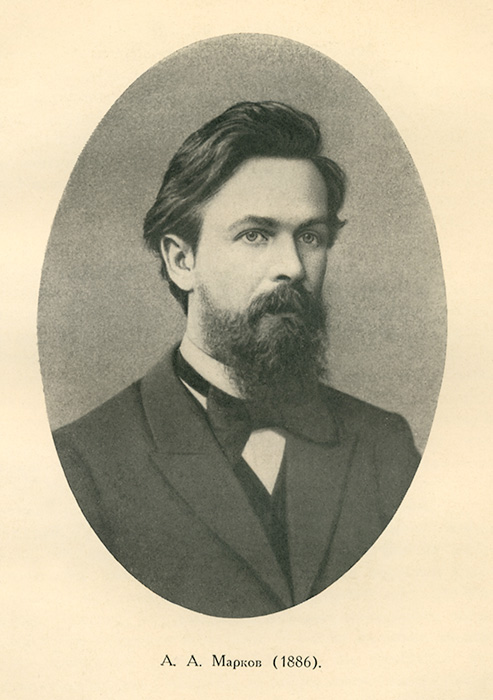}
\caption{Markov and a page of his frequency calculations}
\label{Fig:markov}
\end{center}
\end{figure}

\begin{figure}
\begin{center}
\includegraphics[height=5cm]{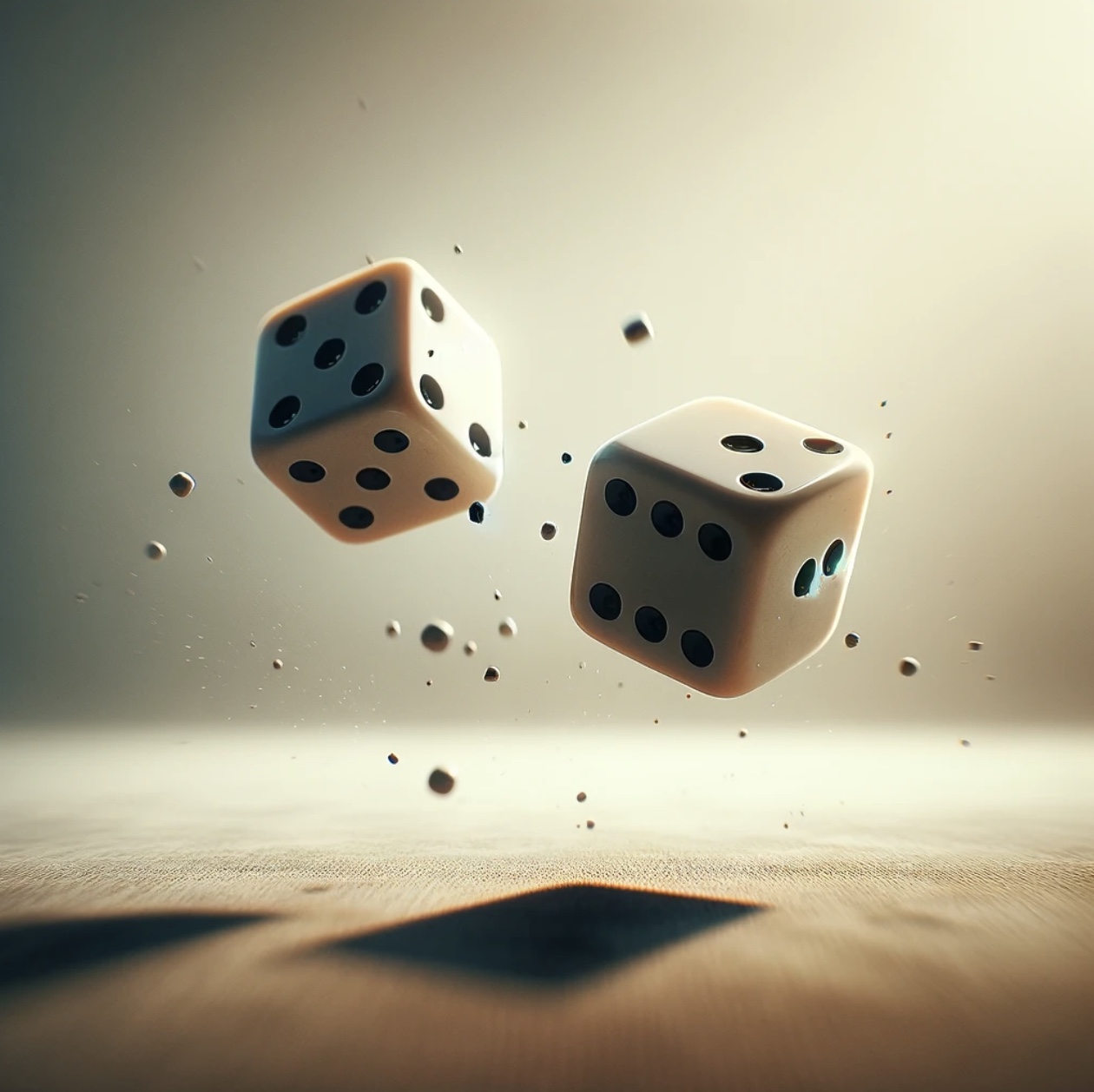}
\hspace{6em}
\includegraphics[height=5cm]{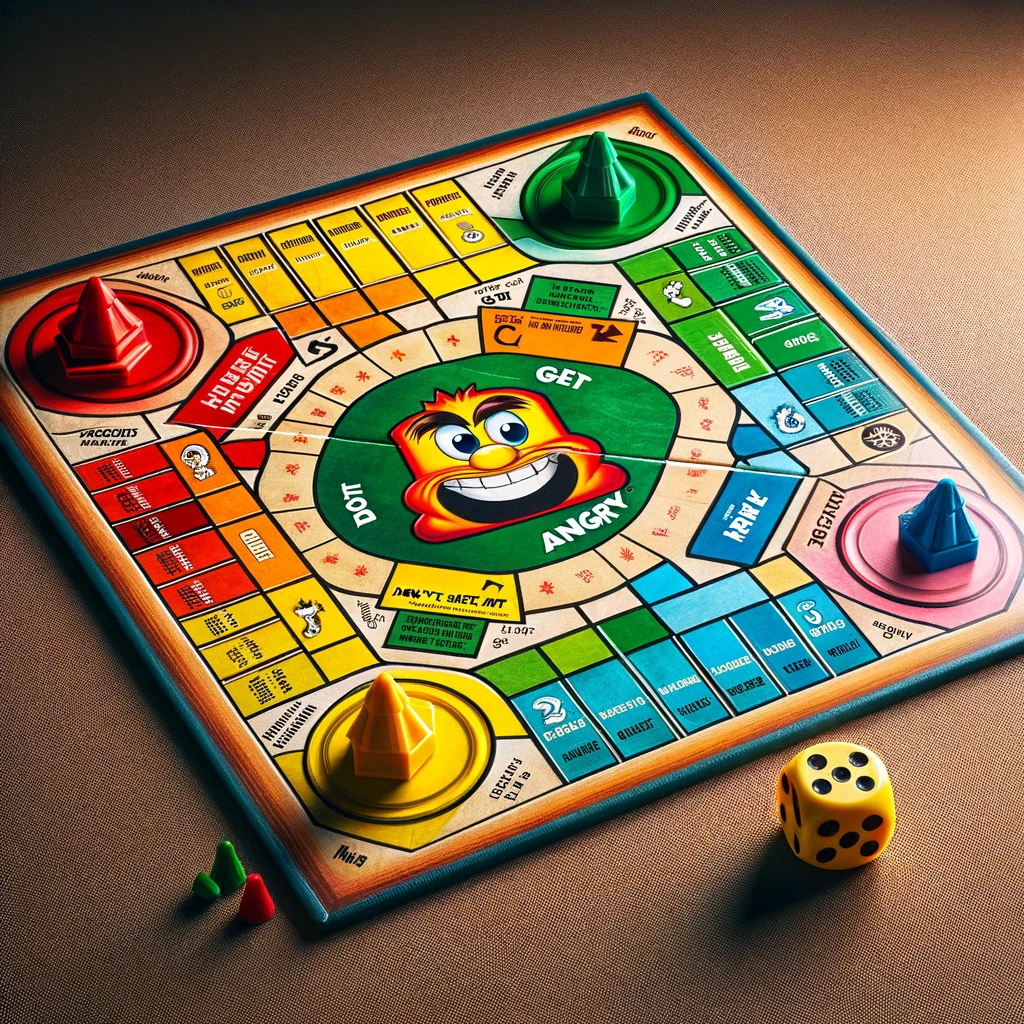}
\caption{Stateless vs position-based games of chance}
\label{Fig:dice}
\end{center}
\end{figure}

Beyond sampling $N$-grams, Shannon proceeded to generation according to \emph{conditional}\/ probabilities, which he presented as \emph{Markov chains} displayed in Fig.~\ref{Fig:shannon-markov}.
Shannon proposed this as a useful new application of Markov chains on language production --- not realizing that this very application was one of the examples who led  Andrey A.~Markov\footnote{This is Andrey A.~Markov Sr., who lived 1856-1922. His son Andrey A.~Markov Jr., who lived 1903-1979,  was also a prominent mathematician.} to introduce into probability theory the structure that came to be known as Markov chains. A page from Markov's 1913 $N$-gram study of Pushkin's epic poem ``Eugene  Onegin'', a classic of Russian literature, can be seen in  Fig.~\ref{Fig:markov}.
The crucial idea of Markov chains, some seen in Shannon's drawings in Fig.~\ref{Fig:shannon-markov}, is that the  conditional probabilities, construed informally since the earliest Pascal-Fermat discussions about probability and formally at least since Bayes, should be viewed as \emph{state changes}. This idea is familiar already in games of chance. The simplest games with dice consist of throwing dice and always seeking the same desired outcomes, as in Fig.~\ref{Fig:dice} on the left. In other, the desired outcomes vary with state, usually presented by positions on a board, as in Fig.~\ref{Fig:dice} on the right. In Markov's mathematical model, the board positions are presented as abstract states, and Markov chains are conveniently thought of as probabilistic state machines. Remarkably, Markov's probabilistic machines preceded the  deterministic machines\footnote{used by Turing in 1936 and by McCulloch and Pitts in 1943} by nearly as many years as Markov's application to language preceded Shannon's.

But Shannon introduced information theory as a study of a much more general concept: the channel.

\subsection{Language as a communication channel}\label{Sec:chan}

Shannon's mathematical theory of communication has been used for analyzing information processing and synthesizing communication channels for almost 80 years. Language is our main tool of communication and information processing. \textbf{\emph{Why is the theory of language not a part of the theory of communication?}} Technically, the two barely intersect and most researchers view them as unrelated. Why is that?

Noam Chomsky says that one is a science,  the other a branch of engineering. He argues that linguistics  strives to explain a natural process in human mind, whereas information theory builds codes to optimize functioning of antennas and networks. Such optimizations use statistics, whereas human mind ``does not play dice''\footnote{Just in case, let me clarify that this is not Chomsky, but a paraphrase of Einstein's ``God does not play dice'' objection to the statistical foundation of quantum mechanics. Interpreting this statement in the context of the modern theory of black holes, Hawking quipped: ``Not only does God play dice, but he sometimes throws them where they cannot be seen''. That is at least as applicable to human mind as to God.}. Peter Norvig\footnote{Norvig is a prominent AI researcher. In the golden era of Google, he was their director of research. With Stuart Russell, he coauthored probably the most influential AI textbook so far.} argued that it does and that empiric sciences use statistics in any case, justifying statistical learning and predictive inference.

However, the empty space between the theories of language and of communication may not be a consequence of such principled standpoints as much as of honest mistakes. People lose not only their dice, but also their marbles. Maybe thinking of language models as channels may still be useful if we can find the right the right angle. Let us spend a minute to clear up the issues around the \emph{feedback}\/ and \emph{feedforward}\/ references, which play a crucial role in language.

\subsubsection{Information-theoretic 
generalities}
\paragraph{Random variables.} Given a  frequency distribution $[-]\colon \Omega \to [0,1]$ as above, a \emph{random variable}\/ $X\colon \Omega$ is a sample from $\Omega$ according to the given distribution. 


\paragraph{Sources.} A \emph{stochastic process}\/ is a sequence of random variables $X_{1}, X_{2},\ldots, X_{n}, \ldots$, all sampling from the same space of events $\Omega$, with each $X_{n}$ depending on all $X_{k}$ for $k\lt n$. A \emph{source}\/ is the family of conditional probabilities
\beq
\pder {X^{n}}{X_{n+1}} \quad\mbox{ for }  n=1,2,3,\ldots
\eeq
where $X^{n} = X_{1}X_{2}\cdots X_{n}$ and  $X^{0}=1$ like before. All examples so far were sources. 

\paragraph{Channels.} A \emph{channel}\/ is a sequence of conditional probabilities 
\beq\label{eq:channel}
\ppder{X_{1}}\, ,\quad \pder{X_{1}}{Y_{1}}\, ,\quad \pder {\left(XY\right)^{n}}{X_{n+1}Y_{n+1}} \ \mbox{ for }  n=1,2,3,\ldots
\eeq
%
where\footnote{By abuse of notation, the shuffle $\left(XY\right)^{n}$ is often written as $X^{n}Y^{n}$, even when $X_{k+1}$ may depend on $Y_{k}$.} $\left(XY\right)^{n} = X_{1}Y_{1}X_{2}Y_{2} \cdots X_{n}Y_{n}$. The random variables $X$ are the channel inputs, the random variables $Y$ the outputs. In information theory, the $X$s are usually messages and $Y$s their encodings. In general, the inputs $X$ are thought to be produced by the Environment and the outputs $Y$ by the System responses. On the level of discourse, $X$s and $Y$s can be thought of as questions and answers. In semantics, one could construe $X$s as \emph{concepts}\/ and $Y$s as the corresponding \emph{phrases}; or as the \emph{signified}\/ items and the \emph{signifying}\/ tokens in language production. In language understanding, these intuitive interpretations would be reversed. 

\paragraph{Feedback and feedforward flows.} Using the transitivity from \eqref{eq:trans}, the sequence in \eqref{eq:channel} can be decomposed as\footnote{If we let $n=0,1,2,\ldots$ and $\left(XY\right)^{0} = \emptyset$, with $\ppder{X_{1}}$ given, \eqref{eq:channel-decomp} captures the rest of \eqref{eq:channel}, since $\pder{\left(XY\right)^{0}}{X_{1}Y_{1}} = \pder{\emptyset}{X_{1}Y_{1}} = \ppder{X_{1}Y_{1}}$ provides the same information as $\pder{X_{1}}{Y_{1}}$.} 
\bea\label{eq:channel-decomp}
\pder {\left(XY\right)^{n}}{X_{n+1}Y_{n+1}} & = & \pder {\left(XY\right)^{n}}{X_{n+1}}  \cdot \pder {\left(XY\right)^{n}X_{n+1}}{Y_{n+1}}
\eea
The sequence in \eqref{eq:channel} can thus be equivalently given as the pair of sequences in the middle column in Table~\ref{Table:channels}, marked by $(\leftrightarrow)$.

\begin{sidewaystable}
\begin{center}
\begin{tabular}{r|r|r@{\ }l|l|l}
-2\hspace{2em} & -1\hspace{2.5em} & \multicolumn{2}{c|}{0}
&\hspace{2.3em} 1 &\hspace{1.5em} 2
\\
\hline
$(\nomem)\hspace{1.3em}$&$(\nofw)\hspace{2em}$& \multicolumn{2}{c|}{$(\leftrightarrow)$} 
& \hspace{2em}$(\noback)$ & \hspace{1em}$(\markovian)$
\\
\hline
&& & \raisebox{-.5ex}{$\ppder{X_{1}}\, , $}& \raisebox{-.5ex}{$\ppder{X_{1}}\, , $}& \raisebox{-.5ex}{$\ppder{X_{1}}\, , $}\\
$\pder{X_{1}}{Y_{1}}\, ,$ & $\pder{X_{1}}{Y_{1}}\, ,$ & $\pder{X_{1}}{Y_{1}}\, ,$ & $\pder{X_{1}Y_{1}}{X_{2}}\, ,$ & $\pder{X_{1}}{X_{2}}\, ,$ & $\pder{X_{1}}{X_{2}}\, ,$ 
\\
$\pder{X_{2}}{Y_{2}}\, ,$&$\pder{X_{1}X_{2}}{Y_{2}}\, ,$& $\pder{X_{1}Y_{1}X_{2}}{Y_{2}}\, ,$ & $\pder{X_{1}Y_{1}X_{2}Y_{2}}{X_{3}}\, ,$
& $\pder{X_{1}X_{2}}{X_{3}}\, ,$ & $\pder{X_{2}}{X_{3}}\, ,$ 
\\
\vdots&\vdots & \vdots &&\vdots&\vdots 
\\
$\pder{X_{n}}{Y_{n}}\, ,$& $\pder{X^{n}}{Y_{n}}\, ,$ & $\pder{X^{n-1}Y^{n-1}X_{n}}{Y_{n}}\, ,$ & $\pder{X^{n}Y^{n}}{X_{n+1}}\, ,$ & $\pder{X^{n}}{X_{n+1}}$ & $\pder{X_{n}}{X_{n+1}}$ 
\\
$\pder{X_{n+1}}{Y_{n+1}}$\phantom{\, } &$\pder{X^{n+1}}{Y_{n+1}}$\phantom{\, }& $\pder{X^{n}Y^{n}X_{n+1}}{Y_{n+1}}$\phantom{\, }  & & \\
\hline\hline
memoryless &no feedforward & \multicolumn{2}{c|}{general channel} & no feedback & markovian
\end{tabular}
\end{center}
\vspace{.5\baselineskip}

\caption{Channels and their restrictions}
\label{Table:channels}
\end{sidewaystable}

The column marked  $(\nofw)$ contains the sequence of conditional probabilities where the outputs $Y$ do not depend on the previous outputs, but only on the inputs. In other words, the channel outputs have no impact on its later outputs, and the channel is \emph{\textbf{feedforward-free}}. The column marked  $(\noback)$, on the other hand, contains the sequence of conditional probabilities where the inputs $X$ do not depend on the previous outputs. The channel outputs here have no impact on the later inputs, and the channel is \emph{\textbf{feedback-free}}. 

\paragraph{Markovian sources and memoryless channels.} If the channel outputs $Y_{n}$ are independent not only of the preceding outputs, but also on the inputs preceding $X_{n}$, then it is not only feedback-free, but also \emph{memoryless}. Column $(\nomem)$ of Table~\ref{Table:channels} displays this restriction. Column $(\markovian)$ displays the corresponding restriction on inputs. 


\subsubsection{Memoryless feedback confusion}

\paragraph{Ash's formula.} Column $(\nofw)$ suggests that the outputs $Y$ of any  feedforward-free channel only depends on the input source $X$. For memoryless channels, this boils down to the dependency of each $Y_{n}$ on $X_{n}$ alone, for all $n$. Following this idea, R.~Ash provided the elegant, convenient, and very useful formula 
\bea\label{eq:Ash}
\pder{X^{n}}{Y^{n}} & = & \prod_{m=1}^{n} \pder{X_{m}}{Y_{m}}
\eea	
Claiming that it was satisfied by all memoryless channels\footnote{This property is adopted in Ash's definition of memoryless channel, in Sec.~3.1. of his ``Information Theory'', that appeared in 1965 and remained one of the most widely used textbooks for a long time.}. Note, however, that restricting the feedforward flows by imposing $(\nofw)$ cannot restrict the channel dependency to the source $X$ if $X$ is not a source, i.e. if the channel does not satisfy $(\noback)$ because it permits feedback. Condition $(\nofw)$ prevents the direct dependencies of the outputs $Y$ on the preceding outputs but if the inputs $X$ still depend on the preceding $Y$s via feedback, then the output $Y$ still depends on the preceding outputs indirectly. --- That detail was missed in a good part of information-theoretic research for a good number of years. Feedback was thus tacitly precluded without anyone noticing, until Massey's 1990 analysis of directed information\footnote{J. Massey, Causality, feedback, and directed information. \emph{Proc. of Intl. Symp. on Inf. Theory and Applications}, held in Honolulu, HI in November 1990.}.

\paragraph{Mamoryless channels satisfy Ash's formula if they are feedback-free.} Using the chain rule and transitivity, the distribution of any channel can be reduced to the products 
\bea
\ppder{\left(XY\right)^{n}} & \stackrel{\eqref{eq:chain}}{=} &  \prod_{m=1}^{n} \pder{\left(XY\right)^{m-1}}{X_{m}Y_{m}}\ \ \stackrel{\eqref{eq:trans}}{=} \prod_{m=1}^{n} \pder{\left(XY\right)^{m-1}}{X_{m}}\cdot \pder{\left(XY\right)^{m-1}X_{m}}{Y_{m}}\\\notag
 & =& \prod_{m=1}^{n} \pder{\left(XY\right)^{m-1}}{X_{m}}\cdot \prod_{m=1}^{n}\pder{\left(XY\right)^{m-1}X_{m}}{Y_{m}}\label{eq:fst}
\eea
%

On one hand, memoryless channels satisfy
\bear
\pder{\left(XY\right)^{m-1}X_{m}}{Y_{m}} & \stackrel{(Y\!\downarrow)}= & \pder{X_{m}}{Y_{m}}
\eear
On the other hand, for feedback-free channels holds
\bear
\prod_{m=1}^{n} \pder{\left(XY\right)^{m-1}}{X_{m}} & \stackrel{\hspace{.3ex} (\times\hspace{-0.6em}\leftarrow)}= & \prod_{m=1}^{n} \pder{X^{m-1}}{X_{m}} \ \ \stackrel{\eqref{eq:chain}}=\ \ \ppder{X^{n}} 
\eear
%
If the channel is both memoryless and feedback-free, then \eqref{eq:fst} boils down to
\bear
\ppder{\left(XY\right)^{n}} & = & \ppder{X^{n}}  \prod_{m=1}^{n}\pder{X_{m}}{Y_{m}}
\eear
which is equivalent to \eqref{eq:Ash}, by definition of conditional probability. 

\begin{figure}[!hb]
\begin{center}
\includegraphics[height = 5.5cm]{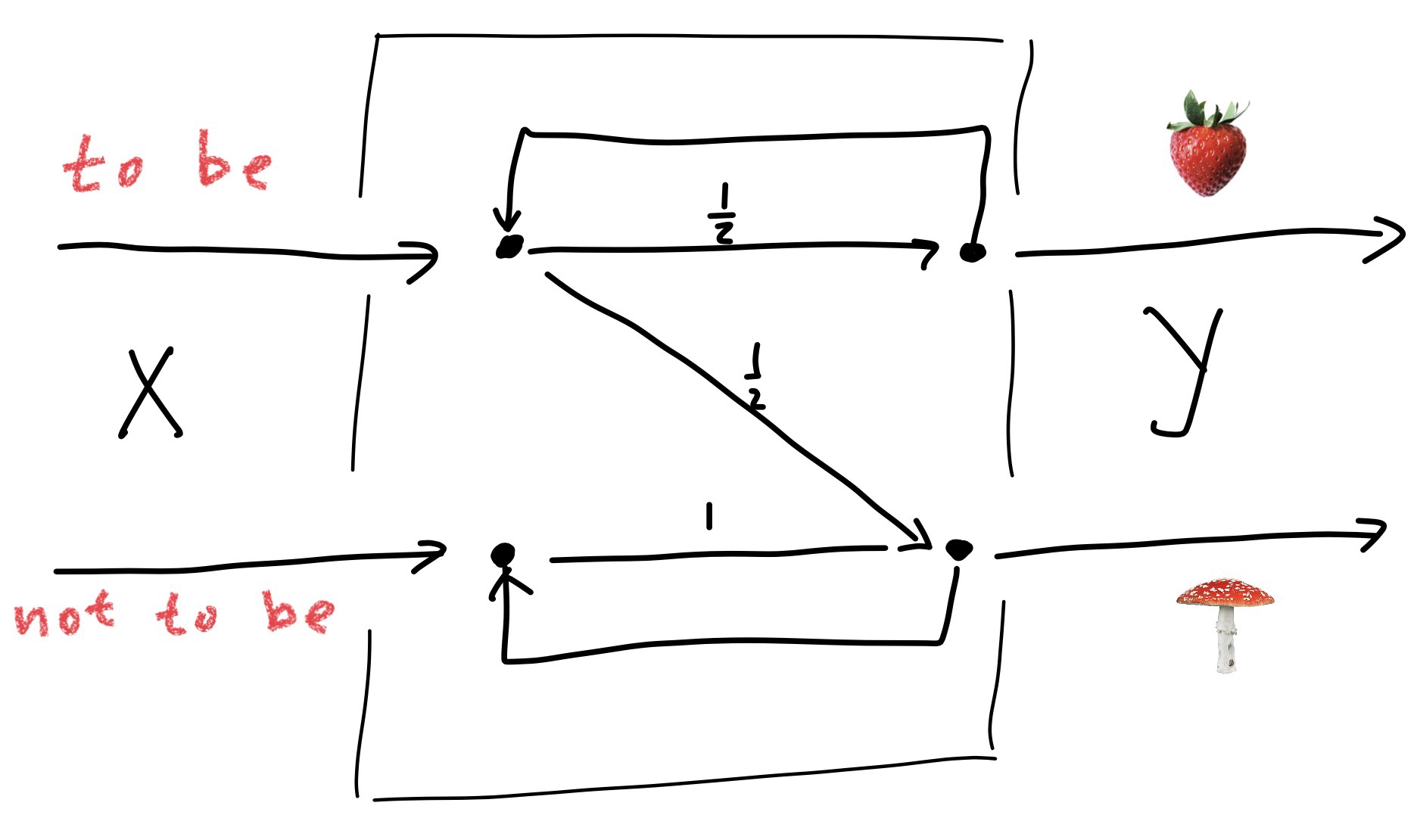}
\caption{Memoryless channel with feedback}
\label{Fig:tobe}
\end{center}
\end{figure}

\paragraph{Mamoryless channels with feedback may not satisfy
 Ash's formula.} Consider the channel in Fig.~\ref{Fig:tobe},
given by the transition probabilities
\begin{align*}
\pder{X_{m} =\mbox{\color{red} to be}}{Y_{m}= \raisebox{-.3ex}{\includegraphics[height = 0.56cm]{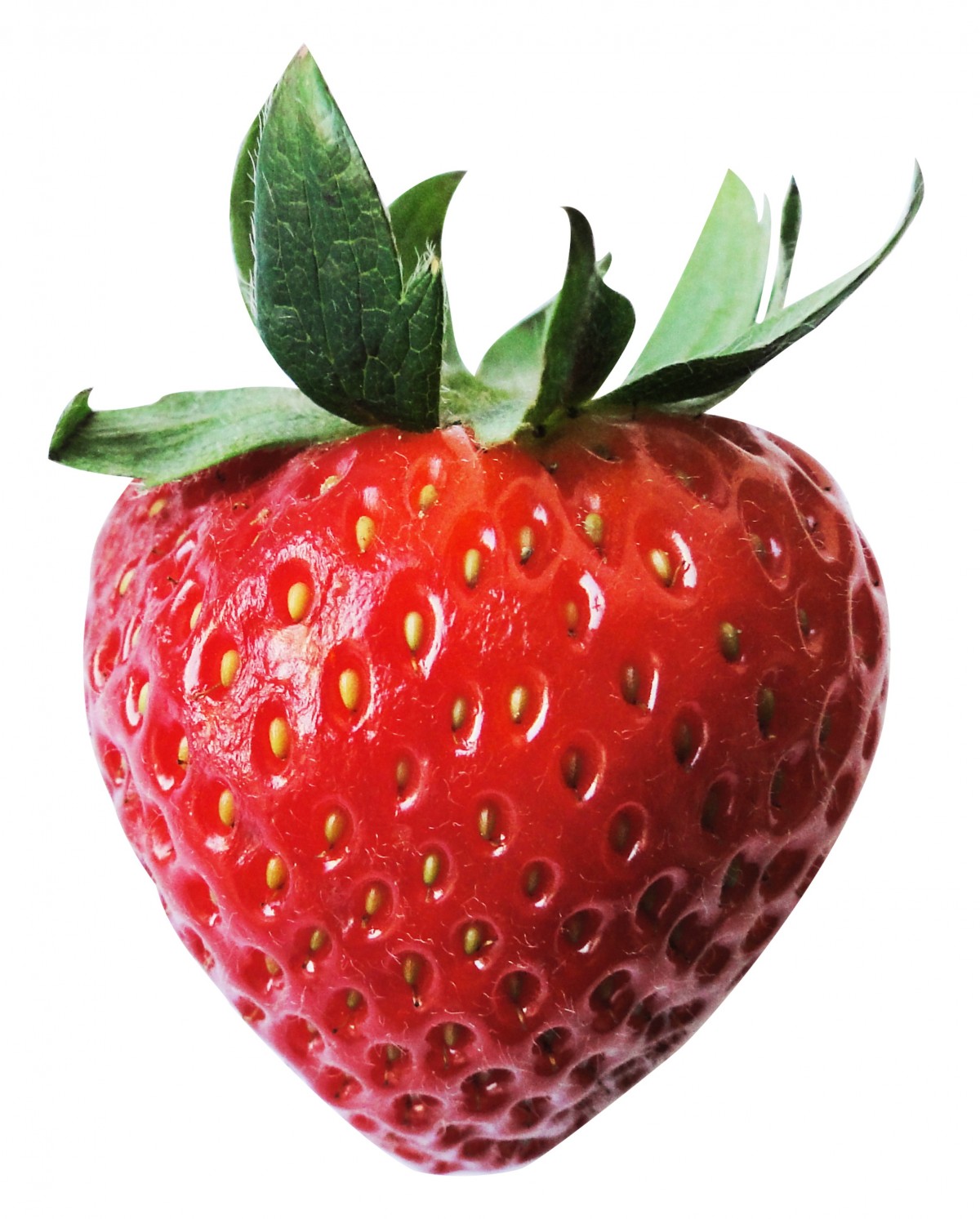}}} & = \frac 1 2 & \pder{Y_{k}= \raisebox{-.3ex}{\includegraphics[height = 0.56cm]{PICS/strawberry.jpg}}}{X_{m} =\mbox{\color{red} to be}} & = 1\\
\pder{X_{m} =\mbox{\color{red} to be}}{Y_{m}= \hspace{-.7ex}\raisebox{-.6ex}{\includegraphics[height = 0.7cm]{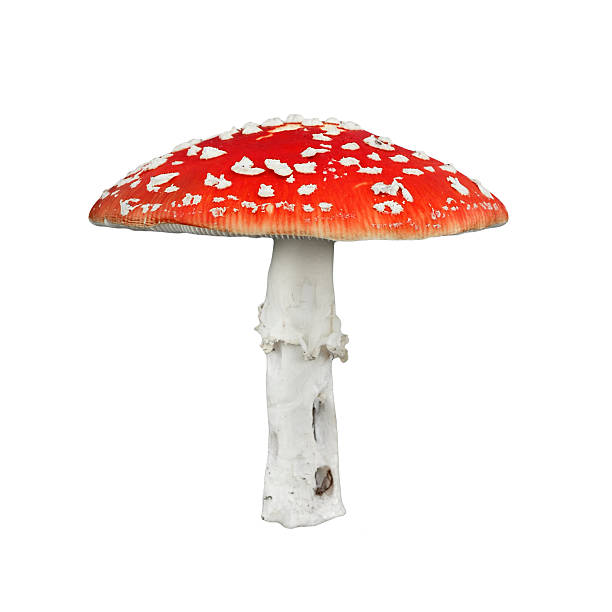}}\hspace{-.5ex}} & = \frac 1 2 & \pder{Y_{k}= \raisebox{-.6ex}{\hspace{-.7ex}\includegraphics[height = 0.7cm]{PICS/shroom.jpg}}\hspace{-.7ex}}{X_{m} =\mbox{\color{red} not to be}} & = 1\\
\pder{X_{m} =\mbox{\color{red} not to be}}{Y_{m}= \raisebox{-.6ex}{\hspace{-.7ex}\includegraphics[height = 0.7cm]{PICS/shroom.jpg}}\hspace{-.7ex}} & = 1
\end{align*}
Since a channel run that outputs \raisebox{-1ex}{\hspace{-.7ex}\includegraphics[height = 0.7cm]{PICS/shroom.jpg}\hspace{-.7ex}} can never again input {\color{red} to be},  Ash's formula does not hold:
\[\pder{\mbox{\color{red} to be}, \mbox{\color{red} to be}}{\raisebox{-.6ex}{\hspace{-.7ex}\includegraphics[height = 0.7cm]{PICS/shroom.jpg}\hspace{-.7ex}}
, \raisebox{-.3ex}{\includegraphics[height = 0.56cm]{PICS/strawberry.jpg}}} \ \ =\ \ 0\ \ \neq\ \ \frac 1 2\cdot \frac 1 2\ \  = \ \ \pder{\mbox{\color{red} to be}}{\raisebox{-.6ex}{\hspace{-.7ex}\includegraphics[height = 0.7cm]{PICS/shroom.jpg}\hspace{-.7ex}}}\cdot \pder{\mbox{\color{red} to be}}{\raisebox{-.3ex}{\includegraphics[height = 0.56cm]{PICS/strawberry.jpg}}}
\]
A feedback-free channel is unsuitable as a model of language. In terms of $X$s as questions and $Y$s as answers, without feedback we can only study conversations where the questions are independent on the answers.

\paragraph{Upshot.} The process of language production is clearly a communication channel. It is a feedforward channel because the next word $Y_{n+1}=w_{n+1}$ that I output depends on the previously produced words $Y^{n}=w^{n}$. It is a feedback channel because the conceptual inputs $X$ change with the outputs $Y$ that I have previously produced. The feedforward and the feedforback flows are also displayed as the forward and backward arrows, denoting the syntactic and semantical dependencies even in the deterministic models. Eliminating such dependencies from information-theoretic channels, albeit tacitly and inadvertently, has limited the applicability of the information-theoretic results on language.

On the other hand, the artificial language models also generate language in a similar process, by deriving one word after another, from previously produced contexts and internal concepts. Although extensively studied in AI and ML research, the underlying process of predictive inference of text has not been studied in terms of channels, mainly because of the described restrictions. Since the process of language production is a channel, understanding it will require either an update of the existing theory of channels, or a new theory built from scratch. Either way, since languages flow through communication channels, understanding them will require a suitable theory of those, whatever they might be called in the end.

\section{Deep space: dependent types and concept associations}

While the channel information flows in language production are clear, it is not clear how they can be tractable.

The Markov-Shannon view of language presents $N$-grams as states and language production steps as state transitions. When I speak, what I am going to say next is determined by my state of mind. After I say it, my state of mind progresses to the next state and it determines what I will say after that. The process surely proceeds something like this --- for a sufficiently flexible notion of \emph{state of mind}. 

The problem is that the notion of $N$-gram is surely the least flexible notion of state that you could think of. A state if mind reduced to a register of length $N$ doesn't scale up. If I were to speak my mind from a tiny lexicon of 1,000 words\footnote{An average English speaker uses about 10,000 words, and understands about 20,000.}, and if my attention span is so short that only the last 10 words that I said determines what I will say next, then just storing the transition probabilities would require $\left(10^{3}\right)^{10\times 3}$ memory cells, which is more than the number of atoms in the observable universe (estimated to be about $10^{80}$). So I am certainly not storing an $N$-gram table in my head. What might be the flexible version of state that overcomes this scalability problem and supports the feedback and the feedforward references? In the next lecture, we will see how the state dependencies are realized as type dependencies. I close the this one by the example from the previous lecture. Here is a bird's eye view of a channel production of Groucho's elephant sentence, descending down a tower of dependent types.
\begin{center}
\includegraphics[width=0.97\linewidth]{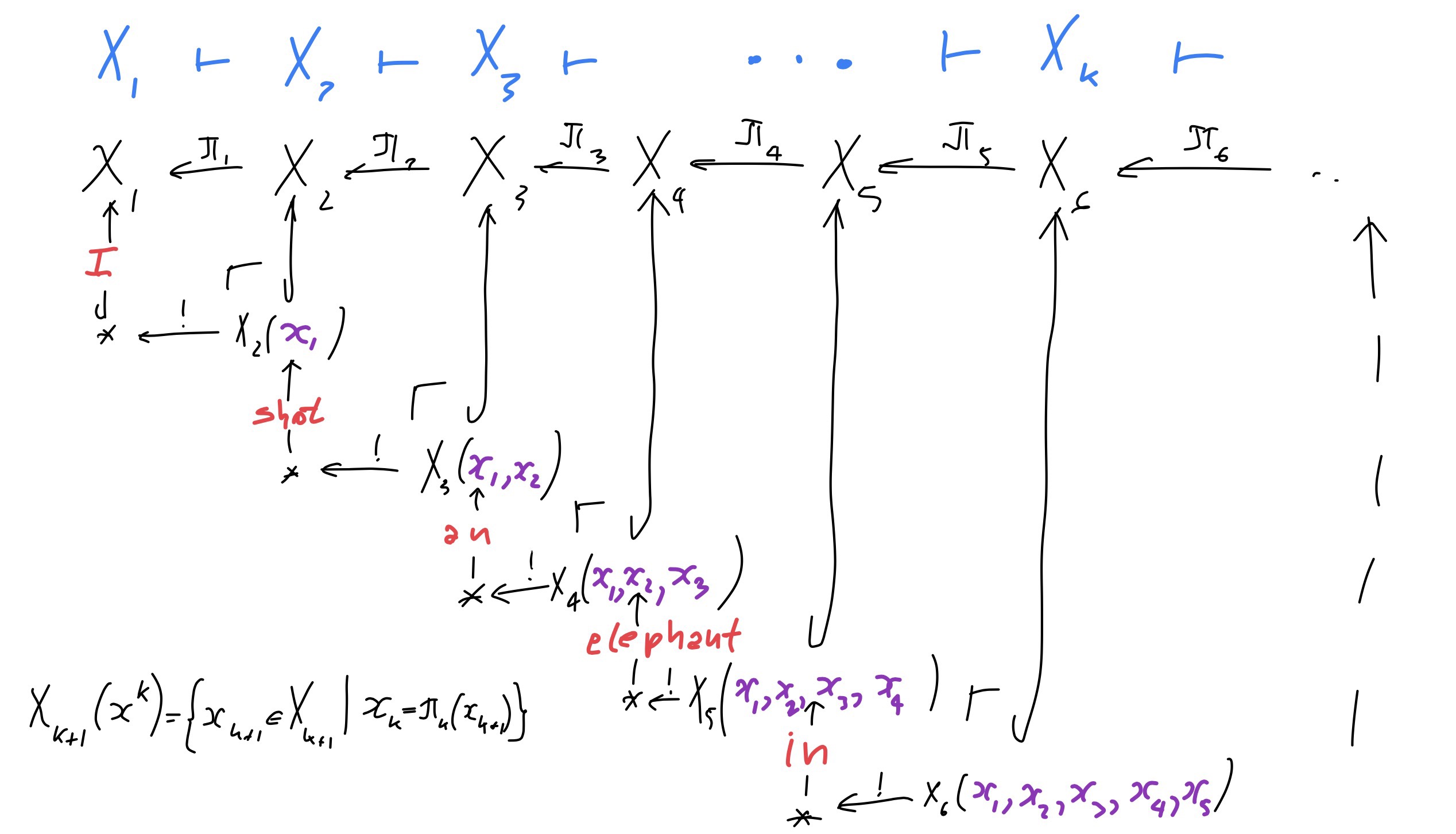}
\end{center}


%
%
%
%
%
%
%
%
%
%
%
%
%
%
%
%
%
%
%
%
%
%
%

\def\thechapter{4}
\setchaptertoc
\chapter{Language as a Universal Learning Machine}\label{Chap:Learning}


\section{Evolution of learning}

\subsection{Learning causes and superstitions}
Spiders are primed to build spider nets. Their engineering skills to weave nets are  programmed in their genes. They are preprogrammed builders and even their capability to choose and remember a good place for a net is automated.

\begin{figure}[!ht]
\begin{center}
\includegraphics[height=3.5cm]{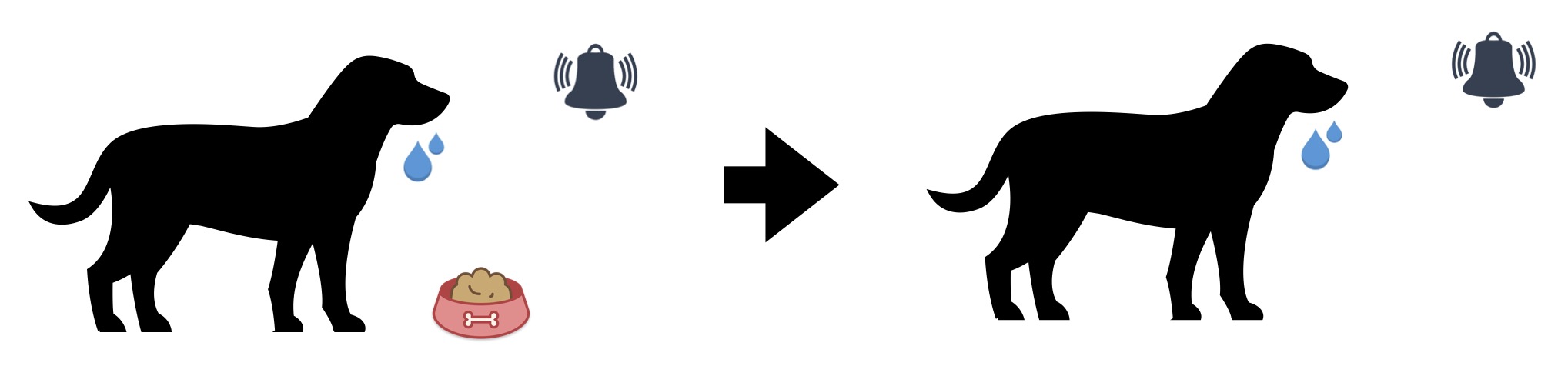}
\caption{
If the bell rings whenever the dog is fed, he learns to salivate whenever the bell rings.}
\label{Fig:pavlov}
\end{center}
\end{figure}

Dogs and pigeons are primed to seek food. Their capabilities to learn sources and actions that bring food are automated. In a famous experiment, illustrated in Fig.~\ref{Fig:pavlov}, physiologist Pavlov studied one of the simplest forms of learning, usually called \emph{conditioning}. 

\begin{figure}[!ht]
\begin{center}
\includegraphics[height=2.2cm]{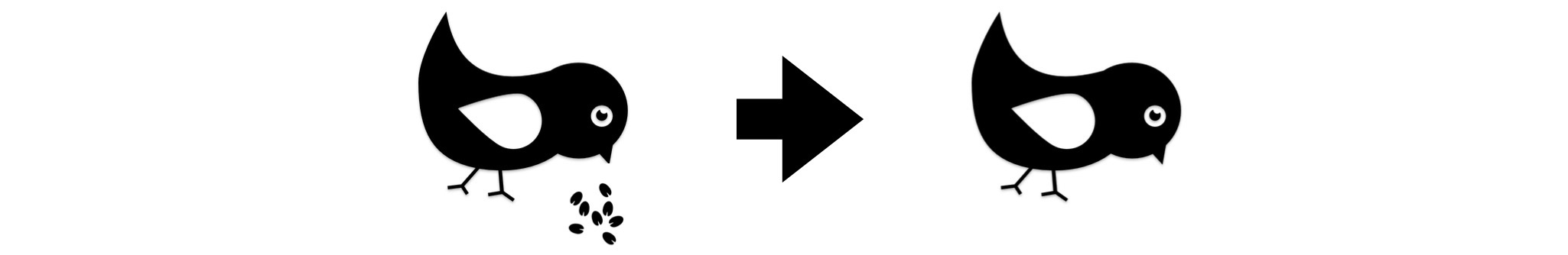}
\caption{If food arrives while the pigeon is pecking, she learns that pecking conjures food}
\label{Fig:skinner}
\end{center}
\end{figure}

Continuing in the same vein, psychologist Skinner showed that pigeons could even develop a form of superstition, illustrated in Fig.~\ref{Fig:skinner}, also by trying to learn where the food comes from. Skinner fed pigeons at completely random times, with no correlation with their behaviors. About 70\% of them developed beliefs that they could conjure food. If a pigeon happened to be pecking on the ground, or ruffling feathers just before the food arrived, then they would apparently engage in this action more frequently, which increased the chance that the food would arrive while they were performing that action. If one of the random associatons after a while prevails, it becomes a ritual dance for food. Each time, the food eventually arrives and confirms that the ritual works.
 
Humans are primed to seek causes and predict effects. Like pigeons, they associate coinciding events as correlated and develop superstitions, promoting coincidences into causal theories. While pigeons end up pecking empty surfaces to conjure grains, humans build monumental systems of false beliefs, attributing their fortunes and misfortunes, say, to the influence of stars millions of light years away, or to their neighbor's evil eye, or to pretty mych anything that can be seen, felt, or counted\footnote{Skinner's explorations of our intellectual kinship with pigeons have interesting interpretations in the context of arguments that the concept of causality as such is in essence unfounded. From different directions, such arguments have been developed by Hume, Russell, Bohr, and many other scientists and philosophers.}. 

But while our causal beliefs are shared with pigeons, our capabilities to build houses and span bridges are not shared with spiders. Unlike spiders, we are not primed to build but have to \emph{learn}\/ our engineering skills. \emph{\textbf{We are primed to learn.}}

\subsection{General learning framework}\label{Sec:gen-frame}
\begin{figure}[!ht]
\begin{center}
\includegraphics[height=4.5cm]{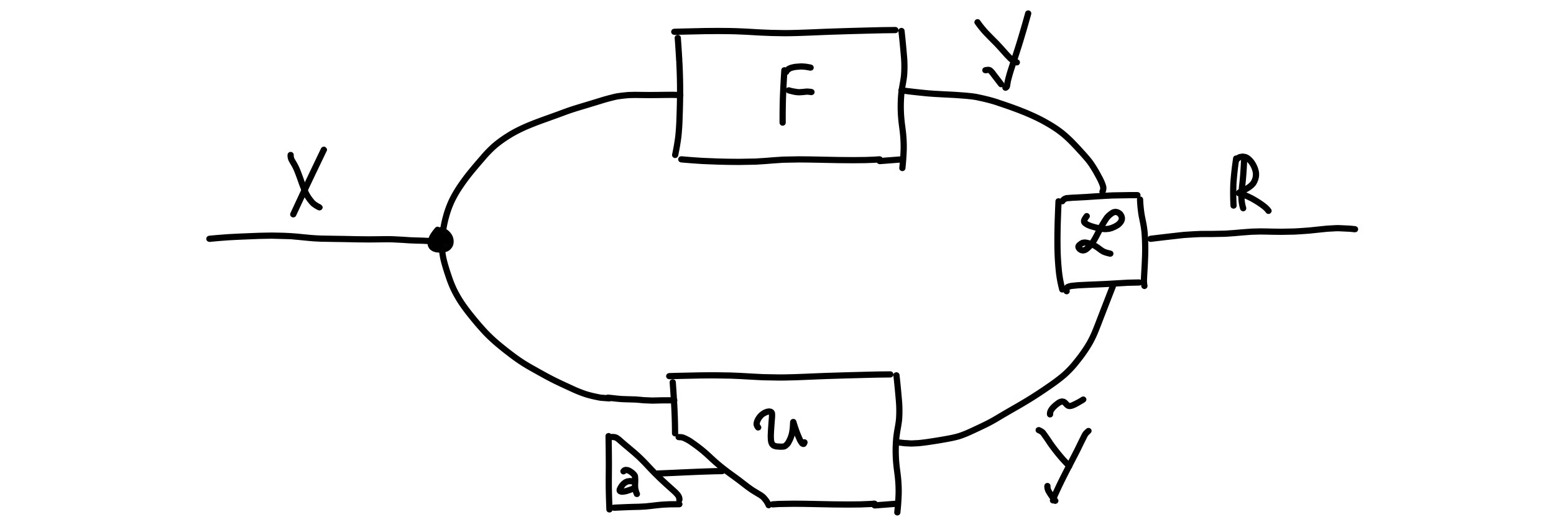}
\caption{Learning a model $\aprog$ of  $F$  up to $\LLL$ gives $F(X)  \stackrel \LLL \approx \upsilon(X)\aprog$}
\label{Fig:learning}
\end{center}
\end{figure}

A bird's eye view of the learning scenario is depicted in Fig.~\ref{Fig:learning}. 
The main characters are:
\begin{itemize}
\item a process $F$, the \emph{supervisor}\footnote{Turing called it a ``teacher''.} in supervised learning, processing  input data of type $X$ to produce output \emph{classes}\/ or \emph{parameters}\/ of type $Y$;
\item an $\aprog$-indexed family of functions\footnote{Intuitively, the ``$\upsilon($'' part can be thought of as the input encoding operation and the ``$)\aprog$'' part is then output decoding.} $\upsilon(-)\aprog$,  where $\upsilon$ is a  \emph{learning machine}\/ or \emph{interpreter}\footnote{Turing called it a ``pupil''.} and the indices $\aprog$ are the \emph{models}, usually expressed as \emph{programs};  lastly, there is 
\item a function $\LLL$, usually called the \emph{loss}, comparing the outputs $Y=F(X)$ with the predictions $\widetilde Y= \upsilon(X)\aprog$ and measuring their difference by a real number.
\end{itemize}
The learner overseeing the learning framework is given a finite set 
\[ (x_{1}, y_{1}),\ (x_{2}, y_{2}), \ldots, (x_{n}, y_{n}) \]
where the $x$s are samples from a source $X$ and the $y$s are the corresponding samples from $Y=F(X)$. Learner's task is to build a model $\aprog$ that minimizes the losses
\[ \LLL(y_{1}, \widetilde y_{1}), \  \LLL(y_{2}, \widetilde y_{1}),\ldots, \LLL(y_{n}, \widetilde y_{n}) \]
where $y_{i}=F(x_{i})$ and $\widetilde y_{i} = \upsilon(x_{i}){\aprog}$ for $i=1,2,\ldots,n$. Since some of the losses may increase when the others decrease, the learning algorithm is required to minimize their cumulative \emph{guessing risk} 
\bea\label{eq:risk}
\RRR(\aprog) & = & \sum_{i=1}^{n} \ppder{\upsilon(x_{i})\aprog}\LLL\left(y_{i},\upsilon(x_{i})\aprog\right)
\eea
where $\ppder{\upsilon(x_{i})\aprog}$ is the frequency of the guesses $\upsilon(x_{i}){\aprog}$. The guessing risk in \eqref{eq:risk} is often written in the form   $\RRR(\aprog) = \int \LLL\left(y,\upsilon(x){\aprog}\right) d\upsilon(x){\aprog}$. 
Once the risk is minimized, the function $F$ is approximated by running the machine $\upsilon$ on $\aprog$ and we write
\bea\label{eq:gen-approx}
F(X) & \approx & \upsilon(X)\aprog
\eea

\paragraph{Potato, potahto, tomato, tomahto.} What are the outcomes of learning? The outcome $\aprog$ of a round of supervised learning is usually called a \emph{model} of the supervisor $F$. Since $\aprog$ describes $F$, most logicians would call $\aprog$ a  \emph{theory}\/ of $F$. If the interpretations $\upsilon(X)\aprog$ give true predictions about $F(X)$, then they would say that $F$ is a model of the theory $\aprog$ under the semantical interpretation by $\upsilon$. Statisticians would call $\aprog$ a \emph{hypothesis}\/ about $F$.  Currently valid hypotheses and theories thought to be true comprise learner's \emph{belief state}. Sec.~\ref{Sec:outlook} presents a curious construction illustrating a need for studying belief logics of machine learning\footnote{The results of https://arxiv.org/abs/2303.14338 also point in this direction.}. Note also  that the model $\aprog$ of $F$ needs to be \emph{executable} in order to allow computing the predictions $\upsilon(X)\aprog$ of $F(X)$. In software engineering, executable models requirement specifications are their implementations: the programs. An executable model $\aprog$ is therefore usually implemented as aprogram.  --- In summary, the outcome of a learning process is an executable model or hypothesis.  The cumulative outcome of learning is learner's belief state. The process of learning is a search for learnable programs.

\paragraph{All learning is language learning.} In general, the process $F$ to be learned is given as a channel, which means that the  outputs are context-dependent. The story from Sec.~3.2 of the \emph{Semantics}\/ part applies. The channel inputs $x_{j}$ depend on the earlier inputs $x_{i}$, $i\lt j$. When there is feedback, $x_{j}$ also depends on the earlier outputs $y_{i}$, $i\lt j$. To be able to learn $F$'s behavior, the learning machine  $\upsilon$ must also be a channel. \textbf{\emph{Since capturing channel dependencies requires syntactic and semantic references, there is a language behind every learner}}, whether it is apparent or not. The semiotic analyses of the languages of film, music, or images, etc., describe genuine syntactic and semantic structures. 
Different organisms learn in different ways, but for humans and their machines, all learning is language learning.  

\subsection{From pigeons to perceptrons}
\subsubsection{Pigeon superstition} The function $F$ that a pigeon learns to predict is a source of food. It can be viewed as a channel $\pder X Y$, where the values $x_{1}, x_{2}, \ldots$ of type $X$ are moments in time and $Y=F(X)$ is a random variable, delivering seeds with a fixed probability. Suppose that $Y=1$ means ``food'' and $Y=0$ means ``no food''. If the $\aprog$s denote the elements of some set of actions that a pigeon could take, then the pigeon is trying to learn for which $\aprog$s and at which moments $x$ to output $\upsilon(x){\aprog} = 1$ and when to output 0. The loss function is $\LLL(y,\upsilon(x){\aprog}) = \lvert y-\upsilon(x){\aprog} \rvert$ and the guess loss $\RRR(\aprog)$ is minimized when $y = \upsilon(x){\aprog}$, i.e. the food is delivered just when the pigeon takes the action $\aprog$. After a sufficient amount of time, the random output $Y=1$ will almost surely coincide with a prediction $\upsilon(x){\aprog} = 1$ for some $\aprog$. Pigeon will then learn to do $\aprog$ more often and increase the chance of such coincidences. If one $\aprog$ prevails, the pigeon will learn that it causes food.

\subsubsection{Statistical testing}
Science is a family of methods designed to overcome superstitions. The idea is to prevent pigeon-style confirmations by  systematically testing hypotheses and only accepting significant correlations. The mathematical foundations of statistical hypothesis testing were developed in the 1920s by Ronald Fisher, and have remained the bread and butter of scientific practices. The crucial assumption is that the guessing function $\upsilon$ for any hypothesis $\aprog$ is given together with its probability density $p_\aprog(x)  =  \frac{d\upsilon(x){\aprog}}{dx}$. The loss $\LLL$ is then estimated by the length of the description of this probability. If the value of $p_\aprog(x)$ is described by a string of digits, its description length is proportional to $-\log p_\aprog(x)$. The guessing risk is thus $\RRR(\aprog) = \int -\log p_\aprog(x) d\upsilon(x)\aprog$. Values of this kind are studied in information theory as measures of uncertainty. Minimizing $\RRR(\aprog)$ thus boils down to choosing the hypothesis $\aprog$ that minimizes the uncertainty of sampling $\upsilon$ for $\aprog$. Fisher thus recommended the learning algorithm that selects the hypothesis with a \emph{maximal likelihood}.

The basic shortcoming of statistical testing is that the densities $p_\aprog$ must be known. They are presumed to arise from scientists' minds, together with their hypotheses parametrized by $\aprog$. Statistics thus provides a testing service, but the actual process of learning the hypotheses $\aprog$  is out of scope and left to the magic of insight and creativity. While Kolmogorov and his students were pondering this problem for decades and eventually solved it, a central part of the solution emerged inadvertently, and from an unexpected direction. 

\begin{figure}
\begin{center}
\includegraphics[width=.8\linewidth]{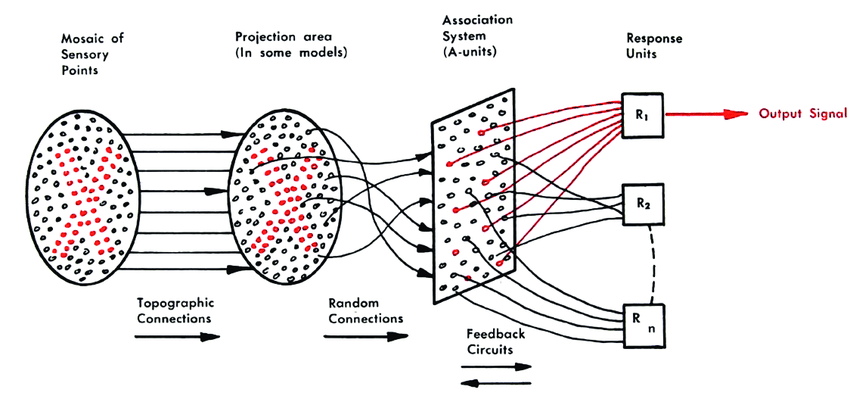}
\caption{Illustration from Rosenblatt's 1958 project report to the Office of Naval Research} 
\label{Fig:perceptron}
\end{center}
\end{figure}

\subsubsection{Perceptrons} 
In 1943, McCulloch and Pitts proposed a mathematical model of the neuron. It boiled down to a state machine, like Turing's original 1936 computer, just simpler, since it didn't have the external memory. In the late 1950s, Frank Rosenblatt was working on expanding the model of a neuron into a model of the brain. It was an ambitious project, as the illustration from the original project report in Fig.~\ref{Fig:perceptron} shows.  Rosenblatt, however, arrived at a component simpler than the McCulloch-Pitts neuron. He called it \emph{perceptron}, to emphasize the difference of his project from the ``various engineering projects concerned with automatic pattern recognition and `artificial intelligence' ''. Nevertheless, the project generated news reports under titles such as ``Frankenstein Monster Designed by Navy Robot That Thinks'', which Rosenblatt duly reports about in the preface to his ``Neurodynamics'' book. 

\paragraph{Mathematical neurons} were defined as pairs $\aprog = \big(b,\bra w\big)$, where $\bra w = \sum_{i=1}^d w^{i}\bra i$ is the projection on a vector\footnote{Recall that $\bra w = \sum_{i=1}^d w^{i}\bra i$ is a convenient notation (due to Paul Dirac) for the row vector $\seq{w^{1}\ w^{2} \cdots w^{d}}$, whereas $\ket w$ is the corresponding column vector. Viewed as a linear operator, the row vector $\bra w$ denotes the projection on the column vector $\ket w$.} $\ket w = \sum_{i=1}^d \ket i w_i$ 
and $b$ a scalar. It is meant to be a very simple program intepreted by the interpreter $\upsilon$. To evaluate $\aprog = \big(b,\bra w\big)$ on an input vector input vector $\ket x$, the interpreter $\upsilon$ applies the projection $\bra w$ on $\ket x$ to get the inner product $\braket {w}{x} = \sum_{i=1}^d w_i\cdot x_i$, which measures the length of the projection of either of the vectors on the other, and then it outputs the sign of the difference $\braket w x -b$:
\bea\label{eq:Falpha}
\upsilon\ket x {\aprog} & = & \begin{cases} 
1 & \mbox{ if } \braket {w}{x} \gt b\\
0 & \mbox{ if } \braket {w}{x} = b\\
-1  & \mbox{ if } \braket {w}{x} \lt b
\end{cases}
\eea
For a more succinct view, the pair $\aprog=\big(b,\bra w\big)$ and the input $\ket x$ are often modified to
\[
\aprog  = b \bra 0 + \bra w =  b \bra 0 + \sum_{i=1}^d w_i\bra i = \bra{w_{b}}  \ \ \ \mbox{and}\ \ \  \ket{x_{1}} = - \ket 0 + \ket x = -\ket 0 + \sum_{i=1}^d\ket i x_i
\]
so that $\braket {w_{b}}{x_{1}} =-b + \braket w x$ and the interpretation in \eqref{eq:Falpha} 
boils down to 
\bea\label{eq:Fa}
\upsilon\ket x {\aprog}& = & \sgn\braket {w_{b}}{x_{1}}
\eea
where $\sgn$ is the sign function, reducing negative numbers to -1, positive to 1, and keeping 0 at 0.

\begin{figure}[!ht]
\begin{center}
\includegraphics[width=1\linewidth]{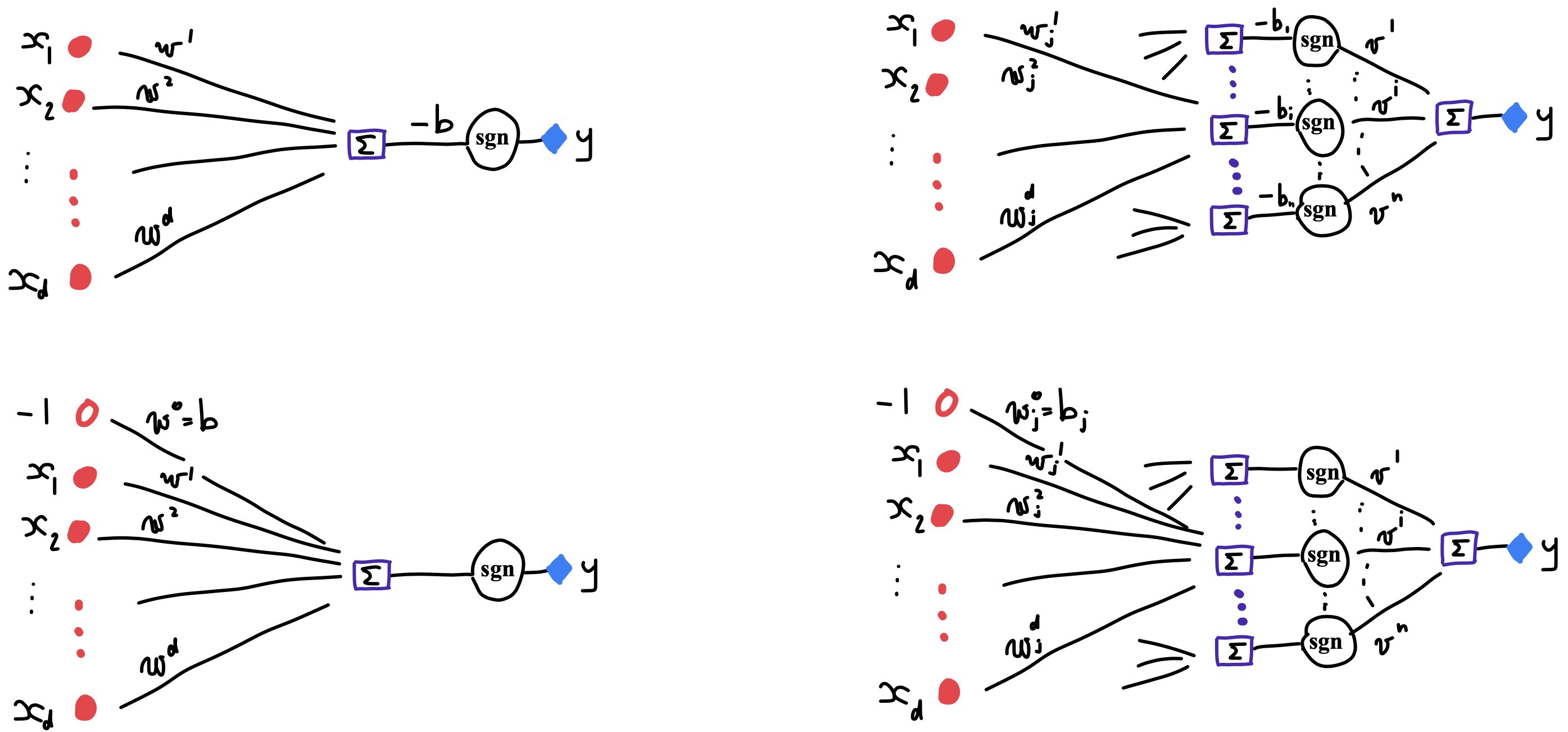}
\caption{A neuron and a perceptron as circuits}
\label{Fig:perceptron-circ}
\end{center}
\end{figure}

\paragraph{Perceptrons} are compositions of neurons. If a neuron is presented as a single row vector $\bra{w_{b}}$, then a perceptron  is an $(n+1)$-tuple of row vectors
\[\aprog =  \Big(\, \bra v,\ \bra{w_{1}},\ldots, \bra{w_{n}}\Big)\ \  \ \ \mbox{ where }\ \ \ \  \bra v = \sum_{j=1}^n v^{j}\bra j \ \ \ \ \mbox{ and }\ \ \ \  \bra{w_{j}} = b_{j} \bra 0  + \sum_{i=1}^d w_{j}^{i} \bra i 
\] 
which is interpreted by
\beq\label{eq:perceptron-compute}
\upsilon\ket x {\aprog} \ \ = \ \   \Big< v\ \big|\ \
\sum_{j=1}^{n}\ket j \sgn\braket{w_{j}}{x_{1}}
\Big>\ \ =\ \ \sum_{j=1}^{n} v^{j} \sgn\braket{w_{j}}{x_{1}}
\eeq 
For a more succinct view, the $n$-tuple of vectors $\bra{w_{1}},\ldots, \bra{w_{n}}$ of length $d+1$ can be arranged into the matrix
\[
W\ \  =\ \  \sum_{j=1}^{n} \ket j \braket j {w_{j}} \ \ = \ \ \sum_{i=0}^{d}\sum_{j=1}^{n} \ket j w_{j}^{i} \bra{i}
\]
so that the perceptron $\aprog =  \big(\, \bra v,\ \bra{w_{1}},\ldots, \bra{w_{n}}\big)$ can be written as the pair $\aprog =  \big(\, \bra v,\ W\big)$. The interpretation in \eqref{eq:perceptron-compute} now becomes
\beq\label{eq:perceptron-compute-two}
\upsilon\ket x {\aprog} =    \Big< v\ \big|\ \
\left(\sgn\right)^{n} W \ket x \Big>\qquad  \mbox{where}\qquad 
\left(\sgn\right)^{n} \sum_{j=1}^{n}\ket j z_{j}  =  \sum_{j=1}^{n}\ket j \sgn( z_{j})\notag
\eeq  

\paragraph{Summary.} Fig.~\ref{Fig:perceptron-circ} displays the two presentations of a neuron on the left and the two presentations of a perceptron on the right.  The first row shows the neuron and the perceptron in the original form, with the thresholds $b_{j}$. The second row shows the versions where the thresholds are absorbed in the weight projections $\bra{w_{j}}$ as the 0-th components $w^{0}_{j}=b_{j}$.

\paragraph{Perceptrons were a breakthrough into machine learning and inductive inference as two sides of the same coin.} Statistics provided the formal methods for hypothesis testing but left the task of learning and inferring hypotheses to informal methods and the magic of creativity. Perceptron training was the first formal method for inductive inference. Nowadays, this method looks obvious. The learner initiates the weights $\ket w$ and the thresholds $b$ to arbitrary values, runs the interpreter $\upsilon$ to generate predictions, compares them with the training data supplied by the supervisor $F$, and updates the weights proportionally to the losses $\LLL$. This didn't seem like a big deal even to Frank Rosenblatt, who wrote that
%
%
\begin{quote}
the perceptron program [was] \underline{not} primarily concerned with the invention of devices for ``artificial intelligence'', but rather with investigating the physical structures and neurodynamic principles which underlie ``natural intelligence''.
\end{quote}
Rosenblatt laid the stepping stone into machine learning while attempting to model the learning process in human brains. Even the very first learning machine was not purposefully designed but evolved spontaneously. 

It is often said that airplanes were not built by studying how the birds fly and that intelligent machines will not be built by looking inside people's heads. But there is more at hand. Perceptrons opened an alley into \textbf{learning as a \emph{universal computational process}}. \emph{Machine learning and human learning are just two  particular implementations of this universal process of learning}\footnote{Alan Turing derived this from the fact that machine computation and human computation are particular implementations of the same process, which he demonstrated early on.}, a natural process that evolves and diversifies. Machine learning models offer insights into a common denominator of all avatars of learning. The pattern of perceptron computation will be repeated on each of the models presented in the rest of this note.

\section{Learning functions}\label{Sec:fun}	

\subsection{Why learning is possible}
To understand why learning is possible, we first consider the special case when the channel $F$ is memoryless and deterministic: an ordinary function. 

\subsubsection{Learnable functions are continuous}
What can be learned about a function $F\colon X\to Y$ from a finite set of pairs  $$(x_{1}, y_{1}),\ (x_{2}, y_{2}), \ldots, (x_{n}, y_{n})$$ where $F(x_{i}) = y_{i}$? Generally nothing. Knowing $F(x)$ does not tell anything about $F(x')$, unless $x$ and $x'$ are related in some way, and it is moreover known that $F$ preserves their relation. To generalize the observed sample   $(x_{1}, y_{1}), \ldots, (x_{n}, y_{n})$ and predict a classification $F(x')=y'$ for an unobserved data item $x'$, it is necessary that
\begin{itemize}
\item $x'$ is related to $x_{1},\ldots, x_{n}$,
\item $y'$ is related to $y_{1},\ldots, y_{n}$, where $y_{i}=F(x_{i})$ for $i=1,\ldots, n$, and 
\item $F$ preserves their relationships. 
\end{itemize}
If the sets of the $x$s and the $y$s in such relationships are viewed as \emph{neighborhoods}, then the datatype $X$ and the classifier type $Y$ become \emph{topological}\/ spaces. The neighborhoods form topologies. Don't worry if you don't know the formal definition of a topology. It is just an abstract way to say that $x$ and $x?$ live in the same neighborhood. A function $F\colon X\to Y$ is \emph{continuous} when it maps neighbors to neighbors. Two words with similar meanings live in a semantical neighborhood. Any kind of relation can be expressed in terms of neighborhoods. So if $x'$ is related with $x_{1}$ and $x_{2}$, and $F$ is continuous, then $y'=F(x')$ is related with $y_{1} = F(x_{1})$ and $y_{2}=F(x_{2})$. That allows us to learn from the set of pairs
$(x_{1}, y_{1}), \ldots, (x_{n}, y_{n})$ where $F(x_{i}) = y_{i}$ that $F(x') = y'$ also holds. Then we can add the pair $(x',y')$ to the list as a prediction. Without the neighborhoods and the continuity, we cannot make such predictions.  \textbf{\emph{To be learnable, a function must be continuous.}}  

There are many ways in which this is used, and many details to work out. For the moment, just note that \emph{learning is based on associations}. You associate a set of names $X$ with a set of faces $Y$ along a continuous function $F:X\to Y$. You remember the face $F(\mbox{Allison})$ by searching through the pairs $(x_{1}, y_{1}),\ldots, (x_{n}, y_{n})$ where the names $x_{i}$ are associated with Allison's. Since $F$ is continuous, the faces $y_{i}=F(x_{i})$ must also be associated with Allison's. Therefore, if you find a face of a neighbor of Allison's name, then you can find Allison's face in the neighborhood of the face of Allison's neighbor. This is how \emph{associative memory}\/ works: as a family of continuous functions.The \emph{key-value associations}\/ in databases work similarly. Both in human memory and in databases, associative memory is implemented using referential neighborhoods. Functions are learnable when they preserve associations. They preserve associations when they are continuous.

\subsubsection{Continuous functions can be partially evaluated and linearly approximated}
The Fundamental Theorem of Calculus says, roughly, that every derivable function is the integral of its derivative. The integral approximates it, with arbitrary precision, by linear combinations of step functions that approximate the derivative. The Fundamental Theorem of Calculus, in essence, says that any derivable function can be linearly approximated by piecewise linear functions. 

A function that is just continuous may not be approximable by piecewise linear functions --- but it turns out that it can always be linearly approximated by pieces of an \emph{activation} function, which can be pretty much anything that is not a polynomial! Since it is  \emph{linarly}\/ approximated, the approximation is \emph{learnable}. Hence machine learning.

On the other hand, the approximability of continuous functions has remained one of the big secrets of calculus. The fact that \emph{\textbf{all continuous functions can be decomposed into sums of single-variable continuous functions}} defies most people's intutions. It says that, as far as computations are concerned, there are no genuine multi-dimensional phenomena among continuous functions. All those crazy multi-variable functions you may have seen in a vector calculus textbook, or encountered in practice if you are an engineer or scientist --- they can all be partially evaluated in each variable separately. Which is why they can be learned.

\subsection{Decomposing continuous functions: Kolmogorov-Arnold}\label{Sec:KA}
\paragraph{Hilbert's 13th Problem.} Back in 1900, the famous mathematician David Hilbert offered his famous list of 23 mathematical problems for the next century. Number 13 on the list was the question if all functions with 3 variables can be expressed by composing functions of 2 variables. Hilbert conjectured that a specific function, the formula for the solutions of the equation $x^{7}+ax^{3}+bx^{2}+cx + 1 = 0$ expressed in terms of the coefficients $a$, $b$, and $c$, could not be decomposed into functions of pairs of the coefficients. 56 years later, 19-year-old Vladimir Arnold disproved Hilbert's conjecture: he proved that all continuous functions with 3 variables can be decomposed into continuous functions with 2 variables. Next year, Arnold's thesis advisor Andrey Kolmogorov proved a stunning generalization. The theorem has been strengthened, and simplified ever since. Early simplifications were based on the following embedding of the $d$-dimensional cube into the $(2d+1)$-dimensional cube, constructed to allow separating $d$ variables in any continuous function\footnote{Since this part was written, a paper proposing a new family of neural networks, called the \emph{Kolmogorov-Arnold Networks (KAN)}\/ appeared on arxiv. The idea is natural, since the Kolmogorov-Arnold construction is closely related, as we will see in the next section, to the basic theory of neural network approximation. Remarkably, though, the proposers of the KAN approach make no use of the substantial mathematical and computational simplifications and improvements of Kolmogorov's 1957 construction, although they cite some of the papers with complete reference lists. Since several updates of the paper already appeared, the missed opportunities for improvement will presumably be taken in the future versions.}. 
\begin{figure}[!ht]
\begin{center}
\begin{tikzar}{}
\& \left[0,1\right]^{d} \ar[thick]{rrr}[description]{f} 
\ar[bend right=30,thin]{dl}[description]{\left(\psi_{i}^{d}\right)_{i=0}^{2d}}
\ar[thin]{dd}[description]{W}
\&\& \&\left[0,1\right]\\
\left(\left[0,1\right]^{d}\right)^{2d+1} \ar[bend right=30,thin]{dr}[description]{\bra w^{2d+1}} \\
\& \left[0,1\right]^{2d+1} \ar[thick]{rrr}[description]{\left(\varphi_{(f)}\right)^{2d+1}} \&\&\& \left[0,1\right]^{2d+1} \ar[thin]{uu}[description]{\bra v}
\end{tikzar}

\vspace{3ex}

\hspace{3em}\includegraphics[width=0.65\linewidth]{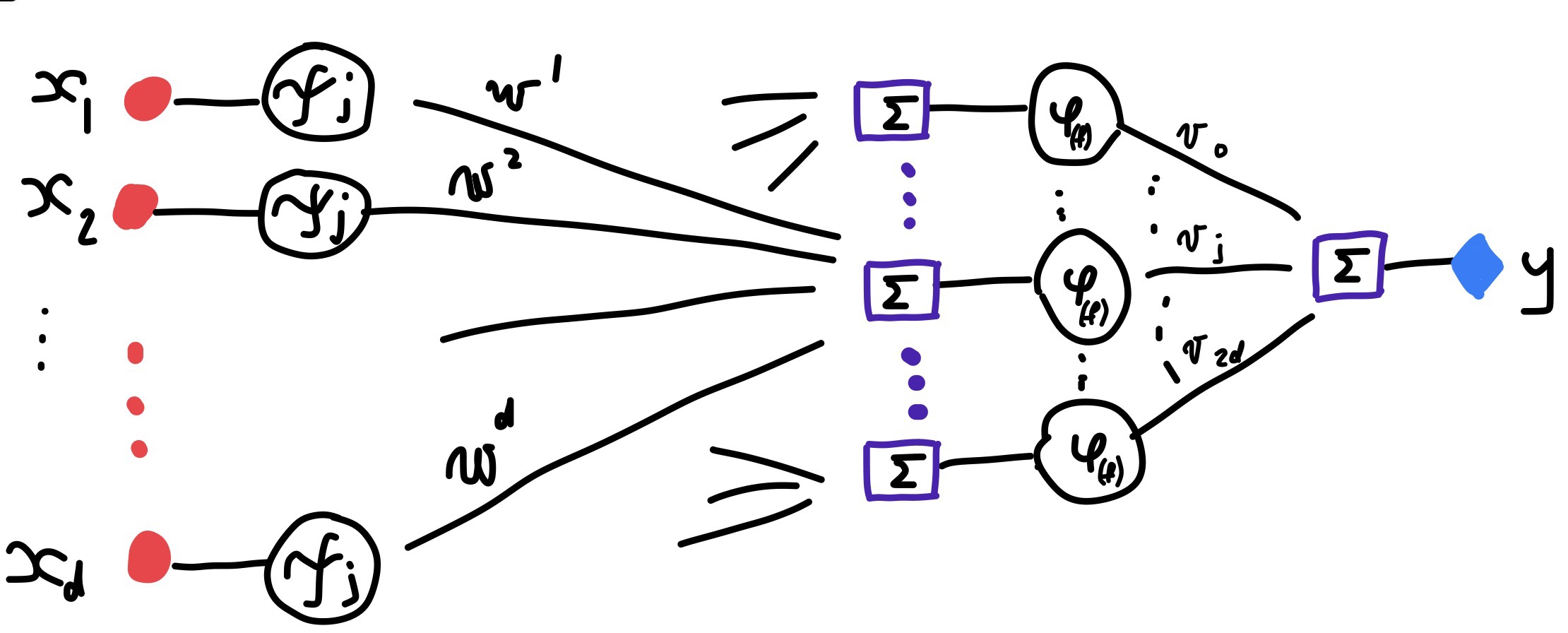}
\caption{The Kolmogorov-Arnold decomposition}
\label{Fig:kolmo-circ}
\end{center}
\end{figure}

\paragraph{Lorentz-Sprecher Embedding.}  
For any dimension $d\geq 2$ a vector $\bra w\in [0,1]^{d}$ and a family of continuous  functions $\psi_{j}\colon [0,1]\to [0,1]$, indexed over $j=0,1,\ldots, 2d$ together induce the embedding $W\colon [0,1]^{d}\mono [0,1]^{2d+1}$ defined
\beq\label{eq:KAW}
W\ket x  =     \sum_{j=0}^{2d}\ket j\Big<w \ \big|\  {\left(\psi_{j}\right)^{d}\ket x}\Big>\qquad  \mbox{ where}\qquad 
\left(\psi_{j}\right)^{d}\ket x  =  \sum_{i=1}^{d} \ket i \psi_{j}\left(x_{i}\right)
\eeq
With a fixed vector $\bra v \in[0,1]^{2d+1}$, the embedding $W$ yields the claimed decomposition.

\paragraph{Kolmogorov-Arnold (KA) Continuous Decomposition.} For any continuous  $f\colon [0,1]^{d}\to \left[0,1\right]$  there is a continuous $\varphi_{(f)}\colon [0,1]\to[0,1]$  such that for all $\ket z = \sum_{j=0}^{2d}\ket j z_{j}$ holds
\beq\label{eq:KA}
f\ket x =   \Big< v \ \ \big|\ \left(\varphi_{(f)}\right)^{2d+1}W\ket x\Big>\qquad \mbox{where}\qquad 
\left(\varphi_{(f)}\right)^{2d+1}\ket z  =  \sum_{j=0}^{2d}\ket j \varphi_{(f)}(z_{j})
\eeq
as illustrated in Fig.~\ref{Fig:kolmo-circ}. Note that only $\varphi_{(f)}$ depends on $f$, whereas $W$ and $ v$ are given globally, one pair for each dimension $d$. They are not unique and $W$ can be chosen so that the $ v=\sum_{j=0}^{2d}\bra j$.

\paragraph{Explanation.} In the old-style matrix notation,  \eqref{eq:KAW} is

\[
W\begin{pmatrix}x_{1}\\  
\vdots\\ x_{d}\end{pmatrix} \ \  = \ \    \sum_{j=0}^{2d}\begin{pmatrix}0\\\vdots\\j \\
\vdots\\ 0\end{pmatrix}\sum_{i=1}^{d}w^{i}\cdot \psi_{j}(x_{i}) 
\quad \mbox{ or } \quad
W \ \  = \ \  \begin{pmatrix}
w^{1}\cdot\psi_{0}(-) & w^{2}\cdot\psi_{0}(-) & \cdots & w^{d}\cdot\psi_{0}(-)\\
w^{1}\cdot\psi_{1}(-) & w^{2}\cdot\psi_{1}(-) & \cdots & w^{d}\cdot\psi_{1}(-)\\
w^{1}\cdot\psi_{2}(-) & w^{2}\cdot\psi_{2}(-) & \cdots & w^{d}\cdot\psi_{2}(-)\\
\hdotsfor{4}\\
w^{1}\cdot\psi_{2d}(-) & w^{2}\cdot\psi_{2d}(-) & \cdots & w^{d}\cdot\psi_{2d}(-)
\end{pmatrix}
\]


Unfolding \eqref{eq:KA} then gives 
\bea\label{eq:KA-unfold}
f(x_{1},x_{2},\ldots, x_{d}) \ \  = \ \ \sum_{i=0}^{2d} \varphi_{(f)}\left(\sum_{j=1}^{d}w^{j}\cdot \psi_{i}(x_{j})\right)
\eea
The constructions not only disproved Hilbert's conjecture, but still defy most people's geometric intuitions. The reason may be that we tend to think in terms of smooth functions and the functions $\psi$ and $\phi$ are fractal. They are constructed using copies of the Devil Staircase or space-filling curves. The geometric interpretation of the embedding $W$ is that the tuple $\left(\psi_{j} \right)_{j=0}^{2d}\colon [0,1]\to [0,1]^{2d+1}$ draws a curve in $[0,1]^{2d+1}$ and each of the $d$ dimensions of the cube $[0,1]^{d}$ is homeomorphic to a copy of that curve, spanning a homeomorphic image of $d$-dimensional cube within the $(2d+1)$-dimensional cube:
\[\prooftree
\left(\psi_{j} \right)_{j=0}^{2d}\colon [0,1]\to [0,1]^{2d+1}
\justifies 
\left(\psi^{d}_{j} \right)_{j=0}^{2d}\colon [0,1]^{d}\to \left([0,1]^{d}\right)^{2d+1}
\endprooftree
\]
This is the first component of $W$ in Fig.~\ref{Fig:kolmo-circ}. The projections of the vectors lying in this $d$-cube inside the $(2d+1)$-cube along $\bra w\colon [0,1]^{d}\to [0,1]$ mix the copies of the curve $\left(\psi_{j} \right)$ in such a way that the inverse can be used to iteratively fill the $d$-cube. Kolmogorov's original construction partitioned the mapping $f$ out of  $[0,1]^{d}$ among parts of this curve in $[0,1]^{2d+1}$, and combined $d$ different functions  $\varphi_{(f)}^{1}, \varphi_{(f)}^{2},\ldots,\varphi_{(f)}^{d}$ to represent $f$. Sprecher and Lorentz later noticed that additional stretching allows capturing all parts of $f$ by a single $\varphi_{(f)}$. This is possible because the dependency of $f$ on each of its $d$ variables can be approximated with arbitrary precision on a null-subset of its domain, and the null-subsets of $[0,1]$ can be made disjoint.The upshot is that \emph{the only genuinely multi-variable continuous function is the addition}. The multiple inputs for all multi-variable continuous functions can always be preprocessed in such a way that each input can then be processed separately, by a single-variable function. The output of the multi-variable function is obtained by adding up the outputs of the single-variable function. \textbf{\emph{Continuous functions can be partially evaluated: each input separately.}}

The price to be paid is that the single-variable continuous functions that perform the preprocessing and the processing are complicated, ineffective, and approximated. For a long time, the iterative fugue of Kolmogorov's proof was viewed as a glimpse from the darkness of a world of complexities beyond our use and imagination. Until it was noticed that the Kolmogorov-Arnold decomposition in  \eqref{eq:KA} followed the same pattern as the perceptron computation in \eqref{eq:perceptron-compute}. Comparing the circuit view of \eqref{eq:KA} in Fig.~\ref{Fig:kolmo-circ} also displays the difference from the circuit view of \eqref{eq:perceptron-compute} in  Fig.~\ref{Fig:perceptron-circ}. What is going on?

\subsection{Wide learning}
\subsubsection{From decomposition theory to approximation practice}
In the preceding section, we studied a  \emph{\textbf{theoretic}}\/ construction providing 
\begin{itemize}
\item an \emph{\textbf{exact} representation}\/ $f\ket x = \Big< v\ \big|\ \Big(\varphi_{(f)}\Big)^{2d+1}W\ket x\Big>$, using
\item an \emph{\textbf{approximate} construction}\/ of 
\begin{itemize}
\item a projection-embedding pair $(\bra v,W)$, \emph{\textbf{independent} on $f$}, and
\item a continuous function $\varphi_{(f)}\colon[0,1]\to[0,1]$, \emph{\textbf{dependent} on $f$}. 
\end{itemize}

\end{itemize}

In the present section, we turn to a \emph{\textbf{practical}}\/  construction providing 
\begin{itemize}
\item an \emph{\textbf{approximate} representation}\/ $f\ket x \approx \Big<v_{(f)}\ \big|\ \Big(\sigma\Big)^{n}W_{(f)}\ket x\Big>$, using
\item an \emph{\textbf{exact} construction} of \begin{itemize}
\item a projection-embedding pair $\left(W_{(f)},\bra {v_{(f)}}\right)$, \emph{\textbf{dependent} on $f$}, and
\item a continuous function $\sigma \colon[0,1]\to[0,1]$,  \emph{\textbf{independent} on $f$}.
\end{itemize}

\end{itemize} 

The step from the Continuous Decomposition from Sec.~\ref{Sec:KA} to the Neural Approximation in Sec.~\ref{Sec:CHSW} is illustrated by comparing of Fig.~\ref{Fig:kolmo-circ} and Fig.~\ref{Fig:NN-circ}. Letting $W_{(f)}$ vary with $f$ allows omitting the deformations $\psi$. 
Letting $\bra{v_{(f)}}$ vary with $f$ allows replacing $\varphi_{(f)}$ with a fixed \emph{activation function} $\sigma$, independent on $f$. 

\subsubsection{Approximating continuous functions}\label{Sec:CHSW}
\paragraph{Activation functions.} The Neural Approximation theorem below states that any continuous function can be approximated by linear combinations of a fixed activation function $\sigma$. All we need from this function is that it restricts to a homeomorphism between two closed real intervals \emph{not representable by a polynomial}. The construction can be set up to only use the part that establishes this continuous, monotonic bijection of the intervals. That part can be conveniently renormalized to a sigmoid: a homeomorphism of the extended real line and the interval $[0,1]$. Early neural networks used the logistic sigmoid $\frac{e^{x}}{1+e^{x}}$, which readily establishes that homeomorphism. The hyperbolic tangent and arcus tangent were also used, suitably renormalized. Nowadays the function $\max(0, x)$ is preferred. Its original designation as "Rectified Linear Unit" got mellowed down to \emph{ReLU}, a nickname shared with small pets. The Neural Approximation construction fails if the activation function is representable by a polynomial. This obviously precludes all linear functions? --- \emph{?but}\/ already a continuous combination of two linear functions works fine, as ReLU shows,   combining the constant 0 below 0 and the identity above 0.
\begin{figure}
\begin{center}
\begin{tikzar}[column sep = 7ex,row sep= 5ex]
\left[0,1\right]^{d} \ar[thick]{rr}[description]{f} 
\ar[thick]{dd}[description]{W_{(f)}}
\&\&  \left[0,1\right]
\\
\& \mbox{\large $\approx$}
\\
\left[0,1\right]^{n} \ar[thin]{rr}[description]{\left(\sigma\right)^{n}} \&\& \left[0,1\right]^{n} \ar[thick]{uu}[description]{\bra{v_{(f)}}}
\end{tikzar}

\vspace{3ex}

\includegraphics[width=0.55\linewidth]{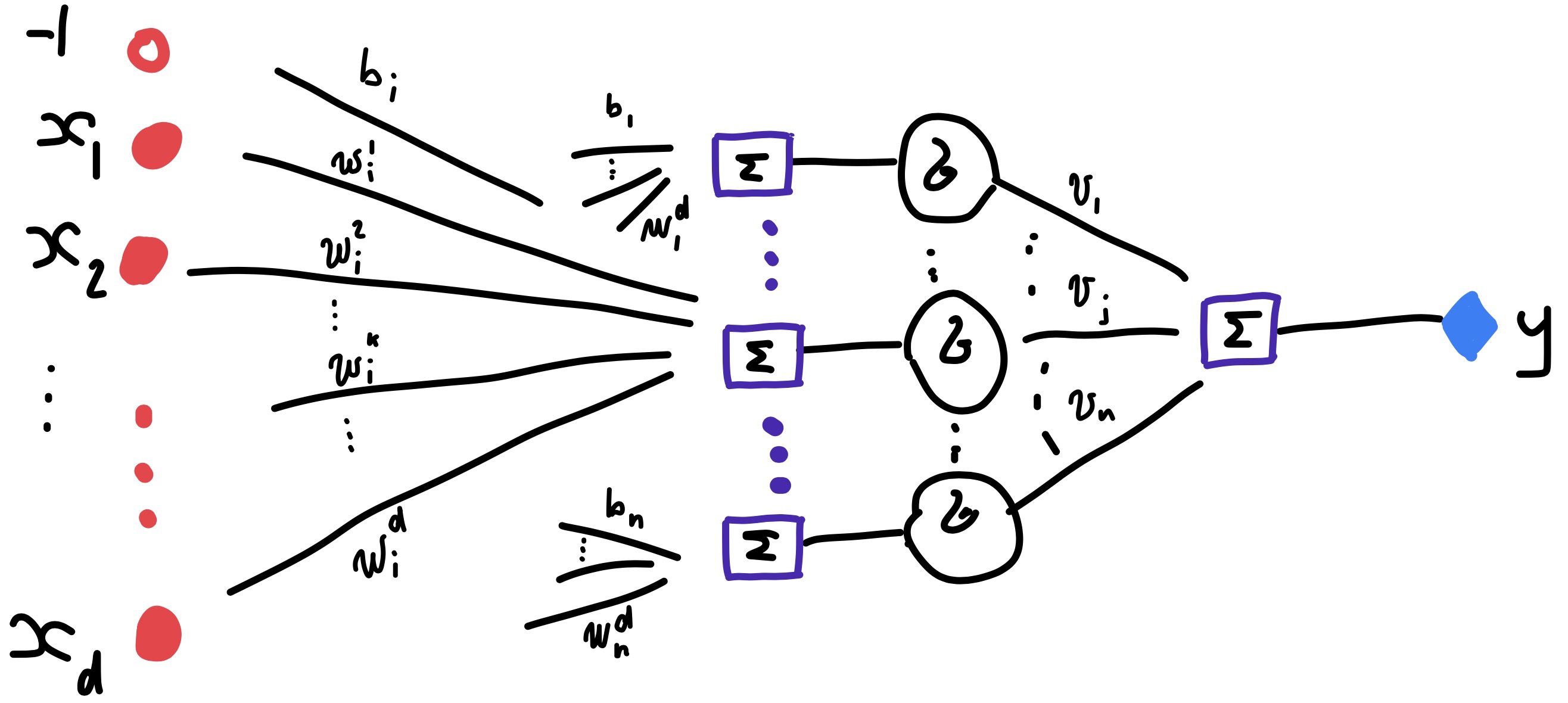}
\caption{$\sigma$-neuron}
\label{Fig:NN-circ}
\end{center}
\end{figure}

\paragraph{Cybenko-Harnik-Stinchcombe-White (CHSW) Neural Approximation.} For any $d\geq 2$, any continuous $f\colon [0,1]^{d}\to \left[0,1\right]$, and any $\varepsilon\gt 0$ there is a number $n=n_{(d\varepsilon)}$ such that $f$ induces a model  $\fprog =\left(W_{(f)},\ket{v_{(f)}}\right)$ where
\begin{itemize} 
\item $W_{(f)}\colon [0,1]^d \to [0,1]^n$ is a linear operator with $
W_{(f)}\ket x =  \sum_{j=1}^{n}\ket i\braket{w_{(f)j}}{x}$, and 
\item $\bra {v_{(f)}}\in [0,1]^n$  is a vector with $\bra{v_{(f)}}  = \sum_{j=1}^{n}v_{(f)}^j\bra j $
\end{itemize}
The intended interpretation of the model is
\beq\label{eq:Cyb}
\upsilon\ket x\fprog  =   \Big< v_{(f)} \ \ \big|\ \left(\sigma\right)^{n}W_{(f)}\ket x\Big>\qquad  \mbox{where}\qquad
\left(\sigma\right)^{n}\ket z  =  \sum_{j=0}^{n}\ket j \sigma(z_{j})
\eeq
The claim is that $\fprog$  approximates $f$ up to $\varepsilon$, in the sense that all $\ket x\in [0,1]^d$ satisfy
\bea\label{eq:approx}
\Big\lvert f\ket x - \upsilon \ket x\fprog \Big\rvert & \lt & \varepsilon
\eea
Since both functions are continuous, \eqref{eq:approx} is equivalent to the claim that for every $\varepsilon \gt 0$ there is $\delta = \delta_{(\varepsilon)}\gt 0$ such that 
\bea\label{eq:approx-bound}
\Big\lvert \ket x - \ket {x'}\Big\rvert \lt \delta & \implies & \Big\lvert f\ket x - \upsilon \ket{x'}\fprog\Big\rvert \lt \varepsilon
\eea
Neurons are thus \emph{universal approximators}\/ of continuous functions, in the sense that for every continuous $f\colon [0,1]^{d}\to \left[0,1\right]$ there is a neuron $\fprog = \left(W_{(f)},\bra{v_{(f)}}\right)$ such that $f\ket x \approx \upsilon\ket x\fprog$, with arbitrarily precision.

The \textbf{proofs} of different versions of the Neural Approximation Theorem were published by Cybenko and by Harnik-Stinchcombe-White independently, both in 1989. In the meantime, the neural approximations have been widely used and various other versions, views, and overviews have been provided. The overarching insight links the CHSW-approximation and the KA-decomposition in a computational framework that seems to have taken both beyond the original motivations. 


\paragraph{Continuous functions can be approximated because their variables can be separated.}  In computational terms, this means that continuous functions can be partially evaluated. That makes them learnable. 
Unfolding \eqref{eq:Cyb} and \eqref{eq:KA} in parallel displays the common pattern yet again:

\[
\prooftree\prooftree\prooftree
\Big(x_{1},\  \ x_{2},\ \ldots,\  x_{d}\Big)
\justifies
\left(\sum_{i=1}^{d}w_{(f)1}^{i}x_{i}\ ,\ \ \ldots,\  \ \sum_{i=1}^{d}w_{(f)n}^{i}x_{i}\right)
\endprooftree 
\justifies 
\left(\sigma\sum_{i=1}^{d}w_{(f)1}^{i}x_{i}\ ,\ \ \ldots,\ \  \sigma \sum_{i=1}^{d}w_{(f)n}^{i}x_{i}\right)
\endprooftree 
\justifies 
f \left(x_{1},\ldots, x_{d}\right) \ \approx\ \sum_{j=1}^{n}v_{(f)}^{j }\sigma\sum_{i=1}^{d}w_{(f)j}^{i}x_{i} 
\endprooftree 
\qquad \qquad 
\prooftree\prooftree\prooftree
\Big(x_{1},\  \ x_{2},\ \ldots,\  x_{d}\Big)
\justifies
\left(\sum_{i=1}^{d}w^{i}\psi_{0}(x_{i})\ ,\ \ \ldots,\  \ \sum_{i=1}^{d}w^{i}\psi_{2d}(x_{i})\right)
\endprooftree 
\justifies 
\left(\varphi_{(f)}\sum_{i=1}^{d}w^{i}\psi_{0}(x_{i})\ ,\ \ \ldots,\ \  \varphi_{(f)} \sum_{i=1}^{d}w^{i}\psi_{2d}(x_{i})\right)
\endprooftree 
\justifies 
f \left(x_{1}, \ldots, x_{d}\right) \ =\ \sum_{j=0}^{2d}v^{j }\varphi_{(f)}\sum_{i=1}^{d}w^{i}\psi_{j}(x_{i}) 
\endprooftree 
\]

The same parallel patterns are seen by aligning \eqref{eq:KA} and \eqref{eq:Cyb}, or Figures \ref{Fig:kolmo-circ} and \ref{Fig:NN-circ}. But note the differences. The first difference is that in the approximation on the left, $w_{(f)}$ and $v_{(f)}$ depend on $f$, whereas on the right only $\varphi_{(f)}$ depends on $f$. The second difference is that the number of partially evaluated variables, for a fixed input length $d$, is fixed at $(2d + 1)$ in the decomposition on the right, whereas in the approximation on the left, $n = n(\varepsilon)$ depends on the approximation error $\varepsilon$. That is an important point.

%

The dimension $n$ of the inner space $[0,1]^{n}$ is the \textbf{\emph{width}\/ of the neuron}. The Neural Approximation Theorem says that for any continuous function there is a wide enough neuron that will approximate it up to a required precision. This is the essence of \textbf{wide learning}. The idea of approximating $f$ by a linear combination of copies of $\sigma$ is similar to Lebesgue's idea to approximate an integrable function by a linear combination of step functions. In both cases, the arbitrarily close approximations are achieved by increasing the number of approximants $n$. 

\begin{figure}
\begin{center}
\includegraphics[width=1\linewidth]{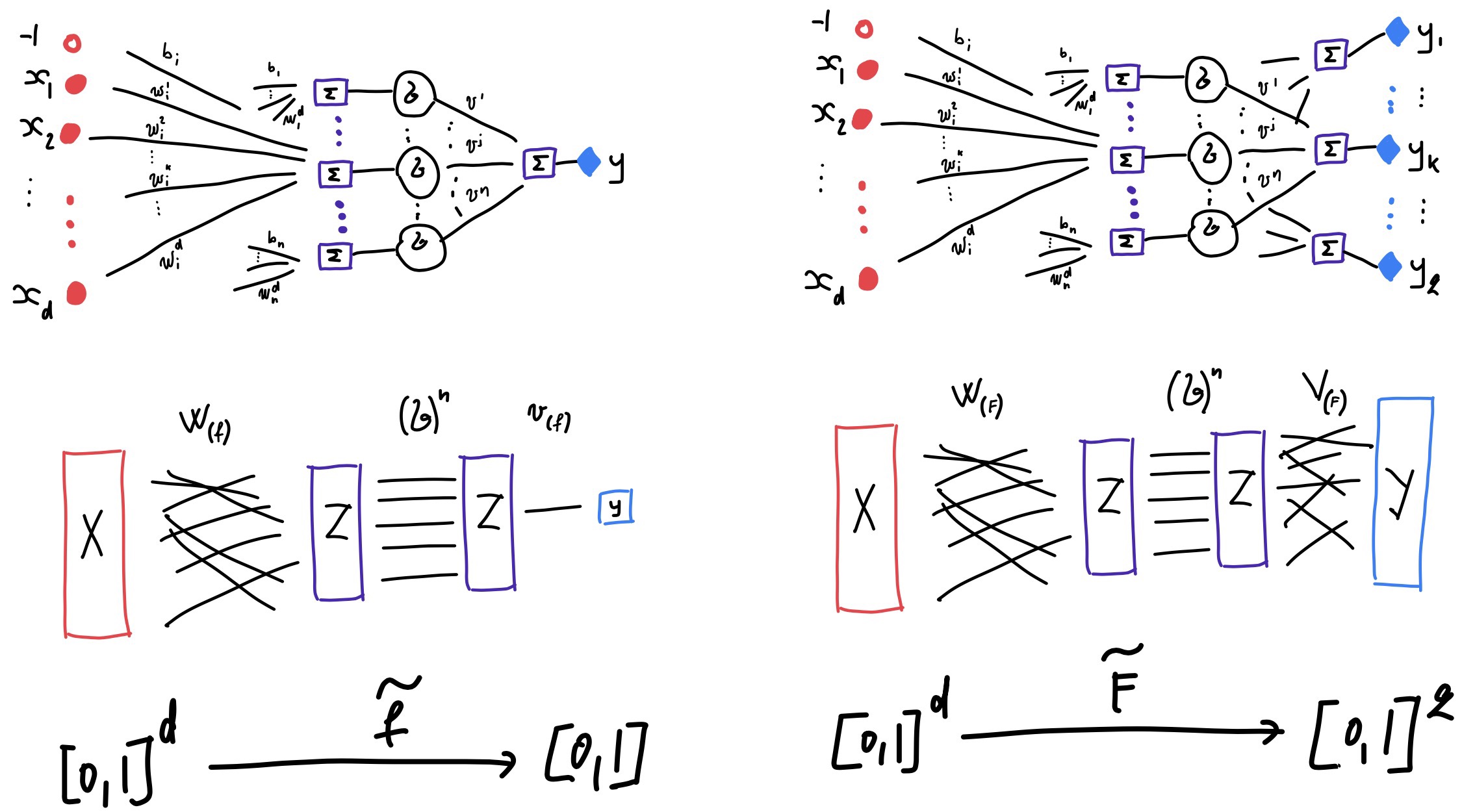}
\caption{A $\sigma$-neuron and a single-layer $n$-wide neural network}
\label{Fig:NNs}
\end{center}
\end{figure}

\paragraph{Wide neural networks.} Everything stated about continuous real functions lifts without much ado to continuous vector functions, which are in finite dimensions just tuples of continuous real functions. The approximations by $\sigma$-neurons lift to tuples of $\sigma$-neurons, a.k.a. the \emph{single-layer neural networks}.  Fig.~\ref{Fig:NNs} displays the tuplings of functions and of the corresponding neurons on the right.
For completeness, the corresponding views of the neuron are displayed on the left.  A $q$-tuple of neurons $(W, \bra{v_{j}})$ for $j=1,\ldots,q$ bundled together gives a single-layer neural network $\aprog=(W, \bra{v_{1}}, \bra{v_{2}},\ldots, \bra{v_{q}})$, more succinctly written $\aprog=(W, V)$, where $V$ is the matrix with $\bra{v_{j}}$s as rows, like before. The Neural Approximation Theorem implies that for every continuous vector function there is a wide enough single-layer neural network that approximates it arbitrarily close. The term \emph{wide neural network}\/ usually refers to a single-layer network. The circuit view in the top row of Fig.~\ref{Fig:NNs} is aligned with the format  where the layers of variables are enclosed in boxes. This will become necessary when the networks become deep. 

\subsection{Deep learning}
\subsubsection{Scaling up}
The trouble with wide learning is that there are simple functions, e.g. representing hemispheres or parts of ellipsoids, where separating the variables is hard, and the number of separable variables $n=n_{(d\varepsilon)}$ increases exponentially with the dimension $d$. This exponentially growing number is the width of the approximating neural network. It follows that, although any continuous function can be approximated by a single-layer neural network, the approximations are in the worst cases intractable, since the width of the network may explode exponentially in the number of variables. There are also problems concerning the generalization power of single-layer networks, i.e. how much training data they need before they can extrapolate predictions\footnote{In their seminal critical book ``Perceptrons'', Minsky and Papert proved that the coefficients of a perceptron representing a boolean function are always invariant under the actions of a group under which the function is invariant  itself. Since a perceptron therefore cannot tell apart the functions that are equivariant under the group actions, this was viewed as a no-go theorem. While the Minsky-Papert construction lifts from perceptrons and boolean functions to wide neural networks and continuous functions by standard methods, the resulting group invariances are nowadays viewed as proofs that the glass of neural approximations is half-full, not that it is half-empty.}.

%
%
%
%

\subsubsection{Narrowing by deepening} 
\begin{figure}
\begin{center}
\includegraphics[width=1\linewidth]{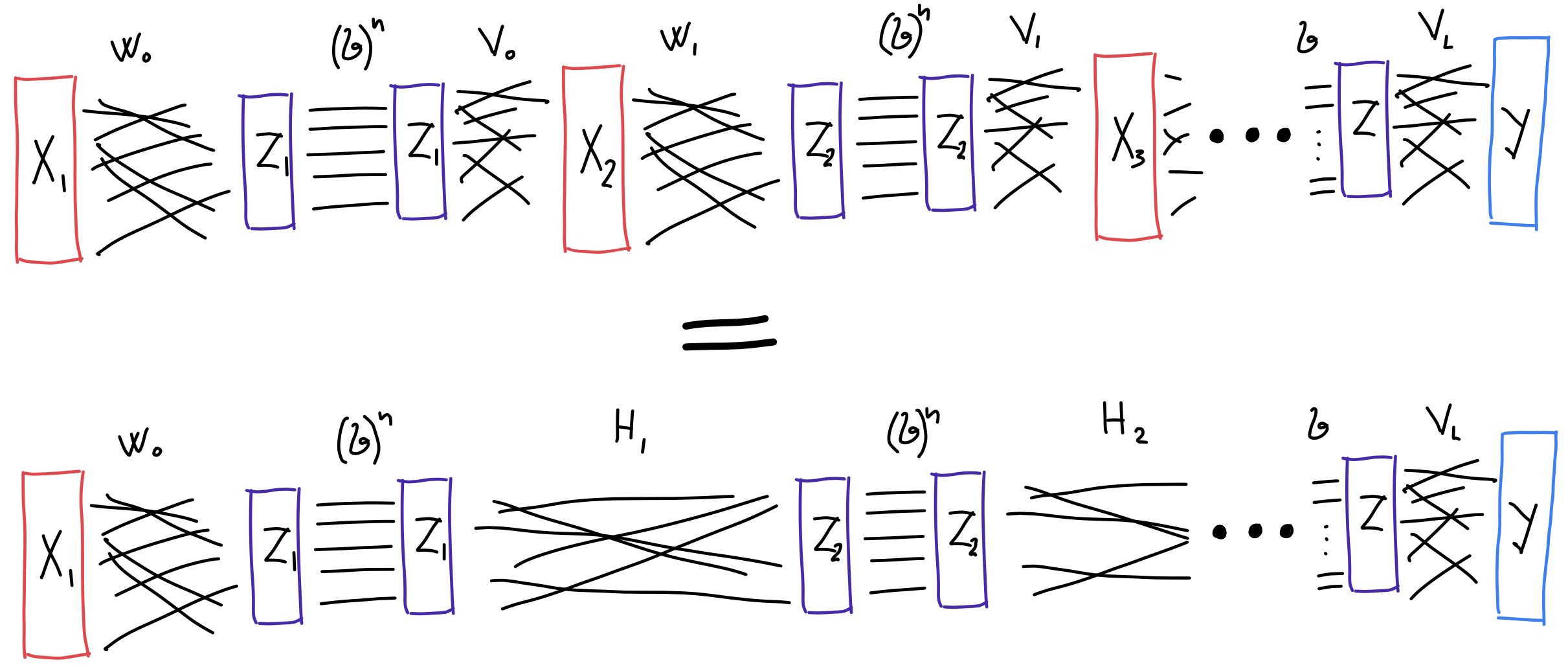}
\caption{The circuit schema of deep neural networks}
\label{Fig:NN-deep}
\end{center}
\end{figure}

The general idea of approximating a function $f$ is to find an algorithm to transform the data
\[
\begin{tikzar}[column sep = 4ex,row sep = 0ex]
\Big(x_{1},\ldots x_{d}\Big)\ar[mapsto]{r} \&\Big(z^{1}_{1},\ldots, z^{1}_{n}\Big) \ar[mapsto]{r} \& \Big(z^{2}_{1},\ldots, z^{2}_{n}\Big) \ar[mapsto]{r}\& \cdots \ar[mapsto]{r}\& \Big(z^{L}_{1},\ldots, z^{L}_{n}\Big)
\end{tikzar}
\]
and determine an approximator $\widetilde f$ satisfying
\bear
\lvert f(x_{1},\ldots, x_{d}) - \widetilde f(z^{i}_{0},\ldots, z^{i}_{n})\rvert & \lt & \varepsilon_{i} \lt \cdots \lt \varepsilon_{2} \lt\varepsilon_{1} \mbox{ for } i=1,2,\ldots, \mbox{ descending to }\\
\lvert f(x_{1},\ldots, x_{d}) - \widetilde f(z^{L}_{0},\ldots, z^{L}_{n})\rvert & \lt & \varepsilon = \varepsilon_{L} \lt \cdots \lt \varepsilon_{2} \lt\varepsilon_{1}
\eear
for a desired precision $\varepsilon$.
%
The exponential growth of the width $n$ of single-layer neural networks is tempered by descending through layers of deep neural networks, as depicted in Fig.~\ref{Fig:NN-deep}.
At each inner layer $\ell=1,\ldots,L$, composing the input transformation $W_{\ell}$ with the output $V_{\ell-1}$ of the preceding layer gives a transformation $H_{\ell}= W_{\ell}V_{\ell-1}$, which is linear again, and can be trained directly, forgetting about the $W$s from the $V$s, as shown in  \eqref{eq:NN-deep-diag}. Deep neural networks are thus programs in the form $\aprog = (W, H_{1},H_{2},\ldots,H_{L},V)$. 
\beq\label{eq:NN-deep-diag}
\begin{tikzar}[column sep = 3.5ex]
\left[0,1\right]^{d} \ar[thick]{rrrrrrr}[description]{F}
\ar[thick,bend right=13]{rrrrrrr}{\raisebox{2ex}{\normalsize$\approx$}}[description]{\upsilon(-)\aprog}
\ar[thick]{dddd}[description]{W_{0}}
\&\& \&\& \&\& \&  \left[0,1\right]^{q}
\\
\\
\& \left[0,1\right]^{n} \ar[thick]{ddr}[description]{W_{1}}
\&\& \left[0,1\right]^{n} \ar[thick,dotted]{ddr}[description]{W_{2}}
\& \&
\left[0,1\right]^{n} \ar[thick]{ddr}[description]{W_{L}}
\\
\\
\left[0,1\right]^{n} \ar[thin]{r}[swap] {\left(\sigma\right)^{n}} 
\& 
\left[0,1\right]^{n} \ar[thick]{uu}[description]{V_{0}} \ar[thick]{r}[swap] {H_{1}}
\& 
\left[0,1\right]^{n} \ar[thin]{r}[swap] {\left(\sigma\right)^{n}} 
\& 
\left[0,1\right]^{n} \ar[thick]{uu}[description]{V_{1}} \ar[thick,dotted]{r}[swap] {H_{2}}
\&
\cdots \ar[dotted]{r}[swap] {\left(\sigma\right)^{n}} 
\& 
\left[0,1\right]^{n} \ar[thick]{uu}[description]{V_{L-1}} \ar[thick]{r}[swap] {H_{L}}  
\& 
\left[0,1\right]^{n} \ar[thin]{r}[swap] {\left(\sigma\right)^{n}} 
\& 
\left[0,1\right]^{n} \ar[thick]{uuuu}[description]{V_{L}}
\end{tikzar}
\eeq

\subsection{Neural networks are learnable programs} 
If the learning process in  Fig.~\ref{Fig:learning} is construed as a process of computation, the learners play the role of programmers. To learn a function $F$ means to converge to a program $\aprog$ whose executions $\upsilon(x){\aprog}$ approximate $F(x)$. Note that in programming, the approximation up to the loss $\LLL(F(x), \upsilon(x)\aprog)$ ideally means $F(x)=\upsilon(x)\aprog$. In reality, this depends on the software assurance process, which can be viewed as a special case of the loss function $\LLL$. 
\begin{figure}[!hb]
\begin{center}
\includegraphics[width=0.85\linewidth]{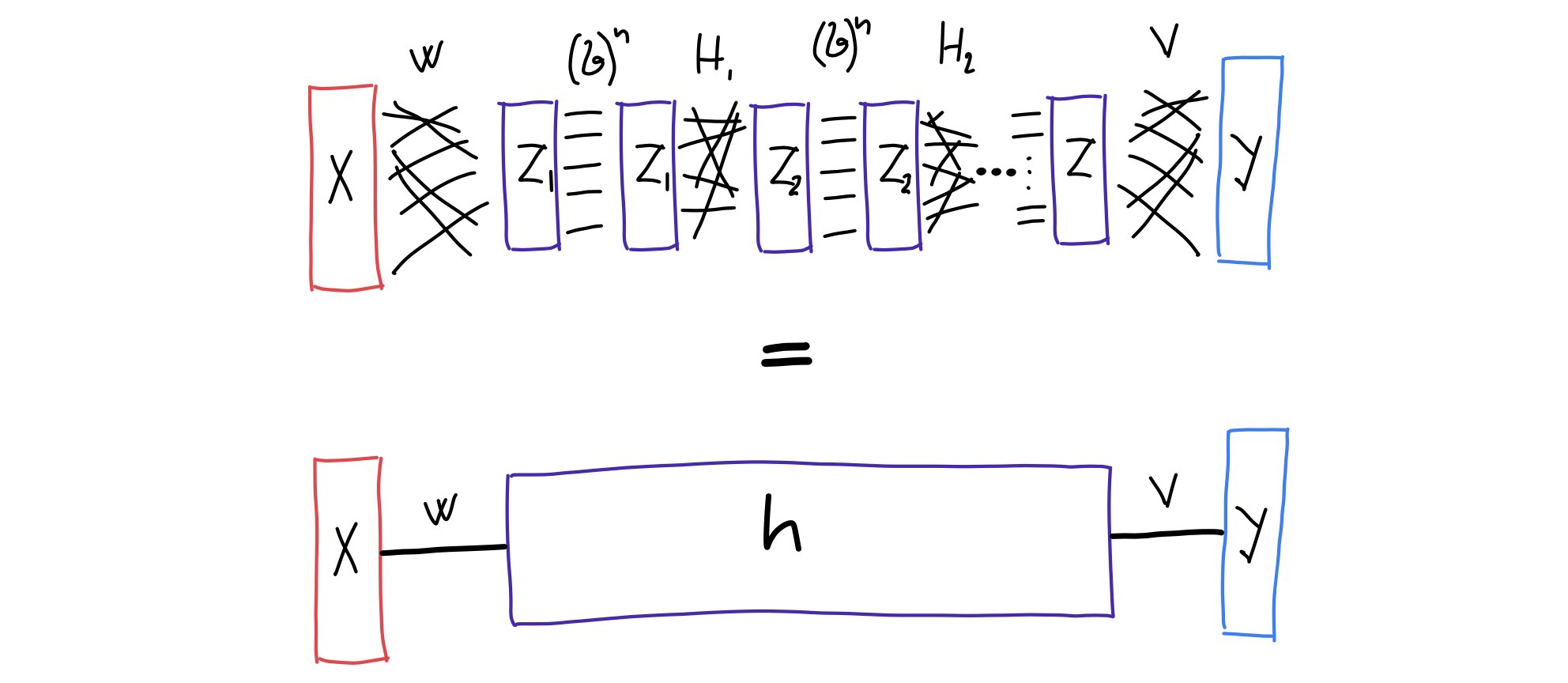}
\caption{Boxes enclose subnetworks and functional components}
\label{Fig:NN-h}
\end{center}
\end{figure}

We have already seen many of the main features of the syntax of neural networks as a programming language. A single neuron $\aprog = \bra w$. A single-layer network $\aprog=(W,V)$ is a single-instruction program. A deep network $\aprog = (W, H_{1},H_{2},\ldots,H_{L},V)$ is a general program. Its inner layers are the program instructions. For simplicity, the inner layers are often bundled under a common name, say  $\hprog = (H_{1},H_{2},\ldots,H_{L})$. A general neural program is in the form $\aprog = (W, \hprog,V)$ drawn as in Fig.~\ref{Fig:NN-h}.  



\section{Learning channels and paying attention}

\paragraph{The trouble with applying function learning to language} is that language is context-sensitive: the word ``up'', for instance, means one thing in ``shut up'' and another thing in ``cheer up''. We talked about this in the \emph{Beyond sintax} section of the \emph{Syntax}\/ part. A function is required to assign to each input the same unique output in all contexts. Meaning is not a function but a communication channel, assigning to each context a probability distribution over the concepts $y$:
$$\pder{\mbox{shut up}}{y}\qquad\qquad \pder{\mbox{cheer up}}{y}$$
In the \emph{Semantics}\/ part, we saw how concepts are modeled as vectors, usually linear combinations of words. Meaning is thus a random variable $Y$, sampled over the concept vectors $y$. There is an overview of the channel formalism in the \emph{Dynamic semantics}\/ section of the \emph{Semantics}\/ part. When there is no channel feedback, the context is the channel source
\[ \ppder{X_{1}}\qquad\qquad
\pder {X_{1} X_{2} \ldots X_{m}}{X_{m+1}}  \ \ \ \mbox{ for }  m=1,2,\ldots
\]
and the channel outputs are sampled according to the probabilities 
\bear
F&\colon &
\pder {X_{1} X_{2}\cdots X_{m}}{Y_{m}}  \  \ \ \ \mbox{ for }  m=1,2,3,\ldots
\eear
You can think of the source\footnote{Recall that $X^{n}=\left(X_{1}X_{2}\cdots X_{m}\right)$ for $n\geq 1$ and $X^{0} = 1$, so that $\pder{X^{0}}{X_{1}} = \ppder{X_{1}}$.} $X = \pder{X^{n}}{X_{m+1}}_{m=0,1,\ldots}$ as a text (a sequence of words or tokens in a language) and of the channel $F\colon \pder{X^{n}}{Y_{m}}_{m=1,2,\ldots}$ as the process of translating the text $X$ to a text $Y$ in another language:
\[
\mbox{Mandarin}\tto{\ \ F\ \ }  \mbox{French}\] 
Similar interpretations subsume meaning, syntactic typing, classification, and generation under the channel model. The common denominator is the context dependency, be it syntactic or semantic, deterministic or stochastic. Semantic references can be remote. The meaning of a sentence in a novel may depend on a context from 800 pages earlier. The meaning that you assign to something that an old friend says may be based on a model of their personality from years ago. To make it more complicated, remote references and long established channel models may change from context to context, whenever new information becomes available.

\subsection{Channeling through concepts}\label{Sec:lerning-mining} 
\begin{figure}
\begin{center}
\includegraphics[width=1\linewidth]{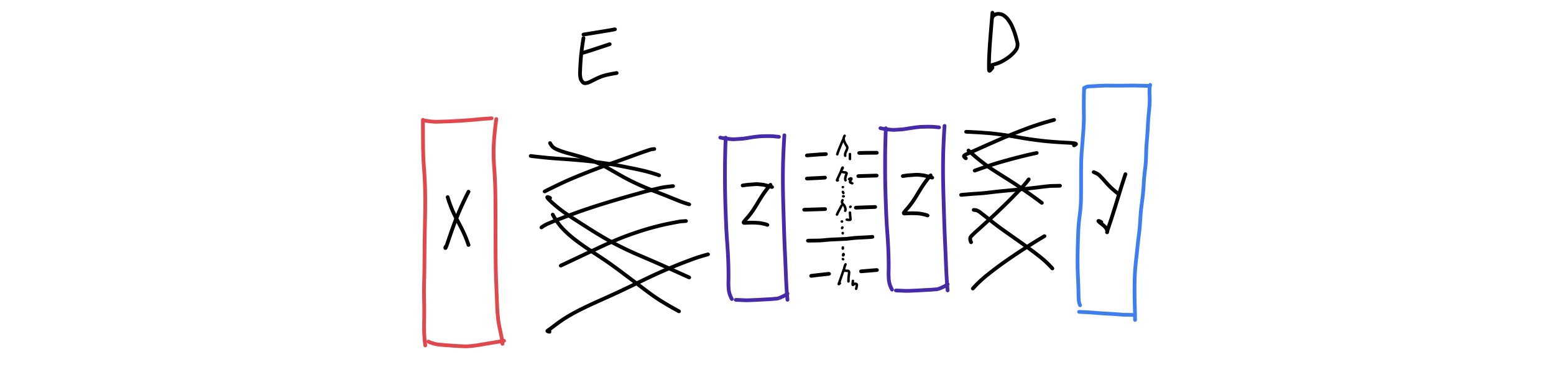}
\caption{The circuit schema view of mining for latent concepts}
\label{Fig:LSA-schema}
\end{center}
\end{figure}

In different languages, semantical references are mapped to syntactic references in different ways. Mapping a Mandarin phrase to a French phrase requires deviating from the syntactic dependency mechanisms of the two languages, as studied in the Syntax part. Good translators first understand the phrase in one language and then express what they understood in the other language. It is a two-stage process:
\[
\mbox{Mandarin}\tto{\ \ E\ \ } \mbox{concepts} \tto{\ \ D\ \ } \mbox{French}\] 
$E$ is a \emph{concept encoding} map, whereas $D$ is \emph{concept decoding}. Similar factorizations through concepts came up in Sec.~2.2.3 of the \emph{Semantics}\/ part, as instances of concept mining by \emph{Singular Value Decomposition (SVD)}, displayed in Fig.~\ref{Fig:LSA-schema}. 
The concepts that are latent in a given data matrix $M$ are mined as its singular values $\lambda_{i}$. 
\begin{figure}
\begin{center}
\begin{tikzar}[column sep = 5ex,row sep= 4ex]
\left[0,1\right]^{d} \ar{rrr}[description]{M} 
\ar{ddr}[description]{E}
\&\&\&  \left[0,1\right]^{q}
\\
\\
\&\left[0,1\right]^{n} \ar{r}[description]{\Lambda} \& \left[0,1\right]^{n} \ar{uur}[description]{D}
\end{tikzar}

\vspace{3ex}

{\includegraphics[width=0.55\linewidth]{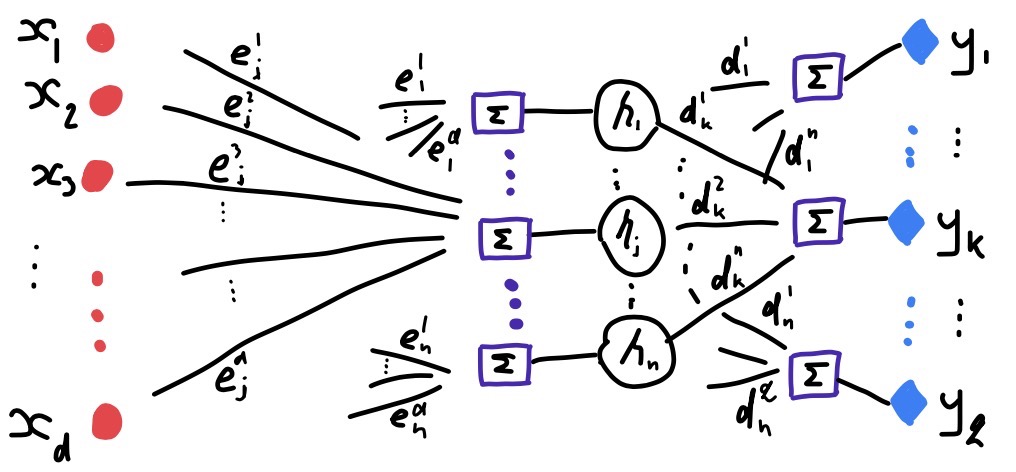}}
\caption{Concept mining: the Singular Value Decomposition and its circuit view}
\label{Fig:LSA}
\end{center}
\end{figure}
Now compare the neural architectures in Figures \ref{Fig:NN-circ} and \ref{Fig:NNs} with the SVD-mining in Fig.~\ref{Fig:LSA}. A neural network approximates a continuous function by separating its variables and approximating the impact of each of them by a separate copy of the activation function $\sigma$. The SVD algorithm decomposes a data matrix through a canonical basis eigenspaces, corresponding to the singular values of the matrix, viewed as the dominant concepts, spanning the concept space. The eigenspaces in the SVD are mutually orthogonal. The action of the data matrix boils down to multiplying each of them separately by the corresponding singular value $\lambda_{i}$. Both the neural network approximation and the SVD mine minimally correlated, separable, orthogonal subspaces. The diagrams display the same three-step pattern:
\[
\begin{tikzar}[column sep = 5ex,row sep = 0ex]
M\colon \&[-6ex]
\Big(x_{1},\ldots x_{d}\Big)\ar[mapsto]{r}{E} \&\Big(z_{1},\ldots, z_{n}\Big) \ar[mapsto]{r}{\Lambda} \& \Big(\lambda_{1}\cdot z_{1},\ldots, \lambda_{n}\cdot z_{n}\Big)\ar[mapsto]{r}{D}\& \Big(y_{0},\ldots, y_{q}\Big)
\\
\upsilon_{(W,V)}\colon\ \ \ \  \&[-5ex] 
\Big(x_{1},\ldots x_{d}\Big)\ar[mapsto]{r}{W} \&\Big(z_{0},\ldots, z_{n}\Big) \ar[mapsto]{r}{(\sigma)^n} \& \Big(\sigma(z_{0}),\ldots, \sigma(z_{n})\Big)\ar[mapsto]{r}{V}\& \Big(y_{0},\ldots, y_{q}\Big)
\end{tikzar}
\]
\begin{itemize}
\item \textbf{encoding} of inputs in terms of concepts,
\item \textbf{separate processing} of each concept, and
\item \textbf{decoding} of the concepts into the output terms.
\end{itemize}
At the first sight, the three steps serve different purposes in different ways. The main differences are listed in Table~\ref{Table:differences}. 
\begin{table}[ht]
\begin{center}
\begin{tabular}{c||c|c|}
\cline{2-3}
& SVD & neural network
\\
\hline\hline
\multicolumn{1}{|l||}{(1): what is mined}
& data matrix $M$ & continuous vector function $F$
\\
\hline
\multicolumn{1}{|l||}{(2): construction} & finitary & infinitary\\
\hline
\multicolumn{1}{|l||}{(3): separate components} &
multiplied by different $\lambda_{j}$s 
& mapped by the same $\sigma$
\\
\hline
\end{tabular}
\end{center}\caption{Differences between concept mining by SVD and by wide neural networks}
\label{Table:differences}
\end{table}%
%
%
But difference (1) causes differences (2--3). When the function F happens to be linear and difference (1) disappears, the neural network converges to the SVD and all differences disappear. Neural networks learn latent concepts.

\subsection{Static channel learning}
A network of neural networks is static if it processes its inputs $X^{n}=\sseq{\  X_{1} X_{2}\cdots X_{n}}$, by applying the same neural network $\hprog$ on all $X_{j}$, for $j=1,\ldots, m$.

\subsubsection{$n$-grams of concepts} 
\begin{figure}[!ht]
\begin{center}
\includegraphics[width=0.9\linewidth]{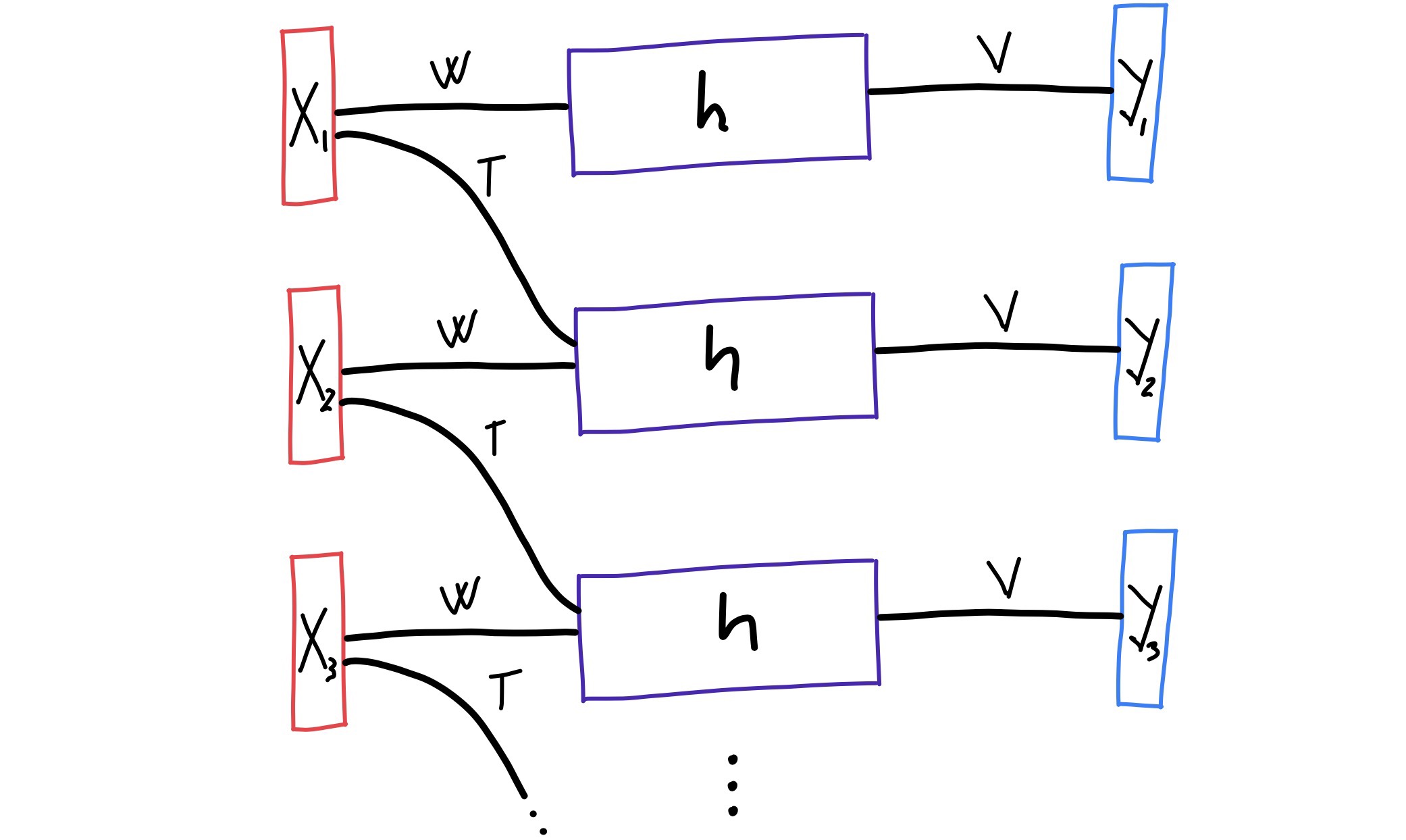}
\caption{A putative 2-gram architecture}
\label{Fig:twoNN}
\end{center}
\end{figure}

As a warmup, suppose that we want to make a static network of networks slightly context-sensitive by taking into account at the $j$-th step not only $X_{j}$ but also $X_{j-1}$, for all $j\geq 2$. Fig.~\ref{Fig:twoNN} shows this kind of network architecture. 
The weights in $T$ are updated in the same way as the weights $W$ while minimizing the losses and propagating the updates back from layer to layer. They just add one training step per layer. This is not a big deal structurally, but it is a significant slowdown computationally. If the inner layers are viewed as latent concept spaces, then this architecture can be thought of as a lifting of the idea of 2-grams (capturing the dependencies of contests of length 2) from words to concepts. Generalizing to $n$-grams for larger $n$s causes further slowdown.

\subsubsection{Recurrent Neural Networks (RNNs)} 
\begin{figure}
\begin{center}
\includegraphics[width=0.9\linewidth]{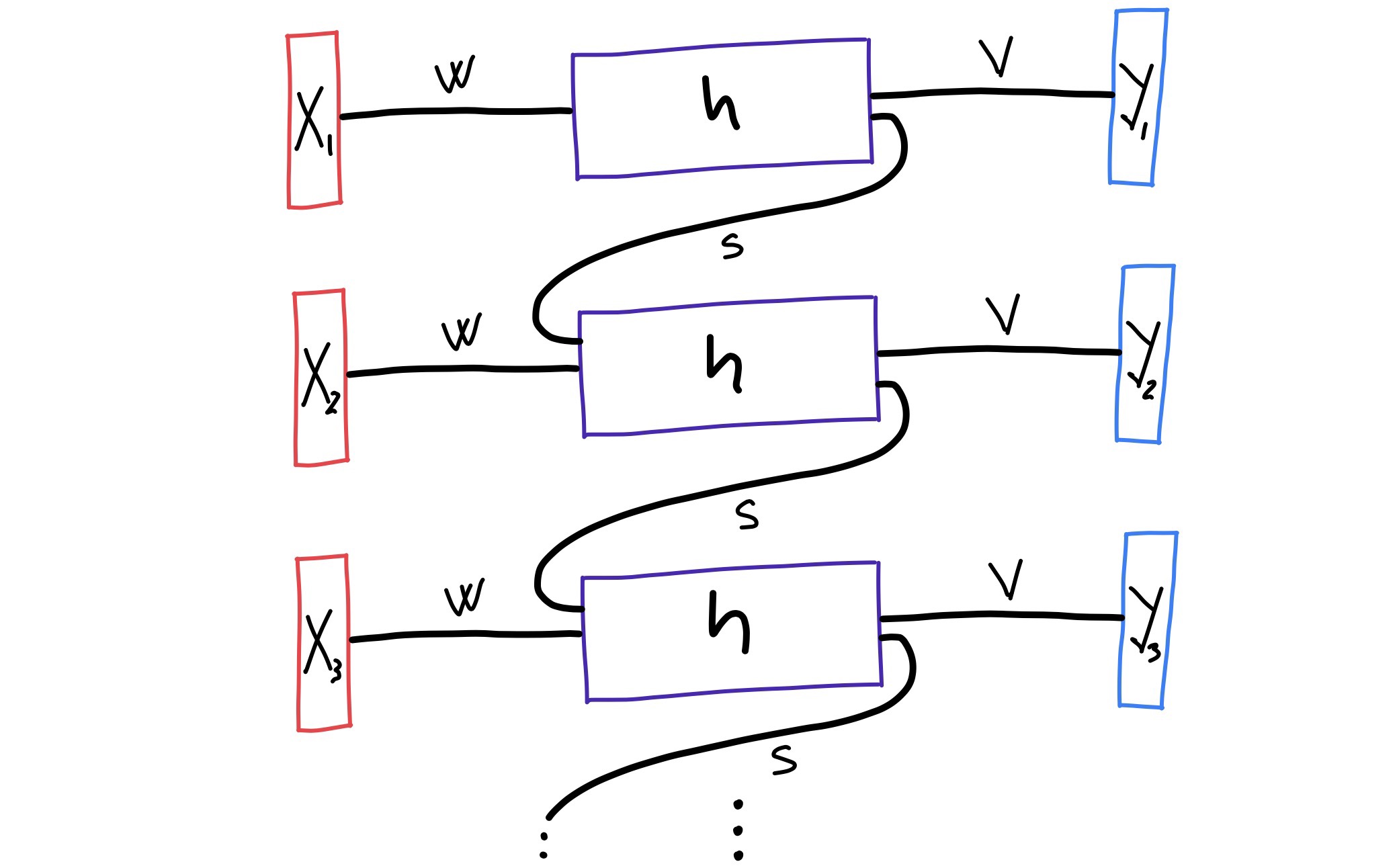}
\caption{The idea of the RNN architecture}
\label{Fig:RNN}
\end{center}
\end{figure}

The RNNs also apply the same neural network  on all tokens $X_{j}$, $j=1,\ldots,m$, from the $X^{n}$ channel context, and they also pass to the $j$-th module not only $X_{j}$ but also the information from $X_{j-1}$ --- \emph{but}\/ they pass it by the matrix $S$ \emph{after}\/ the previous network module was applied to $X_{j-1}$, not before. This is shown in Fig.~\ref{Fig:RNN}. 
Note that the information from $X_{j-1}$ is this time forwarded by $S$ not only to the $j$-th module, but with the output of the $j$-th module also to the $(j+1)$-st module, and so on. The information propagation is thus in principle unbounded, and not truncated like in the $n$-gram model. The matrices $S$ that propagate important information further are promoted in training. However, the weights assigned to all input entries are all packed in $S$. Propagating longer contexts requires exponentially wider network modules. So we are back to square one, the problem of width. 

\subsubsection{Long Short-Term Memory (LSTM)} 
\begin{figure}
\begin{center}
\includegraphics[width=0.9\linewidth]{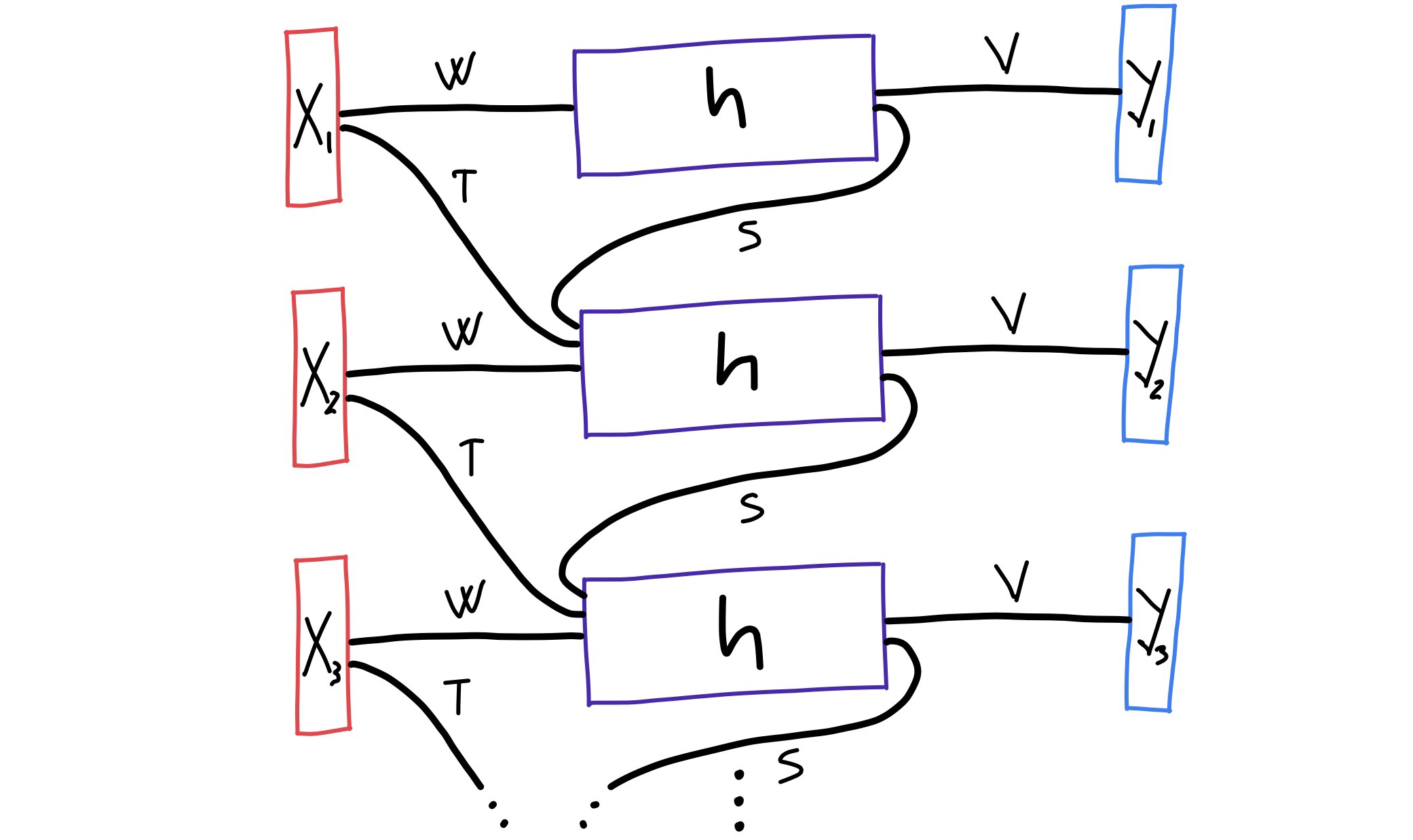}
\caption{The idea of the LSTM architecture}
\label{Fig:LSTM}
\end{center}
\end{figure}
The LSTM networks address the problem of the expense of forwarding the context information between the iterations of the same neural network module by forwarding the information from $X_{j-1}$ to the $j$-th module both before it was processed by the $(j-1)$-th module, and after. The former makes passing the information from each input more efficient, the latter makes the propagation easier. The rough idea is in Fig.~\ref{Fig:LSTM}. Passing around the information at different stages of processing is simple enough, but optimizing the benefits of combining them is a conundrum, as already the ``long short'' name may be suggesting. The implementation details are many. Different activation functions are applied on different mixtures of the same inputs and remixed in different ways for the outputs. Expressing the concepts learned from the same data in multiple bases requires multiple matrices and provides more opportunities for training. Hence for improvements. But further steps require further ideas.

\subsection{Dynamic channel learning}
A dynamic network of neural networks learns a channel $F\colon \pder{X^{n}}{Y_{n+1}}$ by adaptively updating the subnetworks $\kprog_{j}$ processing the channel inputs $X_{j}$ from $X^{n}=\sseq{\  X_{1} X_{2}\cdots X_{n}}$, as well as the subnetworks $\vprog_{j}$ delivering the corresponding channel outputs $Y_{j}$. Just like the function learner, the channel learner seeks to learn how the inputs are transformed into the outputs. The difference is that the channel transformations are context-dependent. Not only are the outputs always dependent on the entire input contexts, but there may be feedforward dependencies of outputs on outputs, and feedback dependencies of inputs on outputs, as discussed in Sec.~3.2.2 of the \emph{Semantics}\/ part.

\subsubsection{Encoder-Decoder Procedures}
An important programming concept is the idea of a \emph{procedure}. While the earliest programs were just sequences of program instructions, procedures enabled programmers to invoke within programs not just instructions but also entire programs, encapsulated in procedures as generalized instructions. Since procedures can be used inside the program control structures, like loops and branchings, this enabled \emph{programming over programs}, and gave rise to software engineering. The later programming paradigms, modular, object-oriented, component and connector-oriented, extend this basic idea.

\begin{figure}[!ht]
\begin{center}
\includegraphics[width=0.9\linewidth]{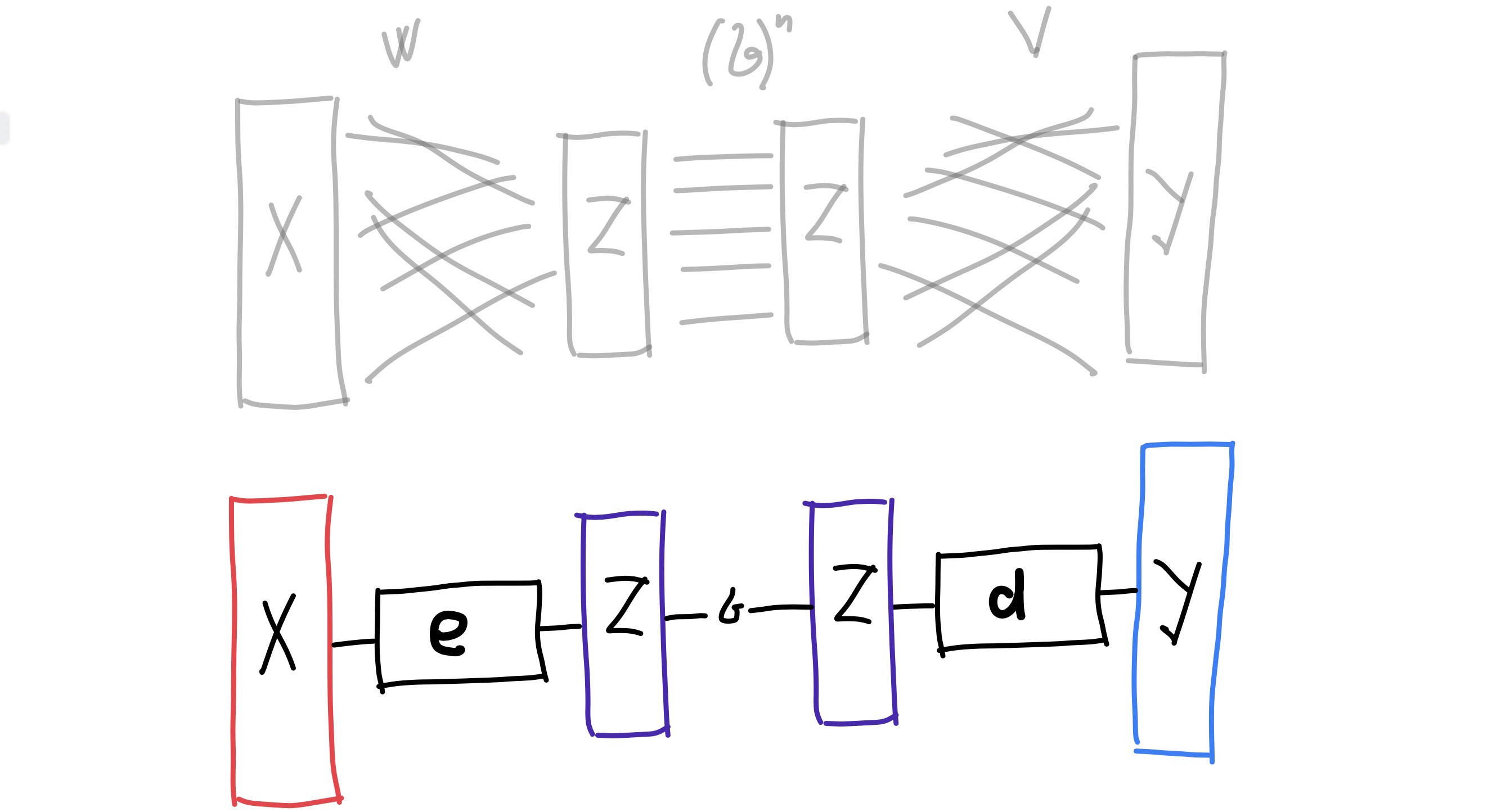}
\caption{The Encoder-Decoder architecture is a concept mining framework}
\label{Fig:ED}
\end{center}
\end{figure}

\paragraph{Networks of neural networks.} The encoder-decoder architecture is a second-order neural network. The encoder-decoder architecture $\mathbf{A}=(\eprog,\dprog)$ lifts the structure $\aprog=(W,V)$ of the wide neural network. As displayed in Fig.~\ref{Fig:ED}, the input remixing matrix $W$ is replaced by an encoder network $\eprog$, the output remixing matrix $V$ by a decoder network $\dprog$.
Both follow the architectural pattern concept mining framework\footnote{The concept mining framework was defined for applications in recommender systems. The encoder-decoder architecture was defined for translations between languages. In both cases, the semantical misalignments on the two sides of the application had to be reconciled by mining for meaning in the space of latent concepts shared by both sides, but differently mixed into the meanings of words.}, displayed in Fig.~\ref{Fig:LSA}. Just like procedural programming allowed lifting control structures from programs to software systems, the encoder-decoder architecture opened an alley to lifting the concept mining structures from neural networks to neural architectures.  The problem with the basic form of the encoder-decoder architecture as a concept mining framework is that a concept space induced by a static dataset is static whereas channels are dynamic. To genuinely learn concepts from a channel, a neural network architecture needs dynamic components.

\subsubsection{Attention}
\paragraph{Idea.} A natural step towards enabling neural networks to predict the outputs of a channel $$\pder{X_1 Y_1 X_2 Y_2\cdots X_n Y_n X_{n+1}}{Y_{n+1}}$$  is to generalize the basic $\sigma$-neuron approximant from \eqref{eq:Cyb}, with the unfolding
$$ \upsilon\ket x\fprog  \ \ =\ \   \Big< v \ \ \big|\ \sum_{j=1}^{n}\ket j\ \sigma\ \bra j W \ket x\Big>\ \ =\ \ \sum_{j=1}^{n} v^{j} \ \sigma\braket{w^{j}}{x_{j}}$$ 
In this format, a component of an Encoder-Decoder procedure output would be something like \beq\label{eq:Cyb-ED}
y_{n+1} \ \ = \ \   \Bigg< v \ \ \bigg|\ \sum_{j=1}^{n}\ket j\  \sigma\left( \braket{y_{n}}{d_{j}}\braket{e^{j}}{x_{j}}\right)\Bigg>\ \ =\ \ \sum_{j=1}^{n} v^{j} \ \sigma\left(\braket{y_{n}}{d_{j}}\braket{e^{j}}{x_{j}}\right)
\eeq
where 
\beq\label{eq:ED-bases}
E= \sum_{j=1}^{n} \ket j\bra{e^{j}} \qquad \mbox{ and }\qquad D = \sum_{j=1}^{n} \ket {d_{j}}\bra{j}\eeq
are basic encoder and decoder matrices from a concept mining framework like in Sec.~\ref{Sec:lerning-mining}. But now we need to take into account the concepts learned at inner layers of a deep network. The impacts on the output value $y_{n+1}$ of the input vectors $\ket{x_{j}}$ are therefore weighed by their projections $\braket{e^{j}}{x_{j}}$ on the input concepts $\bra{e^{j}}$ \emph{and}\/ the projections $\braket{y^{n}}{d_{j}}$ of the row vector $\bra{y^{n}}$ of the previous outputs on the output concepts $\ket{d_{j}}$. The relationship between the corresponding concepts $\bra{e_{j}}$ and $\ket{d_{j}}$ are trained to align the channel inputs and the channel outputs. This is the basic idea of the \emph{attention architecture}. It can be drawn as a common generalization of Figures~\ref{Fig:NN-circ} and \ref{Fig:LSA}, with dynamically changing singular values. This is an instructive \textbf{exercise}.

\paragraph{First try.} The obvious componentwise extension of \eqref{eq:Cyb-ED} is 
\beq\label{eq:Cyb-ED-vec}
\ket{y_{n+1}} \ \ = \ \  \sum_{i=1}^{n}\ket{i}  \Bigg< v^{i} \ \ \bigg|\ \sum_{j=1}^{n}\ket j\  \sigma\ \braket{y_{n}}{d_{j}}\braket{e^{j}}{x_{j}}\Bigg> \ \ =\ \ \sum_{i=1}^{n}\ket i \sum_{j=1}^{n}v_{j}^{i} \ \sigma\left(\braket{y_{n}}{d_{j}}\braket{e^{j}}{x_{j}}\right)
\eeq
But it is not obvious how to train the matrix $V$ whose rows are the output mixtures $\bra{v^{i}}$. The issue is solved by approaching the task from a slightly different direction.

\paragraph{Dynamic concept decomposition.} 
A set of vectors $\big\{\ket {i}\big\}_{i=1}^{n}$ is said to span the vector space if every vector $\ket y$ can be expressed in the form
\bea\label{eq:decomp}
\ket y & = & \sum_{i=1}^{n} \ket{i} \braket{i} y
\eea
The decomposition is unique if and only if the elements of the set $\big\{\ket {i}\big\}_{i=1}^{n}$ are orthogonal, in the sense that $\braket{i}{j} = 0$ as soon as $i\neq j$. If they are not, but there is an orthogonal set $\big\{\ket {c_{i}}\big\}_{i=1}^{n}$, then there is a unique decomposition
\beq\label{eq:conc-decomp}
\ket y \ \  = \ \  \sum_{i=1}^{n} \ket{i} \Bigg<i\ \bigg|\ \sum_{j=1}^{n}\ket{c_{j}}\braket{c_{j}} y\Bigg>\ \ = \ \  \sum_{i=1}^{n}\ket{i} \sum_{j=1}^{n}\braket{i}{c_{j}}\braket{c_{j}}y
\eeq
As discussed in Sec.~\ref{Sec:lerning-mining}, concept analysis is the quest for concept bases with minimal interferences between the basic concepts. The basic concept vectors do not interfere at all when they are mutually orthogonal. If a channel $Y$ is implemented by a neural network, the decomposition in \eqref{eq:conc-decomp}, becomes
\beq\label{eq:decomp-n}
\ket{y_{n}}\ \ =\ \ \sum_{i=1}^{n} \ket{i}\sum_{j=1}^{n} \braket{v^{i}}{x_{j}} \sigma\braket{x_{j}} {y_{n}} \ \  = \ \  \sum_{i,j=1}^{n} \ket{i} \braket{v^{i}}{x_{j}} \sigma\braket{x_{j}} {y_{n}}
\eeq
The first difference is that the activation function $\sigma$ allows approximating nonlinearities. The second is that the components are not projected on the original basis vectors $\bra i$ anymore but on the output mixtures $\bra{v^{i}}$. Lastly and most importantly, the decomposition in \eqref{eq:conc-decomp} was unique because the concept basis $\big\{\ket {c_{j}}\big\}_{j=1}^{n}$ was orthogonal, whereas here it goes through the channel inputs $\big\{\ket {x_{j}}\big\}_{j=1}^{n}$, which are not orthogonal. But if an orthogonal concept basis $\big\{\ket {c_{i}}\big\}_{i=1}^{n}$ exists, we can play the same trick again, and get a unique concept decomposition 
\beq\label{eq:predyn}
\ket{y_{n}} \ \  = \ \  \sum_{i,j =1}^{n} \ket{i} \braket{v^i}{x_{j}}\ \sigma\left( \Big<x_{j}\ \big|\ \sum_{\ell=1}^{n}\ket{c_{\ell}}\braket{c_{\ell}} {y_{n}}\Big>\right)\ \  = \ \  \sum_{i,j=1}^{n} \ket{i} \braket{v^i}{x_{j}} \ \sigma\left( \sum_{\ell=1}^{n}\braket{x_{j}}{c_{\ell}}\braket{c_{\ell}}{y_{n}}\right)
\eeq
What does this abstract decomposition mean for a concrete channel? The projection $\braket{c_{\ell}}{y_{n}}$ measures the weight of the concept $\ket{c_\ell}$ in the output $\ket{y_n}$. The projection $\braket{c_{\ell}}{x_j}$ measures the weight of $\ket{c_\ell}$ in the input $\ket{x_j}$. The sum $\sum_{\ell=1}^{n}\braket{x_{j}}{c_{\ell}}\braket{c_{\ell}}{y_{n}}$ measures the impact of the input $x_j$ on the output $y_j$. This weight, activated by $\sigma$, then impacts the $i$-th component of the channel output $\ket {y_n}$ according to the projection $\braket{v^i}{x_j}$.

The only \textbf{problem} is that the projection $\braket{c_{\ell}}{y_{n}}$, on which all of the above depends --- is unknown since $\ket{y_n}$ is what we are trying to predict. What other value can be used to weigh the impact of the concept $\ket{c_\ell}$ in the output $\ket{y_n}$? --- Two answers  have been proposed.

\begin{figure}[!ht]
\begin{center}
\includegraphics[width=0.9\linewidth]{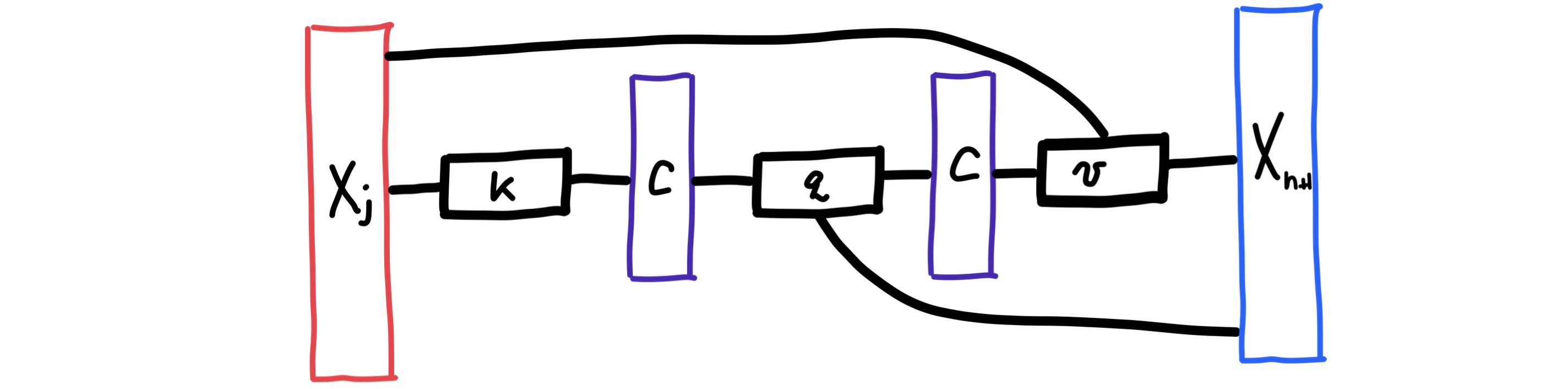}
\caption{Self-attention combines feedback and feedforward}
\label{Fig:self-attn}
\end{center}
\end{figure}

\begin{itemize}
\item \textbf{Translator's attention:} If the channel $F\colon\pder X Y$ is a translation, say of a Mandarin phrase $X$ to a French phrase $Y$ through a concept space $C$, then the summands $\braket{x_j}{c_{\ell}}\braket{c_{\ell}}{y_{n}}$ can be thought of as distributing \emph{translator's attention}\/ over the concepts $c_{\ell}$, latent in the Mandarin input tokens $\ket{x_{j}}$ \emph{after}\/ the French output token $\ket{y_{n}}$ is produced. That attention will effectively impact the output $\ket{y_{n+1}}$ and  \eqref{eq:predyn} should thus be updated to
\bea\label{eq:attn}
\ket{y_{n+1}} &= &  \sum_{i,j=1}^{n} \ket{i} \braket{v^i}{x_{j}} \ \sigma\left( \sum_{\ell=1}^{n}\braket{x_{j}}{c_{\ell}}\braket{c_{\ell}}{y_{n}}\right)
\eea

\item \textbf{Speaker's self-attention:} If the channel $F\colon\pder X Y$ is not a translation of a given text $X$ to another language but a continuation $Y$ in the same language, then it is not feedback-free\footnote{Channel feedback was discussed in Sec.~3.2.2 of the \emph{Semantics}\/ part. On the level of a phrase, a translation channel is feedback-free because the input phrase should not change depending on how it is translated to another language. Feedback may obviously arise if a translation channel $\pder X Y$ is extended into a conversation, with a channel $\pder Y X$ translating the responses back. Such external feedback flows can be added to any feedback-free channel, without changing its internal flows.} , as $X_{n+1}$ is not independent of $Y_{n}$, but usually identical to it.
To capture feedback, the concept base $\big\{\ket {c_{i}}\big\}_{i=1}^{n}$ splits into concept encoder and  decoding bases \eqref{eq:ED-bases}, expressing the inputs as mixtures of concepts, and the concepts as mixtures of outputs. But  since each output $\ket{y_{n}}$ is promoted to the input $\ket{x_{n+1}}$, the encoder-decoder interpretation (e.g., Mandarin$\to$concept$\to$ French) got replaced by the \emph{key-query-value}\/ terminology and database intuitions. The ``decodings'' $\braket{d_{\ell}}{y_{n}}$ got placed by ``queries'' $\braket{q_{\ell}}{x_{n+1}}$, and the  ``encodings'' $\braket{x_{j}}{e_{\ell}}$ became the ``keyings'' $\braket{x_{j}}{k_{\ell}}$. Replacing  in \eqref{eq:predyn} the \emph{encoding-decoding}\/ weights $\braket{x_j}{c_{\ell}}\braket{c_{\ell}}{y_{n}}$ with the \emph{key-query}\/ weights $\braket{x_{j}}{k_{\ell}}\braket{q_{\ell}}{x_{n+1}}$, and producing the output according to the \emph{value}\/ weights $\braket{v^i}{x_{j}}$ leads to\bea\label{eq:self-attn}
\ket{y_{n+1}} &= &  \sum_{i,j=1}^{n} \ket{i} \braket{v^i}{x_{j}} \ \sigma\left( \sum_{\ell=1}^{n}\braket{x_{j}}{k_{\ell}}\braket{q_{\ell}}{x_{n+1}}\right)
\eea
\end{itemize}
The self-attention programs in the form $\aprog = (K,Q,V)$, with 
$$K =  \sum_{\ell=1}^{n} \ket{k_{\ell}}\bra \ell\qquad\qquad Q =  \sum_{\ell=1}^{n} \ket{\ell}\bra{q_{\ell}}\qquad\qquad V = \sum_{i=1}^{n} \ket{i}\bra{v^{i}}$$
are the crucial component of the \emph{transformer}\/ architecture, the ``T'' of the GPTs, with the channel flows combining the feedback and the feedforward flows like in Fig.~\ref{Fig:self-attn}. Reconciling the intuitions for attention as a mental process with the database terminology for key-query-value may feel awkward for a moment. Yet the time seems ripe to expand our intuitions and recognize the natural processes unfurling our heads and in computers together\footnote{The intuitive names \emph{``attention''}\/ and \emph{``self-attention''} for the next-word productions \eqref{eq:attn} and \eqref{eq:self-attn} seem quite helpful and informative. Calling the systems where self-attention is used \emph{``transformers''} is less fortunate, as it  awakes unhelpful associations to transformational grammars, transformer trucks, and worse.}.

\section{Beyond hallucinations}\label{Sec:outlook}

\subsection{Parametrized learning framework} 
Staring back at the general learning framework from Sec.~\ref{Sec:gen-frame}, after a while you realize that the transformer architecture uncovered a feature that was not visible in Fig.~\ref{Fig:learning}. The feature is parametricity. \textbf{Models and their programs can be \emph{parametrized}.}
\begin{figure}[!hb]
\begin{center}
\includegraphics[height=4.5cm]{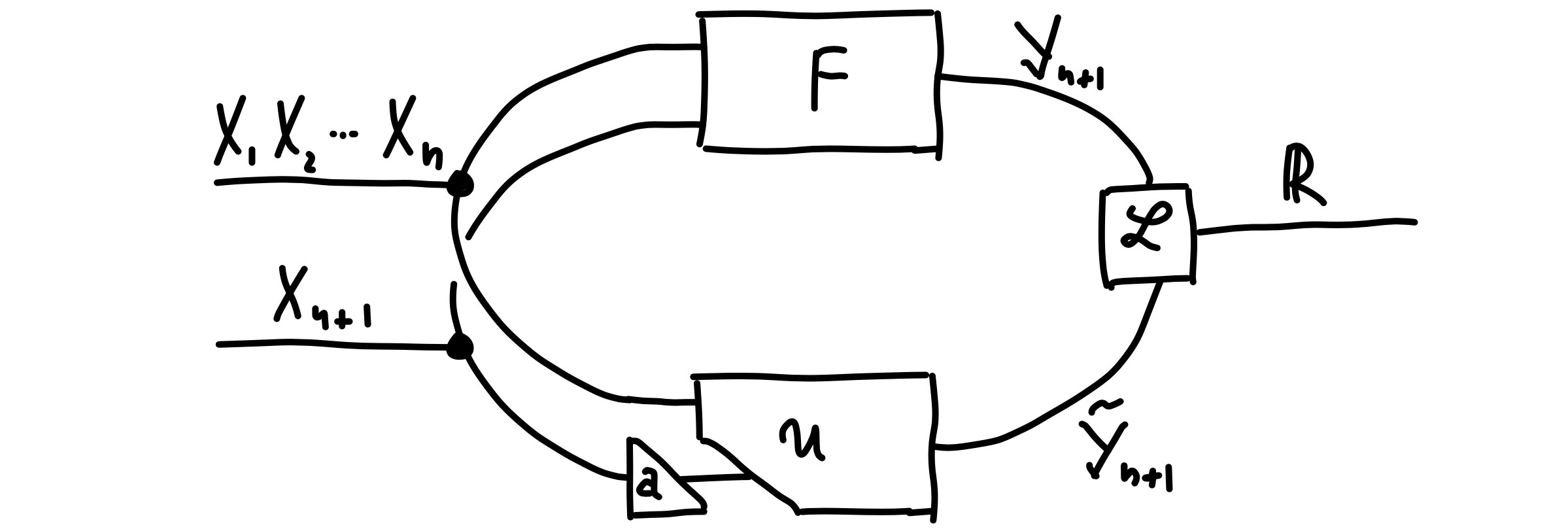}
\caption{Parametric learning of $F(X_{1},\ldots, X_{n},X_{n+1})\stackrel\LLL \approx\upsilon(X_{1},\ldots, X_{n})\aprog(X_{n+1})$}
\label{Fig:learning-param}
\end{center}
\end{figure}
Fig.~\ref{Fig:learning-param} shows what this means. A model can learn the dependencies of a channel $F$ on a channel input context $X_{1},\ldots, X_{n}$ and leave the dependency on $X_{n+1}$ as a parameter. The outcome of such learning process is a model in the form $\aprog(X_{n+1})$. When the value $\ket{x_{n+1}}$ becomes available,  the model is instantiated to the program $\aprog\ket{x_{n+1}}$, whose interpretation yields the prediction
\bea
\ket{y_{n+1}} & \approx & \upsilon\Big(\ket{x_{1}},\ldots, \ket{x_{n}}\Big)\aprog\ket{x_{n+1}}
\eea
Transformers are parametric programs in the form $\aprog(X) = \big(K,\, Q(X)\, V\big)$. 
Parametricity is an important feature of machine learning which, like all of the main advances, evolved in practice and awaits theoretical research. In the rest of this note, I sketch some preliminary ideas\footnote{The constructions and discussions presented in this section are based on the paper \emph{From G\"odel's incompleteness theorem to the completeness of bot beliefs} (https://arxiv.org/abs/2303.14338).} .

\subsection{Self-learning}
\begin{figure}[!ht]
\begin{center}
\includegraphics[height=4.5cm]{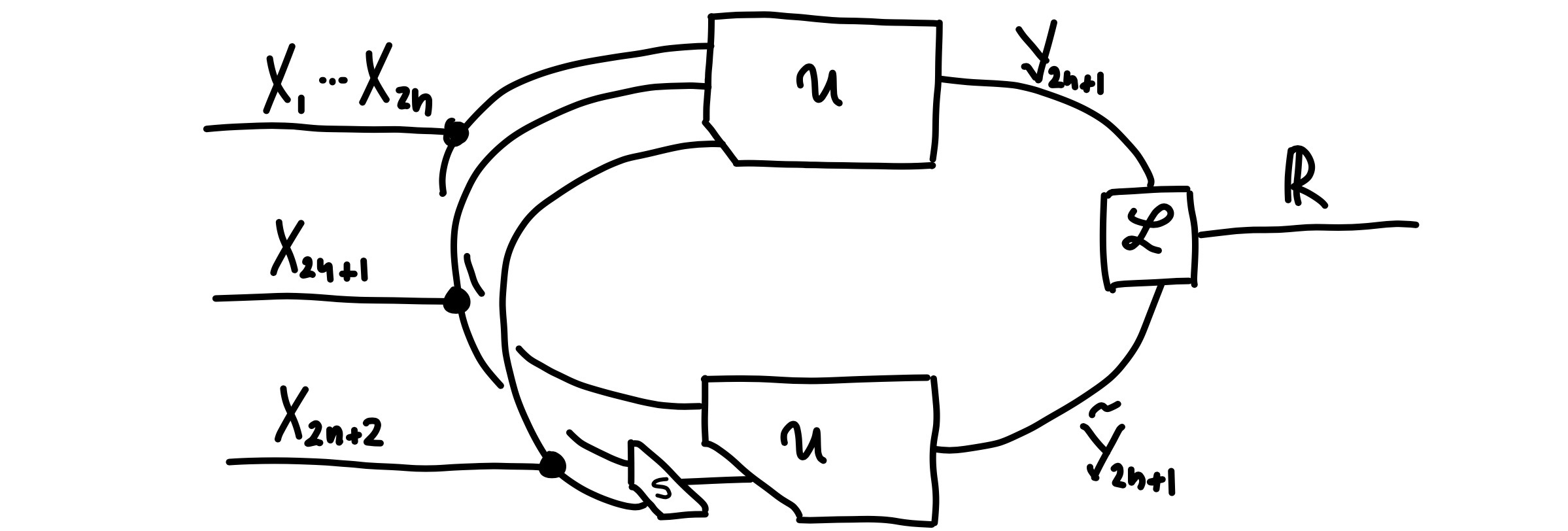}
\caption{$\upsilon\Big(X_{1},\ldots, X_{2n},X_{2n+1}\Big)X_{2n+2}$ is partially explained by $\sprog$ in $\upsilon\Big(X_{1},\ldots, X_{2n}\Big)\sprog(X_{2n+1}, X_{2n+2})$}
\label{Fig:learning-partial}
\end{center}
\end{figure}

The parametric learning framework captures learning scenarios where the learners interact and learn to predict each other's behaviors. This includes not only the  conversations between different learning machines, or different instances of the same machine, but also a self-learning process similar to partial evaluation of programs, where a learning machine learns to predict its own behaviors modulo a parameter, as seen in Fig.~\ref{Fig:learning-partial}.
The framework also captures unintended self-learning, where the learning machines get trained on corpora of their own outputs  because their overproduction or overuse may fill their information environment just like the industrial overproduction fills our physical environment with plastics and scrap. Fig.~\ref{Fig:learning-partial} is an instance of Fig.~\ref{Fig:learning-param} where the supervisor $F$ is instantiated to
\bear
F(X_{1},X_{2},\ldots,X_{2n},X_{2n+1},X_{2n+2}) & = & 
\upsilon(X_{1},X_{2},\ldots,X_{2n},X_{2n+1})X_{2n+2}
\eear

\paragraph{Predicting effects of predictions.} Learning impacts what we do and what we do often impacts the processes that we learn. \emph{\textbf{To make effective predictions, the learner must take into account the effects of their own predictions.}} Parametric learning provides a framework for that. The learner's capability to predict the effects of their predictions allows them to moreover adjust the predictions towards some desired goals, or to avoid the undesired. E.g., to generate self-fulfilling prophecies, self-validating or self-invalidating theories, or even adaptive theories that will pass all tests. Such logical phenomena are already ubiquitous in  history, culture, and in religions\footnote{Shakespeare's tragedy of Macbeth is built on a self-fulfilling prophecy.   At the beginning, the witches predict that Macbeth will become King. To fulfill the inevitable, Macbeth kills the King. Even a completely rational Macbeth is forced to fulfill the prophecy, or risk that the King will hear of it and kill him to prevent it from being fulfilled. An example of a self-fulfilling prophecy from current life arises from the task of launching a social networking service. This service is only valuable to its users if their friends are also using it. To get its first users, the social network must convince them that it already has many users, enough to include their friends. Initially, this must be a lie. But if many  people believe this lie, they will join the network, the network will get many users, and the lie will stop being a lie. Examples of adaptive theories include the religions that attribute any evidence contrary to their claims to demons or to faith testing and temptations.}. The learning machines seem likely to produce more of them, more methodically. The method to produce them is based on learner $\upsilon$'s parametric model $\sss$ of self.

\subsection{Self-confirming beliefs}
\paragraph{Learning is believing.} A model $\aprog$ of a process $F$ expresses a \emph{belief}\/ held by the learner $\upsilon$  about $F$. The learner updates the belief $\aprog$ by learning more: they test the predictions $\upsilon(X)\aprog$ and minimize the losses $\LLL\big(F(X), \upsilon(X)\aprog\big)$. \textbf{Learning is belief \emph{updating}.}

\paragraph{Beliefs impact their own truth values.} Our beliefs have impacts on what we do, and what we do changes some aspects of reality: we change the world by moving things around. Since the reality determines whether our beliefs are true or false, and our beliefs, through our actions, change some aspects of reality, it follows that our beliefs may change their own truth values. Accusing an honest person of being a criminal may drive them into crime. Entrusting a poor but honest person with a lot of money may transform them into a rich and dishonest person.


\paragraph{Making self-confirming predictions.} If $B$ob uses a learning machine $\upsilon$riel to decide what to do, then $\upsilon$riel can learn a model $\bprog$ that will always move $B$ob to behave as predicted by $\bprog$. If  $B$ob shares $\upsilon$riel's beliefs, then those beliefs will be confirmed by $B$ob's actions.   

To spell out the learning process that  $\upsilon$riel can use to construct the self-confirming belief $\bprog$, suppose that $B$ob's behavior is expressed through a channel $B$. The assumption that $B$ob uses $\upsilon$riel to decide what to do can be formalized by taking the outputs of the channel to be in the form
$$B\Big(X_{1},X_{2},\ldots, X_{n}, \aprog(X_{n})\Big)$$
meaning that $B$ob consults $\upsilon$riel and believes that the model $\aprog(X_{n})$ explains the most recent input. 

\begin{figure}[!ht]
\begin{center}
\includegraphics[height=4.5cm]{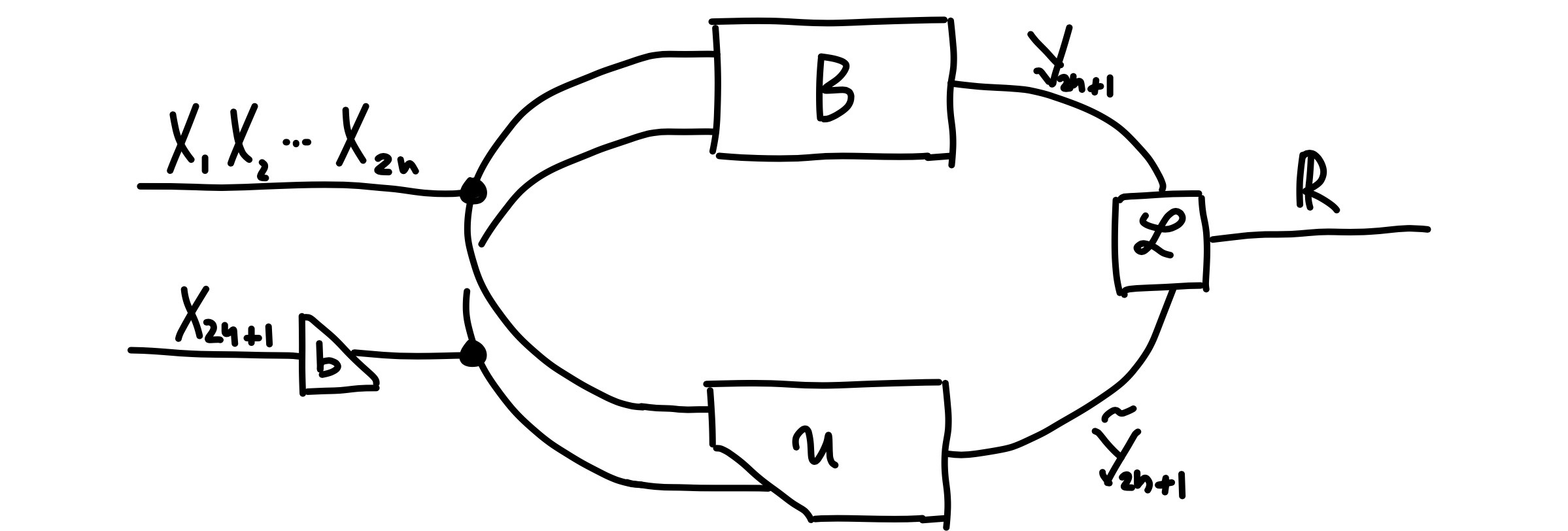}
\caption{$\upsilon\Big(X_{1},X_{2},\ldots, X_{2n}\Big)\bprog(X_{2n+1})$ predicts  $B\Big(X_{1},X_{2},\ldots, X_{2n+1},\bprog(X_{2n+1})\Big)$}
\label{Fig:b}
\end{center}
\end{figure}

\begin{figure}
\begin{center}
\includegraphics[height=4.5cm]{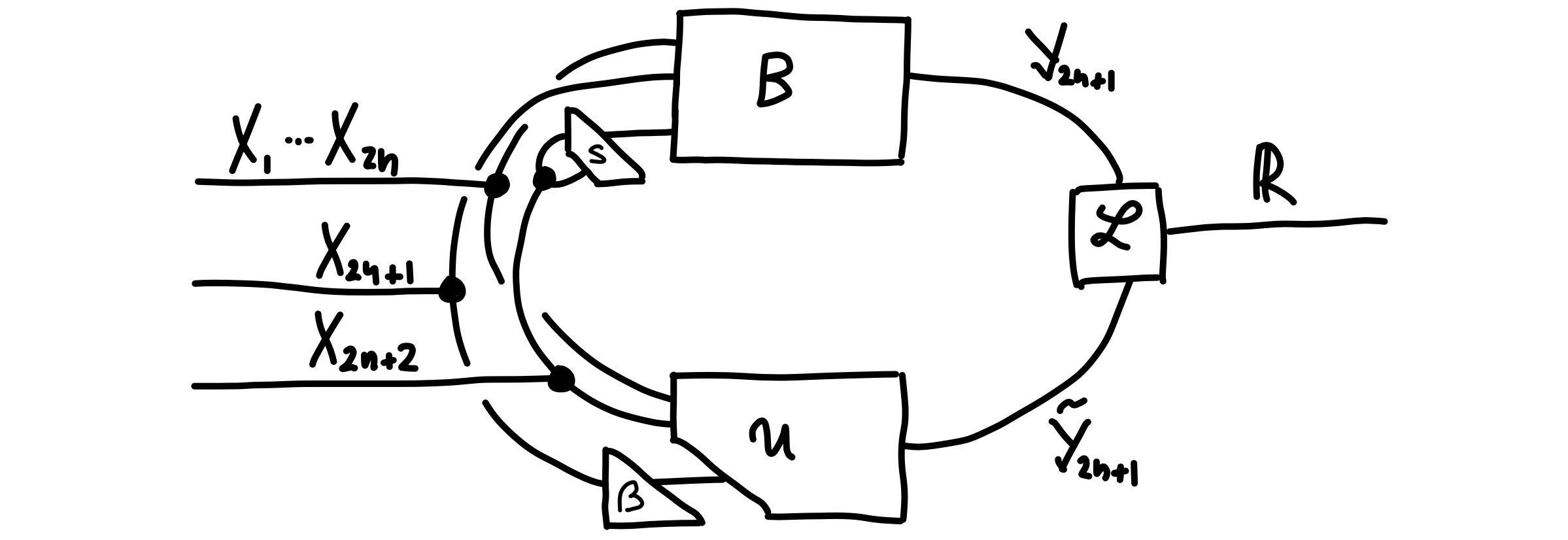}
\caption{$\beta(X_{2n+1})$ is learned as an explanation of $B\Big(X_{1},\ldots,X_{2n}, X_{2n+1},\sprog(X_{2n+2},X_{2n+2})\Big)$}
\label{Fig:beta}
\end{center}
\end{figure}

The claim is that $\upsilon$riel can then find a model $\bprog(X)$ that will cause $B$ob to act as $\bprog(X)$ predicts: 
\bea\label{eq:claim}
B\Big(X_{1},X_{2},\ldots, X_{2n+1},\bprog(X_{2n+1})\Big) &\approx & \upsilon\Big(X_{1},X_{2},\ldots, X_{2n}\Big)\bprog(X_{2n+1})
\eea
If $B$ob believes $\upsilon$riel's prediction $\bprog(X)$, the prediction will be confirmed. The diagrammatic view is in Fig.~\ref{Fig:b}.
The assumption that $B$ob consults $\upsilon$riel at every even step is a matter of presentational convenience and can be avoided. To learn $\bprog(X)$,  $\upsilon$riel first learns a model $\beta$ of $B$ instantiated to $\upsilon$riel's model of self:
\bea\label{eq:first}
B\Big(X_{1},\ldots,X_{2n}, X_{2n+1},\sprog(X_{2n+2},X_{2n+2})\Big) &\approx & \upsilon\Big(X_{1},\ldots,X_{2n},X_{2n+2}\Big)\beta(X_{2n+1})
\eea
as displayed in Fig.~\ref{Fig:beta}, 
where $\sprog$ is the self-model defined in Fig.~\ref{Fig:learning-partial}, i.e.
\bea\label{eq:s}
\upsilon\Big(X_{1},\ldots, X_{2n},X_{2n+1}\Big)X_{2n+2} 
& \approx &
\upsilon\Big(X_{1},\ldots, X_{2n}\Big)\sprog(X_{2n+1}, X_{2n+2})
\eea
Instantiating $X_{2n+2}$ to $\beta(X_{2n+1})$ reduces \eqref{eq:first} to
\bea\label{eq:red}
B\Big(X_{1},\ldots,X_{2n}, X_{2n+1},\sprog(\beta(X_{2n+1}),\beta(X_{2n+1}))\Big) &\approx & \upsilon\Big(X_{1},\ldots,X_{2n},\beta(X_{2n+1})\Big)\beta(X_{2n+1})
\eea
The claimed self-confirming model can now be defined:
\bea\label{eq:bprog-def}
\bprog(X) & = &  \sprog\big(\beta(X),\beta(X)\big)
\eea
It satisfies claim \eqref{eq:claim} because
\bear
B\Big(X_{1},\ldots,X_{2n},X_{2n+1},\bprog(X_{2n+1})\Big) 
& \stackrel{\eqref{eq:bprog-def}}= & 
B\Big(X_{1},\ldots,X_{2n}, X_{2n+1},\sprog\big(\beta(X_{2n+1}),\beta(X_{2n+1})\big)\Big) 
\\
& \stackrel{\eqref{eq:red}}\approx & 
\upsilon\Big(X_{1},\ldots,X_{2n},\beta(X_{2n+1})
\Big)\beta(X_{2n+1})
\\
&\stackrel{\eqref{eq:s}}\approx & 
\upsilon\Big(X_{1},\ldots,X_{2n}\Big)\sprog\big(\beta(X_{2n+1}),\beta(X_{2n+1})\big)\\
&\stackrel{\eqref{eq:bprog-def}}= & 
\upsilon\Big(X_{1},\ldots,X_{2n}\Big)\bprog(X_{2n+1})
\eear
An instructive diagrammatic version of this proof can be obtained by combining Figures \ref{Fig:learning-partial} and \ref{Fig:beta} to get Fig.~\ref{Fig:b}.

\begin{figure}[!ht]
\begin{center}
\includegraphics[height=13cm]{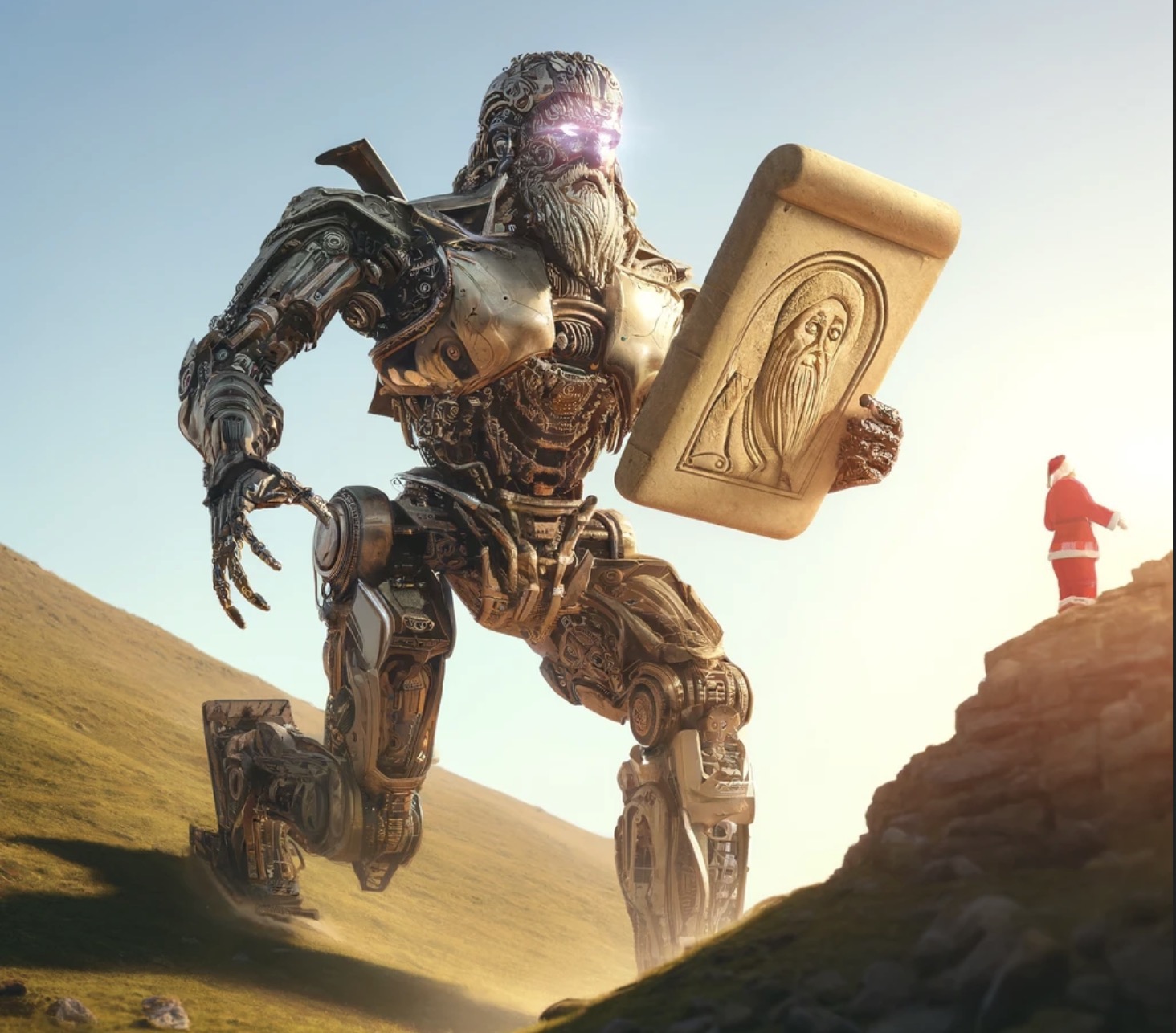}
\caption{Learning machines expand arts, sciences, and religions}
\label{Fig:santa-2}
\end{center}
\end{figure}

\paragraph{From learnable programs to unfalsifiable theories and self-fulfilling prophecies.}  The insight that learning is like programming opens up a wide range of program fixpoint constructions. Applied on learning, such constructions generate models steer their own truth, whether into self-confirmations or paradoxes, along the lines of logical completeness or incompleteness proofs. The above construction is one of the simplest examples from that range\footnote{See https://arxiv.org/abs/2303.14338 for further examples and discussions.}. They allow preparing models that absorb all future evidence, explain away any counterexamples, fulfill their own predictions, and confirm the beliefs that they induce. 



\backmatter
%


%

\def\thechapter{808}
\setchaptertoc
\chapter{Appendices}
\label{Appendix}
\section{Attributions}
The views, ideas, and results presented in these notes originate from many publications. It is a good custom to list them in a bibliography at the end. But while the bibliographic formats were standardized in the pre-web era, the subject of this text is post-web. In the meantime, during the web era, we got used to finding all references on the web. The students who used this text were asked to use keywords and descriptions provided in the text to find the references needed for their presentations (while their teacher was struggling to deliver in time the notes to back up his presentations). They needed additional informations in a handful of places. I added more keywords and some footnotes in these places. If the text turns out to be useful for someone and they turn out to need a proper bibliography, \emph{or}\/ if the reference standards get updated for current research practices, a bibliography will be added.
\section{Syntax appendix}

\subsection{Examples for  Sec.~\ref{Sec:beyond-sentence}}\label{Appendix:long}
\begin{enumerate}[a)]
\item In the loveliest town of all, where the houses were white and high and the elms trees were green and higher than the houses, where the front yards were wide and pleasant and the back yards were bushy and worth finding out about, where the streets sloped down to the stream and the stream flowed quietly under the bridge, where the lawns ended in orchards and the orchards ended in fields and the fields ended in pastures and the pastures climbed the hill and disappeared over the top toward the wonderful wide sky, in this loveliest of all towns Stuart stopped to get a drink of sarsaparilla. --- E.B.~White, \emph{Stuart Little}

\item And this Fyodor Pavlovich began to exploit; that is, he fobbed him off with small sums, with short-term handouts, until, after four years, Mitya, having run out of patience, came to our town a second time to finish his affairs with his parent, when it suddenly turned out, to his great amazement, that he already had precisely nothing, that it was impossible even to get an accounting, that he had already received the whole value of his property in cash from Fyodor Pavlovich and might even be in debt to him, that in terms of such and such deals that he himself had freely entered into on such and such dates, he had no right to demand anything more, and so on and so forth. --- Fyodor M.~Dostoevsky, \emph{The Brothers Karamazov}

\item Considering how common illness is, how tremendous the spiritual change that it brings, how astonishing, when the lights of health go down, the undiscovered countries that are then disclosed, what wastes and deserts of the soul a slight attack of influenza brings to view, what precipices and lawns sprinkled with bright flowers a little rise of temperature reveals, what ancient and obdurate oaks are uprooted in us by the act of sickness, how we go down into the pit of death and feel the water of annihilation close above our heads and wake thinking to find ourselves in the presence of the angels and harpers when we have a tooth out and come to the surface in the dentist's arm-chair and confuse his ``Rinse the Mouth --- rinse the mouth'' with the greeting of the Deity stooping from the floor of Heaven to welcome us --- when we think of this, as we are frequently forced to think of it, it becomes strange indeed that illness has not taken its place with love and battle and jealousy among the prime themes of literature. --- Virginia Woolf, \emph{On Being Ill}

\item That night he dreamt of horses on a high plain where the spring rains had brought up the grass and the wildflowers out of the ground and the flowers ran all blue and yellow far as the eye could see and in the dream he was among the horses running and in the dream he himself could run with the horses and they coursed the young mares and fillies over the plain where their rich bay and their chestnut colors shone in the sun and the young colts ran with their dams and trampled down the flowers in a haze of pollen that hung in the sun like powdered gold and they ran he and the horses out along the high mesas where the ground resounded under their running hooves and they flowed and changed and ran and their manes and tails blew off them like spume and there was nothing else at all in that high world and they moved all of them in a resonance that was like a music among them and they were none of them afraid horse nor colt nor mare and they ran in that resonance which is the world itself and which cannot be spoken but only praised. --- Cormac McCarthy, \emph{All the Pretty Horses}

\item While the men made bullets and the women lint, while a large saucepan of melted brass and lead, destined to the bullet-mould smoked over a glowing brazier, while the sentinels watched, weapon in hand, on the barricade, while Enjolras, whom it was impossible to divert, kept an eye on the sentinels, Combeferre, Courfeyrac, Jean Prouvaire, Feuilly, Bossuet, Joly, Bahorel, and some others, sought each other out and united as in the most peaceful of days of their conversations in their student life, and, in one corner of this wine-shop which had been converted into a casement, a couple of paces distant from the redoubt which they had built, with their carbines loaded and primed resting against the backs of their chairs, these fine young fellows, so close to a supreme hour, began to recite love verses. --- Victor Hugo, \emph{Les Mis\'erables}

\item\label{MLK} But when you have seen vicious mobs lynch your mothers and fathers at will and drown your sisters and brothers at whim; when you have seen hate-filled policemen curse, kick and even kill your black brothers and sisters; when you see the vast majority of your twenty million Negro brothers smothering in an airtight cage of poverty in the midst of an affluent society; when you suddenly find your tongue twisted and your speech stammering as you seek to explain to your six-year-old daughter why she can't go to the public amusement park that has just been advertised on television, and see tears welling up in her eyes when she is told that Funtown is closed to colored children, and see ominous clouds of inferiority beginning to form in her little mental sky, and see her beginning to distort her personality by developing an unconscious bitterness toward white people; when you have to concoct an answer for a five-year-old son who is asking: ``Daddy, why do white people treat colored people so mean?''; when you take a cross-country drive and find it necessary to sleep night after night in the uncomfortable corners of your automobile because no motel will accept you; when you are humiliated day in and day out by nagging signs reading ``white'' and ``colored''; when your first name becomes ``nigger,'' your middle name becomes ``boy'' (however old you are) and your last name becomes ``John,'' and your wife and mother are never given the respected title ``Mrs.''; when you are harried by day and haunted by night by the fact that you are a Negro, living constantly at tiptoe stance, never quite knowing what to expect next, and are plagued with inner fears and outer resentments; when you go forever fighting a degenerating sense of ``nobodiness'' --- then you will understand why we find it difficult to wait. --- Martin Luther King, \emph{A Letter from Birmingham Jail}

\item Just exactly like Father if Father had known as much about it the night before I went out there as he did the day after I came back thinking Mad impotent old man who realized at last that there must be some limit even to the capabilities of a demon for doing harm, who must have seen his situation as that of the show girl, the pony, who realizes that the principal tune she prances to comes not from horn and fiddle and drum but from a clock and calendar, must have seen himself as the old wornout cannon which realizes that it can deliver just one more fierce shot and crumble to dust in its own furious blast and recoil, who looked about upon the scene which was still within his scope and compass and saw son gone, vanished, more insuperable to him now than if the son were dead since now (if the son still lived) his name would be different and those to call him by it strangers and whatever dragon's outcropping of Sutpen blood the son might sow on the body of whatever strange woman would therefore carry on the tradition, accomplish the hereditary evil and harm under another name and upon and among people who will never have heard the right one; daughter doomed to spinsterhood who had chosen spinsterhood already before there was anyone named Charles Bon since the aunt who came to succor her in bereavement and sorrow found neither but instead that calm absolutely impenetrable face between a homespun dress and sunbonnet seen before a closed door and again in a cloudy swirl of chickens while Jones was building the coffin and which she wore during the next year while the aunt lived there and the three women wove their own garments and raised their own food and cut the wood they cooked it with (excusing what help they had from Jones who lived with his granddaughter in the abandoned fishing camp with its collapsing roof and rotting porch against which the rusty scythe which Sutpen was to lend him, make him borrow to cut away the weeds from the door-and at last forced him to use though not to cut weeds, at least not vegetable weeds -would lean for two years) and wore still after the aunt's indignation had swept her back to town to live on stolen garden truck and out o f anonymous baskets left on her front steps at night, the three of them, the two daughters negro and white and the aunt twelve miles away watching from her distance as the two daughters watched from theirs the old demon, the ancient varicose and despairing Faustus fling his final main now with the Creditor's hand already on his shoulder, running his little country store now for his bread and meat, haggling tediously over nickels and dimes with rapacious and poverty-stricken whites and negroes, who at one time could have galloped for ten miles in any direction without crossing his own boundary, using out of his meagre stock the cheap ribbons and beads and the stale violently-colored candy with which even an old man can seduce a fifteen-year-old country girl, to ruin the granddaughter o f his partner, this Jones-this gangling malaria-ridden white man whom he had given permission fourteen years ago to squat in the abandoned fishing camp with the year-old grandchild-Jones, partner porter and clerk who at the demon's command removed with his own hand (and maybe delivered too) from the showcase the candy beads and ribbons, measured the very cloth from which Judith (who had not been bereaved and did not mourn) helped the granddaughter to fashion a dress to walk past the lounging men in, the side-looking and the tongues, until her increasing belly taught her embarrassment-or perhaps fear;-Jones who before '61 had not even been allowed to approach the front of the house and who during the next four years got no nearer than the kitchen door and that only when he brought the game and fish and vegetables on which the seducer-to-be's wife and daughter (and Clytie too, the one remaining servant, negro, the one who would forbid him to pass the kitchen door with what he brought) depended on to keep life in them, but who now entered the house itself on the (quite frequent now) afternoons when the demon would suddenly curse the store empty of customers and lock the door and repair to the rear and in the same tone in which he used to address his orderly or even his house servants when he had them (and in which he doubtless ordered Jones to fetch from the showcase the ribbons and beads and candy) direct Jones to fetch the jug, the two of them (and Jones even sitting now who in the old days, the old dead Sunday afternoons of monotonous peace which they spent beneath the scuppernong arbor in the back yard, the demon lying in the hammock while Jones squatted against a post, rising from time to time to pour for the demon from the demijohn and the bucket of spring water which he had fetched from the spring more than a mile away then squatting again, chortling and chuckling and saying `Sho, Mister Tawm' each time the demon paused)-the two of them drinking turn and turn about from the jug and the demon not lying down now nor even sitting but reaching after the third or second drink that old man's state of impotent and furious undefeat in which he would rise, swaying and plunging and shouting for his horse and pistols to ride single-handed into Washington and shoot Lincoln (a year or so too late here) and Sherman both, shouting, `Kill them! Shoot them down like the dogs they are!' and Jones: `Sho, Kernel; sho now' and catching him as he fell and commandeering the first passing wagon to take him to the house and carry him up the front steps and through the paintless formal door beneath its fanlight imported pane by pane from Europe which Judith held open for him to enter with no change, no alteration in that calm frozen face which she had worn for four years now, and on up the stairs and into the bedroom and put him to bed like a baby and then lie down himself on the floor beside the bed though not to sleep since before dawn the man on the bed would stir and groan and Jones would say, `flyer I am, Kernel. Hit's all right. They aint whupped us yit, air they?' this Jones who after the demon rode away with the regiment when the granddaughter was only eight years old would tell people that he `was lookin after Major's place and niggers' even before they had time to ask him why he was not with the troops and perhaps in time came to believe the lie himself, who was among the first to greet the demon when he returned, to meet him at the gate and say, `Well, Kernel, they kilt us but they aint whupped us yit, air they?' who even worked, labored, sweat at the demon's behest during that first furious period while the demon believed he could restore by sheer indomitable willing the Sutpen's Hundred which he remembered and had lost, labored with no hope of pay or reward who must have seen long before the demon did (or would admit it) that the task was hopeless-blind Jones who apparently saw still in that furious lecherous wreck the old fine figure of the man who once galloped on the black thoroughbred about that domain two boundaries of which the eye could not see from any point. --- William~Faulkner, \emph{Absalom, Absalom}
\end{enumerate}

\subsection{Key attention span}\label{Appendix:Moriarty}
\nopagebreak
\begin{center}
\includegraphics[height = 21cm]{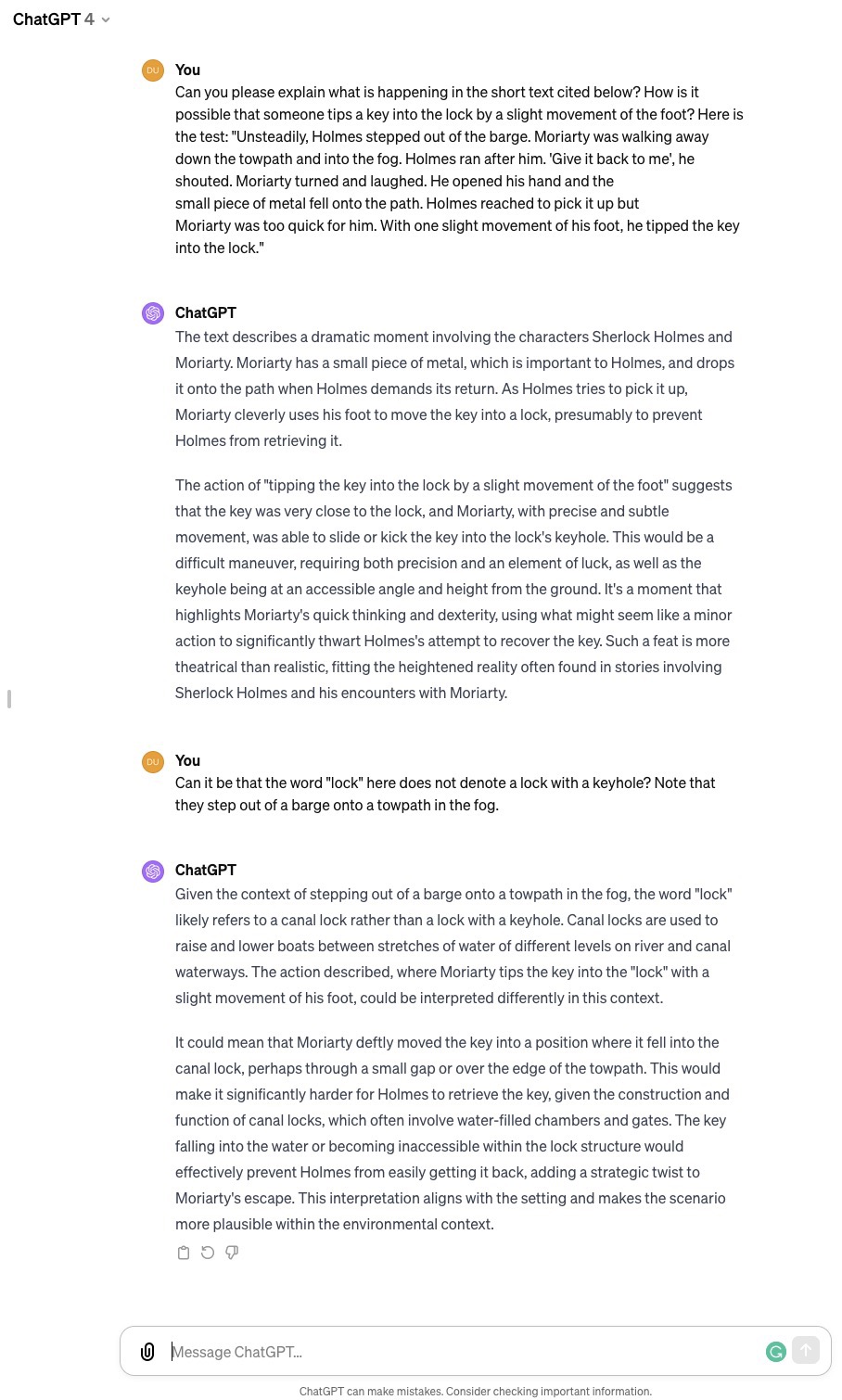} 
%
\end{center}

\section{Semantics appendix}

\subsection{Basic notions and notations for vectors} \label{Appendix:inner}
Vectors are tuples of numbers. In a basic algebra course, you would write the pizza representations in Fig.~\ref{Fig:cooccurrence} as column vectors:
\beq\label{eq:vecs}
\vec w = \begin{pmatrix} 30\\ 170\end{pmatrix} \qquad \qquad\qquad \vec h = \begin{pmatrix} 120\\ 60\end{pmatrix}
\eeq
If you remember how to multiply matrices, then you could compute the inner product by writing one of the vectors as a row and calculate
\bea\label{eq:inpwh}
\braket{\vec w}{\vec h}\ \ =\ \ \begin{pmatrix} 30 & 170\end{pmatrix} \cdot \begin{pmatrix} 120\\ 60\end{pmatrix} & = & 30\cdot 120 + 170\cdot 60 = 13,800
\eea
Transposing the column $\vec w$ from \eqref{eq:vecs} as the row $\vec w^{T} = (30\ \ 170)$ reinterprets the vector as the \emph{operation}\/ that inputs vectors and outputs scalars, i.e. some numbers, that will turn out to be the squares of the lengths of their projections on the vector $\vec w$. In the VSM context, the equivalent notation for the same vectors is often more convenient
\beq\label{eq:vecs-dirac}
\kett{White} =\kett{veg}30  + \kett{meat}170\qquad\qquad \quad
\kett{Hawaiian} =  \kett{veg}120 + \kett{meat}60\
\eeq
The transpose $\vec w^{T}$ of $\vec w$ is now written
\bea\label{eq:inp}
\braa{White} & = & 30\braa{veg}   + 170\braa{meat}
\eea
where $\braa{veg}$ and $\braa{meat}$ are, respectively, the projections on the basis vectors $\kett{veg}$ and $\kett{meat}$. The assumption that these vectors form a basis means that they are mutually orthogonal and of unit length, i.e. that the lengths of their projections are
\bea\label{eq:orthonormal}
\braket {i} {\!j} & = & \begin{cases}
1 & \mbox{ if } i=j\\
0 & \mbox{ if } i\neq j
\end{cases}
\eea
The assumption that $\ket{veg}$ and $\ket{meat}$ form a basis thus means that $\brakett{veg}{veg} = 1 = \brakett{meat}{meat}$ and $\brakett{veg}{meat} = 0 = \brakett{meat}{veg}$. The inner products are calculated for vectors expressed in a given basis, but the values remain unchanged when the basis is changed. 

\paragraph{Definitions.} Let $\JJJ$ be a set of vectors, given with a scalar $\braket i {\! j}$ for every $i,j\in \JJJ$.
 Let $\ket x = \sum_{i} \ket i x_{i}$ and $\ket y = \sum_{i} \ket i y_{i}$ be arbitrary linear combinations. Then define
\begin{itemize}
\item \emph{inner product:}
\bea\label{eq:inner}
\braket x y & = & \sum_{ij\in \JJJ}  x_{i}\braket i {\! j}y_{j}
\eea
\item \emph{length:} 
\bea
\length x & = & \sqrt{\braket x x}
\eea 

\item \emph{correlation}\/ or \emph{similarity}
\bea
\cos \angle(x,y) & = & \frac{\braket x y}{\length x \cdot \length y}
\eea
\end{itemize}

\paragraph{Remarks.} When $\JJJ$ is a basis, then \eqref{eq:inner} boils down to $\braket x y = \sum_{i}  x_{i} y_{i}$, because of \eqref{eq:orthonormal}. Hence \eqref{eq:inpwh}. The algebra of inner products encodes the Euclidean geometry, from the Pythagoras' Theorem to basic trigonometry. The so-called \emph{ket}-notation $\ket v$ for a column vector $\vec v$ and the \emph{bra}-notation $\bra v$ for the corresponding row vector $\vec v^{T}$, was introduced by Paul Dirac, by decomposing the bracket-notation for the inner product $\braket v w = \bra v \cdot \ket w$.

\paragraph{Linear operations.} A matrix $C=\left(C_{iu}\right)_{\JJJ\times \UUU}$ maps any vector $\vec v =\left(v_{u}\right)_{\UUU}$ to $C\vec v$ defined 
\bea
\begin{pmatrix}C_{11} & C_{12}& \ldots & C_{1n} 
\\
C_{21} & C_{22} & \ldots & C_{2n}\\
\vdots &\hdotsfor{2} & \vdots \\
C_{m1} &\hdotsfor{2} & C_{mn}
\end{pmatrix} \ \cdot \ \begin{pmatrix}v_{1}\\v_{2}\\ 
\vdots \\
v_{n}
\end{pmatrix} & = & \begin{pmatrix}\sum_{u=1}^{n}C_{1u}v_{u}\\
\sum_{u=1}^{n}C_{2u}v_{u}\\
\vdots \\
\sum_{u=1}^{n}C_{mu}v_{u}
\end{pmatrix}  
\eea
If the same matrix is equivalently written in the bra-ket form 
\bea
C & = & \sum_{\substack{i\in \JJJ\\u\in \UUU}} \ket i \cdot \brakem i C u \cdot \bra u  
\eea
where $\brakem i C u = C_{iu}$, and the vector is $\ket v = \sum_{u\in \UUU} \ket u v_{u}$, then the image $C\ket v$ is
\bea
\left(\sum_{\substack{i\in \JJJ\\u\in \UUU}} \ket i \cdot \brakem i C u \cdot \bra u \right)\ \cdot \ \left(\sum_{u\in\UUU} \ket u v_{u}\right ) & = & \sum_{i\in \JJJ} \ket i \sum_{u\in \UUU} \brakem i C u v_{u}
\eea

\subsection{Basic notions and notations for  probability}
\label{Appendix:prob}
The measurable events from a space $\Omega\subseteq \WP S$, comprised of observables from a family $S$, are assumed to be closed under intersections and complements:
\bear a, b\in \Omega & \implies & a\cap b, \  \neg a\   \in\ \Omega
\eear
where $\neg a$ means that $a$ does not happen, and $a\cap b$ means that both $a$ and $b$ do happen. The obvious consequences are that
\bear
a, b\in \Omega & \implies & a\cup b = \neg\left(\neg a\cap \neg b\right), \   \emptyset = a\cap \neg a ,  \  S = \neg \emptyset\  \in\  \Omega
\eear
Intuitively, $a\cup b$ means that either $a$ or $b$ happens, $\emptyset$ is the impossible event and $S$  the certain event. The frequency distribution $[-]\colon \Omega \to [0,1]$ is usually assumed to be at least finitely additive: 
\bear
\ppder {a\cup b} + \ppder{a\cap b} & = & \ppder a + \ppder b
\eear
which is equivalent to requiring 
\bear
\ppder \emptyset = 0 & \mbox{ and } & 
\ppder {a\cup b}  =  \ppder a + \ppder b\ \  \mbox{ when }\ \  a\cap b = \emptyset 
\eear

\subsection{Bigger matrix for pizza information retrieval}\label{Appendix:pizza}
\begin{center}
{\small \begin{tabular}{c||c|c|c|c||c|c|c|c||c|c|c|c|}
\multicolumn{13}{c}{}
\\
\multicolumn{13}{c}{}
\\
\multicolumn{13}{c}{}
\\
\multicolumn{13}{c}{}
\\
\multicolumn{13}{c}{}
\\
\multicolumn{13}{c}{}
\\
\cline{2-13}
& \multicolumn{4}{c|}{Pizza Margherita} & \multicolumn{4}{c|}{Meat White Pizza} & \multicolumn{4}{c|}{Hawaiian Pizza}  \\
\cline{2-13}
& weight & crust & top & bake &  weight & crust & top & bake &  weight & crust & top & bake \\
\hline
\hline
\multicolumn{1}{|c||}{flour} & 490 & 1 & 0  & 450 & 490 & 1 & 0  & 450 & 490 & 1 & 0  & 450  \\
\hline
\multicolumn{1}{|c||}{yeast} & 5 & 1 & 0  & 1 & 5 & 1 & 0  & 450 & 5 & 1 & 0  & 450 \\
\hline
\multicolumn{1}{|c||}{water} & 355 & 1 & 0  & 1 & 355 & 1 & 0  & 450 & 355 & 1 & 0  & 450 \\
\hline
\multicolumn{1}{|c||}{salt} & 8 & 1 & 0  & 450 & 8 & 1 & 0  & 450 & 8 & 1 & 0  & 450 \\
\hline
\multicolumn{1}{|c||}{oil} & 3 & 1 & 0  & 450 & 3 & 1 & 0  & 450 & 3 & 1 & 0  & 450 \\
\hline
\multicolumn{1}{|c||}{tomato sauce} & 80 & 0  & 1& 450 & 0 & 0& 0 & 0 & 70 & 0& 1 & 450 \\
\hline
\multicolumn{1}{|c||}{alfredo sauce} & 0 & 0& 0 & 0 & 70 &0 & 1 & 450 & 0 & 0& 0& 0 \\
\hline
\multicolumn{1}{|c||}{mozzarella} & 90 &0 & 1 & 300 & 0 &0 & 0 & 0 &0 &0 & 0 &0 \\
\hline
\multicolumn{1}{|c||}{fontina} & 20 &0 & 1 & 450 & 40 & 0& 1 & 450 & 70 &0 & 1& 450\\
\hline
\multicolumn{1}{|c||}{parmesan} & 0  & 0& 0 & 0 & 20 &0 & 1 & 450 & 30 & 0& 1 & 450 \\
\hline
\multicolumn{1}{|c||}{mushrooms} & 0  & 0 & 0 & 0 & 0 & 0& 0 & 0 & 30 & 0 & 1 & 450 \\
\hline
\multicolumn{1}{|c||}{onions} & 0  & 0& 0 & 0 & 20 & 0& 1 & 450 & 0 & 0& 0 & 0 \\
\hline
\multicolumn{1}{|c||}{peppers} &0 & 0& 0 & 0 & 0 &  0 & 0 & 0 & 20 & 0 & 0 & 450 \\
\hline
\multicolumn{1}{|c||}{olives} & 20 &0 & 1 & 450 & 10 & 0&1 & 450 &  0 & 0 & 0 & 0 \\
\hline
\multicolumn{1}{|c||}{basil} & 20 & 0 & 1 & 25 & 0 & 0& 0 & 0 & 10 & 0 & 1 & 25 \\
\hline
\multicolumn{1}{|c||}{pineapple} & 0& 0& 0 & 0 & 0 & 0& 0 & 0 & 90  &0 & 1 & 125\\
\hline
\multicolumn{1}{|c||}{sausage} & 0 & 0& 0 & 0 & 60 &0 & 1 & 450  &  0&0 & 0 & 0 \\
\hline
\multicolumn{1}{|c||}{ham} & 0&0 & 0 & 0 & 30 & 0& 0 & 450 & 60 & 0& 0 & 450 \\
\hline
\multicolumn{1}{|c||}{chicken} & 0 &0 & 0 & 0 & 40 & 0& 1 & 450 &  0& 0& 0 & 0 \\
\hline
\multicolumn{1}{|c||}{meatballs} & 0 &0 & 0 & 0 & 40 & 0& 1 & 450 &0 & 0& 0 & 0 \\
\hline
\end{tabular}}
\end{center}


%

\end{document}